\documentclass{article}
\pdfoutput=1

\usepackage[final]{neurips_2023}

\usepackage[utf8]{inputenc} 
\usepackage[T1]{fontenc}    
\usepackage{url}            
\usepackage{booktabs}       
\usepackage{nicefrac}       
\usepackage{microtype}      
\usepackage{xcolor}         

\usepackage{fancyhdr}       
\usepackage{graphicx}       
\graphicspath{{media/}}     
\usepackage{adjustbox}

\usepackage{wrapfig,lipsum}

\usepackage{amssymb,latexsym,amsfonts,amsmath,amsthm}

\usepackage{thmtools}
\usepackage{xspace}

\usepackage{booktabs}
\usepackage{mathtools}
\usepackage{tikz} 
\usepackage{dsfont}

\usepackage{amsmath}

\usepackage{mathtools}

\usepackage{dsfont}

\usepackage{graphicx} 
\usepackage{booktabs} 

\usepackage{color,soul}
\usepackage{bm}
\usepackage{graphicx}
\usepackage{enumitem}
\usepackage{pythonhighlight}
\usepackage{nccmath}
\usepackage{algorithm}
\usepackage{algpseudocode}

\usepackage{booktabs}
\usepackage{hyperref}
\usepackage{float,caption,hypcap}

\usepackage{booktabs}
\usepackage{pifont}

\usepackage{scalerel}
\usepackage{comment}

\usepackage{neurips_2023}
\usepackage[capitalize]{cleveref}
\usepackage[textsize=tiny]{todonotes}

\newtheorem{theorem}{Theorem}
\newtheorem{assumption}[theorem]{Assumption}

\newtheorem{corollary}[theorem]{Corollary}
\newtheorem{Lemma}[theorem]{Lemma}

\newtheorem{proposition}[theorem]{Proposition}
\newtheorem{definition}[theorem]{Definition}
\newtheorem{remark}[theorem]{Remark}

\renewcommand*{\proofname}{\textcolor{black}{Proof}}
\renewcommand{\Cref}[1]{\cref{#1}}
 
\renewcommand{\b}[1]{{\bf #1}}

\newcommand{\X}{\b{X}}
\newcommand{\Hx}{\b{H}}
\newcommand{\hx}{\b{h}}
\newcommand{\tHx}{\widetilde{\Hx}}

\newcommand{\A}{\b{A}}
\newcommand{\Aint}{\b{\tilde{A}}}

\newcommand{\U}{\b{U}}
\newcommand{\V}{\b{V}}
\newcommand{\wV}{\widetilde{\b{V}}}
\newcommand{\wc}{\widetilde{c}}
\newcommand{\tu}{\widetilde{u}}

\newcommand{\td}{\widetilde{d}}

\newcommand{\W}{\b{W}}
\newcommand{\Q}{\b{Q}}
\newcommand{\Z}{\b{Z}}

\newcommand{\tP}{\widetilde{\b{P}}}

\newcommand{\blambda}{\bm \Lambda}

\newcommand{\D}{\b D}
\newcommand{\F}{\b F}

\DeclareMathOperator{\prob}{\mathbb{P}}
\DeclareMathOperator{\E}{\mathbb{E}}

\DeclareMathOperator{\Tr}{Trace}

\newcommand{\simplex}{\Delta_{\widetilde{d}}}

\DeclareMathOperator*{\argmin}{arg\,min}
\DeclareMathOperator*{\argmax}{arg\,max}

\definecolor{mygreen}{RGB}{0,153,0}
\definecolor{light-gray}{gray}{0.93}
\definecolor{mid-gray}{gray}{0.88}

\newcommand{\ip}{\b{H_{\scaleto{{p}}{5pt}}}}
\newcommand{\iq}{\b{H_{\scaleto{{q}}{5pt}}}}
\newcommand{\hp}{{\b H_p}}
\newcommand{\hq}{{\b H_q}}
\newcommand{\bhn}{\b H_n}

\newcommand{\hpr}{\b{H_{\scaleto{{p_r}}{5pt}}}}
\newcommand{\hpopt}{\b{H_{\scaleto{{q_r^*}}{8pt}}}}
\newcommand{\hpemp}{\b{H_{\scaleto{{\widehat{p}_r^*}}{8pt}}}}

\newcommand{\hqr}{\b{H_{\scaleto{{q_r}}{5pt}}}}

\newcommand{\vecgrad}{\vec{\nabla} \ell_{(x,y)}(\theta_*)}
\newcommand{\gradloss}{\nabla\ell_{(x,y)}(\theta_*)}

\newcommand{\emp}{Q_n}
\newcommand{\emptheta}{\theta_n}

\newcommand{\tayloremp}{\widetilde{\theta}_n}
\newcommand{\taylorQ}{\widetilde{z}_n}

\newcommand{\mpn}{\b{M}_{p,n}}
\newcommand{\mqn}{\b{M}_{q,n}}

\newcommand{\vecnabla}{\vec{\nabla}}
\newcommand{\vecDelta}{\vec{\Delta}}

\newcommand{\tx}{\widetilde{x}}
\newcommand{\tz}{\widetilde{z}}

\newcommand{\ball}{\Theta_{\overline{r}}(\theta_*)}
\newcommand{\ttheta}{\widetilde{\theta}}
\newcommand{\dtilde}{\widetilde{d}}
\newcommand{\bSigma}{\bm{\Sigma}}
\newcommand{\bLambda}{\bm{\Lambda}}

\newcommand{\bPsi}{\bm{\Psi}}
\newcommand{\bGamma}{\bm{\Gamma}}

\newcommand{\tI}{\tHx}
\newcommand{\B}{\b B}
\newcommand{\tD}{\widetilde{\b D}}

\makeatletter
\newcommand*{\rom}[1]{\expandafter\@slowromancap\romannumeral #1@}
\makeatother

\newcommand{\COMMENT}[2][.45\linewidth]{%
  \leavevmode\hfill\makebox[#1][l]{//~#2}}

\makeatletter
\newenvironment{proof*}[1][\proofname]{\par
  \pushQED{\qed}%
  \normalfont \partopsep=\z@skip \topsep=\z@skip
  \trivlist
  \item[\hskip\labelsep
        \itshape
    #1\@addpunct{.}]\ignorespaces
}{%
  \popQED\endtrivlist\@endpefalse
}
\makeatother

\newcommand{\nbone}{\ding{182}\xspace}
\newcommand{\nbtwo}{\ding{183}\xspace}
\newcommand{\nbthree}{\ding{184}\xspace}

\makeatletter
\newcommand{\crefnames}[3]{%
  \@for\next:=#1\do{%
    \expandafter\crefname\expandafter{\next}{#2}{#3}%
  }%
}
\makeatother
\crefnames{part,chapter,section}{\S}{\S\S}

\author{%
  Youguang Chen\hspace{4em}
  George Biros \\
  Oden Institute for Computational Engineering and Sciences\\
  The University of Texas at Austin \\
}

\begin{document}
\title{FIRAL: An Active Learning Algorithm for Multinomial Logistic Regression}

\maketitle

\allowdisplaybreaks

\begin{abstract}
We investigate theory and algorithms for pool-based active learning for multiclass classification using multinomial logistic regression.  Using finite sample analysis, we prove that the Fisher Information Ratio (FIR)  lower and upper bounds  the excess risk. Based on our theoretical analysis, we propose an  active learning algorithm that  employs regret minimization to minimize the FIR. To verify our derived excess risk bounds, we conduct experiments on synthetic datasets. Furthermore, we compare FIRAL with five other methods and found that our scheme  outperforms them: it consistently produces the smallest classification error in the multiclass logistic regression setting, as demonstrated through experiments on MNIST, CIFAR-10, and 50-class ImageNet. 
\end{abstract}

\section{Introduction}

Active learning is of interest in applications with large pools of unlabeled data for which labeling is expensive. In pool active learning, we're given a set of unlabeled points $U$, an initial set of labeled points $S_0$, and a budget of new points $b$, our goal is to algorithmically select  $b$ new points to label in order to minimize the log-likelihood error over the unlabeled points. Equivalently instead of selecting points directly, we seek to find a probability density function that we can use to sample the $b$ points. Informally (precise formulation in ~\Cref{sec:formulation}), let $x$ denote a data point and $p(x)$ denote the distribution density  of unlabeled points. Let $q(x)$ be the sampling distribution we will use to select the new $b$ points for labeling. We will choose $q(x)$ in order to minimize the generalization error (or excess risk) of the classifier over $p(x)$.  Our theory is classifier specific: it assumes  multinomial logistic regression with parameters $\theta$. The expectations of the Hessian---with respect to $\theta$---of the classifier  loss function over $p(x)$ and $q(x)$ distributions  are  denoted by $\hp$ and $\hq$ respectively. Using finite sample analysis, our first result (\Cref{thm:sub-thm} in \Cref{sec:theory}) is to show that the unlabeled data excess risk is bounded below and above by  
the \emph{Fisher information ratio} $\text{Trace}(\hq^{-1}\hp)$,  subject to the assumption of $p$ being a sub-Gaussian distribution. Our second result (\Cref{thm:mtd-sub-performance} in~\Cref{sec:method}) is to propose and analyze a point selection algorithm based on regret minimization that allows us to bound the generalization error.



There is a large body of work on various active learning methods based on  uncertainty estimation (\cite{Joshi2009,Li2013,Settles2009}), sample diversity (\cite{Sener2017,Gissin2019discriminative}), Bayesian inference ( \cite{Gal2017,Pinsler2019bayesian}), and many others (\cite{ren2021survey}). Here we just discuss the papers closest to our scheme.  Zhang and Oles \cite{zhang-2000} claimed without proof that FIR is asymptotically proportional to the log-likelihood error of unlabeled data. Sourati et~al.~\cite{sourati-2017}  proved that FIR is an upper bound of the expected variance of the asymptotic distribution of the log-likelihood error. Chaudhuri et~al.~\cite{chaudhuri-2015} proved  non-asymptotic results indicating that FIR is closely related to the expected log-likelihood error of an Maximum Likelihood Estimation-based classifier in bounded domain. In this work, we use finite sample analysis to establish FIR-based bounds for the excess risk in the case of multinomial logistic regression with sub-Gaussian assumption for the point distributions. 

Algorithmically finding points to  minimize FIR is an NP-hard combinatorial optimization problem. There have been several approximate algorithms proposed for this problem. Hoi et~al. \cite{hoi-2006} studied the binary classification problem and approximated the FIR using a submodular function and then used a greedy optimization algorithm. Chaudhuri et~al.~\cite{chaudhuri-2015} proposed an algorithm that  first solves a relaxed continuous convex optimization problem, followed by randomly sampling from the weights. Although they derived a performance guarantee for their approach, it needs a substantial number of samples to approach near-optimal performance solely through random sampling from the weights, and no numerical experiments results were provided using such approach. Ash et~al. \cite{ash-2021} adopted a forward greedy algorithm to initially select an excess of points and then utilized a backward greedy algorithm to remove surplus points. But such approach has no performance guarantee. Hence, there is still a need for computationally efficient algorithms that can optimize FIR in a multi-class classification context while providing provable guarantees.

Our  proposed algorithm, FIRAL, offers a locally near-optimal performance guarantee in terms of selecting points to optimize FIR. In our algorithm we have two steps: first we solve a continuous convex relaxation of the  original problem in which we find selection weights for all points in $U$.  Then given these weights, we select $b$ points for labeling by a regret minimization approach. This two-step scheme is inspired by Allen-Zhu et al.~\cite{Allen-2017} where a similar approach was used selecting points for linear regression.  Extending this approach to active learning for multinomial logistic regression has two main challenges. Firstly, we need to incorporate the information from previously selected points in each new round of active learning. Additionally, while the original approach selects features of individual points, in logistic regression, we need to select a Fisher information matrix ($\hq$), which complicates the computation and derivation of theoretical performance guarantees.  In Section~\ref{sec:method}, we present our approach in addressing these challenges.

\paragraph{Our Contributions.} \nbone In \Cref{sec:theory} we prove that FIR is a lower and upper bound of the excess risk for multinomial logistic regression under sub-Gaussian assumptions. \nbtwo In \Cref{sec:method} we detail our FIR Active Learning algorithm (FIRAL) and  prove it selects $b$ points that lead to a bound to the excess risk. \nbthree In \Cref{sec:experiments} we evaluate our analysis empirically on synthetic and real world datasets: MNIST, CIFAR-10, and ImageNet using a subset of 50 classes. We compare FIRAL with several other methods for pool-based active learning.

\section{Problem Formulation}\label{sec:formulation}

We denote a labeled sample as a pair $(x,y)$, where $x\in\mathbb{R}^d$ is a data point, $y \in \{1,2,\cdots, c \}$ is its label, and $c$ is the number of classes. Let $\theta\in\mathbb{R}^{(c-1) \times d}$ be the parameters of a $c$-class logistic regression classifier. Given $x$ and $\theta$, the likelihood of the label $y$ is defined by
\begin{align}\label{eq:setup-conditional}
    p(y|x,\theta) = \begin{cases}
   \frac{\exp(\theta_y^\top x)}{1 + \sum_{l\in[c-1]} \exp(\theta_l^\top x)} ,\qquad y \in [c-1]\\
    \frac{1}{1 + \sum_{l\in[c-1]} \exp(\theta_l^\top x)},\qquad y = c.
    \end{cases}
\end{align}
We use the negative log-likelihood as the loss function: $\ell_{(x,y)}(\theta) \triangleq -\log p(y|x,\theta)$. To simplify notation we define $\td = d(c-1)$. We derive standard  expressions for the gradient  $\nabla \ell_{(x,y)}(\theta) \in \mathbb{R}^{(c-1)\times d}$ and Hessian $\nabla^2\ell_{(x,y)}(\theta) \in \mathbb{R}^{\td \times \td}$  in the Appendix~\ref{appendix:grad-hessian-loss} (Proposition~\ref{prop:grad-and-hessian}).
We assume  there exists $\theta_*$ such that $p(y|x) = p(y|x, \theta_*)$.
Then, given  $p(x)$,  the joint $(x,y)$ distribution is given by
\begin{align}\label{eq:setup-joint}
   \pi_p(x,y) = p(y|x, \theta_*) p(x).
\end{align}
Then, the expected loss at $\theta$ is defined by
\begin{align}\label{eq:setup-generalization-error}
    L_p(\theta) \triangleq \E_{(x,y) \sim \pi_p}[\ell_{(x,y)}(\theta)] = \E_{x\sim p(x)} \E_{y\sim  p(y|x,\theta_*)}[\ell_{(x,y)}(\theta)].
\end{align}
The excess risk  of $p(x)$ at $\theta$ is defined as $R_p(\theta)=L_p(\theta) - L_p(\theta_*)$. Note that $R_p(\theta)\geq 0$.

\paragraph{Notation:} The inner product between two matrices is $ \A \cdot \B = \text{Trace}(\A^\top \B)$. For a matrix $\A\in \mathbb{R}^{m\times n}$,  let  $\|\A\|$ be the spectral norm of $\A$, let $\mathrm{vec}(\A)\in\mathbb{R}^{mn}$ be the vectorization of $\A$ by stacking all rows together, i.e. $\mathrm{vec}(\A) = (\A_1^\top, \cdots, \A_m^\top)^\top$ where $\A_i$ is $i$-th row of $\A$.  Given a positive definite matrix $\A \in \mathbb{R}^{d\times d}$, we define norm $\| \cdot\|_\A$ for $x\in \mathbb{R}^d$ by $\|x\|_\A = \sqrt{x^\top \A x}$. For integer $k\geq 1$, we denote by $\b I_k$ the $k$-by-$k$ identity matrix. For any point distribution $p(x)$ we define $\V_p \triangleq \E_{x\sim p(x)}[x x^\top]$ to be its covariance matrix,  $\hp(\theta) \triangleq \nabla^2 L_p(\theta)$ be the Hessian matrix of $L_p(\theta)$, define $\hp \triangleq \hp(\theta_*)$. 


\paragraph{Active learning.} Let  $U = \{x_i\}_{i=1}^m$, be the set of unlabeled points and $S_0$ be the set of $n_0$ initially labeled samples. In particular, we denote the set of points in $S_0$ as $X_0$. Let $\theta_0$ be the solution of training a classifier with $S_0$, i.e., $\theta_0 \in \argmin_{\theta} \frac{1}{n_0}\sum_{(x,y)\in S_0}\ell_{(x,y)}(\theta)$. We select a set of $b$ points $X\subset U$, query their labels $y \sim p(y| x,\theta_*)$, $\forall x \in X$, and train a new classifier 
$\theta_n \in \argmin_{\theta} \frac{1}{n}\sum_{(x,y)\in S}\ell_{(x,y)}(\theta)$,
where $S$ is the set of $S_0$ with new labeled points and $n=n_0 + b$.

%
%
%
%
Our goal is to optimize the selection of $X$  so that we can minimize the excess risk on the original unlabeled set $U$, i.e. $L_p(\theta_n) - L_p(\theta_*)$.  In this context, we define two problems:
\begin{enumerate}[label= Problem \arabic*,leftmargin=*,noitemsep]
\item\!\!\!: \normalcolor Given $X$ or equivalently $q(x)$,  can we bound $L_p(\theta_n) - L_p(\theta_*)$? \label{item:problem1}
 \item\!\!\!:  \normalcolor Can we construct an efficient algorithm for finding $X$ that minimizes $L_p(\theta_n) - L_p(\theta_*)$?\label{item:problem2}
\end{enumerate}

\section{Excess Risk Bounds}\label{sec:theory}
In this section, we develop our theory to address~\ref{item:problem1}. Our plan is to endow $p(x)$ and $q(x)$ with certain properties (sub-Gaussianity or finite support) and derive FIR bounds for $L_p(\theta_n) - L_p(\theta_*)$. 
%
 Let $\emptheta$ be the empirical risk minimizer (ERM) obtained from $n$ i.i.d. samples  drawn from $\pi_q(x,y)$:
 \begin{align}\label{eq:ERM-q}
    \emptheta\in \argmin_{\theta}  \frac{1}{n}\sum_{i=1}^n \ell_{(x_i,y_i)}(\theta), \qquad \forall i\in[n], \quad(x_i,y_i)\stackrel{\text{i.i.d.}}{\sim} \pi_q(x,y).
 \end{align}

We assume that both $p(x)$ and $q(x)$ are sub-Gaussian distributions. \Cref{appendix:sub-Gaussian} gives a brief review of definitions and basic properties of  sub-Gaussian random variables (vector). We define the $\psi_2$-norm of a sub-Gaussian random variable $x \in \mathbb{R}$ as $\|x\|_{\psi_2}\triangleq \inf \{t>0:\E \exp(x^2/t^2) \leq 2\}$. For a sub-Gaussian random vector $x\in\mathbb{R}^d$, $\| x\|_{\psi_2} = \sup\{\|u^\top x \|_{\psi_2}: \|u\|_2 \leq 1 \}$. We formalize our assumption for $p(x)$ and  $q(x)$ in~\Cref{assume:sub-gaussian}. Based on this assumption, we can derive some properties for the gradient and Hessian of  $\ell_{(x,y)}(\theta)$ shown in\Cref{lm:sub-constants} (proof can be found in~\cref{appendix:parameter}). We present the results for $q$ (thus the subscript in the $K$ constants); exactly the same results, with different constants hold for $p$. 
\begin{assumption}\label{assume:sub-gaussian}
Let  $q(x)$  be a sub-Gaussian distribution for $x\in\mathbb{R}^b$, we assume that  $\V_q$ is positive definite. We assume that there exists $r\gtrsim 1$ such that for any $\theta \in \mathcal{B}_{r}(\theta_*) = \{\theta: \|\theta-\theta_* \|_{2,\infty} \leq r \}$, $\hq(\theta)$ is positive definite, where $\|\cdot\|_{2,\infty}$ denotes the maximum row norm of a matrix.
\end{assumption}

\begin{Lemma}\label{lm:sub-constants}
    If Assumption~\ref{assume:sub-gaussian} holds for $q(x)$, then for $(x,y) \sim \pi_q(x,y)$:
\begin{enumerate}[label={(\arabic*)},leftmargin=*,noitemsep]
\item   \label{assume-sub-x} There exists $K_{0,q}>0$ s.t. $ \| \V_q^{-1/2} x\|_{\psi_2} \leq K_{0,q}$.
   \item \label{assume-sub-grad} There exists $K_{1,q} >0$ s.t. $\| \hq^{-1/2} \mathrm{vec}(\nabla \ell_{(x,y)}(\theta_*)) \|_{\psi_2} \leq K_{1,q} $.
       
    \item \label{assume-sub-hessian} There exists $K_{2,q}(r)>0$ s.t. for any $\theta$ in the  ball $ \mathcal{B}_{r}(\theta_*) = \{\theta: \|\theta-\theta_* \|_{2,\infty} \leq r \}$,  
\begin{align}
    \sup_{u\in\mathcal{S}^{\td-1}}\| u^\top \hq (\theta)^{-1/2} \nabla^2 \ell_{(x,y)}(\theta) \hq (\theta)^{-1/2}u\|_{\psi_1}\leq {K}_{2,q}(r),\label{eq:sub-assume-hessian}
    \end{align}
    where $\mathcal{S}^{\td-1}$ is the unit sphere in $\mathbb{R}^{\td}$, norm $\|\cdot\|_{\psi_1}$ is the norm defined for a sub-exponential random variable $z\in \mathbb{R} $ by $\|z\|_{\psi_1} = \inf\{t>0: \E\exp(|z|/t) \leq 2\}$.
\end{enumerate}
\end{Lemma}

Our main result of this section is \Cref{thm:sub-thm}. Under the sample bound given by \Cref{eq:sub-thm-nbound}, we derive high probability bounds for the excess risk in \Cref{eq:sub-thm-risk}.   Details and the proof of Theorem~\ref{thm:sub-thm} can be found in  Appendix~\ref{appendix:sub-thm}. 

\begin{theorem}\label{thm:sub-thm}
    Suppose Assumption \ref{assume:sub-gaussian} holds for both $p(x)$ and $q(x)$. Let $\sigma$ and $\rho>0$ be constants such that $\hp \preceq \sigma \hq$ and $\b I_{c-1}\otimes \V_p \preceq \rho \hp(\theta_*)$ hold. There exit  constants $C_1, C_2$ and $C_3 >0$, such that for any $\delta\in (0,1)$, whenever
\begin{align}\label{eq:sub-thm-nbound}
    n \geq \max \left\{C_1 \td \log(ed/\delta),\  C_2\sigma\rho\Big(\td+ \sqrt{\td} \log(e/\delta)\Big) \right\},
\end{align}
where $\td\triangleq d(c-1)$,  we have with probability at least $1-\delta$,
{
\begin{align}
    \frac{e^{-\alpha} + \alpha -1}{\alpha^2}\, \frac{\hq^{-1} \cdot \hp}{n} &\lesssim   \E [L_p(\emptheta)] - L_p \lesssim \frac{e^{\alpha} - \alpha -1}{ \alpha^2} \, \frac{\hq^{-1}\cdot \hp}{n}. \label{eq:sub-thm-risk}
\end{align}
}
Here $\hp=\hp(\theta_*)$ and $\hq=\hq(\theta_*)$; and $\E$ is the expectation over ${\{y_i\sim p(y_i|x_i, \theta_*)\}_{i=1}^n}$.  Furthermore,
\begin{align}\label{eq:sub-thm-alpha}
   \qquad  \alpha = C_3 \sqrt{\sigma\rho} \sqrt{\big( \td+ \sqrt{\td} \log(e/\delta)\big) /n}.
\end{align}
\end{theorem}



\begin{wrapfigure}{r}{0.44\textwidth}
\centering
\vspace*{-10pt}
\includegraphics[width=0.8\linewidth]{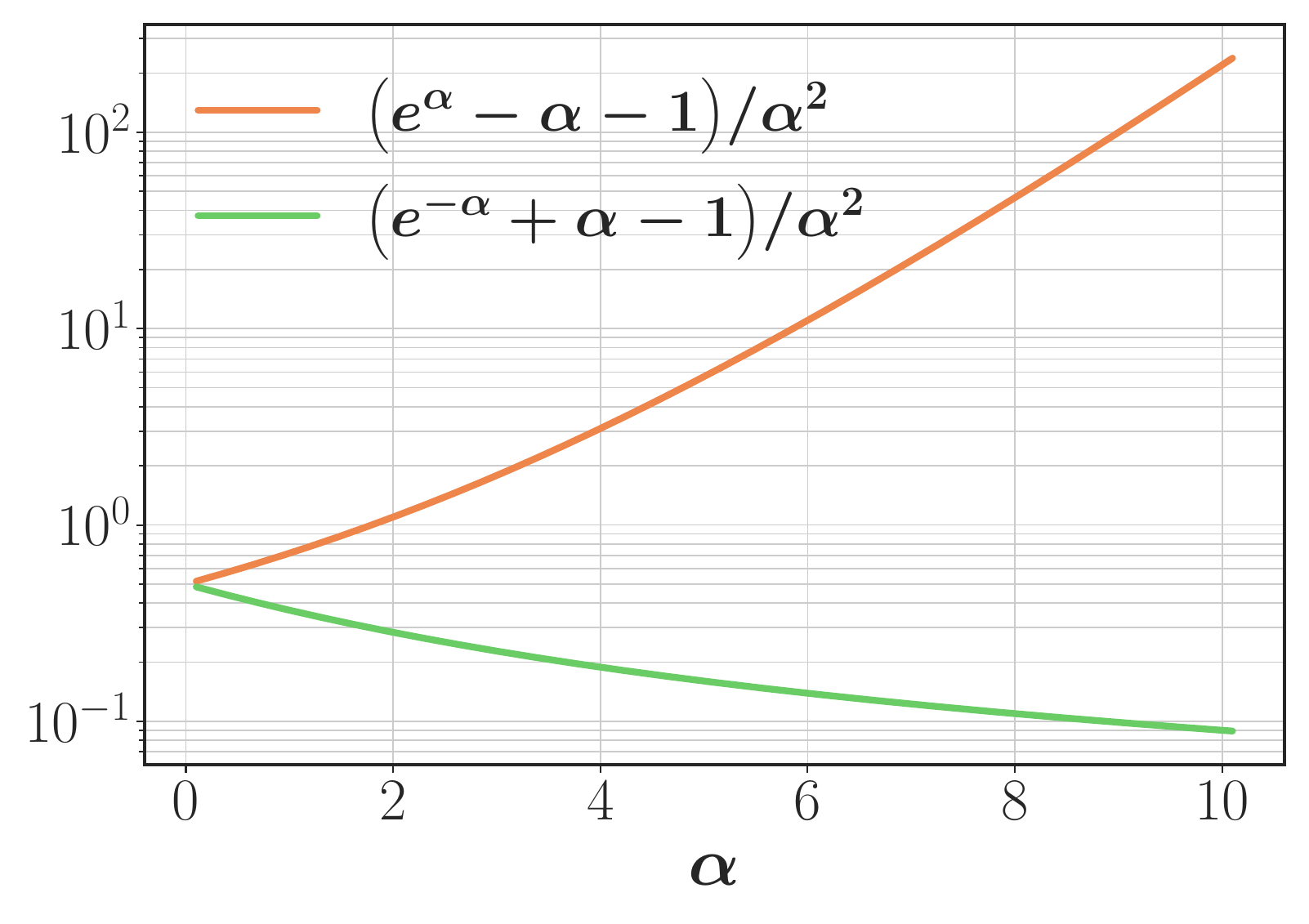}
\caption{\it FIR prefactors in \Cref{eq:sub-thm-risk}.}
\label{fig:fn}
\vspace*{-20pt}
\end{wrapfigure}

From Eq.~\eqref{eq:sub-thm-risk}, we observe that FIR  ($\hq^{-1}\cdot \hp$)  appears in both the lower and upper bounds for $R(\theta_n)$. In other words, it is  essential for controlling the excess risk.  To the right we show how the prefactors $\frac{e^{\alpha}-\alpha-1}{\alpha^2}$ and $\frac{e^{-\alpha} + \alpha -1}{\alpha^2}$ change as a function of $\alpha$. 
Constants $C_1, C_2$ and $C_3$  depend on constants defined in Lemma~\ref{lm:sub-constants} for both $p(x)$ and $q(x)$. 
In  \Cref{appendix:parameter}, we derive bounds for $K_{1,p}$ and $K_{2,p}(r)$ in Proposition~\ref{prop:param-bounds}. For a Gaussian design $x \sim \mathcal{N}(0,\V_p)$, we derive bounds for $\rho$, $K_{0,p}$, $K_{1,p}$ and $K_{2,p}(r)$ in Proposition~\ref{prop:gaussian-design}.

\paragraph{Bounded domain.} If the domain of $x$  is bounded, Chaudhuri et al.\cite{chaudhuri-2015} provided lower and upper bounds for the excess risk of $p(x)$ (Lemma 1 in~\cite{chaudhuri-2015}). Their conclusion is similar to ours, namely that FIR is crucial in controlling the excess risk of $p(x)$. It is worth noting that when the domain is bounded, both $p(x)$ and $q(x)$ are always sub-Gaussian. Thus, our assumption is more general. For the sake of completeness, we provide a detailed derivation of  the excess risk bounds for $p(x)$ in Theorem~\ref{thm:heavy} when $x$ is bounded with \Cref{assumption:heavy}.


\section{Active Learning via Minimizing Fisher Information Ratio}\label{sec:method}
We now discuss the FIRAL algorithm that addresses  \ref{item:problem2}. We can use the theoretical analysis derived in the previous section to guide us for the point selection. Let $p(x)$ be the empirical distribution on unlabeled pool $U$ with $|U|=m$, and $q$  the distribution for the $n=n_0+b$ labeled points. \Cref{eq:sub-thm-risk} inspires us to select points to label such that we can minimize the FIR $\hq^{-1} \cdot \hp$, where $\hq= \hq(\theta_*)$, $\hp=\hp(\theta_*)$. However, we cannot directly use this as $\theta_*$ is unknown. Instead, we will use $\theta_0$, the solution by training the classifier with the initial labeled set $S_0$.\footnote{Such a set can be generated by an alternative method that only uses $p(x)$ to select points, e.g., K-means.} That is, we will find $q$ by minimizing $\hq(\theta_0)^{-1}\cdot\hp(\theta_0)$.

In \Cref{section:mtd-opt}, we formalize our optimization objective in \Cref{eq:mtd-obj-2}. Solving \Cref{eq:mtd-obj-2} exactly is NP-hard \cite{vcerny-2012}. Inspired by \cite{Allen-2017}, we approximate the solution in two steps: \textit{(1) we solve a continuous convex optimization problem in Eq.~\eqref{eq:mtd-obj-relax} (\Cref{section:relaxed}),  (2) and use the results in a regret minimization algorithm to select points by \Cref{eq:mtd-select-direct} (\Cref{section:rounding}).} In \Cref{algo:round} we summarize the scheme. 

We state theoretical guarantees for the algorithm in~\Cref{section:mtd-performance}, where we prove that it achieves $(1+\epsilon)$-approximation of the optimal objective value in Eq.\eqref{eq:mtd-obj-2} with sample complexity $b=\mathcal{O}(\tilde{d}/\epsilon^2)$, as stated in \Cref{thm:mtd-selection}.
Finally, we obtain the excess risk bound for unlabeled points $U$ by accounting for the fact that we use $\theta_0$ instead of $\theta_*$ in the objective function. The overall result is summarized  in  the following theorem.
\begin{theorem}\label{thm:mtd-sub-performance}
   Suppose that \Cref{assume:sub-gaussian} holds. Let $\epsilon\in(0,1)$, $\delta\in(0,1)$, and $b$  the number of points to label. Then with probability at least $1-\delta$, the $\theta_n$---computed by fitting a multinomial logistic regression classifier on the labeled points selected using FIRAL (\Cref{algo:round}) with  learning rate $\eta = 8 \sqrt{\dtilde}/\epsilon$,   $b\geq 32 \dtilde/\epsilon^2 + 16 \sqrt{\dtilde}/\epsilon^2$, and $n=n_{0} +b$ satisfying  \Cref{eq:sub-thm-nbound}---results in 
{
  \begin{align}\label{eq:sub-algo-risk}
      \E[L_{p}(\theta_n)]  - L_{p}(\theta_*) \lesssim (1+\epsilon)\ 2 e^{2\alpha_{0}}\ \frac{e^{\alpha_n} - \alpha_n -1}{ \alpha_n^2} \ \frac{ \text{OPT} }{n}.
  \end{align}
}
Here $\text{OPT}$ is the minimal  $\hq^{-1}\cdot  \hp$,  attained by selecting the best $b$ points from $U$;  $\E$ is the expectation over ${\{y_i\sim p(y_i|x_i,\theta_* )\}_{i=1}^{n}} $; and $\alpha_{0}$ and $\alpha_n$ are constants. 
\end{theorem}

\subsection{Optimization objective}\label{section:mtd-opt}
First we define the precise expression  for $\hq(\theta_0)^{-1} \cdot \hp(\theta_0)$.
We define the Fisher information matrix  $\Hx(x,\theta)  = \nabla^2\ell_{(x,y)}(\theta)$.  By \Cref{eq:loss-hessian} in \Cref{prop:grad-and-hessian} (Appendix~\ref{appendix:B}), we find that for multinomial logistic regression
\begin{align}\label{eq:setup-hessian}
   \Hx(x,\theta)   =   \big[ \text{diag}(\hx(x,\theta)) - \hx (x,\theta) \hx(x,\theta)^\top \big] \otimes (x x^\top),
\end{align}
where $\otimes$ represents the matrix Kronecker product, $\hx(x,\theta)$ is a $(c-1)$-dimensional vector whose $k$-th component is $\hx_k(x,\theta) = p(y=k|x,\theta)$. In \Cref{eq:setup-hessian} we can see that the  Hessian of $\ell_{(x,y)}(\theta)$ does not depend on the class label $y$. 
Following our previous definitions, $\hp(\theta_0) = \nabla^2 L_{p}(\theta_0) = \frac{1}{m}\sum_{x\in U} \Hx(x,\theta_0)$ and $\hq(\theta_0) = \nabla^2 L_{q}(\theta_0) = \frac{1}{n}\sum_{x\in X_0\cup X} \Hx(x,\theta_0)$. For notational simplicity we also define 
\begin{align}
  \Hx(x) \triangleq \Hx(x,\theta_0) + \frac{1}{b}\sum_{x^\prime \in X_{0}}  \Hx(x^\prime,\theta_0) &  \label{eq:mtd-Ix}\\
  \bSigma(z) \triangleq \sum_{i\in[m]} z_i \Hx(x_i),\quad   z_i \mbox{~scalar}. & \label{eq:sigma} 
\end{align}
Then minimizing $\hq(\theta_0)^{-1} \cdot \hp(\theta_0) $ is equivalent to 
\begin{align}\label{eq:mtd-obj-2}
    \min_{\substack{z \in \{0,1\}^m \\\|z\|_1 = b}}  f(z)\triangleq f\Big( \bSigma(z) \Big)  \triangleq \Big(\bSigma(z) \Big)^{-1} \cdot \Hx_p(\theta_0).
\end{align}
We define $z_*$ be the optimal solution  of \Cref{eq:mtd-obj-2} and $f_* \triangleq f(z_*)$. In the following, with some  abuse of notation, we will consider $f$  being a function of either  a vector  $z$ or a positive semidefinite matrix $f(\bSigma)$, depending on the context. 
\Cref{lm:method-f} lists key properties of $f$ when viewed as a matrix function; we use them in~\cref{section:relaxed} to prove the optimality of FIRAL.
 \begin{Lemma}\label{lm:method-f}
  $f: \{ \A \in \mathbb{R}^{\tilde{d}\times \tilde{d}}: \A \succeq \b 0\} \rightarrow \mathbb{R}$  defined in \Cref{eq:mtd-obj-2} satisfies the following properties:
\begin{enumerate}[label={(\arabic*)},leftmargin=*,topsep=0pt,itemsep=-1ex,partopsep=1ex,parsep=1ex,]
    \item \label{mtd-f-1}convex: $f(\lambda \b A + (1-\lambda) \b B) \leq \lambda f(\b A) + (1-\lambda) f(\b B)$ for all $\lambda\in [0,1]$ and $\b A, \b B\in\mathbb{S}^{\widetilde{d}}_{++}$
    \item  \label{mtd-f-2}monotonically non-increasing: if $\b{A} \preceq \b{B}$ then $f({\b A}) \geq f({\b B})$,
    \item \label{mtd-f-3}reciprocally linear:  if $t>0$ then $f(t{\b A})) = t^{-1} f(\b A)$.
\end{enumerate}
 \end{Lemma}

\subsection{Relaxed problem}\label{section:relaxed}

As a first step in solving~\cref{eq:mtd-obj-2} we relax the constraint $z\in \{0,1\}^m$ to $z\in[0,1]^m$. Then we obtain the following convex programming problem:
\begin{align}\label{eq:mtd-obj-relax}
    z_{\diamond} = \argmin_{\substack{z \in [0,1]^m \\\|z \|_1 = b}} f(\bSigma(z)).
\end{align}
Since both the objective function and the constraint set are convex, conventional convex programming algorithm can be used to solve \Cref{eq:mtd-obj-relax}. We choose to use a mirror descent algorithm in our implementation (outlined in the Appendix, \Cref{algo:relax}). Since the integrality constraint is a subset of the relaxed constraint we obtain the following result.
 \begin{proposition}\label{prop:mtd-relax}
$f(z_{\diamond})  \leq f_*$.
 \end{proposition}

In what follows, we use matrices $\bSigma_{\diamond}$ and $\tI(x_i)$ ($i\in[m]$) defined by
\begin{align}\label{eq:mtd-tIx}
     \bSigma_{\diamond} \triangleq \bSigma(z_{\diamond}) \quad \mbox{~and~}\quad \tI(x_i) \triangleq  \bSigma_{\diamond}^{-1/2} \Hx(x_i)\bSigma_{\diamond}^{-1/2}, \quad i \in [m].
\end{align}

\subsection{Solving Sparsification problem via Regret Minimization}\label{section:rounding}

\paragraph{Goal of sparsification.} Now we introduce our method of sparsifying $z_{\diamond}$ (optimal solution to \Cref{eq:mtd-obj-relax}) into a valid integer solution to \Cref{eq:mtd-obj-2}. To do so, we use an online optimization algorithm in which we select one point at a time in sequence until we have $b$ points. Notice that alternative techniques like thresholding $z_{\diamond}$ could be used but it was unclear to us how to provide error estimates for such a scheme. Instead, we use an alternative scheme that we describe below. 

Let $i_t \in [m]$ be the point index selected at step $t\in[b]$. We can observe that if $\lambda_{\min} \big( \sum_{t\in[b]} \tI(x_{i_t}) \big) \geq \tau$ for some $\tau >0$, then $\sum_{t\in [b]}  \Hx(x_{i_t}) \succeq \tau \bSigma_{\diamond}$. By \Cref{lm:method-f}-\Cref{mtd-f-3} and \Cref{prop:mtd-relax}, we obtain the following result.
\begin{proposition}\label{prop:mtd-min-eigen-f-relation}
    Given $\tau\in(0,1)$, we have
    \begin{align}\label{eq:mtd-whitening} 
           \lambda_{\min} \Big(\sum_{t\in [b]}  \tI(x_{i_t}) \Big) \geq \tau &\Longrightarrow f\Big(\sum_{t\in [b]}  \Hx(x_{i_t}) \Big) \leq \tau^{-1} f_*.
    \end{align}
\end{proposition}
From \Cref{eq:mtd-whitening}, a larger $\tau$ value indicates that  $f$ is closer to $f_*$. Therefore, our objective is to choose points in such a way that $ \lambda_{\min} \big(\sum_{t\in [b]} \tI(x_{i_t}) \big) $ is maximized.

\paragraph{Lower bound minimum eigenvalue via Follow-The-Regularized-Leader (FTRL).} We apply FTRL, which is a popular method for online optimization~\cite{online-survey}, to our problem because it can yield a lower bound for $ \lambda_{\min} \big(\sum_{t\in [b]} \tI(x_{i_t}) \big) $ in our setting. FTRL takes $b$ steps to finish.  At each step $t \in [b]$, for a fixed learning rate $\eta>0$, we generate a matrix $\A_t$ defined by 
\begin{align}\label{eq:mtd-At-closed}
   \A_1 = \frac{1}{\td} \b I_{\td},\qquad \A_t = \big(\nu_t \b I_{\tilde{d}} + \eta \sum_{l=1}^{t-1} \tI({x_{i_l}})\big)^{-2} \quad(t\geq 2).
\end{align}
Here $\nu_t$ is the unique constant such that $\Tr(\A_t) =1$. Using \Cref{eq:mtd-At-closed} we can guarantee a lower bound for $ \lambda_{\min} \Big(\sum_{s\in [t]}  \tI(x_{i_t}) \Big)$, which is formalized below:
\begin{proposition}\label{prop:mtd-FTRL}
  Given $A_l$, $l\in[b]$, defined by \Cref{eq:mtd-At-closed} and for all $t\in [b]$ 
    \begin{align}\label{eq:mtd-FTRL-bound}
        \lambda_{\min} \Big(\sum_{l=1}^t \tI(x_{i_l})\Big) 
      &\geq -\frac{2\sqrt{ \dtilde}}{\eta} +  \frac{1}{\eta} \sum_{l=1}^t \Tr\big[\A_l^{1/2} - \big(\A_l^{-1/2}+ \eta \tI(x_{i_l})\big)^{-1}\big].
    \end{align}
\end{proposition}

\paragraph{Point selection via maximizing the lower bound in \Cref{eq:mtd-FTRL-bound}.} 
Now we discuss our choice of point selection at each time step based on \Cref{eq:mtd-FTRL-bound}. Recall that our sparsification goal is to make $ \lambda_{\min} (\sum_{s=1}^t \tI(x_{i_s}) $ as large as possible. Since \Cref{eq:mtd-FTRL-bound} provides a lower bound for such minimum eigenvalue, we can choose $i_t\in [m]$ to maximize the lower bound, which is equivalent to choose
\begin{align}\label{eq:mtd-select-direct}
    i_t \in \argmin_{i\in[m]} \Tr[\big(\A_t^{-1/2}+ \eta \tI(x_{i})\big)^{-1}].
\end{align}
Solving \Cref{eq:mtd-select-direct} directly can become computationally expensive when the dimension $d$, number of classes $c$, and the pool size $n$ are large. This is due to the fact that the matrix $\A_t^{-1/2}+\eta\tI(x_i) \in \mathbb{R}^{\dtilde \times \dtilde}$ (where $\dtilde = d(c-1)$), requiring $n$ eigendecompositions of a $\dtilde \times \dtilde$ matrix to obtain the solution. Fortunately, we can reduce this complexity \textit{without losing accuracy}. 
First, by \Cref{eq:setup-hessian} and \Cref{eq:mtd-Ix}, we have for any $i\in[m]$,
\begin{align}\label{eq:mtd-Ix-new}
   \Hx(x_i) = \underbrace{\frac{1}{b}\sum_{x\in X_{0}}  \Hx(x,\theta_{0})}_{\triangleq\D} + \underbrace{\big[ \text{diag}(\hx(x_i,\theta_0)) - \hx (x_i,\theta_0) \hx(x_i,\theta_0)^\top \big]}_{\triangleq \V_i \blambda_i \V_i^\top} \otimes (x _i x_i^\top) ,
\end{align}
where $\V_i \blambda_i \V_i^\top$ is the eigendecomposition of $ \text{diag}(\hx(x_i,\theta_0)) - \hx (x_i,\theta_0) \hx(x_i,\theta_0)^\top $. Define matrix $\b Q_i\triangleq \V_i \blambda_i^{1/2}$, then $ \tI(x_i)  = \D + (\b Q_i \b Q_i^\top) \otimes  (x _i x_i^\top) $. Substitute this into \Cref{eq:mtd-tIx}, we have a new expression for transformed Fisher information matrix $\tI(x_i)$:
\begin{align}\label{eq:mtd-tIx-new}
    \tI(x_i)&=\underbrace{(\bSigma_{\diamond})^{-1/2}\D(\bSigma_{\diamond})^{-1/2}}_{\triangleq\tD} +  \underbrace{(\bSigma_{\diamond})^{-1/2} (\b Q_i\otimes x_i)}_{\triangleq \tP_i} (\b Q_i \otimes x_i)^\top(\bSigma_{\diamond})^{-1/2}  = \tD + \tP_i \tP_i^\top. 
\end{align}
Now define  $\B_t\in\mathbb{R}^{\dtilde\times \dtilde}$ s.t. $\B_t^{-1/2} = \A_t^{-1/2} + \eta \tD$. By \Cref{eq:mtd-tIx-new}, we have $\A_t^{-1/2} + \eta \tI(x_i) = \B_t^{-1/2} + \eta \tP_i\tP_i^\top$. Applying Woodbury's matrix identity, we have
\begin{align}
    (\A_t^{-1/2} + \eta \tI(x_i))^{-1} = \B_t^{1/2} - \eta \B_t^{1/2} \tP_i (\b I_{c-1} +\eta \tP_i^\top \B_t^{1/2} \tP_i)^{-1}\tP_i^\top \B_t^{1/2}.
\end{align}
Now our point selection objective \Cref{eq:mtd-select-direct} is equivalent to
\begin{align}\label{eq:mtd-select-new}
     i_t\gets \argmax_{i \in [m]} \Big( \b I_{c-1} + \eta \tP_i^\top\B_t^{1/2}  \tP_i\Big)^{-1} \cdot  \tP_i^\top\B_t \tP_i.
\end{align}
 Since $ (\b I_{c-1} + \eta \tP_i^\top\B_t^{1/2}  \tP_i )\in \mathbb{R}^{(c-1)\times (c-1)}$, solving \Cref{eq:mtd-select-new} is faster than solving \Cref{eq:mtd-select-direct}. We summarize FIRAL for selecting $b$ points in \Cref{algo:round}.

 \paragraph{Connection to regret minimization.} Our algorithm is derived as the solution of a regret minimization problem in the adversarial linear bandits setting. We give a brief introduction in Appendix~\ref{appendix:regre-minimization}. Readers who are interested in this topic can refer to Part VI of~\cite{Lattimore-2020}. In our case the action matrix is constrained to  $\{\A \in \mathbb{R}^{\td\times \td}: \A \succeq 0, \Tr(\A) =1 \}$ and is chosen by \Cref{eq:mtd-At-closed}; the loss matrix is constrained to the set of the transformed Fisher information matrices  $\{ \tI(x_i)\}_{i=1}^m$ and is chosen by minimizing \Cref{eq:mtd-select-new}. 

\paragraph{Algorithm complexity.} Our algorithm has two steps: convex relaxation (line 2 in \Cref{algo:round}) and sparsification (lines 3--11). Let $T_{\mathrm{eigen}}(\td)$ be the complexity of eigendecomposition of a $\td$-dimensional symmetric positive definite  matrix. Given an unlabeled point pool $U$ with $m=|U|$, the complexity of solving the convex relaxation problem by mirror descent (\cref{algo:relax}) is $\mathcal{O}\big(m\td^2 \log m+ T_{\mathrm{eigen}}(\td) \log m \big)$, where we assume that the number of iterations is $\mathcal{O}(\log m)$ according to \cref{thm:md-convergence}. Given sample budget $b$, the complexity of solving the sparsification problem is $\mathcal{O}\big( T_{\mathrm{eigen}}(\td)b + T_{\mathrm{eigen}}(c-1) b m\big)$.

\begin{algorithm}[t]
\caption{\textproc{FIRAL}($b$, $U$, $S_0$, $\theta_0$)}
\label{algo:round}
 \hspace*{\algorithmicindent} \textbf{Input:} sample budget $b$, unlabeled pool $U = \{ x_i\}_{i\in [m]}$, labeled set $S_0$, initial ERM $\theta_0$
 
\hspace*{\algorithmicindent} \textbf{Output:} selected points $X$ 
\begin{algorithmic}[1]
\State $X \gets \emptyset$  
\State $z_{\diamond}\gets $ solution of \Cref{eq:mtd-obj-relax}, $\bSigma_{\diamond} \gets \sum_{i=1}^n z_{*,i} \Hx(x_i)$ \hfill \textcolor{lightgray}{\# continuous convex relaxation}
    \State $\V_i \blambda_i \V_i^\top\gets $ eigendecomposition of $\text{diag}(\hx(x_i, \theta_0))-\hx(x_i, \theta_0) \hx(x_i, \theta_0)^\top $, $\forall i \in [m]$
    \State $\tP_i\gets \bSigma_{\diamond}^{-1/2} \big(x_i \otimes (\V_i \blambda_i^{1/2})\big)$, $\forall i \in [m]$ 
\State $\tD \gets$ defined in \Cref{eq:mtd-tIx-new},  $\A_1^{-1/2}\gets \sqrt{\dtilde} \b I_{\dtilde}$, $\B_1^{1/2}\gets (\A_1 ^{-1/2} + \eta \tD)^{-1}$
\For {$t = 1$ to $b$}
    \State $i_t\gets$ solution of \Cref{eq:mtd-select-new}, $X \gets X \cup \{x_{i_t}\}$
    \State $\V \blambda \V^\top \gets $ eigendecomposition of $\eta \sum_{s=1}^t \tI(x_{i_s}) = \eta \sum_{s=1}^t (\tD + \tP_{i_s} \tP_{i_s}^\top )$
    \State find $\nu_{t+1} $ s.t. $\sum_{j\in[\td]}(\nu_{t+1} + \lambda_j)^{-2}=1$
    \State $\A_{t+1}^{-1/2}\gets \V (\nu_{t+1} \b I_{\dtilde} + \blambda)\V^\top$, $\B_{t+1}^{1/2}\gets (\A_{t+1}^{-1/2} + \eta \tD)^{-1}$
\EndFor
\end{algorithmic}
\end{algorithm}

\subsection{Performance guarantee}\label{section:mtd-performance}

We intend to lower bound $\lambda_{\min} \Big(\sum_{t\in[b]} \tI(x_{i_t})\Big)$ through lower bounding the right hand side of \eqref{eq:mtd-FTRL-bound}. First, since our point selection algorithm selects point $x_i$ at each step to maximize $\Tr [\A_t^{1/2} - (\A_t^{-1/2} + \eta \tI(x_i))^{-1}]$, we establish a lower bound for this term at each step, as demonstrated in Proposition \ref{prop:mtd-selection}. 

\begin{proposition}\label{prop:mtd-selection}
 At each step $t\in[b]$, we have
   \begin{align}\label{eq:mtd-trace-bound}
       \max_{i\in[m]} \frac{1}{\eta}\Tr [\A_t^{1/2} - (\A_t^{-1/2} + \eta \tI(x_i))^{-1}] \geq \frac{1-\frac{\eta}{2b}}{b + \eta \sqrt{\dtilde}}.
   \end{align}
\end{proposition}

The derivation is elaborated in Appendix~\ref{append:mtd-selection-prop}. We remark that there is a similar lower bound derived for the optimal design setting in \cite{Allen-2017} (Lemma 3.2), where a rank-1 matrix $\tilde{x}_{i_t} \tilde{x}_{i_t}^\top$ ($i_t\in[m]$ and $\tilde{x}_{i_t}\in \mathbb{R}^d$) is selected at each step. In contrast, in our active learning setting, the selected matrix $\tI(x_{i_t})$ possesses a minimum rank of $c-1$ and can even be a full-rank matrix, contingent upon the labeled points from prior rounds. The distinction between the characteristics of the matrices significantly complicates the derivation of such a general lower bound.

By connecting the observations obtained in this section, we can show that our algorithm can achieve $(1+\epsilon)$-approximation of the optimal objective with sample size $\mathcal{O}(\td/\epsilon^2)$. We conclude our results in Theorem~\ref{thm:mtd-selection}.
\begin{theorem}\label{thm:mtd-selection}
Given $\epsilon\in(0,1)$, let $\eta = 8\sqrt{\dtilde}/\epsilon$, whenever $b\geq 32 \dtilde/\epsilon^2 + 16 \sqrt{\dtilde}/\epsilon^2$, denote the instance index selected by \Cref{algo:round} at step $t$ by $i_t\in[m]$, then the algorithm is near-optimal: $f\Big(\sum_{t\in [b]} \Hx(x_{i_t})\Big) \leq (1+\epsilon)f_*$, where $f$ is the objective function defined in \Cref{eq:mtd-obj-2} and $f_*$ is its optimal value.
\end{theorem}

The excess risk upper bound for unlabeled points can be obtained using our algorithm in Theorem~\ref{thm:mtd-sub-performance} by combining Theorem~\ref{thm:sub-thm} and Theorem~\ref{thm:mtd-selection} while considering the impact of using $\theta_0$ as an approximation for  $\theta_*$. We present the proof in Appendix~\ref{appendix:mtd-sub-performance}.  Comparing Eq.\eqref{eq:sub-algo-risk} to Eq.\eqref{eq:sub-thm-risk}, we observe a factor of $2(1+\epsilon) e^{2 \alpha_0}$ degradation in the upper bound. The $(1+\epsilon)$-term comes from our algorithm, while the $2 e^{2\alpha_0}$-term comes from the use of $\theta_0$ instead of $\theta_*$. This observation suggests that, given a total budget of points to label $b$ we should consider an iterative approach consisting of $r$ active learning rounds. At each round $k$ we label a new batch of size $b/r$ points  and we obtain a new estimate $\theta_k$ that can be used to approximate $\theta_*$.  The prefactor containing $\alpha_0$ will becomes $\alpha_k$  and reduces $\theta_k$ converges to $\theta_*$.  The simplest solution would be to use $r=b$ but this can be computationally expensive. In our tests, we use this batched approach and choose $b/r$ to be a small multiple of $c$.

\section{Numerical Experiments}\label{sec:experiments}

\begin{figure}[!t]
\centering
  \footnotesize
\begin{tikzpicture}
\node[inner sep=0pt] (a1) at (0,0) {\includegraphics[width=3.4cm,height = 2.7cm]{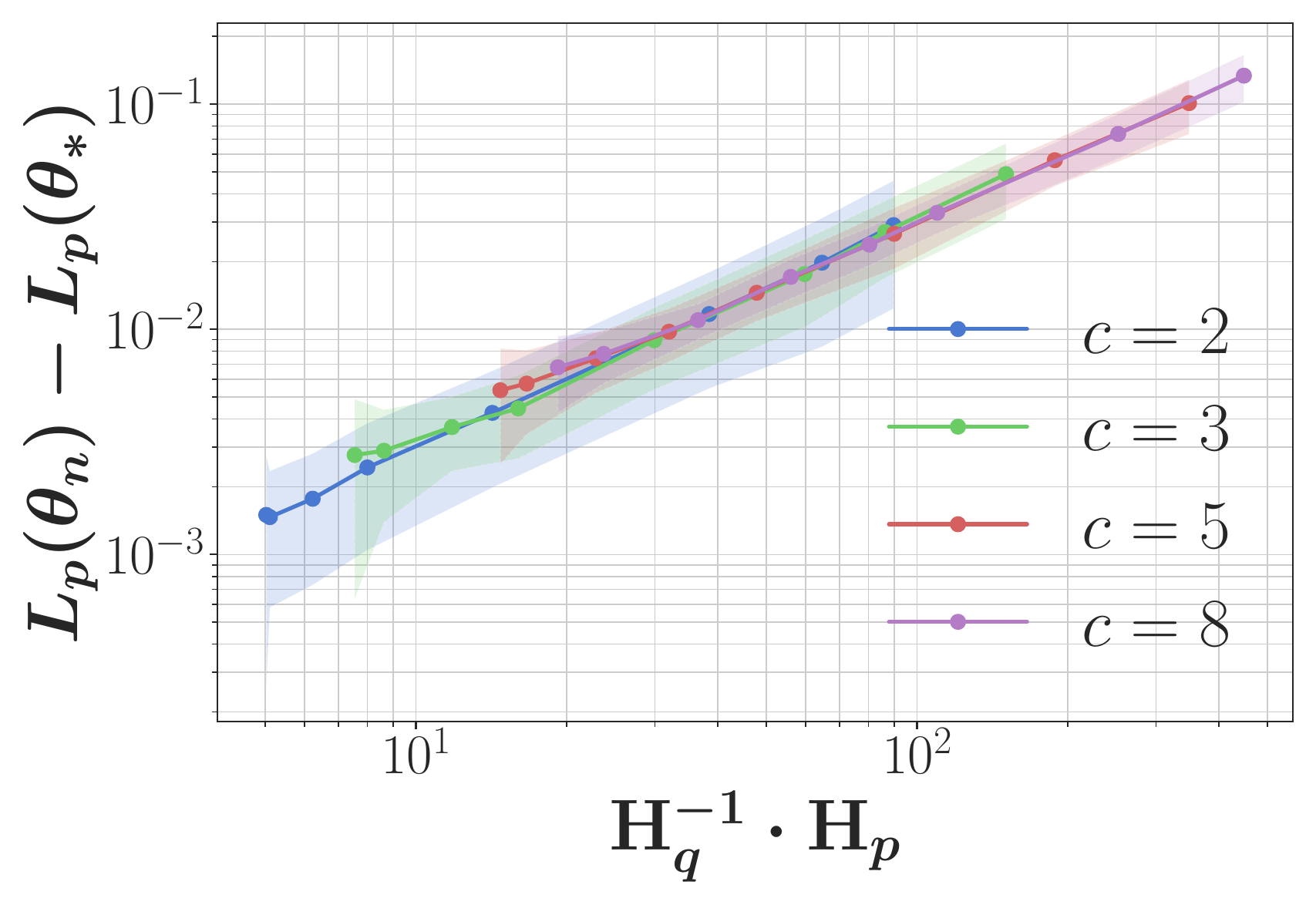}};
\node[inner sep=0pt] (b1) at (3.4,0) {\includegraphics[width=3.4cm,height = 2.7cm]{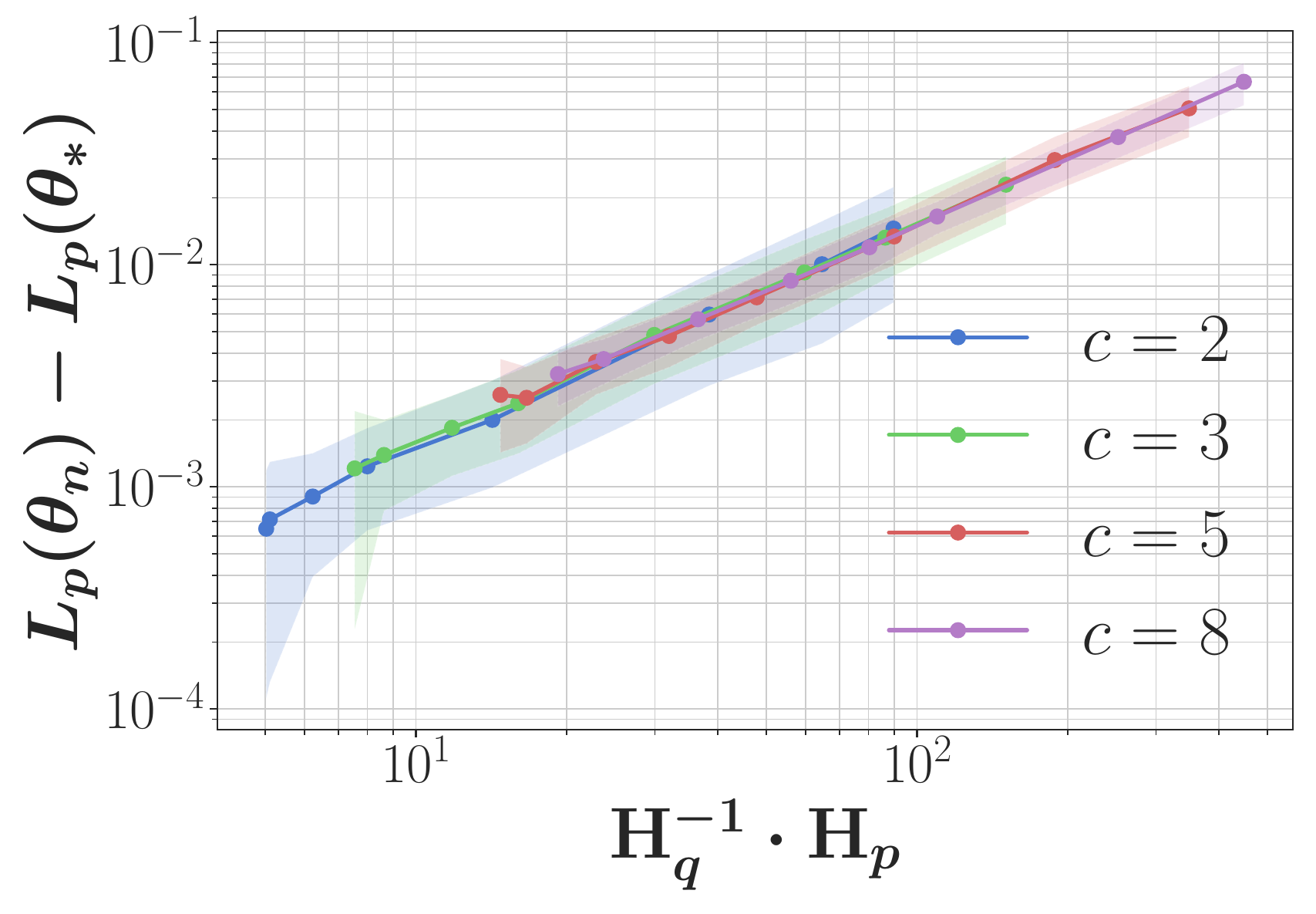}};
\node[inner sep=0pt] (c1) at (6.8,0) {\includegraphics[width=3.4cm,height = 2.7cm]{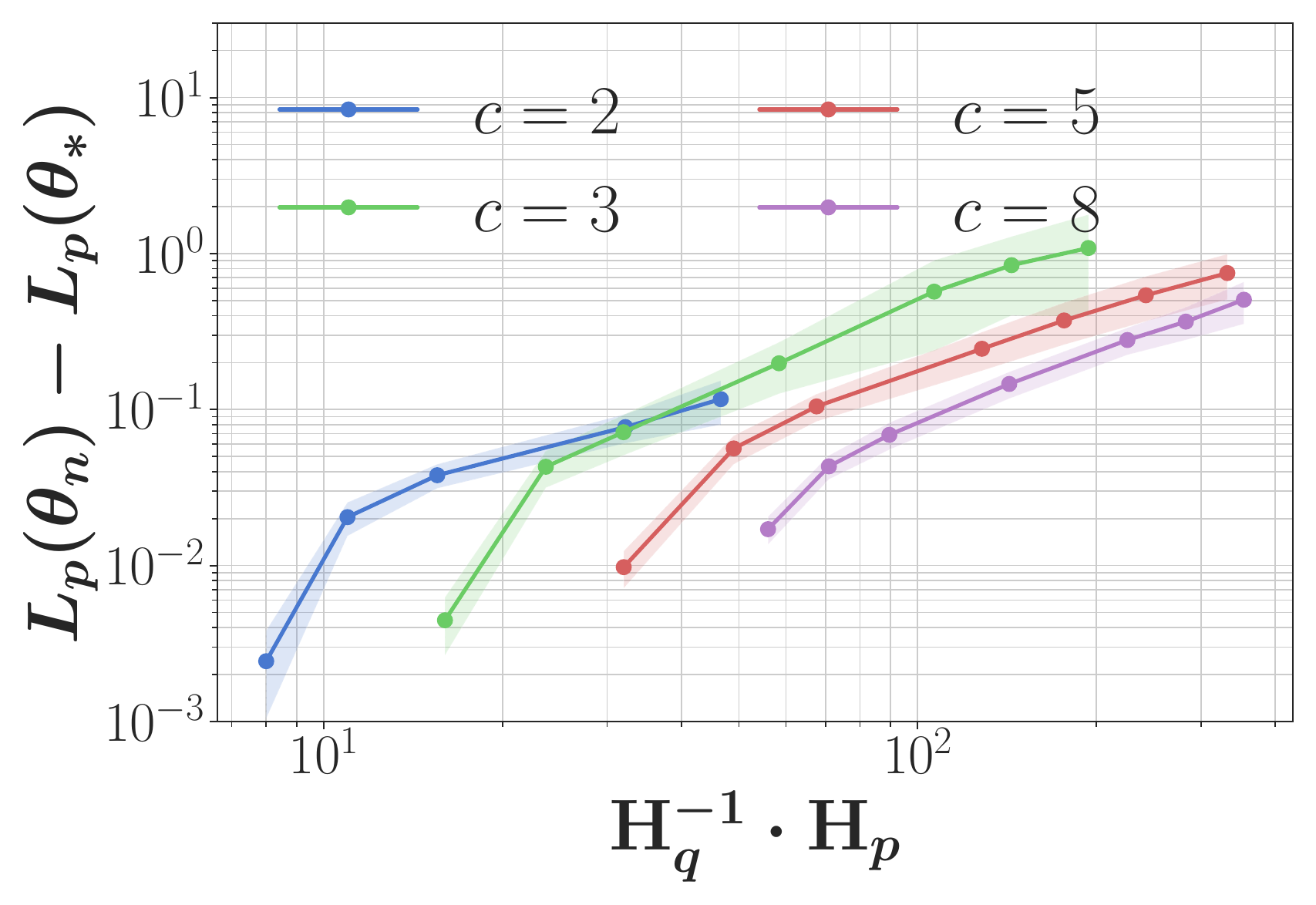}};
\node[inner sep=0pt] (d1) at (10.2,0) {\includegraphics[width=3.4cm,height = 2.7cm]{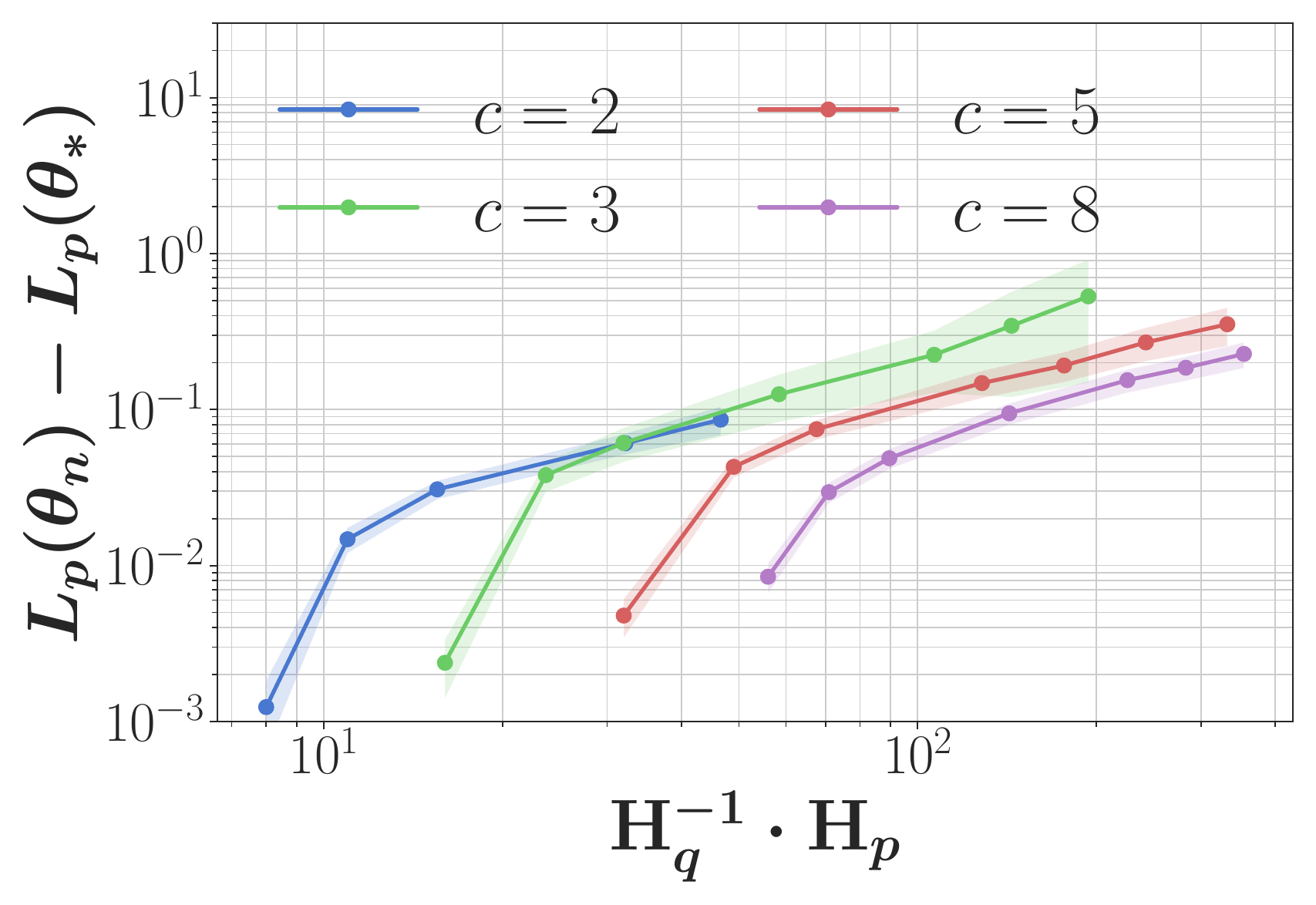}};
\node[]  at (2, 1.8) {\textbf{Dilation} ($\leftarrow:$ increasing $\nu_q$)};
\node[]  at (0.4, 1.4) {$\scriptstyle n=1600$};
\node[]  at (3.7, 1.4) {$\scriptstyle n=3200$};
\node[]  at (8.8, 1.8) {\textbf{Translation} ($\leftarrow:$ decreasing $\tau_q$)};
\node[]  at (7.2, 1.4) {$\scriptstyle n=1600$};
\node[]  at (10.4, 1.4) {$\scriptstyle n=3200$};
\end{tikzpicture}
\caption{\it Synthetic experiments: excess risk of $p(x)$ as a function of the FIR ($\hq^{-1}\cdot \hp$) in  dilation and translation tests.}
\label{fig:syn-risk}
\end{figure}
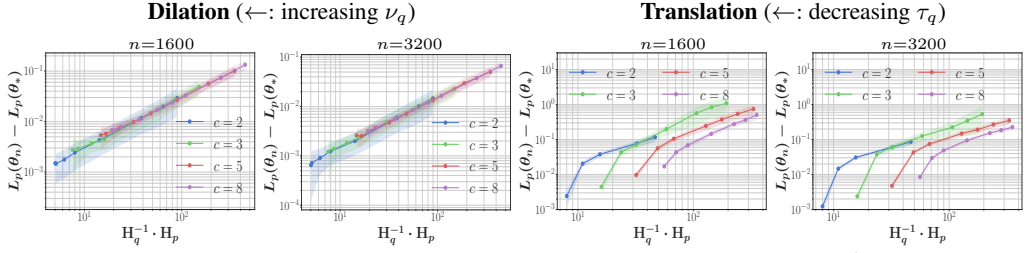

\paragraph{Synthetic datasets.} We use synthetic datasets to demonstrate the excess risk bounds \Cref{eq:sub-thm-risk} derived in \Cref{thm:sub-thm}. We choose $p(x) \sim \mathcal{N}(\b 0, \V_p)$, where $\V_p = 100 \b I_{d}$ and $d=8$.   We explore different numbers of classes denoted by $c\in\{2, 3, 5, 8\}$. We define the ground truth parameter $\theta_*$ in such a way that the points generated from $p(x)$ are nearly equally distributed across the $c$ classes. In \Cref{fig:2d-plot} (\Cref{appendix:synthetic}), we
plot the first two coordinates of the points draw from $p(x)$, where each point is colored by its class id.

We conduct tests using two different types of $q(x)$ based on operations applied to $p(x)$: \textit{dilation} and \textit{translation}.
For the dilation,   $q(x)~\sim \mathcal{N}(\b 0, \nu_q \V_p)$, where $\nu_q \in \mathbb{R}^+$. We vary $\nu_q$ within so that FIR   $(\hq^{-1}\cdot \hp)$ is in $[0.2 \td, 10 \td]$, where $\td = d(c-1)$. For translation,  $q(x) \sim \mathcal{N}(\tau_q \b a, \V_p)$, where $\b a = (1/\sqrt{2}, 1/\sqrt{2}, 0,\cdots, 0)$ and $\tau_q \in \mathbb{R}^+$. We examine various $\tau_q$ values that ensures $\hq^{-1}\cdot \hp \in [ \td, 10 \td]$. For each $c\in\{2,3,5,8\}$, $q(x)$ and $n \in \{1600,3200\}$, we i.i.d. draw $n$ samples from $\pi_q(x)$ and obtain $\theta_n$ defined by \Cref{eq:ERM-q} using these samples. We estimate excess risk $ L_p(\theta_n) - L_p(\theta_*)$ by averaging the log-likelihood error on   $5\times 10^4$ i.i.d. points sampled from $p(x)$. 

 \Cref{fig:syn-risk} displays the excess risk plotted against FIR for both dilation tests (left two plots) and translation tests (right two plots). It is evident that FIR plays a crucial role in controlling the excess risk. In the case of dilation tests, we observe an almost linear convergence rate with respect to FIR. In the translation tests, we observe a faster-than-linear convergence rate, which can be explained by examining the upper bound of \Cref{eq:sub-thm-risk}. As FIR  decreases, $\sigma$ also decreases according to the right plot of \Cref{fig:gaussian-sigma-trace}). By \Cref{prop:gaussian-design}, in our scenario, we have $K_{1,q} \lesssim (100+\tau_q)^{3/4}$. In \Cref{appendix:sub-thm}, it is stated that $C_3 = \mathcal{O}(K_{0,p} K_{1,q}K_{2,p})$. As a result, as FIR decreases, both $C_3$ and $\sigma$ decrease, leading to a decrease in $\alpha$ (as indicated by \Cref{eq:sub-thm-alpha}). Referring to \Cref{fig:fn}, the prefactor of the FIR term in the upper bound decreases as $\alpha$ decreases. Consequently, the upper bound of \Cref{eq:sub-thm-risk} indicates a faster-than-linear convergence rate with respect to the FIR term in the case of translation. We perform similar tests on multivariate Laplace distribution and t-distribution, and the results are consistent with our observations on Gaussian tests. Further details of synthetic experiments are given in \Cref{appendix:synthetic}.

\begin{figure}[!t]
\centering
  \footnotesize
\begin{tikzpicture}
\node[inner sep=0pt] (a0) at (4.6,1.8) {\includegraphics[width=12cm,height = .4cm]{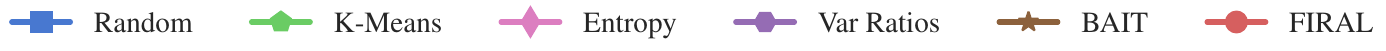}};
\node[inner sep=0pt] (a) at (0,0) {\includegraphics[width=4.5cm,height = 3.3cm]{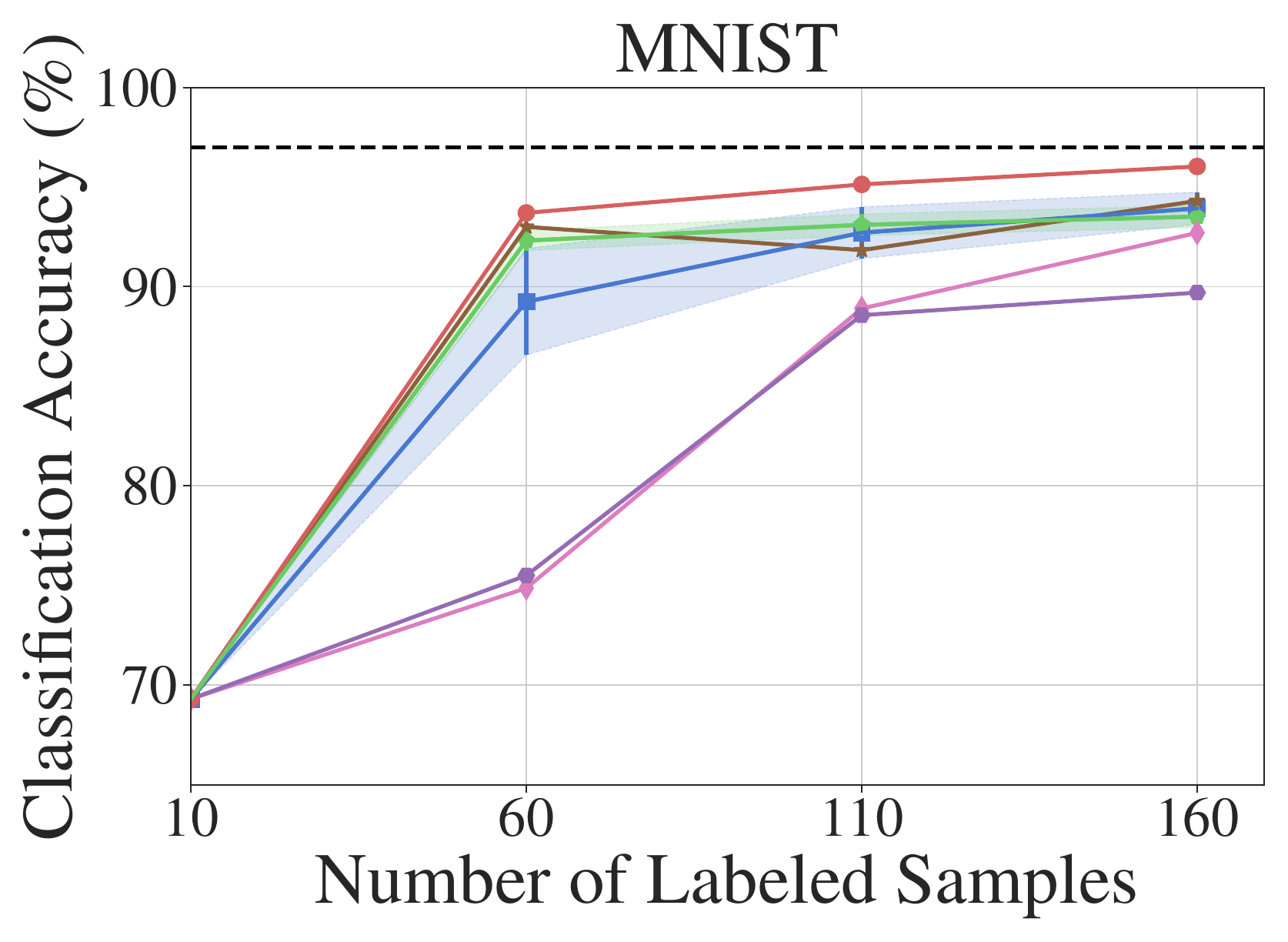}};
\node[inner sep=0pt] (b) at (4.5,0) {\includegraphics[width=4.5cm,height = 3.3cm]{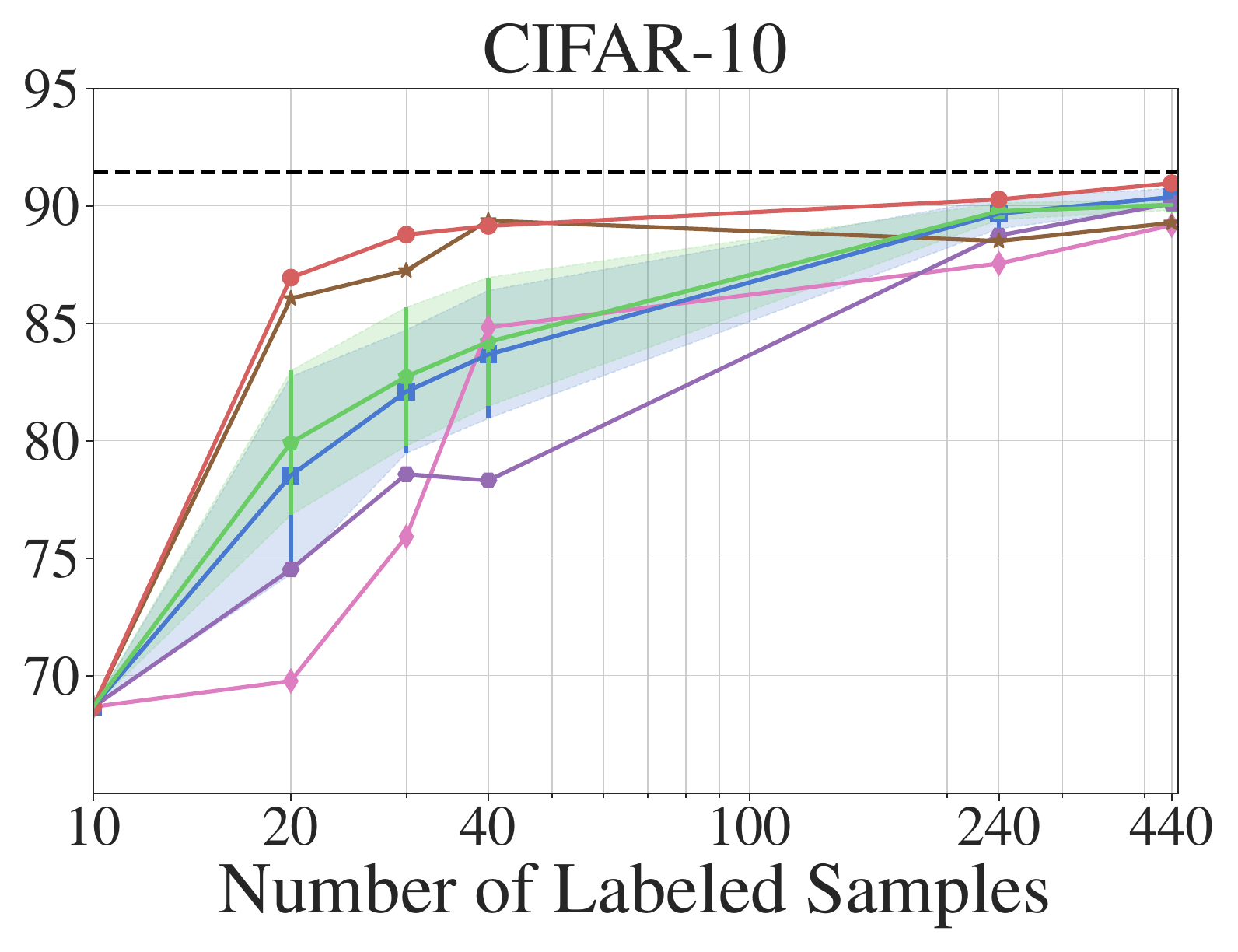}};
\node[inner sep=0pt] (c) at (8.9,0) {\includegraphics[width=4.5cm,height = 3.3cm]{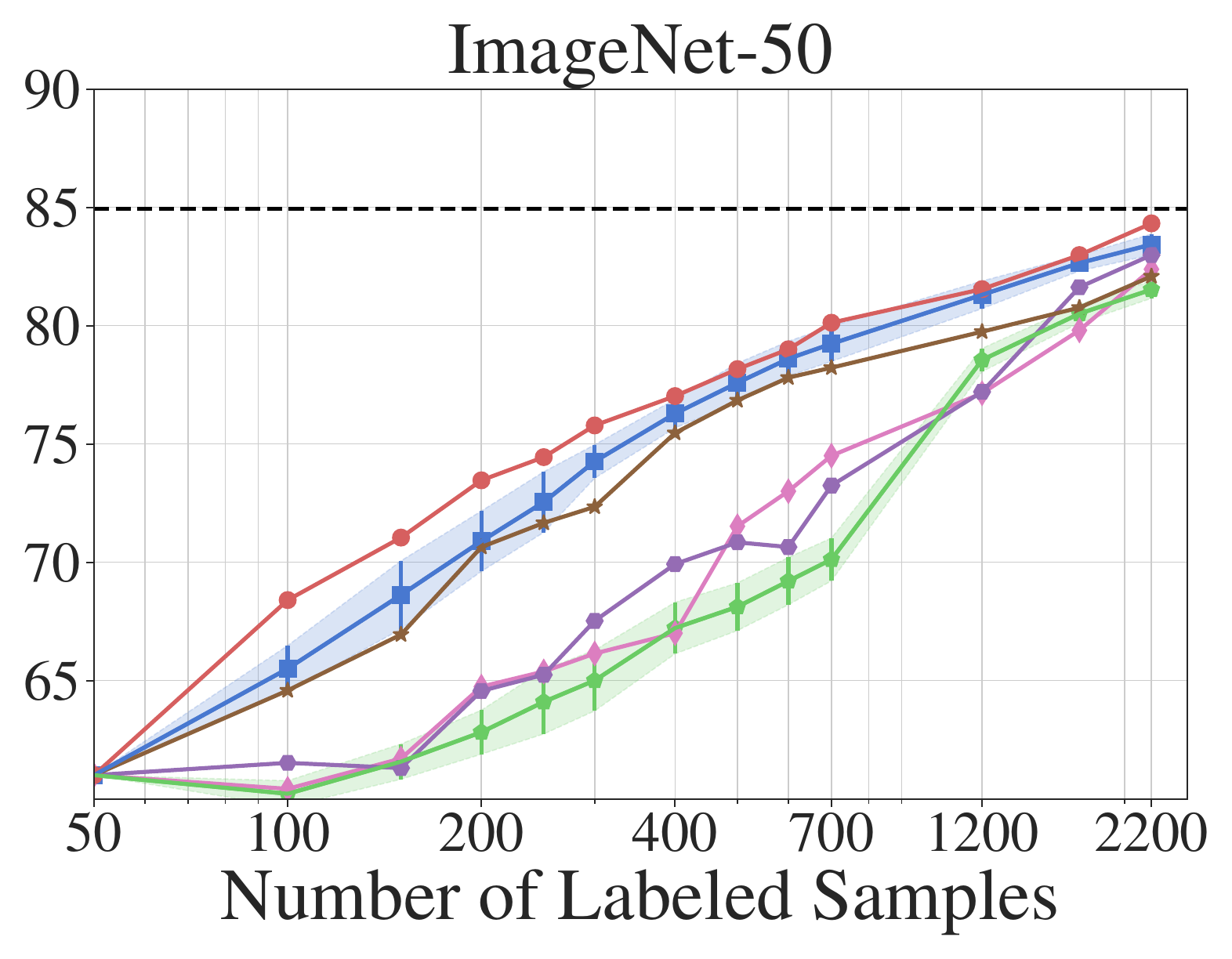}};
\node[inner sep=0pt] (d) at (1,-3.5) {\includegraphics[width=6.5cm,height = 3.3cm]{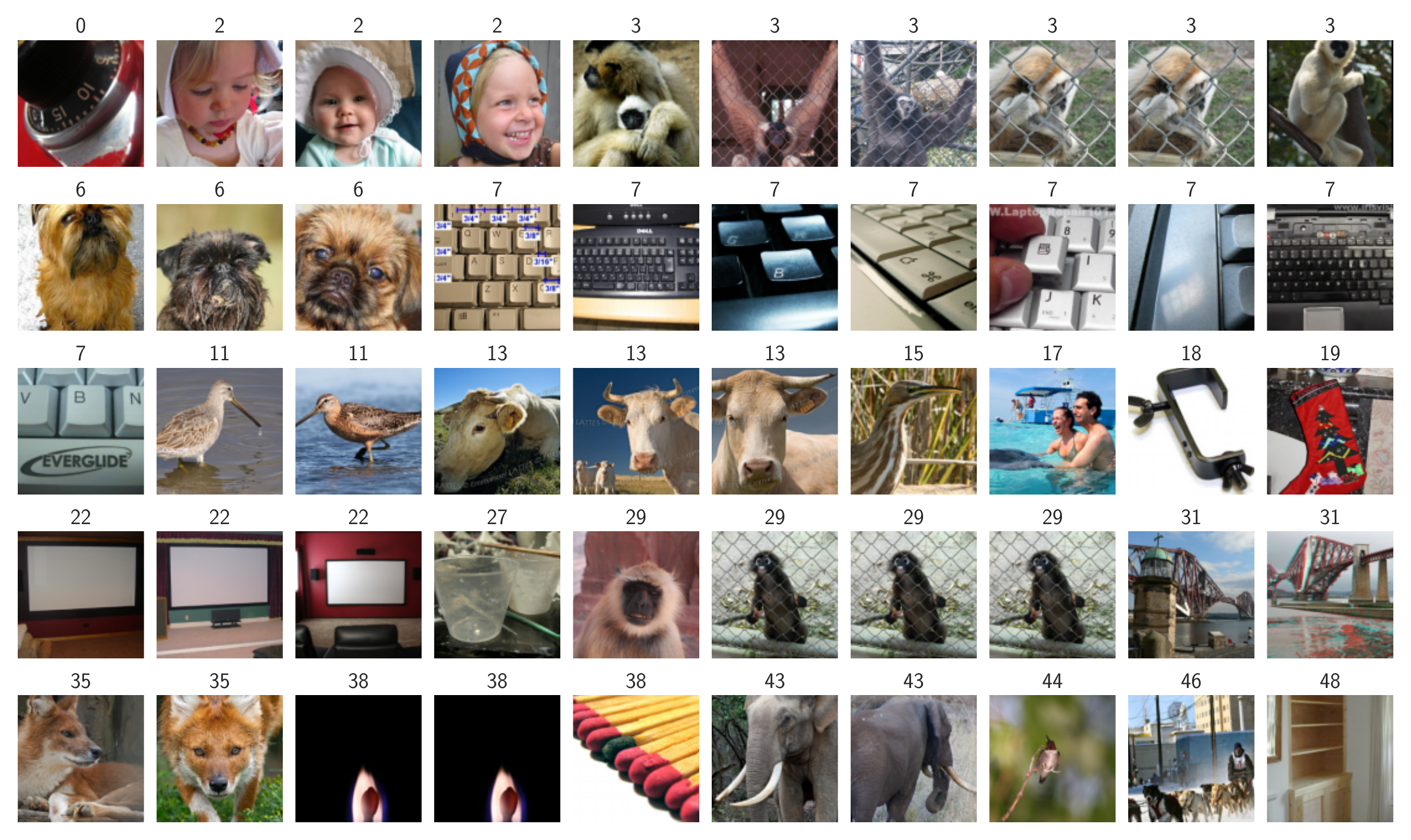}};
\node[inner sep=0pt] (e) at (8,-3.5) {\includegraphics[width=6.5cm,height = 3.3cm]{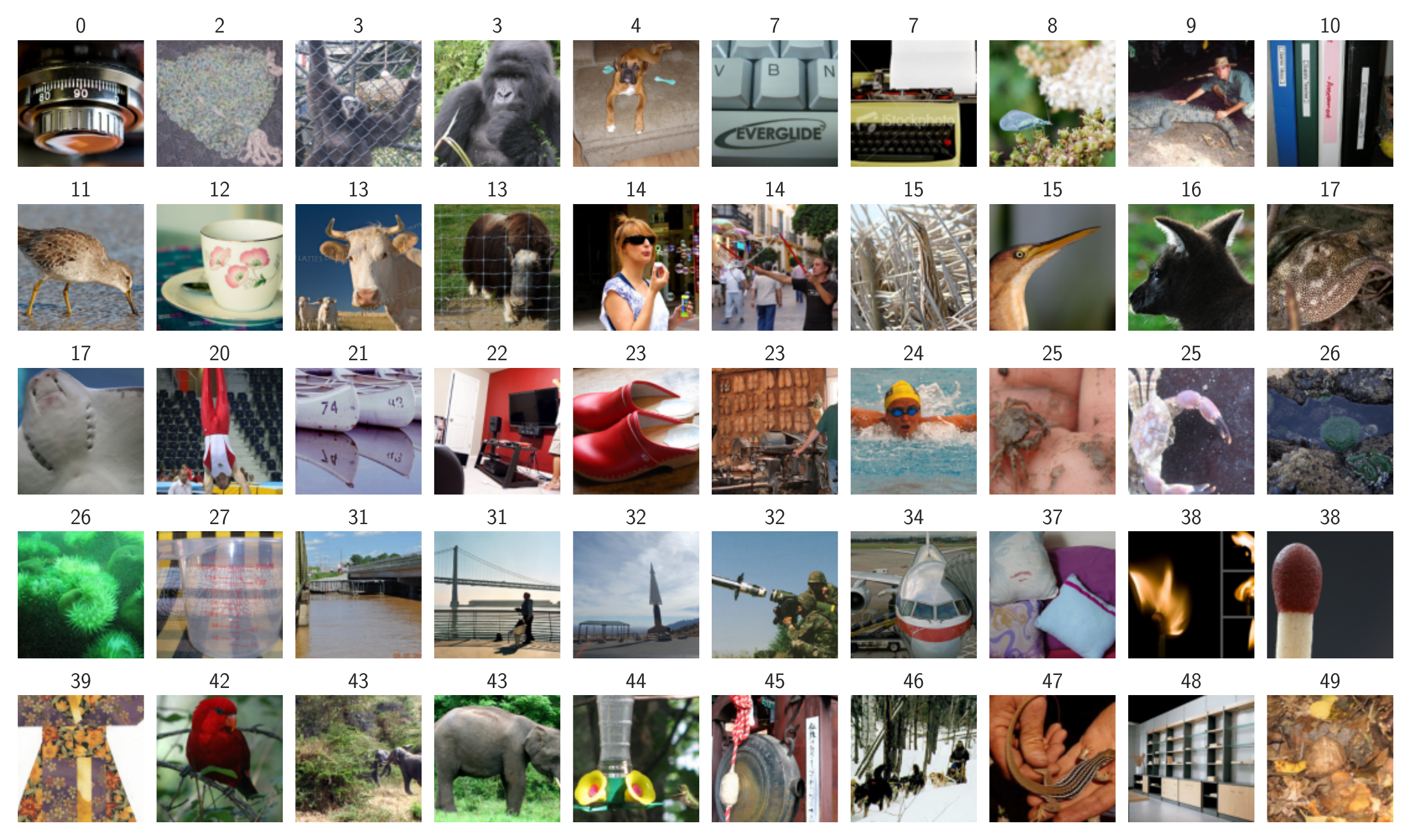}};
\node[]  at (1,-1.8) {BAIT};
\node[]  at (8,-1.8) {FIRAL};
\end{tikzpicture}
\caption{\it Active learning results for MNIST (left) , CIFAR-10 (center) and ImageNet-50 (right). Black dashed lines in the upper row plots are the classification accuracy using all points in $U$ and their labels. The lower row shows 50 images that are selected in the first round of the active learning process for the ImageNet-50 dataset.}
\label{fig:real-accuracy}
\end{figure}

\paragraph{Real-world datasets.}
We demonstrate the effectiveness of our active learning algorithm using three real-world datasets: MNIST~\cite{deng-2012mnist}, CIFAR-10~\cite{cifar10}, and ImageNet~\cite{deng2009imagenet}. In the case of ImageNet, we randomly choose 50 classes for our experiments.  First we use unsupervised learning to extract features and then apply active learning to the feature space, that is, we do \textbf{not} use any label information  in our pre-processing.  For MNIST, we calculate the normalized Laplacian of the training data and use the spectral subspace of the  20 smallest eigenvalues. For CIFAR-10 and ImageNet-50, we use a contrastive learning SimCLR model~\cite{simclr}; then we compute the normalized nearest-neighbor Laplacian and select the subspace of the 20 smallest eigenvalues; For ImageNet-50 we select the subspace of the 40 smallest eigenvalues.   For each dataset, we initialize the labeled data $S_0$ by randomly selecting one sample from each class. Further details about tuning hyperparameter $\eta$ and data pre-processing are given in \Cref{appendix:real-world}.

We compare our algorithm FIRAL with  five methods: (1) Random selection, (2) K-means where $k=b$, (3) Entropy: select top-$b$ points that minimize $\sum_c p(y=c|x)\log p(y=c|x)$ (where $c$ is the  class with the  highest probability), (4) Var Ratios: select top-$b$ points that minimize $p(y=c|x)$ (where $c$ is the  class with the  highest probability), (5) BAIT \cite{ash-2021}: solving the same objective as our method, select $2b$ points and then delete $b$ points, both in greedy way. Random and K-means are non-deterministic, we run each test 20 times. The other methods are deterministic and the only randomness is related to $S_0$. We performed several runs varying $S_0$ randomly and there is no significant variability in the results, so for clarity we only present one representative run. We present the classification accuracy on $U$ in the upper row of \Cref{fig:real-accuracy}. We can observe that our method consistently outperforms other methods across all experiments. K-means, one of the most popular methods due to each simplicity significantly  underperformed FIRAL. It is worth noting that the random selection method serves as a strong baseline in the experiments of ImageNet-50, where our method initially outperforms Random but shows only a marginal improvement in later rounds. But random selection underperforms in CIFAR-10.  In the lower row of \Cref{fig:real-accuracy}, we show the images selected in the first round on ImageNet-50 for BAIT and FIRAL. Images selected in other methods and other datasets can be found in Appendix~\ref{appendix:real-world}. One way to qualitatively compare the two methods is to check the diversity of the samples: in the 50-sample example BAIT samples only 21/50 classes; FIRAL samples 37/50 classes.  This could explain the significant loss of performance of BAIT in the small sample size regime.

\section{Conclusions}\label{sec:conclusion}

We presented FIRAL, a new algorithm designed for the pool-based active learning problem in the context of multinomial logistic regression.  We provide the performance guarantee of our algorithm by deriving a excess risk bound for the unlabeled data. We validate the effectiveness of our analysis and algorithm using experiments on synthetic and real-world datasets.  The algorithm scales linearly in the size of the pool and cubically on the dimensionality and number of classes---due to the eigenvalue solves. The experiments show clear benefits, especially in terms of robustness of performance across datasets,  in the low-sample regime (a few examples per class). 

One limitation of our algorithm is the reliance of a hyperparameter, $\eta$, derived from the learning rate in the FTRL algorithm. There are a large body of work in online optimization about the adaptive FTRL algorithm (e.g., \cite{online-survey}), which eliminates the need for such hyperparameter. In our future work, we will investigate the integration of adaptive FTRL and evaluate its impact on the overall performance of FIRAL.  By exploring this avenue, we aim to enhance the flexibility and efficiency of our algorithm. Another parameter is the number of rounds to use in batch mode, but this we have just set to a small multiple of the number of classes. Other extensions include more complex classifiers and combination with semi-supervised learning techniques.

\section{Acknowledgements}
This material is based upon work supported by NSF award OAC 2204226; by the U.S. Department of Energy, Office of Science, Office of Advanced Scientific Computing Research, Applied Mathematics program, Mathematical Multifaceted Integrated Capability Centers (MMICCS) program, under award number DE-SC0023171; and by the U.S. National Institute on Aging under award number  R21AG074276-01. Any opinions, findings, and conclusions or recommendations expressed herein are those of the authors and do not necessarily reflect the views of the DOE, NIH, and NSF. Computing time on the Texas Advanced Computing Centers Stampede system was provided by an allocation from TACC and the NSF.

 \bibliographystyle{unsrtnat} 
\bibliography{references}  

\begin{thebibliography}{30}
\providecommand{\natexlab}[1]{#1}
\providecommand{\url}[1]{\texttt{#1}}
\expandafter\ifx\csname urlstyle\endcsname\relax
  \providecommand{\doi}[1]{doi: #1}\else
  \providecommand{\doi}{doi: \begingroup \urlstyle{rm}\Url}\fi

\bibitem[Joshi et~al.(2009)Joshi, Porikli, and Papanikolopoulos]{Joshi2009}
Ajay~J. Joshi, Fatih Porikli, and Nikolaos Papanikolopoulos.
\newblock Multi-class active learning for image classification.
\newblock In \emph{2009 IEEE Conference on Computer Vision and Pattern
  Recognition}, pages 2372--2379, 2009.
\newblock \doi{10.1109/CVPR.2009.5206627}.

\bibitem[Li and Guo(2013)]{Li2013}
Xin Li and Yuhong Guo.
\newblock Adaptive active learning for image classification.
\newblock In \emph{2013 IEEE Conference on Computer Vision and Pattern
  Recognition}, pages 859--866, 2013.
\newblock \doi{10.1109/CVPR.2013.116}.

\bibitem[Settles(2009)]{Settles2009}
Burr Settles.
\newblock Active learning literature survey.
\newblock 2009.

\bibitem[Sener and Savarese(2017)]{Sener2017}
Ozan Sener and Silvio Savarese.
\newblock Active learning for convolutional neural networks: A core-set
  approach.
\newblock \emph{arXiv preprint arXiv:1708.00489}, 2017.

\bibitem[Gissin and Shalev-Shwartz(2019)]{Gissin2019discriminative}
Daniel Gissin and Shai Shalev-Shwartz.
\newblock Discriminative active learning.
\newblock \emph{arXiv preprint arXiv:1907.06347}, 2019.

\bibitem[Gal et~al.(2017)Gal, Islam, and Ghahramani]{Gal2017}
Yarin Gal, Riashat Islam, and Zoubin Ghahramani.
\newblock Deep bayesian active learning with image data.
\newblock In \emph{International conference on machine learning}, pages
  1183--1192. PMLR, 2017.

\bibitem[Pinsler et~al.(2019)Pinsler, Gordon, Nalisnick, and
  Hern{\'a}ndez-Lobato]{Pinsler2019bayesian}
Robert Pinsler, Jonathan Gordon, Eric Nalisnick, and Jos{\'e}~Miguel
  Hern{\'a}ndez-Lobato.
\newblock Bayesian batch active learning as sparse subset approximation.
\newblock \emph{Advances in neural information processing systems}, 32, 2019.

\bibitem[Ren et~al.(2021)Ren, Xiao, Chang, Huang, Li, Gupta, Chen, and
  Wang]{ren2021survey}
Pengzhen Ren, Yun Xiao, Xiaojun Chang, Po-Yao Huang, Zhihui Li, Brij~B Gupta,
  Xiaojiang Chen, and Xin Wang.
\newblock A survey of deep active learning.
\newblock \emph{ACM computing surveys (CSUR)}, 54\penalty0 (9):\penalty0 1--40,
  2021.

\bibitem[Zhang and Oles(2000)]{zhang-2000}
Tong Zhang and Frank~J. Oles.
\newblock A probability analysis on the value of unlabeled data for
  classification problems.
\newblock In \emph{17th International Conference on Machine Learning}, 2000.
\newblock URL
  \url{http://www-cs-students.stanford.edu/~tzhang/papers/icml00-unlabeled.pdf}.

\bibitem[Sourati et~al.(2017)Sourati, Akcakaya, Leen, Erdogmus, and
  Dy]{sourati-2017}
Jamshid Sourati, Murat Akcakaya, Todd~K Leen, Deniz Erdogmus, and Jennifer~G
  Dy.
\newblock Asymptotic analysis of objectives based on fisher information in
  active learning.
\newblock \emph{The Journal of Machine Learning Research}, 18\penalty0
  (1):\penalty0 1123--1163, 2017.

\bibitem[Chaudhuri et~al.(2015)Chaudhuri, Kakade, Netrapalli, and
  Sanghavi]{chaudhuri-2015}
Kamalika Chaudhuri, Sham~M Kakade, Praneeth Netrapalli, and Sujay Sanghavi.
\newblock Convergence rates of active learning for maximum likelihood
  estimation.
\newblock \emph{Advances in Neural Information Processing Systems}, 28, 2015.

\bibitem[Hoi et~al.(2006)Hoi, Jin, Zhu, and Lyu]{hoi-2006}
Steven~CH Hoi, Rong Jin, Jianke Zhu, and Michael~R Lyu.
\newblock Batch mode active learning and its application to medical image
  classification.
\newblock In \emph{Proceedings of the 23rd international conference on Machine
  learning}, pages 417--424, 2006.

\bibitem[Ash et~al.(2021)Ash, Goel, Krishnamurthy, and Kakade]{ash-2021}
Jordan Ash, Surbhi Goel, Akshay Krishnamurthy, and Sham Kakade.
\newblock Gone fishing: Neural active learning with fisher embeddings.
\newblock \emph{Advances in Neural Information Processing Systems},
  34:\penalty0 8927--8939, 2021.

\bibitem[Allen-Zhu et~al.(2017)Allen-Zhu, Li, Singh, and Wang]{Allen-2017}
Zeyuan Allen-Zhu, Yuanzhi Li, Aarti Singh, and Yining Wang.
\newblock Near-optimal design of experiments via regret minimization.
\newblock In \emph{International Conference on Machine Learning}, pages
  126--135. PMLR, 2017.

\bibitem[{\v{C}}ern{\`y} and Hlad{\'\i}k(2012)]{vcerny-2012}
Michal {\v{C}}ern{\`y} and Milan Hlad{\'\i}k.
\newblock Two complexity results on c-optimality in experimental design.
\newblock \emph{Computational Optimization and Applications}, 51\penalty0
  (3):\penalty0 1397--1408, 2012.

\bibitem[McMahan(2017)]{online-survey}
H~Brendan McMahan.
\newblock A survey of algorithms and analysis for adaptive online learning.
\newblock \emph{The Journal of Machine Learning Research}, 18\penalty0
  (1):\penalty0 3117--3166, 2017.

\bibitem[Lattimore and Szepesv{\'a}ri(2020)]{Lattimore-2020}
Tor Lattimore and Csaba Szepesv{\'a}ri.
\newblock \emph{Bandit algorithms}.
\newblock Cambridge University Press, 2020.

\bibitem[Deng(2012)]{deng-2012mnist}
Li~Deng.
\newblock The mnist database of handwritten digit images for machine learning
  research.
\newblock \emph{IEEE Signal Processing Magazine}, 29\penalty0 (6):\penalty0
  141--142, 2012.

\bibitem[Krizhevsky and Hinton(2009)]{cifar10}
Alex Krizhevsky and Geoffrey Hinton.
\newblock Learning multiple layers of features from tiny images.
\newblock \emph{Technical Report}, 2009.
\newblock URL \url{https://www.cs.toronto.edu/~kriz/cifar.html}.
\newblock CIFAR-10 dataset.

\bibitem[Deng et~al.(2009)Deng, Dong, Socher, Li, Li, and
  Fei-Fei]{deng2009imagenet}
Jia Deng, Wei Dong, Richard Socher, Li-Jia Li, Kai Li, and Li~Fei-Fei.
\newblock Imagenet: A large-scale hierarchical image database.
\newblock In \emph{2009 IEEE conference on computer vision and pattern
  recognition}, pages 248--255. Ieee, 2009.

\bibitem[Chen et~al.(2020)Chen, Kornblith, Norouzi, and Hinton]{simclr}
Ting Chen, Simon Kornblith, Mohammad Norouzi, and Geoffrey Hinton.
\newblock A simple framework for contrastive learning of visual
  representations.
\newblock In \emph{International conference on machine learning}, pages
  1597--1607. PMLR, 2020.

\bibitem[Vershynin(2010)]{Vershynin-2010}
Roman Vershynin.
\newblock Introduction to the non-asymptotic analysis of random matrices.
\newblock \emph{arXiv preprint arXiv:1011.3027}, 2010.

\bibitem[Vershynin(2018)]{Vershynin-2018}
Roman Vershynin.
\newblock \emph{High-dimensional probability: An introduction with applications
  in data science}, volume~47.
\newblock Cambridge university press, 2018.

\bibitem[Ostrovskii and Bach(2018)]{Bach-2018}
Dmitrii Ostrovskii and Francis Bach.
\newblock Finite-sample analysis of m-estimators using self-concordance, 2018.
\newblock URL \url{https://arxiv.org/abs/1810.06838}.

\bibitem[Hsu et~al.(2012)Hsu, Kakade, and Zhang]{Hsu-2012}
Daniel Hsu, Sham Kakade, and Tong Zhang.
\newblock {A tail inequality for quadratic forms of subgaussian random
  vectors}.
\newblock \emph{Electronic Communications in Probability}, 17\penalty0
  (none):\penalty0 1 -- 6, 2012.
\newblock \doi{10.1214/ECP.v17-2079}.
\newblock URL \url{https://doi.org/10.1214/ECP.v17-2079}.

\bibitem[Kohler and Lucchi(2017)]{Kohler_2017}
Jonas~Moritz Kohler and Aurelien Lucchi.
\newblock Sub-sampled cubic regularization for non-convex optimization, 2017.
\newblock URL \url{https://arxiv.org/abs/1705.05933}.

\bibitem[Bach(2010)]{Bach-2010}
Francis Bach.
\newblock {Self-concordant analysis for logistic regression}.
\newblock \emph{Electronic Journal of Statistics}, 4\penalty0 (none):\penalty0
  384 -- 414, 2010.
\newblock \doi{10.1214/09-EJS521}.
\newblock URL \url{https://doi.org/10.1214/09-EJS521}.

\bibitem[Frostig et~al.(2015)Frostig, Ge, Kakade, and
  Sidford]{frostig2015competing}
Roy Frostig, Rong Ge, Sham~M. Kakade, and Aaron Sidford.
\newblock Competing with the empirical risk minimizer in a single pass, 2015.

\bibitem[Beck and Teboulle(2003)]{beck-2003}
Amir Beck and Marc Teboulle.
\newblock Mirror descent and nonlinear projected subgradient methods for convex
  optimization.
\newblock \emph{Operations Research Letters}, 31\penalty0 (3):\penalty0
  167--175, 2003.

\bibitem[Ruhe(1970)]{Ruhe-1970}
Axel Ruhe.
\newblock Perturbation bounds for means of eigenvalues and invariant subspaces.
\newblock \emph{BIT Numerical Mathematics}, 10\penalty0 (3):\penalty0 343--354,
  1970.

\end{thebibliography}

\newpage
\appendix

\section*{Appendix}
The appendix is organized as follows. In Appendix~\ref{appendix:A}, we provide an introduction to some fundamental probability tools that are utilized in our proofs.  Specifically, we discuss sub-Gaussian and sub-exponential distributions in Appendix~\ref{appendix:sub-Gaussian}, and present Bernstein-type inequalities in Appendix~\ref{appendix:bernstein}. In Appendix~\ref{appendix:B}, we summarize the properties of the multi-class logistic regression model that are needed in our proofs. Specifically, in Appendix~\ref{appendix:log-GLM}, we present the generalized linear model formulation of the multi-class logistic model and in Appendix~\ref{appendix:grad-hessian-loss}, we discuss the gradient and Hessian of the loss function. In Appendix~\ref{appendix:concordance}, we introduce pseudo self-concordant functions.  In Appendix~\ref{appendix:sub-thm}, we present a thorough proof of one of our fundamental results, specifically Theorem~\ref{thm:sub-thm}. In Appendix~\ref{appendix:parameter}, we delve into the properties of some essential constants utilized in constructing the results of Theorem~\ref{thm:sub-thm}. In Appendix~\ref{appendix:bounded-domain}, we provide the excess risk bounds for the case of $p(x)$ having bounded support. The proofs of the main results of Section~\ref{sec:method} are provided in Appendix~\ref{appendix:method}. Finally, in Appendix~\ref{appendix:experiments}, we provide more details of our numerical experiments.

\section{Probability tools}\label{appendix:A}

\subsection{Sub-Gaussian and sub-exponential distributions}\label{appendix:sub-Gaussian}

\begin{definition} [Sub-Gaussian random variable]\label{def:sub-variable}
  A random variable $x$  is sub-Gaussian if there exists $c_1>0$ such that
  $\prob(|x|>t) \leq \exp(1-t^2/c_1^2)$ for all $t \geq 0$.
\end{definition}

\begin{Lemma}[ Proposition~\cite{Vershynin-2010} in \cite{Vershynin-2018}]\label{lm:sub-properties}
 Let $x$ be a sub-Gaussian random variable.   Then the following properties are equivalent, with parameters $c_i>0$:
\begin{enumerate}[label={(\arabic*)},leftmargin=*]
    \item \label{sub-Gaussian-p1} $\prob(|x|>t) \leq \exp(1-t^2/c_1^2),$ for all  $t \geq 0$.
    \item \label{sub-Gaussian-p2}$(\E|x|^p)^{1/p} \leq c_2 \sqrt{p}, $ for all $p\geq 1$.
    \item \label{sub-Gaussian-p3}$\E \exp(x^2/c_3^2) \leq 2$.
\end{enumerate}
\end{Lemma}

\begin{definition} [Sub-Gaussian norm]\label{def:sub-gaussnorm}
    Let $x$ a sub-Gaussian random variable. The sub-Gaussian norm of $x$, denoted $\| x\|_{\psi_2}$, is defined as follows:
\[
    \| x\|_{\psi_2} \triangleq \inf \{t>0 : \E \exp(x^2/t^2)\leq 2 \}.
\]
\end{definition}

\begin{Lemma}[Sub-exponential random variable]\label{lm:sub-exp-properties}
  Let $x$ be a random variable. We say that $x$ is sub-exponential if there exists $c_i>0$ for which one of  following properties is true. Furthermore, these properties are equivalent. 
\begin{enumerate}[label={(\arabic*)},leftmargin=*]
    \item \label{sub-exp-p1}$\prob(|x|>t) \leq \exp(1-t/c_1)$ for all $t \geq 0$.
    \item \label{sub-exp-p2}$(\E|x|^p)^{1/p} \leq c_2 p$ for all $p\geq 1$.
    \item \label{sub-exp-p3}$\E \exp(|x|/c_3) \leq 2$.
\end{enumerate}
\end{Lemma}

\begin{definition} [Sub-exponential norm]\label{def:sub-exp-variable}
The sub-exponential norm of $x$, denoted $\| x\|_{\psi_1}$, is defined as follows:
\[
    \| x\|_{\psi_1} \triangleq \inf \{t>0 : \E \exp(|x|/t)\leq 2 \}.
\]
\end{definition}

\begin{Lemma}[Sub-exponential is sub-Gaussian squared, Lemma 2.7.6 in~\cite{Vershynin-2018}]\label{lm:exp-gaussian-square}
    A random variable $x$ is sub-Gaussian if and only if $x^2$ is sub-exponential. Moreover,
    \[
    \| x^2\|_{\psi_1} = \|x \|_{\psi_2}^2.
    \]
\end{Lemma}

\begin{definition} [Sub-Gaussian random vectors]\label{def:sub-vector}
  A random vector $Z\in \mathbb{R}^d$ is sub-Gaussian if $\langle Z ,u \rangle$ is sub-Gaussian for all $u \in \mathbb{R}^d$, with $\|u\|_2=1$. The sub-Gaussian norm of $Z$ is defined as
\[
    \|Z \|_{\psi_2} \triangleq  \sup_{u \in \mathcal{S}^{d-1}} \|\langle Z,u  \rangle\|_{\psi_2}.
\]
\end{definition}

\begin{Lemma}\label{lm:sub-vector-hoeffding}
    Let $Z_1,\cdots, Z_n$ be independent centered sub-Gaussian random vectors, then $\|\sum_{i=1}^n Z_i \|_{\psi_2}^2 \lesssim \sum_{i=1}^n \| Z_i\|_{\psi_2}^2$.
\end{Lemma}

\begin{Lemma}[Affine transformation of sub-Gaussian vectors, Lemma A.5 in \cite{Bach-2018}]\label{lm:sub-vector-affine}
    Let $X \in \mathbb{R}^d$ such that $\E[X] = 0$, $\bSigma: = \E[XX^\top]$ and $\|\bSigma^{-1/2} X \|_{\psi_2}\leq K$. Then for any $\A \in\mathbb{R}^{d\times d}$ and $b \in\mathbb{R}^d$, $\widehat{X} = \A X + b$ satisfies
    \[
        \| \widehat{\bSigma}^{-1/2} \widehat{X} \|_{\psi_2}\lesssim K,\quad \mbox{where} \quad \widehat{\bSigma} = \E[\widehat{X} \widehat{X}^\top].
    \]
\end{Lemma}

The following lemma gives a high probability bound for the quadratic form $\|x\|_{\bSigma^{-1}}^2$ of a non-centered sub-Gaussian vector $x$, where $\bSigma$ is the covariance of $x$. The result can be viewed as a corollary of Theorem~2.1 in \cite{Hsu-2012}.

\begin{Lemma}[Tail inequalities for quadratic form of sub-Gaussian vectors]\label{thm:sub-quadratic}
Let $\b J\in \mathbb{R}^{d\times d}$ be a symmetric, positive semi-definite  matrix.  For any $\delta \in (0,1)$ the following is true:
\begin{enumerate}[label={(\arabic*)},leftmargin=*]
    \item If  $x \in \mathbb{R}^d$ is a zero-centered sub-Gaussian random vector, i.e. $\E[x]=0$ and there exits $K>0$ such that $ \|x\|_{\psi_2} \leq K$.  Then we have with probability at least $1-\delta$,
    \begin{align}\label{eq:quadratic-center}
        \| x\|_{\b J}^2 \lesssim K^2 \big(\Tr(\b J) + \sqrt{d} \| \b J\| \log (e/\delta)\big).
    \end{align}
    \item If $x \in \mathbb{R}^d$ is a sub-Gaussian random vector with $\| \bSigma^{-1/2} x\|_{\psi_2} \leq K$, where $\bSigma = \E [x x^T]$. Then with probability at least $1-\delta$, 
        \begin{align}\label{eq:quadratic-non-center}
            \|x \|_{\bSigma^{-1}}^2 \lesssim K^2 \big( d + \sqrt{d} \log(e/\delta)\big).
        \end{align}
\end{enumerate}
\begin{proof}
$ $
\begin{enumerate}[label={(\arabic*)},leftmargin=*]
\item By Theorem~2.1 in \cite{Hsu-2012}, we have for all $t>0$,
\begin{align}\label{eq:pf-quratic-1}
    \prob\Big[ \|x \|_{\b J}^2 > K^2\big( \Tr(\b J) +2 \sqrt{\Tr(\b J^2) t} + 2 \| \b J\|t\big)\Big] \leq \exp(-t).
\end{align}
Let $t = \log(1/\delta)$ in \Cref{eq:pf-quratic-1}, since $\sqrt{\Tr(\b J^2)} = \| \b J\|_F \leq \sqrt{d}\|\b J \|$, we can get \Cref{eq:quadratic-center}.
\item Note that we can not directly derive \Cref{eq:quadratic-non-center} from \Cref{eq:quadratic-center} since $x$ is not zero-mean. But we can shift $x$ to an isotropic sub-Gaussian random vector.  Indeed, let $\mu = \E[x]$ and $\bSigma_0 = \E[(x-\mu)(x-\mu)^\top]$. Then $\bSigma_0^{-1/2} (x-\mu)$ is centered isotropic random vector. By Lemma~\ref{lm:sub-vector-affine}, affine transformation of sub-Gaussian random vectors are also sub-Gaussian, i.e. $\bSigma_0^{-1/2} (x-\mu)$ is also sub-Gaussian and 
    \begin{align}
        \| \bSigma_0^{-1/2} (x-\mu)\|_{\psi_2}\lesssim K.
    \end{align}
    Denote $\b J = \bSigma_0^{1/2} \bSigma^{-1}\bSigma_0^{1/2}  $. By Sherman–Morrison formula, we have
    \begin{align}\label{eq:noncenter-quadratic-1}
        \bSigma^{-1} = (\bSigma_0 + \mu\mu^\top)^{-1} = \bSigma_0^{-1} - \frac{\bSigma_0^{-1} \mu \mu^\top \bSigma_0^{-1}}{1 + \mu^\top \bSigma_0^{-1} \mu},
    \end{align}
    and thus
    \begin{align}
        \|\b J\|_{\infty} &\leq 1, \\
        \|\b J\|_2 &= \bigg\| \b I_d - \frac{(\bSigma_0^{-1/2} \mu) (\bSigma_0^{-1/2} \mu)^\top}{1+ \|\bSigma_0^{-1/2} \mu\|_2^2} \bigg \|_2 \leq \|\b I_d \|_2 + \frac{\|\bSigma_0^{-1/2} \mu\|_2^2}{1 +\|\bSigma_0^{-1/2} \mu\|_2^2 } \leq 2,\\
        \Tr(\b J) &=\langle \bSigma^{-1}, \bSigma_0\rangle = \Tr(\b I_d) - \frac{\mu^\top \bSigma_0^{-1} \mu}{1 + \mu^\top \bSigma_0^{-1} \mu} \leq d.
    \end{align}
    By \Cref{eq:quadratic-center}, we have with probability at least $1-\delta$,
    \begin{align}\label{eq:noncenter-quadratic-2}
        \| x - \mu\|_{\bSigma^{-1}}^2 &= \|\bSigma_0^{-1/2} (x-\mu)\|_{\b J}^2 \lesssim \Tr(\b J) + K^2 (\|\b J\|_2\sqrt{\log(1/\delta)} + \|\b J\|_{\infty} \log(1/\delta))\nonumber \\
       & \lesssim  K^2 \Big( d + \sqrt{d}\log(e/\delta)\Big).
    \end{align}
    In addition, by \Cref{eq:noncenter-quadratic-1},
    \begin{align}\label{eq:noncenter-quadratic-3}
        \|\mu \|_{\bSigma^{-1}}^2 = \mu^\top \bSigma^{-1} \mu = \mu^\top \bSigma_0^{-1} \mu - \frac{(\mu^\top \bSigma_0^{-1} \mu )^2}{1 + \mu^\top \bSigma_0^{-1} \mu } = \frac{\mu^\top \bSigma_0^{-1} \mu }{1 + \mu^\top \bSigma_0^{-1} \mu } \leq 1.
    \end{align}
    Combining \Cref{eq:noncenter-quadratic-2,eq:noncenter-quadratic-3}, we obtain 
    \begin{align}
        \|x \|_{\bSigma^{-1}}^2 \leq (\| x-\mu\|_{\bSigma^{-1}} + \| \mu\|_{\bSigma^{-1}})^2 \lesssim  K^2 \Big( d + \sqrt{d}\log(e/\delta)\Big).
    \end{align}
\end{enumerate}
\end{proof}
\end{Lemma}

\subsection{Bernstein-type inequalities}\label{appendix:bernstein}
We give Bernstein-type inequalities for vectors and matrices in the following lemmas. These properties are used in the proof of excess risk bounds in the bounded domain case (\Cref{appendix:bounded-domain}).
\begin{Lemma}[Vector Bernstein inequality; see Theorem 18 in \cite{Kohler_2017}]\label{lm:vector-Bernstein}
Let $x_1, x_2,\cdots, x_n$ be independent random vectors such that
\[
\E[x_i] = 0, \quad \|x_i \|_2 \leq \mu \quad \text{and } \E [\|x_i \|_2^2]\leq \nu, \qquad \forall i \in [n].
\]
Let 
$S = \frac{1}{n}\sum_{i=1}^n x_i $.
Then if  $0 < \epsilon < \nu/\mu$,
\begin{align}
    \prob [\|S \|_2 \geq \epsilon] \leq \exp\big( - \frac{n \epsilon^2}{8\nu} + \frac{1}{4}\big).
\end{align}

\end{Lemma}

\begin{Lemma}[Matrix Bernstein inequality;  see Theorem 19 in \cite{Kohler_2017}]\label{lm:matrix-Bernstein}

Let $\X_1, \X_2,\cdots, \X_n$ be independent random Hermitian matrices with common dimension $d\times d$ such that 
\[
\E[\X_i] = 0, \quad \|\X_i \|_2\leq \mu \quad \text{and } \E [\|\X_i \|_2^2]\leq \nu, \qquad \forall i \in [n].
\]
Let 
${\b S} = \frac{1}{n}\sum_{i=1}^n \X_i$. Then if $0 < \epsilon < 2\nu/\mu$,
\begin{align}
    \prob [\|{\b S} \|_2 \geq \epsilon] \leq 2d \cdot \exp\big( -  \frac{n \epsilon^2}{4\nu}\big).
\end{align}
\end{Lemma}


\section{Multi-class logistic regression and pseudo self-concordance}\label{appendix:B}
In \Cref{appendix:grad-hessian-loss}, we present some properties of the gradient and Hessian of $\ell_{(x,y)}(\theta)$ with respect to $\theta$. In \Cref{appendix:log-GLM}, we show that the multi-class logistic regression model is a Generalized Linear Model. Then we present some properties related with the pseudo-concordance in \Cref{appendix:concordance}.

\paragraph{Notation.} Given $y\in[c]$ and $\eta\in\mathbb{R}^{c-1}$, we define the loss function $\ell(y, \eta)$  by
\begin{align}\label{eq:loss-2-arguments}
    \ell(y,\eta) \triangleq 
\begin{cases}
  -\log\big( \frac{\exp(\eta_y)}{1 + \sum_{l\in[c-1]} \exp(\eta_l)}\big) ,\qquad y \in [c-1]\\
   -\log\big( \frac{1}{1 + \sum_{l\in[c-1]} \exp(\eta_l)}\big),\qquad y = c.
    \end{cases}
\end{align}
where $\eta_y$ is the $y$-th component of $\eta$.  Note that given $x\in\mathbb{R}^d$, $y\in[c]$ and $\theta\in\mathbb{R}^{(c-1)\times d}$, if we let $\eta = \theta x$, then
\[
\ell(y, \eta) =\ell_{(x,y)}(\theta) ,
\]
where $\ell_{(x,y)}\triangleq -\log p(y|x,\theta)$ (\Cref{eq:setup-conditional}). 

To differentiate the derivatives with respect to $\eta$ and $\theta$, we use $\ell^{\prime}(y,\eta)$ to represent the gradient of the loss with respect  to $\eta$, and $\nabla \ell_{(x,y)}(\theta)$ to represent the gradient of the loss with respect to $\theta$.   Similar notations hold for higher order derivatives.

\subsection{Properties of multi-class logistic regression}\label{appendix:grad-hessian-loss}
We present the expressions of  gradient and Hessian of the loss function $\ell_{(x,y)}(\theta)$ with respect to $\theta$ in the following proposition.

\begin{proposition}\label{prop:grad-and-hessian}
    Given a sample point $x\in\mathbb{R}^d$, its label $y\in[c]$, and parameter $\theta\in\mathbb{R}^{(c-1) \times d}$ in the multiclass  logistic regression model. We consider the negative log-likelihood loss $\ell_{(x,y)}(\theta) = -\log p(y|x,\theta)$, where $p(y|x,\theta)$ is defined in Eq.~\eqref{eq:setup-conditional}. Let $\wc \triangleq c-1$, $\td \triangleq d(c-1)$, $\theta_i$ be the $i$-th row of $\theta$. Define vector $\hx(x,\theta) \in \mathbb{R}^{\wc}$ by 
    \begin{align}
         \hx_i(x, \theta) = p(y=i|x,\theta) =\frac{\exp(x^\top \theta_i)}{1 + \sum_{s\in[\wc]} \exp(x^\top \theta_s)}, \qquad \forall i \in [\wc].\nonumber
    \end{align}
    Then the gradient and Hessian of $\ell_{(x,y)}(\theta)$ w.r.t $\theta$ can be expressed in the following ways:
\begin{enumerate}[label={(\arabic*)},leftmargin=*]
\item Gradient $\nabla \ell_{x,y}(\theta) \in \mathbb{R}^{\wc \times d}$ is given by
    \begin{align}\label{eq:loss-grad}
        \nabla\ell_{(x,y)} (\theta) = \begin{bmatrix}
    \beta_1(y,x,\theta) x^\top \\
    \cdots \\
    \beta_{\wc}(y,x,\theta) x^\top
    \end{bmatrix},
    \end{align}
where $ \beta_i(x,y, \theta) = -1_{\{ y=i\}} + \hx_i(x,\theta)$.
\item  Hessian $\nabla^2 \ell_{(x,y)}(\theta) \in \mathbb{R}^{\td \times \td}$ is given by
\begin{align}\label{eq:loss-hessian}
    \nabla^2 \ell_{(x,y)}(\theta) &= \Big(\mathrm{diag}(\hx(x, \theta)) - \hx(x, \theta) \hx(x, \theta)^\top \Big)\otimes (x x^\top)\nonumber\\
    &= \begin{bmatrix}
\alpha_{11}(x,\theta) x x^\top  & \cdots & \alpha_{1\wc}(x,\theta) x x^\top\\
\vdots & \ddots & \vdots\\
\alpha_{\wc 1}(x,\theta) x x^\top  & \cdots & \alpha_{\wc \wc}(x,\theta) x x^\top\\
    \end{bmatrix},
\end{align}
where $\alpha_{i,j}(\theta) = 1_{\{i=j\}}\hx_i(x,\theta) -\hx_i(x,\theta)\hx_j(x,\theta)$.
\end{enumerate}
\end{proposition}

\begin{Lemma}\label{lm:zero-mean-gradient-loss}
    Given a point $x \in \mathbb{R}^d$, $\E_{y\sim p(y|x,\theta_*)} [\nabla \ell_{(x,y)} (\theta_*)] = 0$. In addition, let $p(x)$ be a point distribution and $L_p(\theta)$ be the expected loss at $\theta$,  then 
    \begin{align}\label{eq:grad-expect-loss-0}
        \nabla L_p(\theta_*)  = 0.
    \end{align}
\end{Lemma}
\begin{proof}
Since $\nabla\ell_{(x,y)}(\theta) = -\nabla_{\theta} \log p(y|x,\theta)$, we have
\begin{align}
    \E_{y\sim p(y|x,\theta_*)} [\nabla \ell_{(x,y)} (\theta_*)]&=- \sum_{k\in[c]} p(y=k|x,\theta_*) \nabla_{\theta} \log p(y=k|x,\theta_*)\nonumber\\
    &= - \sum_{k\in[c]} p(y=k|x,\theta_*) \frac{\nabla_{\theta} p(y=k|x,\theta_*)}{p(y=k|x,\theta_*)}\nonumber\\
    & = - \nabla_{\theta} \Big(\sum_{k\in[c]} p(y=k|x,\theta_*)\Big) = -\nabla_{\theta} 1 = 0.
\end{align}
Thus,
\begin{align}\label{eq:zero-grad-1}
    \nabla_{\theta}\big( \E_{y\sim p(y|x,\theta_*)} [\ell_{(x,y)}(\theta)]\big) \big |_{\theta = \theta_*} = \E_{y\sim p(y|x,\theta_*)} [\nabla \ell_{(x,y)} (\theta_*)] = 0.
\end{align}
Since $\nabla L_p(\theta) = \nabla_\theta \int p(x) \E_{y\sim p(y|x,\theta_*)}[\ell_(x,y)(\theta)] dx = \int p(x)  \nabla_\theta \E_{y\sim p(y|x,\theta_*)}[\ell_(x,y)(\theta)] dx$, by Eq.~\eqref{eq:zero-grad-1}, we have
\begin{align}
    \nabla L_p(\theta_*) = \int p(x) \nabla_{\theta}\big( \E_{y\sim p(y|x,\theta_*)} [\ell_{(x,y)}(\theta)]\big) \big |_{\theta = \theta_*} dx = 0.
\end{align}

\end{proof}

The following lemma is a basic property for Fisher information matrix.
\begin{Lemma} \label{lm:fisher-property}
   The Fisher information matrix for a point $x$ at parameter $\theta$ is defined by $ \E_{y\sim p(y|x, \theta_*)}[\nabla\ell_{(x,y)}(\theta)(\nabla\ell_{(x,y)}(\theta))^\top ]$, then 
\begin{align}\label{eq:fisher-property}
   \E_{y\sim p(y|x, \theta_*)}[\nabla\ell_{(x,y)}(\theta_*)(\nabla\ell_{(x,y)}(\theta_*))^\top ]=  \E_{y\sim p(y|x,\theta_*)}[\nabla^2 \ell_{(x,y)}(\theta_*)].
\end{align}
\end{Lemma}
\begin{proof}
 \begin{align}
     \nabla^2 \ell_{(x,y)}(\theta_*) &= -\frac{\nabla^2 p(y|x,\theta_*))}{p(y|x,\theta_*)} + \frac{\nabla p(y|x,\theta_*) \nabla p(y|x,\theta_*)^\top}{p(y|x,\theta_*)^2} \nonumber \\
     &= -\frac{\nabla^2 p(y|x,\theta_*))}{p(y|x,\theta_*)} +  \nabla \ell_{(x,y)}(\theta_*)(\nabla \ell_{(x,y)}(\theta_*))^\top\nonumber
 \end{align}
 Thus,
 \begin{align}
 &\E_{y\sim p(y|x, \theta_*)}[\nabla\ell_{(x,y)}(\theta_*)(\nabla\ell_{(x,y)}(\theta_*))^\top ] \nonumber\\
 =& \E_{y\sim p(y|x,\theta_*)}[\nabla^2 \ell_{(x,y)}(\theta_*)]  + \E_{y\sim p(y|x,\theta_*)}\bigg[\frac{\nabla^2 p(y|x,\theta_*))}{p(y|x,\theta_*)}\bigg] \nonumber\\
     =& \E_{y\sim p(y|x,\theta_*)}[\nabla^2 \ell_{(x,y)}(\theta_*)] + \int {p(y|x,\theta_*)}\frac{\nabla^2 p(y|x,\theta_*))}{p(y|x,\theta_*)}d\sigma\nonumber\\
     =& \E_{y\sim p(y|x,\theta_*)}[\nabla^2 \ell_{(x,y)}(\theta_*)] + \nabla^2 \int p(y|x,\theta_*)d\sigma = \E_{y\sim p(y|x,\theta_*)}[\nabla^2 \ell_{(x,y)}(\theta_*)].\nonumber
 \end{align}
 
\end{proof}

\subsection{Multi-class logistic regression as a Generalized Linear Model (GLM)} \label{appendix:log-GLM}

\begin{definition}[Exponential family model] \label{def:exp-family}
    Suppose $\mu$ is a base measure on space $\mathcal{Y}$ and there exists a sufficient statistic $T: \mathcal{Y}\rightarrow \mathbb{R}^c$. Then the exponential family associated with the function $T(y)$ and measure $\mu$ is defined as the set of distributions with densities $p(y|\eta)$ w.r.t $\mu$, where 
    \begin{align}\label{eq:exp-family}
        p(y|\eta) = \exp(\langle \eta, T(y) \rangle - A(\eta))
    \end{align}
    and $a(\eta)$ is the cumulant function defined by
    \begin{align}
        A(\eta) \triangleq\log \int_{\mathcal{Y}} \exp(\langle \eta, T(y) \rangle ) d\mu(y)
    \end{align}
    whenever $a$ is finite.
\end{definition}

\begin{definition}[Generalized linear model with canonical response function]\label{def:GLM}
    Generalized linear model with canonical response function is a model assuming that:
    \begin{enumerate}
        \item the input $x\in\mathbb{R}^d$ enter into the model via a linear combination $\eta = \theta x$,
        \item the output $y$ is characterized by an exponential family distribution (Definition \ref{def:exp-family}).
    \end{enumerate}
\end{definition}

In the following lemma, we remark that the multi-class logistic regression model is a generalized linear model. The proof is trivial.
\begin{Lemma}\label{lm:log-glm}
    Multi-class logistic regression is a generalized linear model with canonical response function with $\eta$, $ A(\eta)$ and $T(y)$ defined as the followings:
    \begin{align}
        \eta &= [\log(\hx_1/\hx_c), \log(\hx_2/\hx_c),\cdots, \log(\hx_{c-1}/\hx_c) ]^\top\\
        A(\eta) &= -\log \hx_c\\
        T(1) &= [1,0,\cdots,0]^\top, \quad \cdots, \quad T(c-1) = [0,\cdots, 1]^\top,\quad T(c) = [0,\cdots, 0]^\top,
    \end{align} 
    where $\hx_i = p(y=i |x,\theta)$ ($p(y|x,\theta)$ is defined in \Cref{eq:setup-conditional}).
\end{Lemma}

\subsection{Pseudo self-concordance}\label{appendix:concordance}

\begin{Lemma}[pseudo self-concordance of multi-class logistic regression model]\label{lm:log-concordance}
    $\ell(y,\eta)$ is pseudo self-concordant, i.e.
    \begin{align}
       \forall h \in \mathbb{R}^{c-1}, \qquad |\ell^{\prime \prime \prime}(y,\eta)[h,h,h] | \leq 2\|h\|_{\infty} \ell^{\prime \prime}(y,\eta)[h,h].
    \end{align}
\end{Lemma}
\begin{proof}
    By Lemma~\ref{lm:log-glm} and Equation~\eqref{eq:exp-family}, 
    \[ \ell(y,\eta) = -\log p(y, \eta)  = -\langle \eta, T(y) \rangle + A(\eta).
    \]
    From theory of the exponential family distributions, we have
    \begin{align}
        A^\prime(\eta) = \E_{\eta}[T(y)],\quad A^{\prime \prime}(\eta) = \E_{\eta}[(T(y) - \E_\eta[T(y)])^{\otimes 2}],\quad A^{\prime \prime\prime}(\eta) = \E_{\eta}[(T(y) - \E_\eta[T(y)])^{\otimes 3}].
    \end{align}
    where  we denote the $p$th order tensor for a vector $x$ as 
\[
    x^{\otimes p} = \underbrace{x \otimes x\otimes \cdots \otimes x}_{p \text{ times}}.
\]

    Note that $\ell^{(p)}(y,\eta) = A^{(p)}(\eta)$ whenever $p \geq 2$, then we have
    \begin{align}
        \big|\ell^{\prime \prime \prime}(y,\eta)[h,h,h] \big|&= \big| \E\big[(T(y) - \E_\eta[T(y)])^{\otimes 3}[h,h,h]\big]\big| \nonumber\\
        &= \big| \E\big[(T(y) - \E_\eta[T(y)])^{\otimes 2}[h,h] \big\langle T(y) - \E_\eta[T(y)],h  \big\rangle\big]\big|\nonumber\\
        &\leq \sup_{y\in\mathcal{Y}} \big | \big\langle T(y) - \E_\eta[T(y)],h  \big\rangle\big| \ell^{\prime \prime}(y,\eta)[h,h]\nonumber\\
        &\stackrel{(a)}{\leq} 2 \sup_{y\in\mathcal{Y}} \| T(y) \|_1 \|h\|_{\infty} \ell^{\prime \prime}(y,\eta)[h,h]\nonumber\\
        &\stackrel{(b)}{\leq} 2 \|h\|_{\infty} \ell^{\prime \prime}(y,\eta)[h,h], 
    \end{align}
    where (a) follows by Cauchy-Schwarz inequality, triangle inequality, and $\|E_{\eta}[T(y)]\|_2 \leq E_\eta\|T(y)\|_2 \leq \sup_{y\in\mathcal{Y}} \|T(y)\|_2$, (b) follows by the fact that $\| T(y)\|_2 = 1$ for $y\neq c$ and $\| T(y)\|_2 = 0$ for $y=c$ (Lemma~\ref{lm:log-glm}).
\end{proof}

The previous lemma states the pseudo self-concordance of $\ell(y,\eta)$ w.r.t $\eta$. The following proposition states that the empirical loss function is pseudo self-concordant w.r.t $\theta$, which is a corollary of the previous lemma via chain rule.
\begin{proposition}\label{prop:concordance-param}
For multi-class regression model, we fix $\theta_0, \theta_1 \in \mathbb{R}^{(c-1) \times d}$. Let $\theta_t = \theta_0 + t (\theta_1 - \theta_0)$, we define $\phi_n(t)$ by
\begin{align}
\phi_n(t) = = \frac{1}{n}\sum_{i=1}^n \ell_{(x_i,y_i)}(\theta_t).
\end{align}
Then we have
\begin{align}
    |\phi_n^{\prime\prime\prime}(t)| \leq 2 \phi_n^{\prime \prime} (t) \max_{i\in[n]}\|(\theta_1 - \theta_0 )x_i\|_\infty \label{eq:concordance-Pn}
\end{align} 
\begin{proof}
    Denote $\Delta = \theta_1 - \theta_0$, then $\theta_t = \theta_0 = t \Delta$. Following chain rule and the smoothness of $\ell$, we obtain that the derivatives of $\phi(t)$ and $\phi_n(t)$ are given by
    \begin{align}
        \phi_n^{(p)}(t) = \frac{1}{n}\sum_{i=1}^n \ell^{(p)}(y,\theta_t x)[\underbrace{\Delta x, \cdots, \Delta x}_{p \text{ times}}].\nonumber
    \end{align}
    Applying Lemma~\ref{lm:log-concordance}, we can get
    \begin{align}
        |\phi^{\prime \prime \prime}(t) | &\leq \frac{1}{n} \sum_{i=1}^n\big|\ell^{\prime \prime \prime}(y_i,\theta_t x_i)[\Delta x_i, \Delta x_i, \Delta x_i]\big|\nonumber\\
        &\leq \frac{1}{n} \sum_{i=1}^n 2 \| \Delta x\|_\infty \ell^{\prime \prime}(y_i,\theta_t x_i)[ \Delta x_i, \Delta x_i] \nonumber\\
        &\leq 2 \phi_n^{\prime \prime}(t) \max_{i\in[n]}\|(\theta_1 - \theta_0 )x_i\|_\infty.\nonumber
    \end{align}
\end{proof}
\end{proposition}

The following proposition forms the foundation of our proof of~\Cref{thm:sub-thm}. It gives lower and upper bounds to perturbations of pseudo self-concordant function. 
\begin{proposition}[Proposition 1 in \cite{Bach-2010}]\label{prop:taylor-concordance}
    Let $F: \Theta \rightarrow \mathbb{R}$ be a convex $C^3$-mapping. Fix $\theta_0, \theta_1\in\Theta$, let $\Delta = \theta_1 - \theta_0$  and $\theta_t = \theta_0 + t\Delta$ for $t\in\mathbb{R}$. Define function $\phi_F(t) = F(\theta_t)$. Assume that $\b H_0\triangleq \nabla^2 F(\theta_0) \succ 0$ , $|\phi_F^{\prime \prime \prime}(t)| \leq R \|\Delta\|_2 \cdot \phi_F^{\prime\prime}(t)$ for some $R \geq 0$. Denote $S = R\|\Delta\|_2$, we have
    \begin{align}
         \frac{e^{-S} +S -1}{S^2}\|\Delta\|^2_{\b H_0} \leq F(\theta_1) - F(\theta_0) - \big(\nabla F(\theta_0)\big)^\top \Delta &\leq \frac{e^S - S -1}{S^2}\|\Delta\|^2_{\b H_0}, \label{eq:taylor-expansian}\\
        e^{-S}\b H_0 \preceq \nabla^2 F(\theta_1) \preceq e^{S} \b H_0. \label{eq:taylor-Hessian}
    \end{align}
\end{proposition}

\section{Proof of Theorem~\ref{thm:sub-thm}}\label{appendix:sub-thm}

We first give the detailed version of \Cref{thm:sub-thm} in \Cref{section:sub-thm-detail}. In \Cref{section:sub-thm-sketch-proof}, we present a sketch of the proof for the excess risk bounds in \Cref{eq:sub-thm-risk}. In  \Cref{section:sub-thm-tool}, we provide and prove a tail bound for a certain type of random matrices, which is useful in our full proof. Finally, we give the full proof of \Cref{thm:sub-thm} (\Cref{thm:sub-thm-detail}) in \Cref{section:sub-thm-full-proof}.

\paragraph{Notation.}
For the ease of notation, we define the empirical risk over finite samples  $Q_n(\theta)$ and its Hessian $\b H_n(\theta)$ by 
\begin{align}
    \theta_n \in \argmin_{\theta} & Q_n(\theta) \triangleq \frac{1}{n} \sum_{i\in [n]} \ell_{(x_i, y_i)} (\theta), \qquad (x_i,y_i)\stackrel{\mathrm{i.i.d.}}{\sim} \pi_q(x,y),\label{eq:qn}
\\
    &\b H_n(\theta) \triangleq \nabla^2 Q_n(\theta). \label{eq:hn}
\end{align}
In addition, let $\vec{\A} \in\mathbb{R}^{mn}$ be the vectorization of a matrix $\A\in\mathbb{R}^{m\times n}$ by stacking all rows together, i.e. $\vec{\A} = (\A_1^\top, \cdots, \A_m^\top)^\top$ where $\A_i$ is $i$-th row of $\A$.

\subsection{Detailed version of Theorem~\ref{thm:sub-thm}}\label{section:sub-thm-detail}

\begin{theorem}\label{thm:sub-thm-detail}
    Suppose Assumption \ref{assume:sub-gaussian} holds for both $p(x)$ and $q(x)$.Let $\sigma$,  $\rho$ and $\nu >0$ be constants such that $\hp(\theta_*) \preceq \sigma \hq$, $\b I_{c-1}\otimes \V_p \preceq \rho \hp(\theta_*)$ and $\V_q \preceq \nu \V_p$ hold. Whenever
\begin{align}\label{eq:sub-thm-nbound-detail}
    n \gtrsim \max \left\{K_{2,q}^2(r)\td \log(ed/\delta),\  \sigma\rho\nu K_{0,q}^2 K_{1,q}^2 K_{2,q}^2(r)\Big(\td+ \sqrt{\td} \log(e/\delta)\Big) \right\},
\end{align}
where $\td\triangleq d(c-1)$,  we have with probability at least $1-\delta$,
\begin{align}
    L_q(\emptheta) - L_q(\theta_*) &\lesssim K_{1,q}^2\frac{\td + \sqrt{\td}\log(e/\delta)}{n} ,\label{eq:sub-thm-erm-detail}\\
    \frac{e^{-\alpha} + \alpha -1}{\alpha^2}\, \frac{\hq^{-1} \cdot \hp}{n} &\lesssim   \E [L_p(\emptheta)] - L_p \lesssim \frac{e^{\alpha} - \alpha -1}{ \alpha^2} \, \frac{\hq^{-1}\cdot \hp}{n}. \label{eq:sub-thm-risk-detail}
\end{align}
Here $\hp=\hp(\theta_*)$ and $\hq=\hq(\theta_*)$; and $\E$ is the expectation over ${\{y_i\sim p(y_i|x_i, \theta_*)\}_{i=1}^n}$.  Furthermore,
\begin{align}\label{eq:sub-thm-alpha-detail}
   \qquad  \alpha = \mathcal{O}\Big(\sqrt{\sigma\rho}  K_{0,p} K_{1,q} K_{2,p}(r)\sqrt{\big( \td+ \sqrt{\td} \log(e/\delta)\big) /n}\Big).
\end{align}
\end{theorem}

\subsection{Proof sketch of Eq.(\ref{eq:sub-thm-risk})}\label{section:sub-thm-sketch-proof}
Here we present the basics of step 6 in  the full proof of \Cref{thm:sub-thm} (see \Cref{section:sub-thm-full-proof}). Some details 
 of this step are established in the steps 1-5 of the full proof.

Let $\theta_0 = \theta_*$, $\theta_1 = \theta_n$ and $\Delta\triangleq \theta_n - \theta_*$. Define $\phi_p(t) = L_p(\theta_* + t\Delta)$, we first prove that there exits  $\alpha>0$ s.t. $|\phi_p^{\prime \prime \prime}(t)| \leq \alpha \phi_p^{\prime \prime}(t)$. Thus the premise of \Cref{prop:taylor-concordance} is satisfied. By \Cref{eq:taylor-expansian} and the fact that $\nabla L_p(\theta_*) = 0$ (\Cref{lm:zero-mean-gradient-loss}),  we have
\begin{align}\label{eq:pf-sub-sketch-1}
    \frac{e^{-\alpha} +\alpha -1}{\alpha^2}\|\vecDelta\|^2_{\b H_p} \leq L_p(\theta_n) - L_p(\theta_0)  &\leq \frac{e^\alpha - \alpha -1}{\alpha^2}\|\vecDelta\|^2_{\b H_p}
\end{align}
By Taylor theorem, there exists $\Tilde{\theta}$ between $\theta_n$ and $\theta_*$ such that 
\begin{align}\label{eq:pf-sub-sketch-2}
    \vecnabla Q_n(\theta_*) = \vecnabla Q_n(\theta_n) + \b H_n(\Tilde{\theta}) \vecDelta =\b H_n(\Tilde{\theta}) \vecDelta  ,
\end{align}
where the last equality follows by $\vecnabla Q_n(\theta_n) = 0$ because the empirical loss $Q_n$ is convex and $\theta_n$ is its solution. We can prove that if the sample bound \Cref{eq:sub-thm-nbound} holds, 
\begin{align}\label{eq:pf-sub-sketch-3}
    \b H_n(\Tilde{\theta}) \approx \hq,
\end{align}
where  ``$\approx$'' means that there exits $a_1, a_2 >0$ such that $a_1 \hq \preceq \b H_n(\Tilde{\theta}) \preceq  a_2 \hq$. Thus we have
\begin{align}\label{eq:pf-sub-sketch-4}
    \|\vecDelta\|^2_{\hp} = \vecDelta^\top \hp \vecDelta &\approx \vecnabla Q_n(\theta_*)^\top \Big(\hq^{-1}\hp \hq^{-1}  \Big) \vecnabla Q_n(\theta_*)\nonumber\\
    &= \big \langle \hq^{-1} \hp \hq^{-1}, \vecnabla Q_n(\theta_*)\vecnabla Q_n(\theta_*)^\top \big\rangle.
\end{align}
Then we  prove that 
\begin{align}\label{eq:pf-sub-sketch-45}
   \E_{\{y_i\sim p(y_i|x_i,\theta_* )\}_{i=1}^n}\big[\vecnabla Q_n(\theta_*)\vecnabla Q_n(\theta_*)^\top  \big] = \frac{1}{n} \b H_n(\theta_*) \approx\frac{1}{n} \hq.
\end{align}
Substitute this into \Cref{eq:pf-sub-sketch-4}, we have
\begin{align}\label{eq:pf-sub-sketch-5}
      \E_{\{y_i\sim p(y_i|x_i,\theta_* )\}_{i=1}^n}[\|\vecDelta\|^2_{\hp}] \approx  \frac{1}{n} \langle \hq^{-1}, \hp\rangle.
\end{align}
By taking expectation over \Cref{eq:pf-sub-sketch-1} and using \Cref{eq:pf-sub-sketch-5}, we can get \Cref{eq:sub-thm-risk}.

\subsection{Supporting tools}\label{section:sub-thm-tool}
In the following proposition, we present and prove a tail bound for the average sum of independent random matrices $\{\A_i\}_{i\in[n]}$ satisfying $\E[\A_i] = \b I$ and \Cref{eq:sub-matrix-0}.

\begin{proposition}\label{thm:sub-matrix-covariance}
    Let $\A_1, \cdots, \A_n$ be $\td\times \td$ be independent symmetric matrices such that $\E[\A_i] = \b I_{\td}$. There is constant $K>0$ such that for any $i\in[n]$,
    \begin{align}\label{eq:sub-matrix-0}
        \sup_{u\in\mathcal{S}^{\td-1}} \|u^\top\A_i u\|_{\psi_1}\leq K,
    \end{align}
    where $\mathcal{S}^{\td-1}$ is the unit sphere in $\mathbb{R}^{\td}$, $\|\cdot\|_{\psi_1}$ is the norm for sub-exponential random variable (\Cref{def:sub-exp-variable}). Define matrix $\b S_n = \frac{1}{n}\sum_{i=1}^n \A_i$. Then for every $t\geq 0$, with probability at least $1-2\exp(-c_K t^2)$ we have
    \begin{align}\label{eq:sub-matrix-1}
        \|\b S_n - \b I_{\td} \|\leq \max\{a,a^2 \}, \qquad \text{where } a = \frac{C_K\sqrt{\td} + t}{\sqrt{n}}.
    \end{align}
    Here $c_K, C_k $ are constants that depend on $K$.
\begin{proof}
    The proof follows a covering argument. We consider $1/4-$net $\mathcal{N}$ of the unit sphere $\mathcal{S}^{\td-1}$. By Lemma 5.2 in~\cite{Vershynin-2010}, $|\mathcal{N}|\leq 9^{\td}$.  Since $\b S_n$ is symmetric, we can use Lemma 5.4 in~\cite{Vershynin-2010} to bound matrix operator norm using points in $1/4-$net $\mathcal{N}$:
    \begin{align}
        \|\b S_n - \b I_{\td} \| \leq 2 \max_{x \in \mathcal{N}} \Big|\Big\langle\big(\b S_n - \b I_{\td}\big) x, x \Big\rangle\Big| = 2 \max_{x\in\mathcal{N}} \Big|  x^\top \b S_n x -1\Big|,
    \end{align}
    where the last equality follows by $\|x\|_2 = 1$ on $\mathcal{N}$. Thus it is sufficient to prove with the given probability, 
    \begin{align}\label{eq:sub-matrix-2}
       2 \max_{x\in\mathcal{N}} \Big|  x^\top \b S_n x -1\Big|\leq \max\{a, a^2\}\triangleq \epsilon.
    \end{align}
    Pick an arbitrary $x \in \mathcal{N}$, then
    \begin{align}
   n x^\top \b S_n x = \sum_{i=1}^n x^\top \A_i x \triangleq \sum_{i=1}^n Z_i^2,
     \end{align}
     where we define random variable $Z_i \triangleq x^\top \A_i x$. We have the following properties for  $Z_i$:
    \begin{align}
        \E [Z_i]& = \E[x^\top \A_i x] = \langle  x^\top, \E[\A_i] x\rangle = 1,\nonumber\\
        \| Z_i\|_{\psi_1} &=\| x^\top \A_i x\|_{\psi_1} \stackrel{(a)}{\leq} K ,\nonumber\\
        \| Z_i - 1\|_{\psi_1}& = \|Z_i - \E[Z_i] \|_{\psi_1}\stackrel{(b)}{\leq} 2 \| Z_i\|_{\psi_1} \leq 2 K,\nonumber
    \end{align}
    where inequality (a) follows by \Cref{eq:sub-matrix-0}, inequality (b) follows by Jensen's inequality. 
    
    Thus  $Z_1-1, Z_2-1, \cdots, Z_n -1$ are independent centered sub-exponential random variables. Using Corollary 5.17 in \cite{Vershynin-2010}, we can get
    \begin{align}
        \prob \big(\big|x^\top \b S_n x -1 \big| \geq \frac{\epsilon}{2}\big) &= \prob\big(\big|\frac{1}{n} \sum_{i=1}^n (Z_i -1) \big| \geq \frac{\epsilon}{2} \big)\leq 2 \exp[- \frac{c_1}{K^2} \min (\epsilon, \epsilon^2) n]\nonumber\\
        & \leq2 \exp[- \frac{c_1}{K^2} a^2 n] \leq 2 \exp[-\frac{c_1}{K^2}(C_K^2 \td + t^2)].
    \end{align}
    Take the union bound of all $x\in \mathcal{N}$, let 
    \begin{align}\label{eq:sub-matrix-3}
        c_K = \frac{c_1}{K^2}, \qquad C_K = K \sqrt{\log 9/c_1},
    \end{align}
    we have
    \begin{align}
        \prob \bigg(\max_{x\in\mathcal{N}} \big| x^\top \b S_n x -1 \big| \geq \frac{\epsilon}{2} \bigg) &\leq 9^n \cdot 2 \exp[-\frac{c_1}{K^2}(C_K^2 \td+ t^2)]\nonumber\\
        &\leq 2 \exp\big[ p\log 9 - d_1\log9 -\frac{c_1 t^2}{K^2}\big] \nonumber \\
        &=2 \exp(-\frac{c_1 t^2}{K^2}) = 2\exp(-c_K t^2). 
    \end{align}
    As we noted in \Cref{eq:sub-matrix-2}, this completes the proof.
\end{proof}
\end{proposition}

\begin{corollary} \label{cor:sub-matrix-covariance}
Under the premise of \Cref{thm:sub-matrix-covariance}, whenever 
    \begin{align}
        n\gtrsim K^2 (\td + \log(1/\delta)),
    \end{align}
with probability at least $1-\delta$, 
    \begin{align}
        1/2 \b I_{\td} \preceq \b S_n \preceq 3/2 \b I_{\td}.
    \end{align}
\end{corollary}
\begin{proof}
    Let $t =2 K \sqrt{\log(1/\delta)/c_1}$, by \Cref{eq:sub-matrix-3} we have
    \begin{align}
       2 \exp(-c_K t^2) \leq 2 \exp\big(-\frac{c_1}{K^4} \frac{K^2 \log(1/2\delta)}{c_1}\big) = \delta.
    \end{align}
    Let $n = \frac{32}{c_1} K^2 (\td + \log(1/\delta)$, then
    \begin{align}
        a = \frac{C_K \sqrt{\td} + t}{\sqrt{n}} = \frac{\frac{2}{\sqrt{c_1}} K^2 (\sqrt{\td} + \sqrt{\log(1/\delta)})}{\frac{4\sqrt{2}}{\sqrt{c_1}} K^2 \sqrt{\td + \log(1/\delta)}} \leq \frac{1}{2} ,
    \end{align}
    and thus $\max\{ a, a^2\}\leq 1/2$.  Therefore, with probability at least $1-\delta$, we have
    \begin{align}
        \| \b S_n - \b I_{\td}\|\leq \frac{1}{2},
    \end{align}
    and thus $1/2\b I_{\td}\preceq \b S_n \preceq 3/2 \b I_{\td}$.
\end{proof}


\subsection{Proof of Theorem~\ref{thm:sub-thm} (Theorem~\ref{thm:sub-thm-detail})}\label{section:sub-thm-full-proof}
We present the full proof of \Cref{thm:sub-thm} as the following. Some of the techniques used in the proof are inspired by \cite{Bach-2018}.
\begin{proof}
By the definitions of $\sigma$, $\rho$ and $\nu$ in \Cref{thm:sub-thm}, we have the following basic inequalities. Given vectors $v \in \mathbb{R}^d$ and $u \in \mathbb{R}^{\td}$, we have the following norm relations:
\begin{align}
   & \|v\|_{\V_q} \leq \sqrt{\nu }\| v\|_{\V_p} , \qquad \|v\|_{\V_p^{-1}} \leq \sqrt{\nu }\| v\|_{\V_q^{-1}}, \label{eq:sub-pf-nu}\\
    &\| u\|_{\hp} \leq \sqrt{\sigma} \|u\|_{\hq} ,\label{eq:sub-pf-sigma}\\
   & \| u\|_{\wV_p} \leq \sqrt{\rho} \| u\|_{\hp} ,\label{eq:sub-pf-rho}
\end{align}
where $\wV_p \triangleq \b I_{c-1} \otimes \V_p$.

\textbf{step 1.}
Let $V_n = \sqrt{n }\hp^{-1/2} \vecnabla Q_n(\theta_*)$, then $V_n$ is a centered, isotropic sub-Gaussian random vector. Indeed, since $\nabla Q_n(\theta_*) = \frac{1}{n} \sum_{i\in [n]} \vecnabla \ell_{z_i}(\theta_*)$, we have
\begin{align}
    \underset{\{z_i \sim  \pi_q\}_{i=1}^n}{\E} [V_n] &= \frac{1}{\sqrt{n}} \hq^{-1/2} \sum_{i\in [n]} \E_{z_i \sim \pi_q}[\vecnabla \ell_{z_i}(\theta_*)] = 0\nonumber\\
     \underset{\{z_i \sim  \pi_q\}_{i=1}^n}{\E}  [V_n V_n^\top] &=  \hq^{-1/2}\bigg( \frac{1}{n}\sum_{i\in[n]} \E_{z_i \sim \mathcal{P}}[\vecnabla \ell_{z_i}(\theta_*) \vecnabla \ell_{z_i}(\theta_*)^\top] \bigg) \hq^{-1/2} \nonumber\\
    &= \hq^{-1/2} \hq  \hq^{-1/2}
    = \b I_{\td}.
\end{align}
By \Cref{lm:sub-vector-hoeffding}, 
\begin{align}
    \| V_n\|_{\psi_2}^2 \lesssim \sum_{i\in [n]} \| \frac{1}{\sqrt{n}} \hq^{-1/2} \vecnabla\ell_{z_i}(\theta_*)\|_{\psi_2}^2= K_{1,q}^2.
\end{align}
Now we apply the upper bound for quadratic form of sub-Gaussian random vector derived in Eq.~\eqref{eq:quadratic-center} from  \Cref{thm:sub-quadratic}, we can get
\begin{align}\label{eq:sub-thm-gradient}
    \| \vecnabla Q_n(\theta_*)\|_{\hq^{-1}}^2 = \frac{1}{n} \|V_n \|_2^2 \lesssim \frac{K_{1,q}^2 \Big(\td+ \sqrt{\td} \log(e/\delta)\Big)}{n}.
\end{align}

\textbf{step 2.} W.l.o.g we assume that Assumption~\ref{assume:sub-gaussian}-\ref{assume-sub-hessian} holds with $r =\mathcal{O}(1)$ and denote $\overline{K}_{2,q}\triangleq {K}_{2,q}(r)$ $\overline{K}_{2,p}\triangleq {K}_{2,p}(r)$ for ease of discussion. Now we show that the Hessian $\hq(\theta)$ is a good approximation to  $\hq$ for any $\theta \in \mathcal{B}_{q,\widehat{r}}(\theta_*) = \{ \theta: \| \theta- \theta_*\|_{\V_q, \infty} \leq \widehat{r}\}$, where $\widehat{r}=1/c$ for some constant $c$ depending on $K_{0,q}$ and $\overline{K}_{2,q}$. 

Fix $\theta_0 =\theta_*$ and pick arbitrary $\theta_1 \in \Theta$, let $\theta_t = \theta_0  + t \Delta$, where $\Delta\triangleq\theta_1 - \theta_0$. Define function
\begin{align}\label{eq:sub-pf-phiq}
    \phi_q(t) \triangleq L_q(\theta_t) = \E_{z\sim \pi_q} [\ell_z(\theta_t)]
\end{align}
Our goal is  to show that $\phi_q(t)$ is pseudo self-concordant, i.e. we intend to get some constant $C>0$ s.t. $|\phi_q^{\prime \prime \prime}(t)| \leq C  \phi_q^{\prime \prime}(t)$. First we observe that
\begin{align}\label{eq:sub-s2-1}
    \phi_q^{\prime \prime}(t) &= \E_{(x,y)\sim \pi_q}[\ell^{\prime \prime}(y, \theta_t x)[\Delta x, \Delta x]] = \E_{(x,y)\sim \pi_q}[\vecDelta^\top \big(\nabla^2 \ell_{(x,y)}(\theta_t x)\big) \vecDelta] \nonumber\\
    &= \vecDelta^\top \E_{(x,y)\sim \pi_q}[ \nabla^2\ell_{(x,y)}(\theta_t x)] \vecDelta = \| \vecDelta\|_{\hq(\theta_t)}^2.
\end{align}
Note that $\ell(y, \eta)$ is the loss function defined in \Cref{eq:loss-2-arguments} and $\ell^{\prime \prime}(y,\eta)$ is the Hessian w.r.t $\eta$.

On the other hand, by Lemma~\ref{lm:log-concordance} we have
\begin{align}\label{eq:sub-s2-2}
    |\phi_q^{\prime \prime \prime }(t)|&\leq \E_{(x,y)\sim \pi_q}\big[\big| \ell^{\prime \prime \prime }(y, \theta_t x)[\Delta x,\Delta x,\Delta x ] \big| \big]\nonumber\\
    &\leq 2\E_{(x,y)\sim \pi_q}\big[ \ell^{\prime \prime}(y, \theta_t x)[\Delta x, \Delta x]   \|\Delta x \|_\infty \big]\nonumber\\
    &\leq 2 \sqrt{\E_{(x,y)\sim\pi_q}\big[ \big(\ell^{\prime \prime}(y, \theta_t x)[\Delta x, \Delta x]  \big)^2\big]} \sqrt{\E_{(x,y)\sim\pi_q}\big[ \| \Delta x\|_\infty^2\big]},
\end{align}
where the last inequality follows by Cauchy-Schwartz inequality. 

Now we bound both of the square root terms in Eq.~\eqref{eq:sub-s2-2}. For the first square root term, let $\widehat{\Delta} \triangleq \hq(\theta_t)^{1/2} \vecDelta / \| \vecDelta\|_{\hq(\theta_t)}$,  then 
$\vecDelta = \| \vecDelta\|_{\hq(\theta_t)} \hq(\theta_t) ^{-1/2} \widehat{\Delta}$ and $\| \widehat{\Delta}\|_2 = 1$. We have
\begin{align}\label{eq:sub-s2-3}
    \ell^{\prime \prime}(y, \theta_t x)[\Delta x, \Delta x]& = \vecDelta^\top \nabla^2\ell_{(x,y)}(\theta_t x)\vecDelta \nonumber\\
    &= \| \vecDelta\|_{\hq(\theta_t)}^2  \widehat{\Delta}^\top \hq(\theta_t) ^{-1/2} \nabla^2\ell_{(x,y)}( \theta_t x) \hq(\theta_t)^{-1/2} \widehat{\Delta}.
\end{align}
We claim that $\ell^{\prime \prime}(y, \theta_t x)[\Delta x, \Delta x]$ is a sub-exponential random variable. Indeed,
\begin{align}\label{eq:sub-s2-3-2}
    \bigg\|\ell^{\prime \prime}(y, \theta_t x)[\Delta x, \Delta x]\bigg\|_{\psi_1} &\stackrel{(a)}{\leq}   \| \vecDelta\|_{\hq(\theta_t)}^2 \| \widehat{\Delta}^\top \hq(\theta_t) ^{-1/2} \nabla^2\ell_{(x,y)}(\theta_t x) \hq(\theta_t)^{-1/2} \widehat{\Delta}\|_{\psi_1}\nonumber\\
    & \stackrel{(b)}{\leq}   \| \vecDelta\|_{\hq(\theta_t)}^2 \sup_{u \in\mathcal{S}^{\td-1}}  \| u^\top \hq(\theta_t) ^{-1/2} \nabla^2\ell_{(x,y)}(\theta_t x) \hq(\theta_t)^{-1/2} u\|_{\psi_1}\nonumber\\
    & \stackrel{(c)}{\leq}  \| \vecDelta\|_{\hq(\theta_t)}^2\overline{K}_{2,q},
\end{align}
where (a) follows by Eq.~\eqref{eq:sub-s2-3}, (b) follows by the fact that $\|\widehat{\Delta}\|_2 = 1$, (c) follows by Assumption~\ref{assume:sub-gaussian}-\ref{assume-sub-hessian}. By the property of sub-exponential random variable in Lemma~\ref{lm:sub-exp-properties}-\ref{sub-exp-p1}, we can obtain that
\begin{align}\label{eq:sub-s2-4}
    \E_{(x,y)\sim \pi_q} \big[ \big(\ell^{\prime \prime}(y, \theta_t x)[\Delta x, \Delta x]  \big)^2\big] \lesssim \overline{K}_{2,q}^2  \| \vecDelta\|_{\hq(\theta_t)}^4 \stackrel{Eq.~\eqref{eq:sub-s2-1}}{=}\overline{K}_{2,q}^2  \phi_q^{\prime \prime}(t)^2 .
\end{align}

On the other hand, let $\Delta_i^\top$ be the $i$th row of $\Delta\in\mathbb{R}^{(c-1)\times d}$. For $x\sim q(x)$, define random variable $\xi(x) \triangleq \| \Delta x\|_\infty$, we claim that $\xi(x) $ is sub-Gaussian. Indeed,
\begin{align}\label{eq:sub-s2-6}
    \xi(x) &= \|\Delta x \|_\infty = \max_{i\in[c-1]} |\langle x, \Delta_i \rangle| = \max_{i\in[c-1]} |\langle \V_q^{-1/2} x, \V_q^{1/2} \Delta_{i} \rangle| \nonumber\\
    & = \max_{i\in[c-1]}\|\V_q^{1/2} \Delta_{i} \|_2 \bigg|\bigg\langle \V_q^{-1/2} x, \frac{\V_q^{1/2} \Delta_{i}}{\|\V_q^{1/2} \Delta_{i}\|_2} \bigg\rangle\bigg| \nonumber\\
    &\leq  \big\|\Delta\big\|_{\V_q, \infty} \max_{i \in[c-1]} \bigg|\bigg\langle \V_q^{-1/2} x, \frac{\V_q^{1/2} \Delta_{i}}{\|\Delta_i\|_{\V_q}}  \bigg\rangle \bigg| \triangleq  \big\|\Delta\big\|_{\V_q, \infty} \bigg|\bigg\langle \V_q^{-1/2} x, \frac{\V_q^{1/2} \Delta_{i(x)}}{\|\Delta_{i(x)}\|_{\V_q}}  \bigg \rangle \bigg|
\end{align}
where we define $i(x)$ for each $x$ as the index such that the maximum is attained. Now we have
\begin{align}
  \| \xi(x)\|_{\psi_2}  &\leq  \big\| \Delta\big\|_{\V_q, \infty} \bigg\|\bigg\langle \V_q^{-1/2} x, \frac{\V_q^{1/2} \Delta_{i(x)}}{\|\Delta_{i(x)}\|_{\V_q}}  \bigg \rangle\bigg\|_{\psi_2}\nonumber\\
  &\leq   \big\| \Delta\big\|_{\V_q, \infty} \sup_{u \in \mathcal{S}^{d-1}} \|\langle \V_p^{-1/2} x, u \rangle\|_{\psi_2} =   \big\| \Delta\big\|_{\V_q, \infty} \| \V_q^{-1/2} x\|_{\psi_2}\nonumber\\
  & \leq    \big\| \Delta\big\|_{\V_q, \infty}  K_{0,q},
\end{align}
where the last inequality follows by Assumption~\ref{assume:sub-gaussian}-\ref{assume-sub-x}. Applying Lemma~\ref{lm:sub-properties}-\ref{sub-Gaussian-p2}, we have
\begin{align}\label{eq:sub-s2-7}
    \E_{(x,y)\sim \pi_q} [\| \Delta x\|_\infty^2] = \E_{x\sim q} [|\xi(x)|^2] \lesssim  \big\| \Delta\big\|_{\V_q, \infty}^2  K_{0,q}^2.
\end{align}

Now substitute Eqs.~\eqref{eq:sub-s2-4} and \eqref{eq:sub-s2-7} into Eq.~\eqref{eq:sub-s2-2}, we can prove that $\phi_p(t)$ is pseudo self-concordant:
\begin{align}\label{eq:sub-s2-7-2}
     |\phi_q^{\prime \prime \prime }(t)| \leq C \| \Delta\|_{\V_q,\infty} K_{0,q} \overline{K}_{2,q} \| \vecDelta\|_{\hq(\theta_t)}^2 = C\| \Delta\|_{\V_q,\infty} K_{0,q} \overline{K}_{2,q}\phi_q^{\prime \prime}(t),
\end{align}
where the last equality follows by Eq.~\eqref{eq:sub-s2-1}. We consider the ball $\mathcal{B}_{q,\widehat{r}}(\theta_*) = \{\theta\in \Theta: \|\theta - \theta_* \|_{\V_q,\infty} \leq \widehat{r} \}$, where $\widehat{r}$ is defined by
\begin{align}\label{eq:sub-s2-r-hat}
   \widehat{r} \triangleq \frac{1}{C\log{\sqrt{2}}\cdot K_{0,q} \overline{K}_{2,q} }.
\end{align}
Thus for any $\theta\in\mathcal{B}_{q,\widehat{r}}(\theta_*)$, by Eq.~\eqref{eq:sub-s2-7-2}
\begin{align}
    |\phi_q^{\prime \prime \prime }(t)| \leq \log{\sqrt{2}} \cdot\phi_q^{\prime\prime} (t).
\end{align}
Now we satisfy the premise of  Proposition~\ref{prop:taylor-concordance} by setting $S = \log \sqrt{2}$. With Eq.~\eqref{eq:taylor-Hessian} we can conclude that for  any $\theta\in\mathcal{B}_{q,\widehat{r}}(\theta_*)$,
\begin{align}\label{eq:sub-s2-8}
    1/\sqrt{2} \hq \preceq \hq(\theta) \preceq \sqrt{2}\hq.
\end{align}

\textbf{step 3.} In this step, we consider an $\epsilon$-net $\mathcal{N}_\epsilon$ on ball $\mathcal{B}_{q,\widehat{r}}(\theta_*)$ under metric $\| \cdot\|_{\V_q,\infty}$($\widehat{r}$ is defined in Eq.~\eqref{eq:sub-s2-r-hat}). We intend to approximate empirical Hessian $\bhn(\theta)$ using $\bhn(\theta^\prime)$, where $\theta^\prime \in \mathcal{N}_\epsilon$.

Since $\{x_i \}_{i=1}^n$ are drawn independently from $q(x)$, by \eqref{eq:quadratic-non-center} in \Cref{thm:sub-quadratic}  it holds with probability at least $1-\delta$ that
\begin{align}
    \|x_i \|_{\V_q^{-1}}^2 \lesssim K_{0,q}^2\Big( d + \sqrt{d} \log(e/\delta))\Big).
\end{align}
By union bound and Eq.~\eqref{eq:sub-pf-nu},  with probability at least $1-\delta$ we have 
\begin{align}\label{eq:sub-s3-1}
\max_{i\in [n]} \|x_i \|_{\V_q^{-1}}^2 \lesssim   K_{0,q}^2\Big( d + \sqrt{d} \log(en/\delta))\Big) \triangleq R^2.
\end{align}

Let $\mathcal{N}_\epsilon$ be an  $\epsilon$-net  on ball $\mathcal{B}_{q,\widehat{r}}(\theta_*)$  with $\epsilon$ defined as
\begin{align}
    \epsilon \triangleq \frac{\log{\sqrt{2}}}{2 \cdot R}.
\end{align}
Denote $\mathcal{P}: \mathcal{B}_{q,\widehat{r}}(\theta_*) \rightarrow \mathcal{N}_\epsilon$ as the projection of $\theta\in \mathcal{B}_{q,\widehat{r}}(\theta_*)$ onto the $\epsilon-$net, i.e. $\mathcal{P}(\theta)$ is the closest point in $\mathcal{N}_\epsilon$ to $\theta$ under norm $\| \cdot\|_{\V_q,\infty}$:
\begin{align}\label{eq:sub-s3-2}
    \mathcal{P}(\theta) \in \arg\min_{\theta^\prime \in \mathcal{N}_\epsilon} \|\theta- \theta^\prime \|_{\V_q,\infty}.
\end{align}
We remark that the choice of $\mathcal{P}(\theta)$ does not effect our results. Now pick arbitrary $\theta_1\in\ball$, $\theta_0 = \mathcal{P}(\theta) $, $\theta_t = \theta_0 + t (\theta_1 -\theta_0)$, and $\phi_n(t) = Q_n(\theta_t)$. Using Proposition~\ref{prop:concordance-param}, we have
\begin{align}
    \phi_n^{\prime\prime\prime}(t)| &\leq 2 \phi_n^{\prime \prime} (t) \max_{i\in[n]}\|(\theta_1 - \theta_0 )x_i\|_\infty
    \nonumber\\
    &\leq 2 \phi_n^{\prime \prime}(t) \|\theta_1 - \theta_0 \|_{\V_q,\infty} \max_{i\in[n]} \|x_i \|_{\V_q^{-1}} \nonumber\\
    & \leq 2 R \epsilon \phi_n^{\prime \prime}(t) = \log{\sqrt{2}} \cdot \phi_n^{\prime \prime}(t),
\end{align}
where the last inequality follows by Eqs.~\eqref{eq:sub-s3-1} and \eqref{eq:sub-s3-2}.
Thus $\phi_n(t)$ is pseudo self-concordant, and we can apply Proposition~\ref{prop:taylor-concordance} with $S = \log\sqrt{2}$. By Eq.~\eqref{eq:taylor-Hessian} we have
\begin{align}\label{eq:sub-s3-3}
    1/\sqrt{2} \bhn(\mathcal{P}(\theta)) \preceq \bhn(\theta) \preceq\sqrt{2} \bhn(\mathcal{P}(\theta)), \qquad \forall \theta \in \mathcal{B}_{q,\widehat{r}}(\theta_*).
\end{align}

\textbf{step 4.} In this step we approximate empirical Hessian $\bhn(\theta)$ using $\hq(\theta)$, for all $\theta\in \mathcal{N}_\epsilon$. 
Note that $\bhn(\theta) = \nabla^2 Q_n(\theta)=\frac{1}{n} \sum_{i=1}^n \nabla^2 \ell_{z_i}(\theta x_i)$. For an arbitrary $\theta\in\mathcal{N}_\epsilon$, let $\A_i = \hq(\theta)^{-1/2} \nabla^2 \ell_{z_i}(\theta) \hq(\theta)^{-1/2}$, then $\E[\A_i] = \b I_{\td}$ and 
\begin{align}\label{eq:sub-s4-0}
    \frac{1}{n}\sum_{i\in[n]}\A_i = \hq(\theta)^{-1/2}\bhn(\theta)\hq(\theta)^{-1/2}.
\end{align}
By Assumption~\ref{assume:sub-gaussian}-\ref{assume-sub-hessian}, $\{ \A_i\}_{i=1}^n$ satisfy the premise of \Cref{thm:sub-matrix-covariance}. Applying Corollary~\ref{cor:sub-matrix-covariance} and  then using union bound over all $\theta\in \mathcal{N}_\epsilon$, we obtain that whenever
\begin{align}\label{eq:sub-s4-1}
    n \gtrsim \overline{K}_{2,q}^2 (\td + \log (|\mathcal{N}_\epsilon|/\delta),
\end{align}
where $|\mathcal{N}_\epsilon|$ is the number of points contained in $\mathcal{N}_\epsilon$ , then with probability at least  $1-\delta$, 
\begin{align}\label{eq:sub-s4-3}
    1/2 \b I_{\td} \preceq  \frac{1}{n}\sum_{i\in[n]} \A_i \preceq 3/2  \b I_{\td}, \qquad \forall \theta\in\mathcal{N}_\epsilon.
\end{align}
By Eq.~\eqref{eq:sub-s4-0}, Eq.~\eqref{eq:sub-s4-3} is equivalent to
\begin{align}\label{eq:sub-s4-2}
    1/2 \hq(\theta) \preceq \bhn(\theta) \preceq 3/2 \hq(\theta),\qquad \forall \theta\in\mathcal{N}_\epsilon.
\end{align}

Now we intend to derive a bound for $n$ to satisfy Eq.~\eqref{eq:sub-s4-1}. First we need to estimate an upper bound for $|\mathcal{N}_\epsilon|$. By Proposition 4.2.12 in \cite{Vershynin-2018}, we have $ |\mathcal{N}_\epsilon|\leq (\frac{3\widehat{r}}{\epsilon})^{\td}$. Thus a sufficient condition for \eqref{eq:sub-s4-1} is
\begin{align}
    n \gtrsim \overline{K}_{2,p}^2 \bigg(\td + \td \log\big(\frac{e \widehat{r}}{\epsilon \delta}\big)\bigg) .
\end{align}
Recall that $\widehat{r}= O\bigg( 1/( K_{0,q} \overline{K}_{2,q})\bigg) $, $\epsilon =O\bigg( 1/\Big( K_{0,q}\sqrt{d + \sqrt{d} \log(en/\delta)}\Big)\bigg) $, then
\begin{align}
    \log\big(\frac{e \overline{r}}{\epsilon \delta}\big) = \log \bigg( \frac{e  K_{0,q}\sqrt{d + \sqrt{d} \log(en/\delta)}}{ K_{0,q} \overline{K}_{2,q}}\bigg).
\end{align}
Thus it is sufficient to let 
\begin{align}
    n \gtrsim \overline{K}_{2,q}^2 \td\log(ed/\delta),
\end{align}
which is the first bound at Eq.~\eqref{eq:sub-thm-nbound}.

\textbf{step 5.} Next we prove that if $n$ is larger than the second bound of Eq.~\eqref{eq:sub-thm-nbound}, then $\emptheta \in \mathcal{B}_{q,\widehat{r}}(\theta_*)$ and Eq.~\eqref{eq:sub-thm-erm-detail} holds. First, combining Eqs.~\eqref{eq:sub-s2-8}, \eqref{eq:sub-s3-3} and \eqref{eq:sub-s4-2}, we have with probability at least $1-\delta$,
\begin{align}\label{eq:sub-s5-1}
    \frac{1}{4} \hq \preceq \bhn(
    \theta) \preceq 3 \hq, \qquad \forall \theta \in \mathcal{B}_{q,\widehat{r}}(\theta_*).
\end{align}

Let $\theta_0 = \theta_*$, pick arbitrary $\theta_1 \in \mathcal{B}_{q,\widehat{r}}(\theta_*)$, $\theta_t = \theta_0 + t \Delta$, where $\Delta \triangleq\theta_1 -\theta_0$. By Eq.~\eqref{eq:sub-s2-1}, we already have $\phi_q^{\prime \prime}(0) = \|\vecDelta\|_{\hq}$. On the other hand, we can show that 
\begin{align}
    \phi_n^{\prime \prime}(t) = \frac{1}{n} \sum_{i=1}^n \ell^{\prime \prime }(y_i, \theta x_i)[\Delta x,\Delta x]  = \| \vecDelta\|_{\bhn(\theta_t)},
\end{align}
Thus Eq.~\eqref{eq:sub-s5-1} reduces to
\begin{align}
     \frac{1}{4}\phi_q^{\prime \prime} (0) \leq  \phi_n^{\prime \prime}(t) \leq 3\phi_q^{\prime \prime} (0),\qquad t \in [0,1].
\end{align}
Integrating this twice, we have $\frac{1}{4} \phi_q^{\prime \prime} (0) t^2 \leq \phi_n(t) - \phi_n(0) -\phi_n^\prime (0) t\leq 3 \phi_q^{\prime \prime} (0) t^2$ . Let $t=1$, we can get with probability at least $1-\delta$, 
\begin{align}\label{eq:sub-s5-2}
    \frac{1}{4} \| \vecDelta \|_{\hq}^2 \leq Q_n(\theta) - Q_n(\theta_*) - \langle \vecnabla Q_n(\theta_*), \vecDelta\rangle \leq 3 \| \vecDelta\|_{\hq}^2.
\end{align}
Using Cauchy-Schwartz inequality, we can obtain
\begin{align}\label{eq:sub-s5-3}
    Q_n(\theta) - Q_n(\theta_*) &\geq \frac{1}{4} \| \vecDelta \|_{\hq}^2 + \langle \vecnabla Q_n(\theta_*), \vecDelta\rangle\nonumber\\
    &\geq \frac{1}{4} \| \vecDelta \|_{\hq} \Big(\| \vecDelta \|_{\hq} - 4 \| \vecnabla Q_n(\theta_*) \|_{\hq^{-1}}\Big).
\end{align}
Our goal is to prove that given $n$ lower bounded by the second bound in Eq.~\eqref{eq:sub-thm-nbound}, $\emptheta \in \mathcal{B}_{q,\widehat{r}}$. Since $Q_n(\theta)$ is a convex function and $\ball$ is a convex set, it suffices to show that the  right hand side of Eq.~\eqref{eq:sub-s5-3} is non-negative for all $\theta \in \partial \mathcal{B}_{q,\widehat{r}}$, i.e. $\| \Delta  \|_{\V_q,\infty} = \widehat{r}$. First note that
\begin{align}\label{eq:sub-s5-3-2}
    \| \vecDelta\|_{\hq} &\stackrel{Eq.~\eqref{eq:sub-pf-sigma}}{\geq} \frac{1}{\sqrt{\sigma}} \| \vecDelta \|_{\hp} \stackrel{Eq.~\eqref{eq:sub-pf-rho}}{\geq} \sqrt{\frac{1}{\sigma\rho\nu}} \|\vecDelta \|_{\wV_p} \geq \sqrt{\frac{1}{\sigma\rho}} \|\Delta \|_{\V_p}\nonumber\\ &\stackrel{Eq.~\eqref{eq:sub-pf-nu}}{\geq} \sqrt{\frac{1}{\sigma\rho\nu}} \|\Delta \|_{\V_q} = \sqrt{\frac{1}{\sigma\rho\nu}} \cdot \widehat{r} \geq \frac{1}{C\sqrt{\sigma\rho\nu}K_{0,q} \overline{K}_{2,q}}.
\end{align}
Since we have proved that $\| \vecnabla Q_n(\theta_*)\|_{\hp^{-1}} \lesssim \sqrt{\frac{K_{1,q}^2 \Big(\td+ \sqrt{\td} \log(e/\delta)\Big)}{n}}$ in step 1, connecting this with Eqs.~\eqref{eq:sub-s5-3-2} and \eqref{eq:sub-s5-3}, we have $\emptheta\in \mathcal{B}_{q,\widehat{r}}(\theta_*)$ if
\begin{align}
    n \gtrsim \sigma\rho \nu K_{0,q}^2 K_{1,q}^2  \overline{K}_{2,q}^2  \Big(\td+ \sqrt{\td} \log(e/\delta)\Big).
\end{align}

Now let $\theta_1 = \emptheta$, then $  \vecDelta=\mathrm{vec}(\emptheta -\theta_* ) $. Since $Q_n(\emptheta) \leq Q_n(\theta_*)$, from Eq.~\eqref{eq:sub-s5-3} we can get
\begin{align}\label{eq:sub-s5-4}
    \| \mathrm{vec}(\emptheta -\theta_* )\|_{\hq}^2  \leq \|\vecnabla Q_n(\theta_*) \|_{\hq^{-1}}.
\end{align}

We have proved that $1/\sqrt{2}\hq \preceq \hq(\theta) \preceq \sqrt{2} \hq$ in Eq.~\eqref{eq:sub-s2-8}, it can be reduced to
\begin{align}\label{eq:sub-s5-4-2}
    \frac{1}{\sqrt{2}}\phi_q^{\prime \prime}(0) \leq \phi_q^{\prime \prime}(t) \leq \sqrt{2} \phi_q^{\prime \prime}(0), \quad 0 \leq t\leq 1.
\end{align}
Integrating twice on $[0,1]$, we have $\frac{1}{2\sqrt{2}}\phi_q^{\prime \prime}(0) t^2 \leq \phi_q(t)-\phi_q(0) \leq \frac{\sqrt{2}}{2} \phi_q^{\prime \prime}(0) t^2$. Since $\emptheta \in \mathcal{B}_{q,\widehat{r}}(\theta_*)$, we can assume $\theta_1 = \emptheta$. Let $t=1$, we can get
\begin{align}\label{eq:sub-s5-5}
    L_q(\emptheta) - L_q(\theta_*)  &\stackrel{Eq.~\eqref{eq:sub-pf-phiq}}{=}\phi_q(\emptheta) - \phi_q(\theta_*) \stackrel{Eq.~\eqref{eq:sub-s2-1}}{\leq}\frac{\sqrt{2}}{2} \|  \mathrm{vec}(\emptheta -\theta_* )\|_{\hq}^2 \nonumber\\
    &\stackrel{Eq.~\eqref{eq:sub-s5-4}}{\leq} \frac{\sqrt{2}}{2}\|\vecnabla Q_n(\theta_*) \|_{\hq^{-1}} \stackrel{Eq.~\eqref{eq:sub-thm-gradient}}{\lesssim} \sqrt{\frac{K_{1,q}^2 \Big(\td+ \sqrt{\td} \log(e/\delta)\Big)}{n}}.
\end{align}

\textbf{step 6.} Now we bound the excess risk with respect to $p(x)$, i.e. $L_p(\emptheta) - L_p(\theta_*)$.

Our goal is to use the Taylor expansion property in \Cref{prop:taylor-concordance}. First  we have to show that $L_p(\theta)$ is pseudo self-concordant. Let $\theta_0 = \theta_*$, $\theta_1 = \emptheta$, and $\theta_t = \theta_0  + t \Delta$, where $\Delta = \theta_1 - \theta_0$. Define  
\begin{align}\label{eq:sub-pf-phip}
    \phi_p(t) \triangleq L_p(\theta_t)=\E_{z \sim \pi_p}[\ell_z(\theta_t)].
\end{align}
We can follow the argument from step 2 and obtain that
\begin{align}\label{eq:sub-s6-1}
   | \phi_p^{\prime \prime \prime}(t)| \leq C \| \Delta \|_{\V_p,\infty} K_{0,p} \overline{K}_{2,p} \phi_p^{\prime \prime}(t).
\end{align}
Note that
\begin{align}
    \| \Delta\|_{\V_p,\infty} \leq  \|\vec{\Delta} \|_{\wV_p} \stackrel{Eq.~\eqref{eq:sub-pf-rho}}{\leq} \sqrt{\rho} \|\vec{\Delta} \|_{\hp} &\stackrel{Eq.~\eqref{eq:sub-pf-sigma}}{\leq} \sqrt{\sigma \rho} \|\vec{\Delta} \|_{\hq} \nonumber\\
    &\stackrel{Eq.~\eqref{eq:sub-s5-5}}{\lesssim} \sqrt{\sigma\rho}K_{1,q}\sqrt{\frac{ \td+ \sqrt{\td} \log(e/\delta)}{n}}.
\end{align}
Substitute this into \Cref{eq:sub-s6-1}, we have $| \phi_p^{\prime \prime \prime}(t)| \leq\alpha \phi_p^{\prime \prime}(t)$, where 
\begin{align}
    \alpha = \mathcal{O}\bigg(\sqrt{\sigma\rho}K_{0,p} K_{1,q} \overline{K}_{2,p} \sqrt{\frac{\td + \sqrt{\td} \log(e/\delta)}{n}}  \bigg).
\end{align}
Now we can use Proposition \ref{prop:taylor-concordance} and let $S = \alpha$. Note that $\nabla L_p(\theta_*) = 0$, by \Cref{eq:taylor-expansian} we have
\begin{align}\label{eq:sub-s6-2}
    \frac{e^{-\alpha} + \alpha - 1}{\alpha^2} \| \vecDelta\|_{\hp}^2 \leq L_p(\emptheta) - L_p(\theta_*) \leq \frac{e^{\alpha} -\alpha - 1}{\alpha^2} \| \vecDelta\|_{\hp}^2.
\end{align}

By Taylor theorem, there exits $\ttheta \in \mathcal{B}_{q,\widehat{r}}(\theta_*)$ between $\theta_*$ and $\emptheta$ such that
\begin{align}
    \vecnabla Q_n(\theta_*) = \vecnabla Q_n(\emptheta) + \bhn(\ttheta) \vecDelta.
\end{align}
Since $\vecnabla Q_n(\emptheta) = 0$, we have
\begin{align}\label{eq:sub-s6-3}
    \vecnabla Q_n(\theta_*) = \bhn(\ttheta) \vecDelta.
\end{align}
By Eq.~\eqref{eq:sub-s5-1}, we have $\frac{1}{4}\hq \preceq \bhn(\ttheta) \preceq 3 \hq$. Define $\mqn\triangleq \hq^{1/2} (\bhn(\ttheta))^{-1} \hq^{1/2}$, then
\begin{align}\label{eq:sub-s6-4}
    \frac{1}{3} \b I_{\td} \preceq \mqn \preceq 4 \b I_{\td}.
\end{align}

For the lower bound in Eq.~\eqref{eq:sub-s6-2}, we have with probability at least $1-\delta$,
\begin{align}\label{eq:sub-s6-lower}
    &L_p(\emptheta) -  L_p(\theta_*) {\geq}  \frac{e^{-\alpha} + \alpha - 1}{\alpha^2} \vecDelta^\top \hp \vecDelta\nonumber\\
    &\stackrel{ }{=}  \frac{e^{-\alpha} + \alpha - 1}{\alpha^2}\big( \vecDelta^\top \bhn(\ttheta)\big) \big( \bhn(\ttheta)^{-1}\hp \bhn(\ttheta)^{-1}\big) \big( \bhn(\ttheta) \vecDelta\big) \nonumber\\
    &\stackrel{\Cref{eq:sub-s6-3}}{=} \frac{e^{-\alpha} + \alpha - 1}{\alpha^2} \vecnabla Q_n(\theta_*)^\top \hq^{-1/2}  \mqn \big( \hq^{-1/2}  \hp \hq^{-1/2}\big) \mqn \hq^{-1/2} \vecnabla Q_n(\theta_*) \nonumber\\
    &\stackrel{\Cref{eq:sub-s6-4}}{\geq} \frac{e^{-\alpha} + \alpha - 1}{9\alpha^2} \big\langle \hq^{-1} \hp \hq^{-1}, \vecnabla Q_n(\theta_*) \vecnabla Q_n(\theta_*)^\top\big\rangle.
\end{align}

Similarly, we can derive the upper bound:
\begin{align}\label{eq:sub-s6-upper}
    &L_{p}(\emptheta) -  L_{p}(\theta_*)\leq   \frac{e^{\alpha} -\alpha - 1}{\alpha^2}   \vecDelta^\top \hp \vecDelta \nonumber\\
    &= \frac{e^{\alpha} -\alpha - 1}{\alpha^2} \vecnabla Q_n(\theta_*)^\top \hq^{-1/2}  \mqn \big( \hq^{-1/2}  \hp \hq^{-1/2}\big) \mqn \hq^{-1/2} \vecnabla Q_n(\theta_*) \nonumber\\
    & \leq  16\frac{e^{\alpha} -\alpha - 1}{\alpha^2}\big\langle \hq^{-1} \hp \hq^{-1}, \vecnabla Q_n(\theta_*) \vecnabla Q_n(\theta_*)^\top\big\rangle. 
\end{align}

Given $\{ x_i\}_{i=1}^n\stackrel{i.i.d}{\sim} q(x)$, we have
\begin{align}\label{eq:sub-s6-expectation}
    &\E_{\{y_i\sim p(y_i|x_i,\theta_* )\}_{i=1}^n} [\vecnabla \emp(\theta_*) \vecnabla\emp(\theta_*)^\top ] \nonumber\\
    =& \frac{1}{n^2}\E_{\{y_i\sim p(y_i|x_i,\theta_* )\}_{i=1}^n} \Big[\sum_{i=1}^n \vecnabla \ell_{z_i}(\theta_*) \sum_{j=1}^n (\vecnabla\ell_{z_i}(\theta_*))^\top \Big]\nonumber\\
    =& \frac{1}{n^2} \sum_{i=1}^n \E_{y_i \sim p(y_i|x_i,\theta_* )}[\vecnabla \ell_{z_i}(\theta_*) \vecnabla\ell_{z_i}(\theta_*)^\top]
    + \frac{2}{n^2}\sum_{i\neq j} \E_{\substack{y_i \sim p(y_i|x_i,\theta_* )\\ y_j\sim p(y_j|x_j,\theta_*)}}[\vecnabla \ell_{z_i}(\theta_*) 
    \vecnabla\ell_{z_j}(\theta_*)^\top] \nonumber\\
    \stackrel{(a)}{=} &\frac{1}{n^2} \sum_{i=1}^n \E_{y_i \sim p(y_i|x_i,\theta_* )}[\vecnabla\ell_{z_i}(\theta_*)  \vecnabla\ell_{z_i}(\theta_*) ^\top] 
     \stackrel{(b)}{=} \frac{1}{n^2} \sum_{i=1}^n \E_{y_i \sim p(y_i|x_i,\theta_* )}[\nabla^2\ell_{z_i}(\theta_*) ]  \nonumber\\
     =& \frac{1}{n} \bhn(\theta_*)
\end{align}
where (a) follows by the independence between $y_i$ and $y_j$ and the fact that $\E_{y_i\sim p(y_i|x_i,\theta_*)}[\nabla \ell_{(x_i, y_i)}(\theta_*)] = 0$ from Lemma~\ref{lm:zero-mean-gradient-loss}, (b) follows by Lemma~\ref{lm:fisher-property}.

Similar to the argument in step 4, using Corollary~\ref{cor:sub-matrix-covariance} we have with probability at least $1-\delta$,
\begin{align}\label{eq:sub-s6-4-new}
    \frac{1}{2} \hq \preceq \bhn(\theta_*) \preceq \frac{3}{2} \hq,
\end{align}
where the requirement for $n$ is already satisfied due to the second bound for $n$ in Eq.~\eqref{eq:sub-thm-nbound}. Since $\hq^{-1/2} \hp \hq^{-1/2}$ is symmetric positive definite, we can assume it has eigen-decomposition $\hq^{-1/2} \hp \hq^{-1/2} = \sum_{i=1}^{\td} \lambda_i v_i v_i^\top$. Then
\begin{align}\label{eq:sub-s6-7}
    \big\langle \hq^{-1} \hp \hq^{-1}, \bhn(\theta_*)\big\rangle &=    \big\langle \hq^{-1/2} \hp \hq^{-1/2}, \hq^{-1/2} \bhn(\theta_*) \hq^{-1/2}\big\rangle \nonumber\\
    &= \sum_{i=1}^{d^\prime} \lambda_i v_i^\top \big( \hp^{-1/2} \bhn(\theta_*) \hp^{-1/2}\big) v_i.
\end{align}
Using \Cref{eq:sub-s6-4-new}, we can get upper bound and lower bound of \Cref{eq:sub-s6-7}:
\begin{align}\label{eq:sub-s6-8}
    \frac{1}{2} \langle\hq^{-1}, \hp\rangle\leq \big\langle \hq^{-1} \hp \hq^{-1}, \bhn(\theta_*)\big\rangle \leq \frac{3}{2} \langle\hq^{-1}, \hp\rangle.
\end{align}

Combining Eqs.~\eqref{eq:sub-s6-8} and \eqref{eq:sub-s6-expectation}, we have
\begin{align}
    \frac{\langle\hq^{-1}, \hp\rangle}{2n}&\leq \E_{\{y_i\sim p(y_i|x_i,\theta_* )\}_{i=1}^n} \Big\langle \hq^{-1} \hp \hq^{-1}, \vecnabla Q_n(\theta_*) \vecnabla Q_n(\theta_*)^\top\Big\rangle \nonumber\\
    &= \frac{1}{n} \Big\langle \hq^{-1} \hp\hq^{-1}, \bhn(\theta_*)\Big\rangle \leq \frac{3\langle\hq^{-1}, \hp\rangle}{2n}.
\end{align}

Combining this with the upper bound Eq.~\eqref{eq:sub-s6-upper} and lower bound Eq.~\eqref{eq:sub-s6-lower}, we can obtain with probability at least $1-\delta$,
\begin{align}
    \frac{e^{-\alpha} + \alpha -1}{18 \alpha^2} \frac{\langle\hq^{-1}, \hp\rangle}{n} \leq  \E[L_p(\theta_n)] - L_p(\theta_*) \leq \frac{24(e^{\alpha} - \alpha -1)}{ \alpha^2}\frac{\langle \hq^{-1}, \hp\rangle}{n}.
\end{align}
where the expectation $\E$ is w.r.t $\{y_i \sim p(y_i|x_i, \theta_*)\}_{i=1}^n$.

\end{proof}

\section{Parameter discussion}\label{appendix:parameter}
In this section, we discuss the constants introduced in \Cref{lm:sub-constants}.
In \Cref{prop:param-bounds}, we derive upper bounds for  $K_{1,p}$ and $K_{2,p}(r)$ when \Cref{assume:sub-gaussian} holds. If we additionally assume that $p(x)\sim \mathcal{N}(\b 0,\V_p)$, then  we can derive bounds for $\rho$, $K_{0,p}$, $K_{1,p}$ and $K_{2,p}(r)$ in \Cref{prop:gaussian-design}. Note that we  discuss constants for $p(x)$ here as example, but the results can be similarly extended to $q(x)$ if the same assumption holds for $q(x)$.

\begin{proposition}\label{prop:param-bounds}
Suppose Assumption~\ref{assume:sub-gaussian} holds for $p(x)$. $\rho$ is the minimum constant  defined in Theorem~\ref{thm:sub-thm} such that $\b I_{c-1} \otimes \V_p \preceq \rho \hp$. Then
\begin{enumerate}[label={(\arabic*)},leftmargin=*]
\item For $K_{1,p}$ defined in \Cref{lm:sub-constants}-\ref{assume-sub-grad}, we have
\begin{align}\label{eq:k1-bound}
    K_{1,p} < 2\sqrt{\rho} K_{0,p}.
\end{align}
\item For $K_{2,p}(r)$ defined in \Cref{lm:sub-constants}-\ref{assume-sub-hessian}, let $\rho(\theta) >0$ be constant s.t. $\b I_{c-1} \otimes \V_p \preceq \rho(\theta) \hp(\theta)$ for $\theta \in \mathcal{B}_{r}(\theta_*)$, we have
    \begin{align}\label{eq:k2-bound}
         K_{2,p}(r) < 2 \sup_{\theta \in \mathcal{B}_r(\theta_*)} \rho(\theta) K_{0,p}^2.
    \end{align}
\end{enumerate}
\end{proposition}

\begin{proof}
For the ease of notation, we use $\wc = c-1$ and $\td = d(c-1)$.  We define ${\b h}(x,\theta) \mathbb{R}^{\wc}$ for a given $x\in\mathbb{R}^d$ and $\theta\in\mathbb{R}^{\wc \times d}$ by 
\begin{align}
    {\b h}_i(x, \theta) = \frac{\exp(x^\top \theta_i)}{1 + \sum_{s \in [\wc]} \exp(x^\top \theta_s)}, \qquad \forall i \in[\wc]
\end{align}
where $\theta_i$ is the $i$-th row of $\theta$.
\begin{enumerate}[label={(\arabic*)},leftmargin=*]
\item 
Denote $\wV_p \triangleq \b I_{\wc}\otimes \V_p$, then $\wV_p \preceq \rho \hp$ and $\hp^{-1/2} \preceq \sqrt{\rho} \wV_p^{-1/2}$. Thus
    \begin{align}\label{eq:k1-1}
        \|\hp^{-1/2}\vecgrad\|_{\psi_2}\leq \sqrt{\rho} \|\wV_p^{-1/2} \vecgrad \|_{\psi_2}.
    \end{align}
By \Cref{prop:grad-and-hessian},  the $i$-th row ($i\in[\wc]$) of matrix $\gradloss$ is
\begin{align}
    [\gradloss]_i = \frac{\partial \ell_{(x,y)}(\theta_*)}{\partial \theta_{*,i}} = \beta_i(x,y) x,\nonumber
\end{align}
where $\beta_i(x,y) \triangleq -1_{\{y=i\}} + {\b h}_i(x,\theta_*)$.

Therefore $\Big(\vecgrad\Big)^\top = [\beta_1(x,y) x^\top, \beta_2(x,y) x^\top, \cdots, \beta_{\wc}(x,y) x^\top]$ and thus 
\begin{align}\label{eq:k1-2-0}
    &\big(\wV_p^{-1/2}\vecgrad\big)^\top \nonumber\\
    =& \big[\beta_1(x,y)( \V_p^{-1/2} x)^\top, \beta_2(x,y) ( \V_p^{-1/2} x)^\top, \cdots, \beta_{\wc}(x,y)( \V_p^{-1/2} x)^\top\big] .
\end{align}
We also observe that for any $(x,y)$, 
\begin{align}\label{eq:k1-2}
\sum_{i\in [\wc]}| \beta_i(x,y)|  \leq 1 + \frac{\sum_{j\in[\wc]} \exp(x^\top \theta_j^*)}{1 + \sum_{j\in[\wc]} \exp(x^\top \theta_j^*)} < 2.
\end{align}
By definition of the sub-Gaussian vector norm we have 
\begin{align}\label{eq:k1-3}
    \|\wV_p^{-1/2}\vecgrad\|_{\psi_2}\triangleq\sup_{u\in \mathcal{S}^{d\wc -1}} \|\langle \wV_p^{-1/2}\vecgrad,u \rangle \|_{\psi_2}
\end{align} 
where $\mathcal{S}^{\td -1}$ is the unit sphere in $\mathbb{R}^{\td}$. For any $u\in \mathcal{S}^{d\wc -1}$, we represent $u^\top = [u_1^\top, u_2^\top, \cdots, u_{\wc}^\top ]$, where $u_i \in \mathbb{R}^{d}$ for each $i\in[\wc]$. Then for any $y\in[c]$, by \Cref{eq:k1-2-0} we have
\begin{align}
    \|\langle \wV_p^{-1/2}\vecgrad,u \rangle \|_{\psi_2} = \Big\| \sum_{i\in[\wc]} \beta_{i}(x,y) u_i^\top \V_p^{-1/2} x \Big\|_{\psi_2}.
\end{align}
For a given $x$ and $u\in \mathcal{S}^{\td -1}$, define 
\begin{align}
    u(x) \in  \argmax_{u_i, i\in[\wc]} |u_i^\top \V_p^{-1/2} x|,
\end{align}
where the choice of $u(x)$ does not effect our result. By Eq.~\eqref{eq:k1-2},
\begin{align}\label{eq:k1-4}
    \|\langle \wV_p^{-1/2}\vecgrad,u \rangle \|_{\psi_2} < 2 \|(u(x))^\top \V_p^{-1/2} x  \|_{\psi_2}.
\end{align}
Since $\| u(x)\| \leq 1$, by combining Eqs.~\eqref{eq:k1-4} and \eqref{eq:k1-3} we can get
\begin{align}
     \|\wV_p^{-1/2}\vecgrad\|_{\psi_2} < 2 \sup_{v \in \mathcal{S}^{d-1}} \|v^\top \V_p^{-1/2} x \|_{\psi_2} = 2 \| \V_p^{-1/2} x\|_{\psi_2} \leq 2 K_{0,p}.
\end{align}
\item
   Let $\W_p(\theta) \triangleq \wV_p ^{1/2} \hp(\theta)^{-1/2}$, then $\W_p(\theta) \preceq \sqrt{\rho(\theta)} \b I_{\td}$.  First, we observe that
\begin{align}\label{eq:k2-bound-1}
& \sup_{u \in \mathcal{S}^{\td-1}} \|u^\top \hp(\theta)^{-1/2} \nabla^2\ell_{(x,y)}(\theta) \hp(\theta)^{-1/2} u \|_{\psi_1}  \nonumber\\
= & \sup_{\substack{v \triangleq \W_p(\theta) u \\ \|u\|_2 \leq 1}} \|v^\top \wV_p^{-1/2} \nabla^2\ell_{(x,y)}(\theta)\wV_p^{-1/2} v \|_{\psi_1} \nonumber\\
\stackrel{(a)}{\leq} & \sup_{\| u\|_2\leq 1} \| (\sqrt{\rho(\theta)} u)^\top \wV_p^{-1/2} \nabla^2\ell_{(x,y)}(\theta) \wV_p^{-1/2} (\sqrt{\rho(\theta)} u)\|_{\psi_1}\nonumber\\
\leq & \rho(\theta) \sup_{u \in \mathcal{S}^{\td -1}} \| u^\top \wV_p^{-1/2} \nabla^2\ell_{(x,y)}(\theta) \wV_p^{-1/2} u\|_{\psi_1},
\end{align}
where (a) follows by the fact that $\lambda_{\max}(\W_p(\theta)) \leq \sqrt{\rho(\theta)})$ and thus $\{v = \W_p(\theta)) u: \|u\|_2 \leq 1 \} \subset \{\sqrt{\rho(\theta)} u: \|u\|_2 \leq 1 \}$. 

By \Cref{prop:grad-and-hessian}, we have the Hessian  $\nabla^2\ell_{(x,y)}(\theta)\in\mathbb{R}^{\td \times \td}$ with the following form:
\begin{align}\label{eq:k2-bound-hessian}
    \nabla^2 \ell_{(x,y)}(\theta) = \begin{bmatrix}
\alpha_{11}(x,\theta) x x^\top  & \cdots & \alpha_{1\wc}(x,\theta) x x^\top\\
\vdots & \ddots & \vdots\\
\alpha_{\wc 1}(x,\theta) x x^\top  & \cdots & \alpha_{\wc \wc}(x,\theta) x x^\top\\
\end{bmatrix}
\end{align}
where
\begin{align}\label{eq:k2-bound-alpha}
    \alpha_{i,j}(\theta) = 1_{\{i=j\}}{\b h}_i(x,\theta) -{\b h}_i(x,\theta){\b h}_j(x,\theta).
\end{align}
For any $u\in\mathcal{S}^{\td-1}$, we decompose it into $\wc$ chunks with dimension $d$, i.e. $u^\top = [u_1^\top, \cdots, u_{\wc}^\top]$ and $u_i \in\mathbb{R}^d$. Since $\wV_p = \b I_{\wc}\otimes \V_p$, we have $\wV_p^{-1/2} = \b I_{\wc} \otimes \V_p^{-1/2}$. Define $\tu_i\triangleq \V_p^{-1/2} u_i$, $\tu \triangleq \wV_p^{-1/2} u$, then $\tu^\top = [\tu_1^\top, \cdots, \tu_{\wc}^\top]$. For the ``$\sup$'' term in Eq.~\eqref{eq:k2-bound-1}, we have
\begin{align}\label{eq:k2-bound-2}
   & \sup_{u \in \mathcal{S}^{\td -1}} \| u^\top \wV_p^{-1/2} \nabla^2\ell_{(x,y)}(\theta) \wV_p^{-1/2} u\|_{\psi_1} =\sup_{u \in \mathcal{S}^{\td -1}} \| \tu^\top  \nabla^2\ell_{(x,y)}(\theta) \tu\|_{\psi_1}\nonumber\\
    \stackrel{(a)}{=} &\sup_{u \in \mathcal{S}^{\td -1}} \Big\| \sum_{i\in[\wc]} \sum_{j \in[\wc]} \alpha_{ij}(x,\theta) \tu_i^\top x x^\top \tu_j \Big\|_{\psi_1} \nonumber\\
    \stackrel{(b)}{=} &\sup_{u \in \mathcal{S}^{\td -1}} \Big\| \sum_{i\in[\wc]} \sum_{j \in[\wc]} \alpha_{ij}(x,\theta) u_i^\top (\V_p^{-1/2} x) (\V_p^{-1/2} x) ^\top u_j\Big\|_{\psi_1} ,
\end{align}
where (a) follows by Eq.~\eqref{eq:k2-bound-alpha}, (b) follows by $\tu_i = \V_p^{-1/2} u_i$.

Now we  intend to upper bound Eq.~\eqref{eq:k2-bound-2} by using $\|\V_p^{-1/2} x\|_{\psi_2} \leq K_{0,p}$.  First for any $x\in \mathbb{R}$ and $u\in\mathcal{S}^{\td-1}$, we define 
\[u(x) \in \argmax_{u_i, i\in[\wc]} \big|u_i^\top (\V_p^{-1/2} x)(\V_p^{-1/2} x)^\top u_i \big| ,\]
where the choice of $u(x)$ does not effect our result. Since for any $a,b \in \mathbb{R}$, we have inequality $|ab|\leq \frac{a^2 + b^2}{2}\leq \max\{a^2, b^2 \}$, then
\begin{align}\label{eq:k2-bound-3}
    \big|u_i^\top (\V_p^{-1/2} x)(\V_p^{-1/2} x)^\top u_j \big|\leq \big|u(x)^\top (\V_p^{-1/2} x)(\V_p^{-1/2} x)^\top u(x) \big|, \qquad \forall i,j \in[\wc].
\end{align}
On the other hand, by Eq.~\eqref{eq:k2-bound-alpha} we have
\begin{align}
    |\alpha_{ij}(x,\theta) | = \begin{cases}
      {\b h}_i(x,\theta) - {\b h}_i^2(x,\theta) & \text{if $i=j$,}\\
      {\b h}_i(x,\theta){\b h}_j(x,\theta) & \text{otherwise.}
    \end{cases}       
\end{align}
Thus
\begin{align}\label{eq:k2-bound-4}
    \sum_{i\in[\wc]} \sum_{j\in[\wc]} |\alpha_{ij}(x,\theta)|  &= \sum_{i\in[\wc]}\Big[ {\b h}_i(x,\theta) -  {\b h}_i^2(x,\theta)  +  {\b h}_i(x,\theta) [\| {\b h}(x,\theta) \|_1 -  {\b h}_i(x,\theta)]\Big]\nonumber\\
    & = \sum_{i\in[\wc]} \Big[(1+ \| {\b h}(x,\theta)\|_1)  {\b h}_i(x,\theta)- 2  {\b h}_i^2(x,\theta)\Big ]\nonumber\\
    & = (1+ \| {\b h}(x,\theta)\|_1) \| {\b h}(x,\theta)\|_1 - 2 \sum_{i\in[\wc]} {\b h}_i^2(x,\theta)\nonumber\\
    &<2,
\end{align}
where the last inequality follows by the fact that $\| {\b h}(x,\theta)\|_1  = 1 - \frac{1}{1+\sum_{s\in[\wc]} \exp(x^\top \theta_s)} <1$.

Now substitute Eq.~\eqref{eq:k2-bound-2} into Eq.~\eqref{eq:k2-bound-1}, we can obtain that
\begin{align}\label{eq:k2-bound-5}
    & \sup_{u \in \mathcal{S}^{\td-1}} \|u^\top \hp(\theta)^{-1/2} \nabla^2\ell_{(x,y)}(\theta) \hp(\theta)^{-1/2} u \|_{\psi_1}\nonumber\\
    \leq & \rho(\theta) \sup_{u \in \mathcal{S}^{\td -1}} \Big\| \sum_{i\in[\wc]} \sum_{j \in[\wc]} \alpha_{ij}(x,\theta) u_i^\top (\V_p^{-1/2} x) (\V_p^{-1/2} x) ^\top u_j\Big\|_{\psi_1}\nonumber\\
   \stackrel{(a)}{\leq} & \rho(\theta) \sup_{u \in \mathcal{S}^{\td -1}} \Big\|\Big( \sum_{i\in[\wc]} \sum_{j \in[\wc]} |\alpha_{ij}(x,\theta)|\Big) \Big( u(x)^\top(\V_p^{-1/2} x) (\V_p^{-1/2} x) ^\top u(x) \Big)\Big\|_{\psi_1}\nonumber\\
    \stackrel{(b)}{<} &2 \rho(\theta) \sup_{v \in \mathcal{S}^{d -1}} \|(v^\top \V_p^{-1/2} x)^2\|_{\psi_1}\nonumber\\
    \stackrel{(c)}{=} &2 \rho(\theta)\sup_{v \in \mathcal{S}^{d -1}} \|(v^\top \V_p^{-1/2} x)\|_{\psi_2}^2 \nonumber\\
     = &  2 \rho(\theta) \|\V_p^{-1/2}x \|_{\psi_2}^2\stackrel{(d)}{\leq}  2 \rho(\theta) K_{0,p}^2,
\end{align}
where (a) follows by Eq.~\eqref{eq:k2-bound-3}, (b) follows by Eq.~\eqref{eq:k2-bound-4} and the fact that $u(x) \in\mathbb{R}^d$ and $\| u(x)\|_2\leq 1$, (c) follows by Lemma~\ref{lm:exp-gaussian-square}, (d) follows by \Cref{lm:sub-constants}-\ref{assume-sub-x}. Comparing Eq.~\eqref{eq:k2-bound-5} to \Cref{eq:sub-assume-hessian} (in \Cref{lm:sub-constants}-\ref{assume-sub-hessian}), we can get
\begin{align}
     K_{2,p}(r) < 2 \sup_{\theta \in \mathcal{B}_r(\theta_*)} \sqrt{\rho(\theta)} K_{0,p}.
\end{align}
\end{enumerate}
\end{proof}

Before establishing the result for Gaussian design, we provide a  form of Hessian expression of the loss function with respect to $\theta$ in the following lemma.
\begin{Lemma}\label{lm:sub-hessian-decompose}
    For any  $(x,y)$  and parameter $\theta$, $\nabla^2 \ell_{(x,y)}(\theta) = \tx(\theta) \tx(\theta)^\top$, where $\tx(\theta) =  (\ell^{\prime \prime}(y,\theta x))^{1/2}\otimes x$.
\end{Lemma}
\begin{proof}
    The proof is trivial. By chain rule, $\nabla^2 \ell_{(x,y)}(\theta) = \ell^{\prime \prime}(y,\theta x) \otimes xx^\top$.
\end{proof}

In the following proposition, we consider the case for a Gaussian design, i.e. $p(x) \sim \mathcal{N}(\b 0, \V_p)$. In particular, we present the bounds for constants $\rho$, $K_{0,p}$, $K_{1,p}$ and $K_{2,p}(r)$ used in \Cref{thm:sub-thm} by using $\theta_*$, $\V_p$ and $r$. Our bound for $\rho$ is inspired  Proposition D.1 in \cite{Bach-2018}, where the binary logistic regression on Gaussian design is considered.

\begin{proposition}[Gaussian design]\label{prop:gaussian-design}
Suppose $p(x)\sim \mathcal{N}(\b 0, \V_p)$, Assumption~\ref{assume:sub-gaussian} holds for $p(x)$. Suppose that $\rho >0$ is the minimum constant such  that $\wV_p \triangleq \b I_{\wc} \otimes \V_p \preceq  \rho \hp$, then for $\rho$ and constant defined in \Cref{lm:sub-constants}, we have
\begin{align}
    \rho &\lesssim\big(2+ \max_{i\in[\wc]} \|\theta_{*,i}\|_{\V_p}^2 \big)^{3/2}, \label{eq:rho-bound-gaussian}\\
    K_{0,p} & \lesssim 1 ,\label{eq:K0-bound-gaussian}\\
    K_{1,p} &\lesssim \big(2+ \max_{i\in[\wc]} \|\theta_{*,i}\|_{\V_p}^2 \big)^{3/4}, \label{eq:K1-bound-gaussian}\\
    K_{2,p}(r) & \lesssim  \big(2+ r^2 + \max_{i\in[\wc]} \|\theta_{*,i}\|_{\V_p}^2 \big)^{3/4} ,\label{eq:K2-bound-gaussian}
\end{align}
where $\theta_{*,i}$ is the $i$-th row of $\theta_* \in \mathbb{R}^{(c-1) \times d}$.
\end{proposition}

\begin{proof}
$ $
\begin{enumerate}[label={(\arabic*)},leftmargin=*]
\item Proof of Eq.~\eqref{eq:rho-bound-gaussian}.

First, we consider the decorrelated design $z\triangleq \wV_p^{-1/2} x$, thus $z \sim \mathcal{N}(\b 0, \b I_{\wc})$. Define parameter $\xi \triangleq \theta \wV_p^{1/2}$, and denote $\xi_* = \theta_* \wV_p^{1/2}$. Then we have $\theta x = \xi z$. By Lemma~\ref{lm:sub-hessian-decompose}, we have  
\begin{align}\label{eq:rho-0}
    \hp = \hp(\theta_*) = \E_{x}[\tx(\theta_*) \tx(\theta_*)^\top],
\end{align}
where $\tx(\theta) = [\ell^{\prime \prime} (y,\theta x)]^{1/2} \otimes x$, note that Hessian $\ell^{\prime \prime} (y,\theta x) \in \mathbb{R}^{\wc \times \wc}$ has no dependence on label $y$.

Now we define $\tz(\xi) \triangleq \wV_p^{-1/2} \tx(\theta)$, then
\begin{align}\label{eq:rho-1}
    \tz(\xi) &= (\b I_{\wc}\otimes \wV_p^{-1/2} ) ( [\ell^{\prime \prime} (y,\theta x)]^{1/2} \otimes x) =  ([\ell^{\prime \prime} (y,\theta x)]^{1/2}) \otimes (\wV_p^{-1/2} x) \nonumber\\
    &=  [\ell^{\prime \prime} (y,\xi z)]^{1/2} \otimes z.
\end{align}
Then the covariance matrix of $\tz(\xi_*)$ has the following form:
\begin{align}
    \bPsi(\xi_*) &\triangleq \E_{z}[\tz(\xi_*) \tz(\xi_*)^\top] \nonumber\\
    &=\E_{z} [ \ell^{\prime \prime} (y,\xi_* z) \otimes (zz^\top)] \label{eq:rho-2}\\
    &=\wV_p^{-1/2}  \hp \wV_p^{-1/2} , \nonumber
\end{align}
where the last equality follows by definition of $\tz(\xi_*)$ and \Cref{eq:rho-0}. Thus, we can upper bound $\rho$ by finding lower bound of $\lambda_{\min}(\bPsi(\xi_*))$ since by the definition of $\rho$, we have
\begin{align}\label{eq:rho-3}
    \rho \leq \frac{1}{\lambda_{\min}(\bPsi(\xi_*))}.
\end{align}
For any $z \sim \mathcal{N}(\b 0,\b I_{\wc})$, we have
\begin{align}
  \ell^{\prime \prime} (y,\xi_* z) = \bGamma(z) - {\b h}(z) {\b h}(z)^\top,
\end{align}
where ${\b h}(z) \in \mathbb{R}^{\wc}$ and 
\begin{align}
    {\b h}_i(z) = \frac{\exp(z^\top \xi_{*,i})}{1 + \sum_{j \in[\wc]} \exp(z^\top \xi_{*,j})},
\end{align}
and $\bGamma(z) = \mathrm{diag}( {\b h}_1(z), {\b h}_2(z), \cdots, {\b h}_{\wc}(z))$.
Thus for any $z \sim \mathcal{N}(\b 0, \b I_{\wc})$,
\begin{align}
    \ell^{\prime \prime} (y,\xi_* z) &= \bGamma(z)^{1/2}\Big[\b I_{\wc} - \big(\bGamma(z)^{-1/2} {\b h}(z)\big)\big(\bGamma(z)^{-1/2} {\b h}(z)\big)^\top \Big]\bGamma(z)^{1/2}\nonumber\\
    &\succeq (1 - \| \bGamma(z)^{-1/2} {\b h}(z)\|_2^2) \bGamma(z)\nonumber\\
    &= (1 - \|{\b h}(z)\|_1)\bGamma(z),
\end{align}
where the last equality follows by the fact that the $i$-th component of $\bGamma(z)^{-1/2} {\b h}(z)$ is $\sqrt{{\b h}_i(z)}$. Substitute this into \Cref{eq:rho-2}, we can get
\begin{align}
    \bPsi(\xi_*) \succeq \E_{z} \Big[(1 - \|{\b h}(z)\|_1) \bGamma(z) \otimes  (z z^\top)\Big].
\end{align}
Note that $\bGamma(z)$ is a diagonal matrix, we additionally have 
\begin{align}\label{eq:rho-4}
    \lambda_{\min}[\bPsi(\xi_*)]& = \lambda_{\min}\Big( \E_{z}\Big[(1 - \|{\b h}(z)\|_1) \bGamma(z)\otimes (z z^\top)  \Big]\Big)\nonumber\\
    &= \min_{i \in [\wc]} \lambda_{\min}\Big( \E_{z}\Big[{\b h}_i(z)(1 - \|{\b h}(z)\|_1) z z^\top\Big] \Big).
\end{align}

For any arbitrary $i\in [\wc]$, we have
\begin{align}
    {\b h}_i(z)(1 - \|{\b h}(z)\|_1) = \frac{\exp(z^\top \xi_{*,i})}{\Big(1 + \sum_{j \in [\wc]}\exp(z^\top \xi_{*,j}) \Big)^2}.
\end{align}
By the symmetry of $\mathcal{N}(\b 0, \b I_{\wc})$, w.l.o.g. we can assume that $\xi_{*,i}$ is parallel to $e_1$, where $e_1$ is the unit vector of the first coordinate. Thus we have $z^\top \xi_{*,i} = \| \xi_{*,i}\|_2  z_1$ and 
\begin{align}\label{eq:rho-4-2}
    {\b h}_i(z)(1 - \|{\b h}(z)\|_1) = \frac{\exp(t_i z_1)}{\Big(1 + \beta + \exp(t_i z_1)  \Big)^2} \approx \exp(-|t_i z_1|),
\end{align}
where we use  $\approx$ to represent the intersection of $\lesssim$ and $\gtrsim$,  $\beta = \sum_{j\neq i} \exp(z^\top \xi_{*,j})$ and we define $t_i$ by 
\begin{align}\label{eq:rho-ti}
    t_i \triangleq \|\xi_{*,i}\|_2 = \| \theta_* \V_p^{1/2} \|_2 = \|\theta_* \|_{\V_p}.
\end{align}
Now by \Cref{eq:rho-4-2} we have
\begin{align}\label{eq:rho-5}
    \E_{z} \Big[{\b h}_i(z)(1 - \|{\b h}(z)\|_1) z z^\top\Big] &\approx \E_{\{z_i \sim \mathcal{N}(0,1)\}_{i=1}^d}[\exp(- |t_i z_1|) z z^\top]\nonumber\\
    &= \begin{bmatrix}
\kappa & \b 0_{d-1}^\top \\
\b 0_{d-1} & \kappa_{\perp} \b I_{d-1},
\end{bmatrix}
\end{align}
where $\kappa$ and $\kappa_\perp$ have the following forms if we denote the standard one dimensional Gaussian density function as $\phi(\cdot)$:
\begin{align}
    &\kappa = \int_{-\infty}^\infty \exp(-|t_i u|) u^2 \phi(u) du , \label{eq:rho-k1}\\
    &\kappa_{\perp} = \int_{-\infty}^\infty \exp(-|t_i u|) \phi(u) du.\label{eq:rho-k2}
\end{align}
By \Cref{eq:rho-3,eq:rho-4,eq:rho-5}, we can upper bound $\rho$ by finding the lower bounds for $\kappa$ and $\kappa_\perp$. First we denote the Gaussian integral as $G(t)\triangleq \int_{t}^\infty e^{-u^2/2}du$, which has sharp bounds as 
\begin{align}\label{eq:rho-gt}
    \frac{2e^{-t^2/2}}{t + \sqrt{t^2 + 4}} \leq G(t) \leq \frac{2e^{-t^2/2}}{t + \sqrt{t^2 + 8\pi}}, \qquad t\geq 0.
\end{align}
For $\kappa$, we have
\begin{align}\label{eq:rho-k1bound}
    \kappa& = \sqrt{\frac{2}{\pi}}\cdot \int_{0}^\infty e^{-t_i u -u^2} u^2 du = \sqrt{\frac{2}{\pi}} e^{t_i^2/2} \int_{0}^\infty  e^{-(u+t_i)^2/2} u^2 du \nonumber\\
    &= \sqrt{\frac{2}{\pi}}\cdot e^{t_i^2/2} \int_{t_i}^\infty e^{-v^2/2} (v-t)^2 dv\nonumber\\
    & = \sqrt{\frac{2}{\pi}} \cdot e^{t_i^2/2} \big[(1+t_i^2) G(t_i) - t_i e^{-t_i^2/2}\big].\nonumber\\
    & \stackrel{\text{(a)}}{\gtrsim} \frac{2 (t_i^2 +1)}{t_i + \sqrt{t_i^2 +4}} - t_i = \frac{t_i(t_i -\sqrt{t_i^2+4}) + 2}{t_i + \sqrt{t_i^2+4}}\nonumber\\
    & = \frac{2(\sqrt{t_i^2 +4} - t_i)}{(\sqrt{t_i^2 +4} + t_i)^2} = \frac{8}{(\sqrt{t_i^2 +4} + t_i)^3} \geq \frac{1}{(t_i^2 + 2)^{3/2}},
\end{align}
where (a) follows by the lower bound of $G(t_i)$ from \eqref{eq:rho-gt}. 
Similarly for $\kappa_{\perp}$,
\begin{align}\label{eq:rho-k2bound}
    \kappa_\perp &= \sqrt{\frac{2}{\pi}} \cdot \int_{0}^\infty e^{-t_i u - u^2/2} du \nonumber\\
    &=\sqrt{\frac{2}{\pi}} e^{t_i^2/2} \cdot \int_{t_i}^\infty e^{-v^2/2} dv =  \sqrt{\frac{2}{\pi}} e^{t_i^2/2} G(t_i) \nonumber\\
    & \gtrsim \frac{1}{(t_i^2 +2)^{1/2}}.
\end{align}
Combining \eqref{eq:rho-5}, \eqref{eq:rho-k1bound} and \eqref{eq:rho-k2bound}, we can get for each $i\in[\wc]$,
\begin{align}
   \lambda_{\min}\Big( \E_{z}\Big[{\b h}_i(z)(1 - \|{\b h}(z)\|_1) z z^\top\Big] \Big) \gtrsim \min\{\kappa, \kappa_\perp\} \gtrsim \frac{1}{(t_i^2 +2)^{3/2}}.
\end{align}
Substitute this into \eqref{eq:rho-4}, we have
\begin{align}
    \lambda_{\min}[\bPsi(\xi_*)] \gtrsim \min_{i \in [\wc]} \frac{1}{(t_i^2 +2)^{3/2}}.
\end{align}
Combining this with the bound of $\rho$  in \eqref{eq:rho-3} and the definition of  $t_i$ 
 in \eqref{eq:rho-ti},  we can obtain that
\begin{align}
    \rho \leq \frac{1}{\lambda_{\min}[\bPsi(\xi_*)] } \lesssim \max_{i\in[\wc]} (2 + \|\theta_{*,i}\|_{\V_p}^2)^{3/2} = \big( 2 + \max_{i \in [\wc]}\|\theta_{*,i}\|_{\V_p}^2 \big)^{3/2}.
\end{align}
\item Since $x \sim \mathcal{N}(\b 0, \V_p)$, $
\V_p^{-1/2} x \sim \mathcal{N}(\b 0, \b I_{d})$. For any $u\in\mathcal{S}^{d-1}$, $u^\top \V_p^{-1/2} x \sim \mathcal{N}(0,1)$. Thus
\begin{align}
    \| \V_p^{-1/2} x\|_{\psi_2} = \sup_{u \in \mathcal{S}^{d-1}} \|u^\top \V_p^{-1/2} x \|_{\psi_2} \lesssim 1
\end{align}
and $K_{0,p} \lesssim 1$.
\item Substitute Eqs.~\eqref{eq:rho-bound-gaussian} and ~\eqref{eq:K0-bound-gaussian} into Eq.~\eqref{eq:k1-bound}, we have
\begin{align}
    K_{1,p} < 2\sqrt{\rho}K_{0,p} \lesssim \Big(2+ \max_{i\in[\wc]} \|\theta_{*,i}|_{\V_p}^2 \Big)^{3/4}.
\end{align}
\item Substitute Eqs.~\eqref{eq:rho-bound-gaussian} and ~\eqref{eq:K0-bound-gaussian} into Eq.~\eqref{eq:k2-bound},  we have
\begin{align}
    K_{2,p}(r) &< 2 \sup_{\theta \in \mathcal{B}_r(\theta_*)} \rho(\theta) K_{0,p}^2\nonumber\\
    & \lesssim \sup_{\max_{i\in[\wc]}\| \theta_i - \theta_{*,i}\|_{\V_p} \leq r }(2 + \max_{i\in[\wc]} \|\theta_i\|_{\V_p}^2)^{3/4}\nonumber\\
    & \lesssim \big(2 + r^2 + \max_{i\in[\wc]} \|\theta_{*,i}\|_{\V_p}^2\big)^{3/4},
\end{align}
where the last inequality follows by the triangle inequality $\| \theta_i\|_{\V_p} \leq \|\theta_i - \theta_{*,i} \|_{\V_p} + \| \theta_{*,i}\|_{\V_p}$.
\end{enumerate}
\end{proof}

\section{Bounded domain}\label{appendix:bounded-domain}
For the case of bounded domain, we present the assumptions in \Cref{assumption:heavy}, which are similar to the regularity assumptions used in \cite{chaudhuri-2015}. Then we present the excess risk $L_p(\theta_n) - L_p(\theta_*)$ bounds in \Cref{thm:heavy}.  Our proof is inspired by the proof of Theorem 5.1 in \cite{frostig2015competing}.

\begin{assumption}\label{assumption:heavy}
There exist constants $L_1, L_2$ and $L_3 >0$, for any sample $(x,y)$ randomly drawn from distribution $\pi_p(x,y)$ or $\pi_q(x,y)$, the following conditions are satisfied: 
\begin{enumerate}[label={(\arabic*)},leftmargin=*]
\item \label{assume:bound-pd-matrix} $\hp$ and $\hq$ are positive definite.
\item \label{assume:bound-grad-hessian} gradient and Hessian of loss function with respect to $\theta$ at $\theta_*$ are bounded: 
\begin{align}
     \| \mathrm{vec}(\nabla \ell_{(x,y)}(\theta_*) )\|_{\hp^{-1}} \leq L_1 ,\qquad
    \| \hp^{-1/2}\nabla^2 \ell_{(x,y)}(\theta_*)\hp^{-1/2} \| \leq L_2 ,\label{eq:heavy-assume-2}
\end{align}
\item \label{assume:bound-lipschitz} Lipschitz continuity of Hessian: there exits a neighborhood around $\theta_*$ denoted by $\mathcal{B}(\theta_*)$ such that $\forall \theta^\prime \in \mathcal{B}(\theta_*)$, 
\begin{align}
    \Big\| \hp^{-1/2}\Big(\nabla^2 \ell_{(x,y)}(\theta_*) -  \nabla^2 \ell_{(x,y)}(\theta^\prime)\Big) \hp^{-1/2} \Big\| \leq L_3 \|\mathrm{vec}(\theta_* - \theta^\prime) \|_{\hp}.\label{eq:heavy-assume-3}
\end{align}
\end{enumerate}
\end{assumption}

\begin{remark}
    We did not explicitly assume that $x\in \mathbb{R}^d$ is bounded. However, by \Cref{prop:grad-and-hessian}, each row of gradient $\nabla_(x,y)(\theta_*)$ is the scaling of $x$. Thus \Cref{assumption:heavy}-\ref{assume:bound-grad-hessian} assumes that $x$ is bounded implicitly.
\end{remark}


\begin{theorem}\label{thm:heavy}
Suppose Assumption~\ref{assumption:heavy} holds. Let $\sigma>0$  be the constant such that $\hp \preceq \sigma \hq$.   For any $\delta \in (0,1)$, whenever 
\begin{align}\label{eq:heavy-n-bound}
    n \geq 256\max \Big\{ L_2^2 \sigma^2 \log(2d(c-1)/\delta), \log(1/\delta) \sigma^4 L_1^2 L_3^2\Big\},
\end{align}
with probability at least $1-\delta$, we have 
\begin{align}\label{eq:heavy-risk-bound}
    \frac{3}{8} \frac{(1-\epsilon_p)}{(1+\epsilon_q)^2}  \frac{\Tr(\hq^{-1}\hp)}{n} &\leq \E [L_p(\emptheta)] - L_p(\theta_*)
    \leq \frac{5}{8} \frac{(1+\epsilon_p)}{(1-\epsilon_q)^2}  \frac{\Tr(\hq^{-1}\hp)}{n},
\end{align}
where $\E$ is the expectation over ${\{y_i\sim p(y_i|x_i, \theta_*)\}_{i=1}^n}$, $\epsilon_p$ and $\epsilon_q$ are given by
\begin{align}\label{eq:heavy-epsilon}
   \epsilon_p = 2 \sigma^2 L_1 L_3 \sqrt{\frac{2 + 8\log(1/\delta)}{n}}\, \quad\epsilon_q = 4 \sigma L_2 \sqrt{\frac{\log(2d(c-1)/\delta)}{n}} + 2 \sigma^2 L_1 L_3 \sqrt{\frac{2 + 8\log(1/\delta)}{n}}.
\end{align}
\end{theorem}

\begin{remark}
    For \Cref{thm:heavy}, if \Cref{eq:heavy-n-bound} holds, we can upper bound $\epsilon_p$ and $\epsilon_q$. This results in a simpler upper bound for the excess risk with respect to $p(x)$:
    \begin{align}\label{eq:bound-domain-9-5bound}
        \E[L_p(\theta_n)] - L_p(\theta_*) \leq \frac{9}{5}\frac{\Tr(\hq^{-1}\hp)}{n}.
    \end{align}
    We show this at the end of the proof of \Cref{thm:heavy}.
\end{remark}

\begin{proof}[\textbf{proof of Theorem~\ref{thm:heavy}}]
We deploy the notation of $Q_n(\theta)$ and $\bhn(\theta)$ defined in \Cref{eq:qn,eq:hn} for the ease of notation.
\textit{Throughout the whole proof, we treat parameter as vector}, i.e. $\theta\in \mathbb{R}^{\td}$. Denote the samples drawn from $\pi_q(x,y)$ by $\{ z_i = (x_i, y_i) \stackrel{\mathrm{i.i.d}}{\sim} \pi_q(x,y)\}_{i=1}^n$. Since $\hp \preceq \sigma \hq$, for a vector $v\in \mathbb{R}^{\td}$ we have
\begin{align}\label{eq:vec-norm-relations}
     \| v \|_{\hq^{-1}}& \leq \sqrt{\sigma}\|v\|_{\hp^{-1}}, \qquad
    \| v\|_{\hp} \leq \sqrt{\sigma} \|v\|_{\hq}.
\end{align}

For the ease of notation,  we define norms for a matrix $\A\in\mathbb{R}^{\td \times \td}$ by
\begin{align}
    \|\A \|_P\triangleq\|\hp^{-1/2} \A \hp^{-1/2} \|, \qquad\|\A \|_Q \triangleq \|\hq^{-1/2} \A \hq^{-1/2} \|.
\end{align}
Note that for a matrix symmetric semi-positive definite matrix $\A \in \mathbb{S}_+^{\td}$, 
\begin{align}
    \hq^{-1/2} \A \hq^{-1/2} & = (\hq^{-1/2} \hp^{1/2}) (\hp^{-1/2} \A \hp^{-1/2}) (\hp^{1/2} \hq^{-1/2})\nonumber\\
    &\preceq \sigma \hp^{-1/2} \A \hp^{-1/2}
\end{align}
where the last inequality follows by the fact $\lambda_{\max}(\hq^{-1/2}\hp^{1/2}) = \sqrt{\sigma}$. Thus we have the following relation between these two norms:
\begin{align}\label{eq:mat-norm-relation}
     \|\A \|_Q \leq \sigma  \|\A \|_P.
\end{align}

\textbf{step 1.}
We aim to choose a ball $\mathcal{B}_1(\theta_*)$ centered at $\theta_*$ and $n$ sufficiently large such that for any $\theta \in \mathcal{B}_1(\theta_*)$, $\bhn(\theta)$  approximates  $\hq$ in the spectral sense with high probability.

First, we have by triangle inequality that
\begin{align}
    \| \bhn(\theta) -\hq \|_Q &\leq \|\bhn(\theta) - \bhn(\theta_*) \|_Q + \| \bhn(\theta_*) - \hq \|_Q. \label{eq:s1-1}
\end{align}
To bound the first term in \Cref{eq:s1-1}, we can use  \Cref{assumption:heavy}-\ref{assume:bound-lipschitz}, i.e. if $\theta\in \mathcal{B}(\theta_*)$, then
\begin{align}\label{eq:s1-2}
    \|\bhn(\theta) - \bhn(\theta_*) \|_Q \stackrel{\Cref{eq:mat-norm-relation}}{\leq} \sigma  \|\bhn(\theta) - \bhn(\theta_*) \|_P \leq \sigma L_3 \|\theta - \theta_* \|_{\hp}.
\end{align}

Now we consider the second term on the right hand side of  \Cref{eq:s1-1}. Let $\X_i = \hp^{-1/2} \big( \nabla^2 \ell_{z_i}(\theta_*) - \hq \big) \hp^{-1/2}$ for each $i\in [n]$ and $\b S = \frac{1}{n} \sum_{i=1}^n \X_i$. Since $\E[\nabla^2 \ell_{z_i}(\theta_*)] = \nabla^2 L_q(\theta_*) = \hq $, then $\E[\X_i] = 0$. By \Cref{eq:heavy-assume-2}, we have $\| \nabla^2 \ell_{z_i}(\theta_*)\|_P\leq L_2$. Thus  for any $i\in [n]$:
\begin{align}
 \| \X_i\| = \| \nabla^2\ell_{z_i}(\theta_*) - \hq \|_P \leq 2 L_2 ,\nonumber\\
 \|\E(\X_i^2)\| \leq \E\| \X_i^2\| \leq \E \| \X_i \|^2 \leq 4 L_2^2.
\end{align}
Let $\mu = 2L_2$ and $\nu = 4L_2^2$  in  the matrix Bernstein inequality (i.e. \Cref{lm:matrix-Bernstein}), we have with probability at least $1-\delta$, 
\begin{align}
    \|\b  S\| \leq 4 L_2 \sqrt{\frac{\log(2\td/\delta)}{n}} \triangleq\epsilon_1.
\end{align}
Note that $\| \bhn(\theta_*) - \hq\|_P= \| \b S\|$. Then with probability at least $1 - \delta$, 
\begin{align}\label{eq:s1-3}
    \| \bhn(\theta_*) - \hq\|_Q \leq \sigma \| \bhn(\theta_*) - \hp\|_P \leq \sigma \epsilon_1.
\end{align}

Substitute \Cref{eq:s1-2,eq:s1-3}  into \Cref{eq:s1-1}, we can get
\begin{align}\label{eq:s1-4} 
     \| \bhn(\theta) -\hq \|_Q\leq \sigma L_3 \|\theta - \theta_* \|_{\hp} + \sigma \epsilon_1.
\end{align}

Now consider a ball centered at $\theta_*$:
\[
\mathcal{B}_1(\theta_*) \triangleq \{ \theta: \|\theta -\theta_* \|_\hp \leq \frac{1}{4\sigma L_3} \},
\]
 then $\sigma L_3 \|\theta - \theta_* \|_\hq \leq 1/4$ for any $\theta \in \mathcal{B}_1(\theta_*)$. Besides, if we choose $n$ such that
\begin{align}
    n \geq 256 L_2^2\sigma^2 \log (2\td/\delta),
\end{align}
we have 
\begin{align}\label{eq:s1-eps-1}
    \epsilon_1 \leq \frac{1}{4\sigma}.
\end{align}

Substitute \Cref{eq:s1-eps-1} into \Cref{eq:s1-4}, we have
$ \| \bhn(\theta) -\hq \|_Q\leq 1/2$
and thus with probability at least $1-\delta$,
\begin{align}\label{eq:s1-spectral}
    \frac{1}{2}\hq \preceq \bhn(\theta) \preceq \frac{3}{2}\hq.
\end{align}

\textbf{step 2.}
Next we show that when $n$ is large enough, $\emptheta \in  \mathcal{B}_1(\theta_*)$ with high probability. Given $\theta$, by Taylor's expansion there exits $\Tilde{\theta} $ between $\theta$ and $\theta_*$ such that
\begin{align}
    \emp(\theta) = \emp(\theta_*) + \nabla\emp(\theta_*)^\top (\theta - \theta_*) + \frac{1}{2} (\theta - \theta_*)^\top \nabla^2 \emp(\Tilde{\theta})(\theta - \theta_*) . \nonumber
\end{align}
Then for all $\theta \in \mathcal{B}_1(\theta_*)$, 
\begin{align}
    \emp(\theta) - \emp(\theta_*) &= \nabla\emp(\theta_*)^\top (\theta - \theta_*) + \frac{1}{2} \|\theta - \theta_*\|_{\bhn(\Tilde{\theta})}^2 \nonumber\\
    &\stackrel{(a)}{\geq}  \nabla\emp(\theta_*)^\top (\theta - \theta_*) + \frac{1}{4} \|\theta - \theta_*\|_{\hq}^2 \nonumber\\
    &\stackrel{(b)}{\geq} \|\theta - \theta_*\|_{\hq} \bigg(\frac{1}{4} \|\theta - \theta_* \|_{\hq} - \|\nabla\emp(\theta_*) \|_{\hq^{-1}}\bigg) \nonumber\\
    &\stackrel{(c)}{\geq} \|\theta - \theta_*\|_{\hq} \bigg(\frac{1}{4 \sqrt{\sigma}} \|\theta - \theta_* \|_{\hp} - \sqrt{\sigma}\|\nabla\emp(\theta_*) \|_{\hp^{-1}}\bigg)  \label{eq:s2-1}
\end{align}
where (a) follows by \Cref{eq:s1-spectral}, (b) follows by Cauchy-Schwartz inequality, and (c) follows by \Cref{eq:vec-norm-relations}.

Now if we can show for all $\theta \in \partial \mathcal{B}_1\theta_*)$, the right hand side of \Cref{eq:s2-1} is non negative, then $\emptheta \in \mathcal{B}_1(\theta_*)$ because $\emp(\theta)$ is a convex function.  Let $\xi_i = \hp^{-1/2} \nabla \ell_{z_i}(\theta_*)$ and $S = \frac{1}{n}\sum_{i=1}^n \xi_i$. Then $\E[\xi_i] = \hp^{-1/2} \nabla L_p(\theta_*) = 0$ by \Cref{lm:zero-mean-gradient-loss}. By \Cref{assumption:heavy}-\ref{assume:bound-grad-hessian}, for any $i\in[n]$ we have
\begin{align}
    \| \xi_i\| = \| \nabla\ell_{z_i}(\theta_*)\|_{\hp^{-1}} \leq L_1, \nonumber \\
    \E[\|\xi_i \|^2] \leq L_1^2.
\end{align}
Let $\mu = L_1$ and $\nu = L_1^2$ in the vector Bernstein inequality (i.e. \Cref{lm:vector-Bernstein}), with probability at least $1 - \delta$ we have
\begin{align}\label{eq:s2-2}
    \| \nabla \emp(\theta_*)\|_{\hp^{-1}} = \|S\| \leq L_1 \sqrt{\frac{2 + 8 \log(1/\delta)}{n}} \triangleq \epsilon_2.
\end{align}
Now if we choose $n$ such that
\[
    n\geq 256(2 + 8\log(1/\delta) )\sigma^4 L_1^2 L_3^2,
\]
then 
\begin{align}\label{eq:s2-eps-2}
    \epsilon_2 \leq \frac{1}{16L_3\sigma^2}.
\end{align}
Thus for all  $\theta \in \partial \mathcal{B}_1(\theta_*)$, combining \Cref{eq:s2-eps-2,eq:s2-2,eq:s2-1} we have
\begin{align}
    \emp(\theta) - \emp(\theta_*) &\geq \|\theta - \theta_*\|_{\hq} \bigg(\frac{1}{4 \sqrt{\sigma}} \|\theta - \theta_* \|_{\hp} - \sqrt{\sigma}\|\nabla\emp(\theta_*) \|_{\hp^{-1}}\bigg)  \nonumber\\
    &\geq \|\theta - \theta_*\|_{\hq} \bigg(\frac{1}{4 \sqrt{\sigma}}  \frac{1}{4\sigma L_3} - \sqrt{\sigma} \frac{1}{16 \sigma^2 L_3} \bigg) = 0.
\end{align}
Then with probability at least $1 - \delta$,  $\emptheta \in B_1(\theta_*)$.

\textbf{step 3.}
We denote $\Delta \triangleq \emptheta - \theta_*$, then by Taylor's theorem, there exits $\tayloremp$ between $\emptheta$ and $\theta_*$ such that
\begin{align}
    0 = \nabla \emp(\emptheta) = \nabla \emp(\theta_*) + \bhn(\tayloremp) \Delta.
\end{align}
In this step, we get a spectral relation between $\bhn(\tayloremp)$ and $\hq$.

We have ensured that $\bhn(\tayloremp)$ is positive definite in step 1 (by \Cref{eq:s1-spectral}), thus 
\begin{align}\label{eq:Delta}
    \Delta = - \big(\bhn(\tayloremp) \big)^{-1} \nabla \emp(\theta_*),
\end{align}
and with probability at least $1-\delta$ we have
\begin{align}\label{eq:s3-1}
    \| \Delta \|_{\hq}  &=   (\Delta^\top \hq \Delta)^{1/2}  = [\nabla \emp(\theta_*)^\top \big(\bhn(\tayloremp) \big)^{-1}  \hq \big(\bhn(\tayloremp) \big)^{-1}  \nabla \emp(\theta_*) ]^{1/2}\nonumber \\
    &= \bigg[ \bigg( \nabla \emp(\theta_*)^\top \hq^{-1/2} 
 \bigg) \bigg( \hq^{1/2} \big(\bhn(\tayloremp) \big)^{-1}  \hq \big(\bhn(\tayloremp) \big)^{-1}\hq^{1/2}\bigg) \bigg(  \hq^{-1/2}\bhn(\theta_*) \bigg) \bigg] ^{1/2} \nonumber \\
 &\leq \| \hq^{1/2} \big(\bhn(\tayloremp) \big)^{-1}  \hq \big(\bhn(\tayloremp) \big)^{-1}\hq^{1/2}\| ^{1/2} \|\hq^{-1/2} \nabla \emp(\theta_*) \|\nonumber\\
 &\leq \|\hq^{1/2} \big(\bhn(\tayloremp) \big)^{-1}  \hq^{1/2} \|  \| \nabla \emp(\theta_*)\|_{\hq^{-1}}\nonumber \\
 & \stackrel{(a)}{\leq} 2\sqrt{\sigma} \|\nabla \emp (\theta_*) \|_{\hp^{-1}} \nonumber \\
 & \stackrel{(b)}{\leq} 2\sqrt{\sigma} \epsilon_2,
\end{align}
where (a) follows by \Cref{eq:vec-norm-relations} and $1/2\hp \preceq \bhn(\tayloremp) $ from \Cref{eq:s1-spectral} since $\tayloremp \in \mathcal{B}(\theta_*)$, (b) follows by \Cref{eq:s2-2}.

Denote $\widetilde{\Delta}\triangleq \tayloremp - \theta_*$, since $\tayloremp$ lies between $\emptheta$ and $\theta_*$, we have
\begin{align}\label{eq:s3-2}
    \| \widetilde{\Delta}\|_\hq \leq \| \Delta\| _\hq \leq 2 \sqrt{\sigma} \epsilon_2.
\end{align}
Following a similar argument as step 1, we can obtain that
\begin{align}
    \|\bhn(\tayloremp) - \hq\|_Q &\leq \| \bhn(\tayloremp) - \bhn(\theta_*) \|_Q + \|\bhn(\theta_*) - \hq \|_Q\nonumber\\
    &\leq \sigma \|\bhn(\tayloremp) - \bhn(\theta_*)  \|_P + \sigma \epsilon_1\nonumber\\
    &\leq \sigma L_3 \| \widetilde{\Delta}\|_\hp + \sigma \epsilon_1\nonumber\\
    &\stackrel{(a)}{\leq} 2 \sigma^2 L_3 \epsilon_2 + \sigma \epsilon_1 \triangleq \epsilon_q,
\end{align}
where (a) follows by \Cref{eq:s3-2} and the fact that $\| \widetilde{\Delta}\|_\hp \leq \sqrt{\sigma} \|\widetilde{\Delta} \|_\hq$. Note that we can upper bound $\epsilon_q$ by using \Cref{eq:s1-eps-1,eq:s2-eps-2}:
\begin{align}\label{eq:eps-p}
    \epsilon_q = 2 \sigma^2 L_3 \epsilon_2 + \sigma \epsilon_1 \leq \frac{3}{8}.
\end{align}
Thus, with probability at least $1-\delta$, we have
\begin{align}\label{eq:spectral-relation-emp}
    (1-\epsilon_q)\hq \preceq\bhn(\tayloremp) \leq (1+ \epsilon_q) \hq.
\end{align}

\textbf{step 4.} Now we use Taylor's expansion to get bounds for $L_p(\theta_n) - L_p(\theta_*)$. By Taylor's theorem, there exits $\taylorQ $ between $\emptheta$ and $\theta_*$ such that 
\begin{align}\label{eq:s4-1}
   L_p(\emptheta) - L_p(\theta_*) = \frac{1}{2} \|\Delta \|_{\hp(\taylorQ)}^2,
\end{align}
where the first order term vanishes because $\nabla L_p(\theta_*) = 0$ by \Cref{lm:zero-mean-gradient-loss}. 

From the Lipschitz condition \Cref{assumption:heavy}-\ref{assume:bound-lipschitz}, we have 
\begin{align}
    \|\hp(\taylorQ) - \hp \|_P \leq L_3 \| \taylorQ - \theta_* \|_{\hp} \stackrel{(a)}{\leq} 2 \sigma^2 L_3 \epsilon_2 \triangleq \epsilon_p,\nonumber
\end{align}
where inequality (a) follows by 
\begin{align}
    \|\taylorQ - \theta_* \|_{\hp} \leq \|\Delta \|_{\hp} \stackrel{\Cref{eq:vec-norm-relations}}{\leq} \sqrt{\sigma}\|\Delta \|_{\hq}\stackrel{\Cref{eq:s3-1}}{\leq} 2 \sigma^2 \epsilon_2. \nonumber
\end{align}
Note that we can upper bound $\epsilon_p$ by using \Cref{eq:s2-eps-2}:
\begin{align}\label{eq:eps-q}
    \epsilon_p =  2 \sigma^2 L_3 \epsilon_2 \leq \frac{1}{8}.
\end{align}
Thus,
\begin{align}\label{eq:spectral-relation-Q}
    (1-\epsilon_p)\hp \preceq \hp(\taylorQ) \leq (1+ \epsilon_p) \hp.
\end{align}

Define matrices $\mqn$ and $\mpn$ as follows:
\begin{align}
    \mqn &\triangleq \hq^{1/2} \big(\bhn(\tayloremp) \big)^{-1} \hq^{1/2},\nonumber\\
    \mpn &\triangleq \hp^{-1/2} \hp(\taylorQ) \hp^{-1/2}.\nonumber
\end{align}
By \Cref{eq:spectral-relation-emp,eq:spectral-relation-Q}, we have
\begin{align}
    &\lambda_{\max}( \mqn) \leq \frac{1}{1-\epsilon_q}, \qquad \lambda_{\min} (\mqn) \geq \frac{1}{1+\epsilon_q}, \label{eq:eigen-mpn}\\
    &\lambda_{\max}( \mpn) \leq (1+\epsilon_p), \qquad\lambda_{\min} (\mpn) \geq (1-\epsilon_p). \label{eq:eigen-mqn}
\end{align}

Now we can bound the excess risk $L_p(\emptheta) - L_p(\theta_*)$ by using the Taylor expansion in \Cref{eq:s4-1}: 
\begin{align}\label{eq:s5-1}
    L_p(\emptheta) - L_p(\theta_*) &= \frac{1}{2} \Delta^\top \hp(\taylorQ) \Delta \nonumber\\
    & = \frac{1}{2} \Delta^\top \hp^{1/2} \bigg(\hp^{-1/2} \hp(\taylorQ) \hp^{-1/2} \bigg) \hp^{1/2} \Delta\nonumber\\
    & = \frac{1}{2} \Delta^\top \hp^{1/2} \mpn \hp^{1/2}\Delta.
\end{align}
Observe that,
\begin{align}\label{eq:s5-2}
    &\Delta^\top \hp \Delta \nonumber \\
    =& \Delta^\top \bhn(\tayloremp)\hq^{-1/2} \underbrace{\big( \hq^{1/2} \big(\bhn(\tayloremp)\big)^{-1} \hp \big(\bhn(\tayloremp)\big)^{-1} \hq^{1/2} \big)}_{\triangleq \b M}\hq^{-1/2} \bhn(\tayloremp) \Delta,
\end{align}
and
\begin{align}\label{eq:s5-3}
    \b M 
   & = \big( \hq^{1/2} \big(\bhn(\tayloremp)\big)^{-1} \hq^{1/2}\big) \big( \hq^{-1/2} \hp\hq^{-1/2}\big)\big( \hq^{1/2} \bhn(\tayloremp)\big)^{-1} \hq^{1/2}\big)\nonumber\\
   & = \mqn \big( \hq^{-1/2} \hp\hq^{-1/2}\big)\mqn.
\end{align}
Substitute \Cref{eq:s5-3} into \Cref{eq:s5-2}, we have
\begin{align}\label{eq:s5-4}
     \Delta^\top \hp \Delta &=  \big(\Delta^\top \bhn(\tayloremp)\hq^{-1/2}\big)  \mqn \big( \hq^{-1/2}\hp \hq^{-1/2} \big)\mqn \big( \hq^{-1/2} \bhn(\tayloremp) \Delta\big).
\end{align}
Based on the previous steps,  with probability at least $1-\delta$,  we have a lower bound for $L_p(\emptheta) - L_p(\theta_*) $ by \Cref{eq:s5-1}:
\begin{align}\label{eq:s5-lower}
    &L_p(\emptheta) - L_p(\theta_*) \nonumber\\
     \stackrel{}{=} & \frac{1}{2} \Delta^\top \hp^{1/2} \mpn \hp^{1/2}\Delta\nonumber \\
    \geq& \frac{1}{2} \lambda_{\min}(\mpn) \Delta^\top \hp \Delta \nonumber\\
    \stackrel{\eqref{eq:s5-4}}{\geq}&   \frac{1}{2} \lambda_{\min}(\mpn)  \big(\Delta^\top \bhn(\tayloremp)\hq^{-1/2}\big)  \mqn \big( \hq^{-1/2}\hp \hq^{-1/2} \big)\mqn \big( \hq^{-1/2} \bhn(\tayloremp) \Delta\big)\nonumber \\
    \geq & \frac{1}{2} \lambda_{\min}(\mpn) \lambda_{\min}^2 (\mqn) \big(\Delta^\top \bhn(\tayloremp)\hq^{-1} \hp \hq^{-1} \bhn(\tayloremp) \Delta\big)\nonumber \\
    \geq &\frac{1}{2} \frac{(1-\epsilon_p)}{(1+\epsilon_q)^2}  \big\langle \hq^{-1} \hp \hq^{-1} , \nabla \emp(\theta_*) \nabla\emp(\theta_*)^\top \big\rangle,
\end{align}
where the last inequality follows by \Cref{eq:eigen-mpn,eq:eigen-mqn}, and the fact that $\bhn(\tayloremp)\Delta = - \nabla \emp(\theta_*)$ from \Cref{eq:Delta}.

By similar argument, we can get an upper bound:
\begin{align}\label{eq:s5-upper}
     L_p(\emptheta) - L_p(\theta_*) &\leq \frac{1}{2} \lambda_{\max}(\mpn) \lambda_{\max}^2 (\mqn) \big(\Delta^\top \bhn(\tayloremp)\hq^{-1} \hp \hq^{-1} \bhn(\tayloremp) \Delta\big)\nonumber \\
     &\leq \frac{1}{2} \frac{(1+\epsilon_p)}{(1-\epsilon_q)^2}  \big\langle \hq^{-1} \hp \hq^{-1}, \nabla \emp(\theta_*) \nabla\emp(\theta_*)^\top \big\rangle.
\end{align}

Following the same argument as we derive \Cref{eq:sub-s6-expectation} in \Cref{section:sub-thm-full-proof},  given $\{ x_i\}_{i=1}^n$, we have
\begin{align}\label{eq:s5-5}
    \E_{\{y_i\sim p(y_i| x_i,\theta_*)\}_{i=1}^n} [\nabla \emp(\theta_*) \nabla\emp(\theta_*)^\top ] = \frac{1}{n} \bhn(\theta_*).
\end{align}
Now if we take conditional expectation on both sides of \Cref{eq:s5-lower,eq:s5-upper}, we can obtain that
\begin{align}\label{eq:s5-6}
    \frac{1}{2} \frac{(1-\epsilon_p)}{(1+\epsilon_q)^2}  \frac{\big\langle \hq^{-1} \hp\hq^{-1},\bhn(\theta_*) \big\rangle}{n} &\leq \E_{\{y_i\sim p(y_i| x_i,\theta_*)\}_{i=1}^n} [L_p(\emptheta) - L_p(\theta_*)]\nonumber\\
    &\leq \frac{1}{2} \frac{(1+\epsilon_p)}{(1-\epsilon_q)^2}  \frac{\big\langle \hq^{-1} \hp\hq^{-1},\bhn(\theta_*) \big\rangle}{n} .
\end{align}

From the analysis in step 1, we have with probability at least $1-\delta$,
\begin{align}
    \| \bhn(\theta_*) - \hq\|_Q \leq \sigma \epsilon_1 \leq \frac{1}{4},
\end{align}
where the last inequality follows by \Cref{eq:s1-eps-1}. Thus
\begin{align}\label{eq:s5-7}
    \frac{3}{4}\hq \preceq \bhn(\theta_*) \preceq \frac{5}{4} \hq,
\end{align}
and
\begin{align}\label{eq:s5-8}
    \frac{3}{4} \Tr(\hq^{-1}\hp)\leq \big\langle \hq^{-1} \hp\hq^{-1},\bhn(\theta_*) \big\rangle\leq \frac{5}{4} \Tr(\hq^{-1}\hp).
\end{align}
Substitute \Cref{eq:s5-8} into \Cref{eq:s5-6}, we have with probability at least $1-\delta$, 
\begin{align}\label{eq:s5-9}
    \frac{3}{8} \frac{(1-\epsilon_p)}{(1+\epsilon_q)^2}  \frac{\Tr(\hq^{-1}\hp)}{n} &\leq \E [L_p(\emptheta)] - L_p(\theta_*)
    \leq \frac{5}{8} \frac{(1+\epsilon_p)}{(1-\epsilon_q)^2}  \frac{\Tr(\hq^{-1}\hp)}{n},
\end{align}
where $\E$ is the expectation over ${\{y_i\sim p(y_i|x_i,\theta_*)\}_{i=1}^n}$. 

Note that, with the upper bounds given in \Cref{eq:eps-p,eq:eps-q}, we can additionally bound the upper bound of \Cref{eq:s5-9}:
\begin{align}
    \E_[L_p(\emptheta)] - L_p(\theta_*)
    & \leq \frac{5}{8} \frac{(1+\epsilon_p)}{(1-\epsilon_q)^2}  \frac{\Tr(\hq^{-1}\hp)}{n}\nonumber\\
    &\leq\frac{5}{8} \frac{1 + 1/8}{(1-3/8)^2} \frac{\Tr(\hq^{-1}\hp)}{n} \nonumber \\
    &= \frac{9}{5} \frac{\Tr(\hq^{-1}\hp)}{n}.
\end{align}

\end{proof}

\section{Proofs of Section~\ref{sec:method}}\label{appendix:method}

\paragraph{Notation.}

For a positive integer $k$, let $\mathbb{S}^k$ be the cone of  symmetric matrices with dimension $k\times k$, $\mathbb{S}^k_+$ be the cone of  symmetric semi-positive definite matrices with dimension $k\times k$, and $\mathbb{S}^k_{++}$ be the cone of  symmetric positive definite matrices with dimension $k\times k$.

\subsection{Proof of Lemma~\ref{lm:method-f}}
\begin{proof}
\begin{enumerate}[leftmargin=*]
    \item We can verify convexity by considering an arbitrary line, given by $\Z + t \V$, where $ \Z \in \mathbb{S}^{\widetilde{d}}_{++}$ and $\V \in \mathbb{S}^{\widetilde{d}}$. We define $g(t) = f(\Z + t\V)$, where $t$ is restricted to the interval such that $\Z + t \V \in \mathbb{S}^{\widetilde{d}}_{++}$. From covex analysis, it is sufficient for us to prove the convexity of function $g$. We have
    \begin{align}
        g(t)& = \langle (\Z + t\V)^{-1}, \hp(\theta_0)\rangle\nonumber\\
        &= \Tr\big(\Z^{1/2} \hp(\theta_0) \Z^{1/2}\big( \b I + t \Z^{-1/2} \V \Z^{-1/2}\big)^{-1} \big).
    \end{align}
    We can write $\Z^{-1/2} \V \Z^{-1/2}$ in its eigendecomposition form, i.e. $\Z^{-1/2} V \Z^{-1/2} = \Q \bSigma \Q^\top$, where $\bSigma = \text{diag}\{ \lambda_1,\cdots, \lambda_{\widetilde{d}}\}$. Then we have
    \begin{align}
         g(t)& =\Tr\big( \Z^{1/2} \hp(\theta_0) \Z^{1/2} \Q \big(\b I + t \bSigma \big)^{-1} \Q^\top \big)\nonumber\\
         & = \Tr\big(\big(\Q^\top \Z^{1/2} \hp(\theta_0) \Z^{1/2} \Q\big) \big(\b I + t \bSigma \big)^{-1} \big)\nonumber\\
         & = \sum_{i=1}^{\widetilde{d}} \frac{1}{1 + t\lambda_i} \big[ \Q^\top \Z^{1/2} \hp(\theta_0) \Z^{1/2} \Q\big]_{ii},
    \end{align}
    and thus
    \begin{align}\label{eq:pf-f-1}
        g^{\prime \prime}(t)  = \sum_{i=1}^{\widetilde{d}} \frac{2 \lambda_i^2}{(1 + t\lambda_i)^3} \big[ \Q^\top \Z^{1/2} \hp(\theta_0) \Z^{1/2} \Q\big]_{ii}
    \end{align}
    Since $\Z + t \V $ is positive definite, so is $ \b I + t \Z^{-1/2} \V \Z^{-1/2}$. Thus $1 + t \lambda_i >0$ for all $i\in [\widetilde{d}]$. In addition, $\Q^\top \Z^{1/2} \hp(\theta_0) \Z^{1/2} \Q$ is also positive definite, then its diagonals are all positive. Thus $ g(t)^{\prime \prime}\geq 0$ by \Cref{eq:pf-f-1}, we conclude that $g$ is convex, and thus $f$ is convex.
    \item If $\b A \preceq \b B$, then $\b B^{-1} -\b A^{-1}\preceq \b 0$. Thus $\langle \b B^{-1} -\b A^{-1}, \hp(\theta_0)\rangle \leq 0$ since $\hp(\theta_0)$ is positive definite, i.e.
    \begin{align}
        f(\b A) \geq f(\b B).
    \end{align}

    \item Property 3 is trivial to prove.
\end{enumerate}
\end{proof}

\subsection{Solving relaxed problem  by entropic mirror descent}\label{appendix:mtd-mirror-descent}

We present the algorithm for solving relaxed problem \Cref{eq:mtd-obj-relax} using  entropic mirror descent  in \Cref{algo:relax}. Let $z = b \kappa$, then \Cref{eq:mtd-obj-relax} is equivalent to:
\begin{align}
    \kappa_{\diamond} = \argmin_{\substack{\kappa \in \mathbb{R}_{+}^m \\\|\kappa \|_1 = 1}} f(\kappa) \triangleq \big\langle\big(\sum_{i\in[m]} \kappa_i \b H(x_i)\big)^{-1}, \hp(\theta_0) \big\rangle.
\end{align}
Line 5 of the algorithm computes the gradient of $f(\kappa)$:
\begin{align}
    g_i \triangleq \frac{\partial f(\kappa)}{\partial \kappa_i} = - \big\langle \b H(x_i), \bSigma^{-1} \hp(\theta_0) \bSigma^{-1} \rangle,
\end{align}
where $\bSigma = \sum_{i\in[m]} \kappa_i \b H(x_i)$. We present the convergence rate of the algorithm in \Cref{thm:md-convergence}, which is adopted from Theorem 5.1 in \cite{beck-2003}.


\begin{algorithm}[!t]
\caption{\textproc{RelaxSolve}($b$,  $\hp(\theta_0)$, $\{\b H(x_i) \}_{i \in[m]}$)}
\label{algo:relax}
 \hspace*{\algorithmicindent} \textbf{Output:} \nolinebreak
$z_{\diamond}$ 
\begin{algorithmic}[1]
    \State $\kappa = (1/m, 1/m, \cdots, 1/m)\in \mathbb{R}^m$
\For {$t = 1$ to $T$} \COMMENT{$T$ is iteration number}
    \State $\beta_t \gets \mathcal{O}(\sqrt{\frac{\log m}{t}})$
    \State $\bSigma \gets  \sum_{i \in [m]} \kappa_i \b H(x_i)$
    \State $g_i \gets -  \big\langle \b H(x_i), \bSigma^{-1} \hp(\theta_0) \bSigma^{-1}  \big\rangle$, $\forall i \in[m]$
    \State $\kappa_i \gets \kappa_i \exp(-\beta_t g_i)$
    \State $\kappa_i \gets \frac{\kappa_i}{ \sum_{j\in [m]} \kappa_j}$
\EndFor
\State $z_{\diamond} \gets b \kappa$
\end{algorithmic}

\end{algorithm}

\begin{theorem}\label{thm:md-convergence}
Suppose $f: \mathbb{R}^n \supseteq \mathcal{X}\rightarrow \mathbb{R}$ is convex Lipschitz continuous function w.r.t $\|\cdot\|_1$, i.e. $|f(x)- f(y)| \leq L_f\|x-y \|_1$. Consider using entropic mirror descent algorithm with $T$ steps and step size $\eta_t= \frac{1}{L_f}\sqrt{\frac{2\log n}{T}}$, denote solution at step $t$ as $x_t$. Then we have
\begin{align}
    \min_{1\leq t\leq T}f(x_t) - \min_{x\in \mathcal{X}} f(x) \leq L_f\sqrt{\frac{2 \log n}{T}}.
\end{align}
\end{theorem}

\subsection{Proof of Proposition~\ref{prop:mtd-FTRL}}\label{appendix:regre-minimization}

We first introduce the background of the regret minimization problem in \Cref{section:pf-FTRL-background}. Note that in this section, we consider that the loss matrix  $\b F_t$ at each step $t$ can be any symmetric, semi-positive definite matrix (i.e. $\b F_t \in \mathbb{S}^{\td}_+$). This is  more general than the case of   $\b F_t \in \{\tI (x_i)\}_{i=1}^m$ in \Cref{section:rounding}. Then we give the proof of \Cref{prop:mtd-FTRL} in \Cref{section:pf-FTRL-proof}.

\subsubsection{Background of regret minimization}\label{section:pf-FTRL-background}
We introduce a regret minimization  problem in the adversarial linear bandits setting with full information. Consider a game of $b$ rounds. At each round $t\in[b]$:
\begin{itemize}[leftmargin=*]
    \item the player chooses an action $\A_t \in \simplex$, where $\simplex=\{\A\in \mathbb{R}^{\dtilde\times \dtilde}: \A \succeq {\b 0}, \Tr(\A) = 1\}$ 
    \item afterwards, the environment reveals a loss matrix $\F_t \in \mathbb{S}_{+}^{\td}$
    \item the loss $\langle \A_t, \F_t\rangle$ is incurred
\end{itemize}
The goal of the player is to minimize the \textit{regret} over all rounds, which is defined by
 \begin{align}\label{eq:mtd-regret-appendix}
     \text{Regret}(\{\A_t \}_{t=1}^b) \triangleq \sum_{t=1}^b \langle\A_t , \F_t \rangle -  \inf_{\U \in \simplex} \langle \U, \sum_{t=1}^b \F_t \rangle.
 \end{align}
The regret represents the excess loss compared to the loss incurred by a single optimal action $\b U\in\simplex$ in hindsight. In our setting, the loss incurred by a single optimal action is actually the minimum eigenvalue of the summed matrix of the loss matrices. We remark this property in \Cref{lm:mtd-eigen-min}.

\begin{Lemma}\label{lm:mtd-eigen-min}
    For any $\A \in \mathbb{S}_{+}^{\dtilde}$, $\lambda_{\min}(\A) = \inf_{\U\in \simplex} \langle \U, \A \rangle$.
\end{Lemma}
\begin{proof}
    Since $\A \in \mathbb{S}_{+}^{\dtilde}$, we have eigendecomposition $\A = \V\bLambda \V^\top$, where $\bLambda=\text{diag}\{\lambda_1,\cdots, \lambda_{\dtilde}\}$. Assume that $\lambda_1 \geq \cdots\geq \lambda_{\dtilde} \geq 0$ and  $\b v_i$ is the eigenvector asscoiated with eigenvalue $\lambda_i$ for $i\in[\dtilde]$.

    We first show $\lambda_{\min}(\A) \geq \inf_{\U\in \simplex} \langle \U, \A \rangle$. Let $\B = \b  v_{\dtilde}\b  v_{\dtilde} ^\top$, then $\B \succeq \b 0$ and $\Tr(\B)=1$, i.e. $\B\in \simplex$. Thus
    \begin{align}\label{eq:eigen-min-1}
        \inf_{\U\in \simplex} \langle \U, \A \rangle \leq \langle \B, \A \rangle = \b v_{\dtilde} ^\top \V \bLambda \V^\top \b  v_{\dtilde} = \lambda_{\dtilde} = \lambda_{\min}(\A). 
    \end{align}

    On the other hand, for any $\U \in \simplex $, we have
    \begin{align}
        \langle \U,\A\rangle &= \langle \U, \sum_{i\in[\dtilde]} \lambda_i \b  v_i \b 
 v_i^\top \rangle = \sum_{i\in[\dtilde]} \lambda_i \b 
 v_i^\top \U\b  v_i \nonumber\\
        &\geq \lambda_{\dtilde}  \sum_{i\in[\dtilde]} \b v_i^\top \U \b v_i  = \lambda_{\dtilde}   \langle \U, \V \V^\top \rangle = \lambda_{\dtilde}  \Tr(\U) =\lambda_{\dtilde}.\label{eq:eigen-min-2}
    \end{align}
    Since \Cref{eq:eigen-min-2} holds for any $\U \in \simplex$, then
\begin{align}\label{eq:eigen-min-3}
    \lambda_{\min}(\A) \leq \inf_{\U\in \simplex} \langle \U, \A \rangle.
\end{align}
 Combining \Cref{eq:eigen-min-1} and  \Cref{eq:eigen-min-3}, we can get $ \lambda_{\min}(\A) =\inf_{\U\in \simplex} \langle \U, \A \rangle$.
     
\end{proof}

\paragraph{Follow-The-Regularized-Leader (FTRL).}
 FTRL algorithm chooses action $\A_t$ at the beginning of each round based on the previous loss matrices $\{\F_l \}_{l=1}^{t-1}$. In particular, for a given regularizer $w(\cdot)$ and learning rate $\eta >0$.
\begin{align}\label{eq:mtd-At-appendix}
     \A_1 = \argmin_{\A \in \simplex} w(\A) , \qquad   \A_{t} = \argmin_{\A \in \simplex} \bigg\{ \eta \sum_{l=1}^{t-1} \langle \A,\F_l \rangle + w(\A)\bigg\}\quad  (t\geq 2).
\end{align}
We deploy the $\ell_{1/2}$-regularizer introduced by \cite{Allen-2017}:$  w(\A) = -2\Tr(\A^{1/2})$. Under such a regularizer, we can derive the closed form for $\A_t$, i.e. \Cref{eq:mtd-At-closed}.

\subsubsection{Proof of Proposition~\ref{prop:mtd-FTRL}}\label{section:pf-FTRL-proof}
\begin{proof}
    By Theorem 28.4 in \cite{Lattimore-2020}, we have an upper bound for regret as following:
    \begin{align}\label{eq:FTRL-1}
       \mathrm{Regret}(\{\A_t \}_{t=1}^b) \triangleq\sum_{t=1}^b \langle\A_t , \F_t \rangle -  \inf_{\U \in \simplex} \langle \U, \sum_{t=1}^b \F_t \rangle \leq \frac{\mathrm{diam}_w(\simplex)}{\eta} + \frac{1}{\eta} \sum_{t=1}^b D_w(\A_t, \Aint_{t+1}),
    \end{align}
where $\mathrm{diam}_w(\simplex)\triangleq \max_{\A ,\B \in \simplex}w(\A) - w(\B)$ is the diameter of $\simplex$ with respect to $w$ , $D_w$ is $w$-induced Bregman divergence,  and $\Aint_{t+1}$ is defined by
\begin{align}\label{eq:FTRL-2}
    \Aint_{t+1} = \argmin_{\A \succeq {\b 0}} \big\{ \eta \langle\A, \F_{t} \rangle + D_w(\A, \A_{t})\big\}.
\end{align}
Since the regularizer $w(\A) = -2 \Tr(\A^{1/2})$ for any $\A\succeq \b 0$, $w(\A)$ is differentiable and it has gradient $\nabla w(\A) = -\A^{-1/2}$. By definition of Bregman divergence, we have for any $\A, \B \succeq \b 0$:
\begin{align}\label{eq:FTRL-3}
    D_w(\A, \B) &= w(\A) - w(\B) - \langle\A- \B, \nabla w(\B) \rangle\nonumber\\
    &= -2 \Tr(\A^{1/2} + 2\Tr(\B^{1/2}) +\langle \A-\B, \B^{-1/2} \rangle\nonumber\\
    &= \langle \A, \B^{-1/2}\rangle + \Tr(\B^{1/2}) - 2 \Tr(\A^{1/2}).
\end{align}

Substitute \Cref{eq:FTRL-3} into \eqref{eq:FTRL-2}, we can get 
\[
\Aint_{t+1} = \argmin_{\A \succeq {\b 0}} \big\{ \eta \langle\A, \F_{t} \rangle + \langle \A, \A_t^{-1/2}\rangle + \Tr(\A_t^{1/2}) - 2 \Tr(\A^{1/2}) \big\} \triangleq g(\A).
\]
By the first order optimality condition of convex optimization, we have
\[\eta \F_t + \A_t^{-1/2} - \Aint_{t+1}^{-1/2} = 0,\]
and thus $\Aint_{t+1} = (\A_t^{-1/2} + \eta\F_t)^{-2}$. Therefore, by \Cref{eq:FTRL-3}
\begin{align}\label{eq:FTRL-4}
    D_w(\A_t, \Aint_{t+1}) &= \langle \A_t, \Aint_{t+1}^{-1/2} \rangle + \Tr(\Aint_{t+1}^{1/2}) -2 \Tr(\A_t^{1/2})\nonumber\\
    &=\langle\A_t, \A_t^{-1/2}+ \eta \F_t\rangle + \Tr[(\A_t^{-1/2}+ \eta \F_t)^{-1}] -2 \Tr(\A_t^{1/2})\nonumber\\
    & = \langle \A_t, \eta\F_t\rangle + \Tr[(\A_t^{-1/2}+ \eta \F_t)^{-1} - \A_t^{1/2}]. 
\end{align}
Substitute \Cref{eq:FTRL-4} into \Cref{eq:FTRL-1}, we  can get
\begin{align}\label{eq:FTRL-5}
   \lambda_{\min} (\sum_{t=1}^b \F_t) \stackrel{(a)}{=}\inf_{\U \in \simplex}\langle \U, \sum_{t=1}^b \F_t\rangle 
      &\geq -\frac{\text{diam}_w(\simplex)}{\eta} +  \frac{1}{\eta} \sum_{t=1}^b \Tr[\A_t^{1/2} - (\A_t^{-1/2}+ \eta \F_t)^{-1}]\nonumber\\
      &\stackrel{(b)}{\geq}  -\frac{2 \sqrt{\dtilde}}{\eta} +  \frac{1}{\eta} \sum_{t=1}^b \Tr[\A_t^{1/2} - (\A_t^{-1/2}+ \eta\F_t)^{-1}],
\end{align}
where equality (a) follows by \Cref{lm:mtd-eigen-min} and inequality (b) follows by the fact that $\text{diam}_w(\simplex) \leq 2 \sqrt{\dtilde}$.

Since \Cref{eq:FTRL-5} holds for any $\F_t \in \mathbb{S}_+^{\td}$, then let $\F_t \in \{\tI(x_i)\}_{i\in[m]}$ and \Cref{eq:mtd-FTRL-bound} is proved.
\end{proof}

\subsection{Proof of Proposition~\ref{prop:mtd-selection}}\label{append:mtd-selection-prop}

In \Cref{section:support-prop-selection}, we present some key inequalities that we need for the proof. In \Cref{section:pf-selection}, we present the full proof of \Cref{prop:mtd-selection}. It is worth noting that a similar property to Proposition \ref{prop:mtd-selection} is proven in \cite{Allen-2017}. However, in their setting, the loss matrices are rank-1 matrices, specifically of the form $\widetilde{x}_i \widetilde{x}_i^\top$, where $\widetilde{x}_i$ is a vector. On the other hand, in our setting, the loss matrices are transformed Fisher information matrices  (i.e. $\tI(x_i)$, as defined in Equation \ref{eq:mtd-tIx}). This distinction significantly complicates the derivation of a general result such as \Cref{eq:mtd-trace-bound} in Proposition \ref{prop:mtd-selection}. The proof is by no means trivial. We remark that we do \textit{not} assume special structure on points from unlabeled pool $U = \{x_i\}_{i\in[m]}$ and the ground truth parameter $\theta_*$ in our proof to \Cref{prop:mtd-selection}.

\subsubsection{Supporting Lemmas}\label{section:support-prop-selection}
\begin{Lemma}\label{lm:inequ-1}
    For any $i\in [m]$, $a_i > 0 $, $b_i >0$, $\pi_i \geq 0$, then $\max_{i \in [m]} \frac{a_i}{b_i} \geq \frac{\sum_{i \in [m] }\pi_i a_i}{\sum_{i \in [m] }\pi_i b_i}$.
\end{Lemma}
\begin{proof}
    We can use induction to prove the inequality. If $n=2$, without loss of generality, we can assume $a_1/b_1 \geq a_2/b_2$, then
    \begin{align}
        a_1 b_2 &\geq a_2 b_1 \nonumber\\
        \pi_1 a_1 b_1  + \pi_2 a_1 b_2 &\geq         \pi_1 a_1 b_1  + \pi_2 a_2 b_1\nonumber
    \end{align}
and 
    \begin{align}
        \max\{\frac{a_1}{b_1}, \frac{a_2}{b_2}\} = \frac{a_1}{b_1} \geq \frac{\pi_1 a_1 + \pi_2 a_2}{\pi_1 b_1 + \pi_2 b_2}.\nonumber
    \end{align}
Suppose the inequality is satisfied when $n=m-1$, i.e.
\begin{align}
    \max_{i\in[m-1]} \frac{a_i}{b_i} \geq \frac{\sum_{i\in [m-1]} \pi_i a_i}{\sum_{i\in [m-1]} \pi_i b_i}.
\end{align}
When $n=m$, 
\begin{align}
    \max_{i\in [m]} \frac{a_i}{b_i} =  \max\big\{ \max_{i\in[m-1]}\frac{a_i}{b_i}  , \frac{a_m}{b_m} \big\} &\geq \max \bigg\{ \frac{\sum_{i\in [m-1]} \pi_i a_i}{\sum_{i\in [m-1]} \pi_i b_i}, \frac{a_m}{b_m}\bigg\}\nonumber\\
    &\geq \frac{\sum_{i\in [m]} \pi_i a_i}{\sum_{i\in [m]} \pi_i b_i} .\nonumber
\end{align}

The last inequality follows by the previous derivation when $n=2$. Thus by induction, the inequality is proved for any positive integer n.
\end{proof}

\begin{Lemma}\label{lm:inequ-2}
    For any $i\in [m]$, $a_i \geq 0 $, $b_i \geq0$, then $\sum_{i \in [m]} \frac{a_i}{1 + b_i} \geq \frac{\sum_{i \in [m] } a_i}{1 + \sum_{i \in [m]}b_i}$.
\end{Lemma}
\begin{proof}
    We can use induction to prove this inequality. When $n=2$, 
    \begin{align}
        &[a_1(1+b_2) + a_2 (1+b_1)] (1+b_1+b_2) \nonumber \\
        &= a_1(1+b_2)(1+b_1) + a_1 b_2(1+b_2) + a_2(1+b_1)(1+b_2) + a_2 b_1 (1+b_1) \nonumber \\
        &= (a_1+a_2) (1+b_1)(1+b_2) + a_1 b_2(1+b_2)+ a_2 b_1 (1+b_1) \nonumber \\
        & \geq (a_1+a_2) (1+b_1)(1+b_2). \label{eq:n=2}
    \end{align}
    Divide $(1+b_1) (1+b_2) (1+b_1+b_2)$ on both sides of \Cref{eq:n=2}, we can get 
    \begin{align}
        \frac{a_1}{1+b_1} + \frac{a_2}{1+b_2} &= \frac{[a_1(1+b_2) + a_2 (1+b_1)] (1+b_1+b_2) }{(1+b_1) (1+b_2) (1+b_1+b_2)}\nonumber\\
        &\stackrel{\Cref{eq:n=2}}{\geq} \frac{(a_1+a_2) (1+b_1)(1+b_2)}{(1+b_1) (1+b_2) (1+b_1+b_2)} = \frac{a_1 + a_2}{1+b_1+b_2}.  \label{eq:int-1}
    \end{align}
Suppose the inequality is satisfied when $n=m-1$, i.e.
\begin{align}\label{eq:int-2}
    \sum_{i \in m-1} \frac{a_i}{1 + b_i} \geq \frac{\sum_{i \in [m-1] } a_i}{1 + \sum_{i \in [m-1]}b_i}.
\end{align}
When $n=m$, 
\begin{align}
    \sum_{i\in[m]} \frac{a_i}{1 + b_i} = \sum_{i\in[m-1]} \frac{a_i}{1+ b_i} + \frac{a_m}{1+b_m} &\stackrel{\Cref{eq:int-2}}{\geq}  \frac{\sum_{i\in [m-1]} a_i}{1+ \sum_{i\in [m-1]} b_i} + \frac{a_m}{1+b_m} \nonumber \\
    &\stackrel{\Cref{eq:int-1}}{\geq}  \frac{\sum_{i\in [m]} a_i}{ 1 + \sum_{i\in[m]} b_i}.
\end{align}
\end{proof}

\begin{Lemma}\label{lm:inequ-3}
    For any matrices $\A, \B\in \mathbb{S}_{+}^p$, we have
    \begin{align}
        \langle (\b I + \B)^{-1}, \A\rangle\geq \frac{\Tr(\A)}{1 + \Tr(\B)}.
    \end{align}
\begin{proof}
Denote eigenvalues of matrix $\A$ as $\alpha_1\geq \alpha_2 \geq \cdots \geq \alpha_p\geq 0$ and eigenvalues of matrix $\B$ as $\beta_1 \geq \beta_2 \geq \cdots \geq \beta_p \geq 0$. Then eigenvalues of $(\b I+\B)^{-1}$ are $0 \leq 1+\beta_1)^{-1} \leq (1+\beta_2)^{-1} \leq \cdots \leq (1+\beta_p)^{-1}$. Thus we have
\begin{align}
    \langle (\b I + \B)^{-1}, \A\rangle &\stackrel{(a)}{\geq}  \sum_{i=1}^p \frac{\alpha_i}{1+\beta_i}\nonumber\\
    &\stackrel{(b)}{\geq}  \frac{\sum_{i=1}^p \alpha_i}{1+\sum_{i=1}^p \beta_i} = \frac{\Tr(\A)}{1+ \Tr(\B)},
\end{align}
where inequality (a) follows by the lower bound of Von Neumann’s trace inequality \cite{Ruhe-1970}, inequality (b) follows by Lemma~\ref{lm:inequ-2}.

    \end{proof}
\end{Lemma}


\subsubsection{Proof of Proposition~\ref{prop:mtd-selection}}\label{section:pf-selection}

\begin{proof} 
Recall that in \Cref{section:rounding}, we define $\B_t$ by
\begin{align}\label{eq:mtd-Bt}
    \B_t^{-1/2} = \A_t^{-1/2} + \eta \tD,
\end{align}
where $\tD = (\bSigma_{\diamond})^{-1/2} \D(\bSigma_{\diamond})^{-1/2}$. In addition, we have
\begin{align}\label{eq:mtd-identity}
  \b I_{\td} \stackrel{\Cref{eq:mtd-tIx}}{=} \sum_{i\in[m]} z_{\diamond, i} \tI(x_i) \stackrel{\Cref{eq:mtd-tIx-new}}{=} \sum_{i\in [m]} z_{\diamond, i} \tD + \sum_{i\in [m]} z_{\diamond, i} \tP_i \tP_i^\top = b \tD +\sum_{i\in [m]} z_{\diamond, i} \tP_i \tP_i^\top.
\end{align}

\textbf{step 1.} We first decompose $\frac{1}{\eta}\Tr[\A_t^{1/2} -(\A_t^{-1/2} + \eta \tI(x_i))^{-1}]$ for any $i\in[m]$ into the sum of two inner products between matrices. By Woodbury's matrix identity, we have 
\begin{align}\label{eq:pf-key-1}
     (\A_t^{-1/2} + \eta \tI(x_i))^{-1} &=   (\B_t^{-1/2} + \eta \tP_i \tP_i^\top)^{-1} \nonumber\\
     &=\B_t^{1/2} -\eta \B_t^{1/2} \tP_i (\b I + \eta \tP_i ^\top \B_t^{1/2} \tP_i)^{-1} \tP_i^\top \B_t^{1/2}.
\end{align}
Thus 
\begin{align}\label{eq:pf-key-2}
    &\frac{1}{\eta}\Tr [\A_t^{1/2} -(\A_t^{-1/2} + \eta \tI(x_i))^{-1}]\nonumber
    \\
    =& \frac{1}{\eta}\Tr(\A_t^{1/2} -\B_t^{1/2} ) + \Big\langle (\b I + \eta \tP_i^\top \B_t^{1/2} \tP_i)^{-1}, \tP_i^\top \B_t \tP_i \Big\rangle.
\end{align}
 We apply Woodbury's matrix identity to $\B_t^{1/2}$ in \Cref{eq:mtd-Bt}, then
\begin{align}\label{eq:pf-key-3}
    \B_t^{1/2} &= (\A_t^{-1/2} + \eta(\bSigma_{\diamond})^{-1/2} \D(\bSigma_{\diamond})^{-1/2} )^{-1}\nonumber\\
    &= \A_t^{1/2} - \eta \A_t^{1/2} \underbrace{(\bSigma_{\diamond})^{-1/2} \Big[ \D^{-1} + \eta (\bSigma_{\diamond})^{-1/2} \A_t^{1/2} (\bSigma_{\diamond})^{-1/2}\Big]^{-1} (\bSigma_{\diamond})^{-1/2}}_{\triangleq \b E} \A_t^{1/2} .
\end{align}
Thus
\begin{align}\label{eq:pf-key-4}
   &\frac{1}{\eta}\Tr(\A_t^{1/2} -\B_t^{1/2} )\nonumber\\
   = &\Big\langle\Big(\D^{-1} + \eta (\bSigma_{\diamond})^{-1/2} \A_t^{1/2} (\bSigma_{\diamond})^{-1/2}\Big)^{-1} ,  (\bSigma_{\diamond})^{-1/2} \A_t (\bSigma_{\diamond})^{-1/2}\Big\rangle\nonumber\\
   =& \Big\langle \D^{1/2}\Big(\b I + \eta  \D^{1/2}(\bSigma_{\diamond})^{-1/2} \A_t^{1/2} (\bSigma_{\diamond})^{-1/2}  \D^{1/2}\Big)^{-1}  \D^{1/2},  (\bSigma_{\diamond})^{-1/2} \A_t (\bSigma_{\diamond})^{-1/2}\Big\rangle\nonumber\\
    =& \Big\langle \Big(\b I + \eta  \D^{1/2}(\bSigma_{\diamond})^{-1/2} \A_t^{1/2} (\bSigma_{\diamond})^{-1/2}  \D^{1/2}\Big)^{-1}   ,\D^{1/2}(\bSigma_{\diamond})^{-1/2} \A_t (\bSigma_{\diamond})^{-1/2} \D^{1/2} \Big\rangle.
\end{align}
Substitute \Cref{eq:pf-key-4} into \Cref{eq:pf-key-2}, we can get
\begin{align}\label{eq:pf-key-5} 
    &\frac{1}{\eta}\Tr [\A_t^{1/2} -(\A_t^{-1/2} + \eta \tI(x_i))^{-1}] \nonumber\\
    =& \Big\langle \Big(\b I + \eta  \D^{1/2}(\bSigma_{\diamond})^{-1/2} \A_t^{1/2} (\bSigma_{\diamond})^{-1/2}  \D^{1/2}\Big)^{-1}   ,\D^{1/2}(\bSigma_{\diamond})^{-1/2} \A_t (\bSigma_{\diamond})^{-1/2}  \D^{1/2}\Big\rangle\nonumber\\
    +&  \Big\langle (\b I + \eta \tP_i^\top \B_t^{1/2} \tP_i)^{-1}, \tP_i^\top \B_t \tP_i \Big\rangle.
\end{align}

\textbf{step 2.}  Now we intend to find a lower bound for $\max_{i\in[m]}\frac{1}{\eta}\Tr[\A_t^{1/2} -(\A_t^{-1/2} + \eta \tI(x_i))^{-1}]$ using \Cref{eq:pf-key-5}. For the first inner product on the right hand side of \Cref{eq:pf-key-5}, we can apply \Cref{lm:inequ-3}:
\begin{align}\label{eq:pf-key-6}
    &\Big\langle \Big(\b I + \eta  \D^{1/2}(\bSigma_{\diamond})^{-1/2} \A_t^{1/2} (\bSigma_{\diamond})^{-1/2}  \Big)^{-1}   ,\D^{1/2}(\bSigma_{\diamond})^{-1/2} \A_t (\bSigma_{\diamond})^{-1/2}  \D^{1/2}\Big\rangle\nonumber\\
    \geq & \frac{\Tr(\D^{1/2}(\bSigma_{\diamond})^{-1/2} \A_t (\bSigma_{\diamond})^{-1/2} \D^{1/2})}{1 + \eta \Tr(\D^{1/2}(\bSigma_{\diamond})^{-1/2} \A_t^{1/2} (\bSigma_{\diamond})^{-1/2} \D^{1/2} )}\nonumber\\
    = &\frac{\langle \A_t, \tD\rangle}{1 + \eta \langle \A_t^{1/2} ,\tD\rangle}.
\end{align}
Similarly, applying Lemma~\ref{lm:inequ-3} to the second term on the right hand side of \eqref{eq:pf-key-5}, we can get
\begin{align}\label{eq:pf-key-7}
    \Big\langle (\b I + \eta \tP_i^\top \B_t^{1/2} \tP_i)^{-1}, \tP_i^\top \B_t \tP_i \Big\rangle \geq \frac{\Tr(\tP_i^\top \B_t \tP_i)}{1 + \eta \Tr(\tP_i^\top \B_t^{1/2} \tP_i)} = \frac{\langle \B_t,\tP_i \tP_i^\top\rangle}{1+ \eta \langle\B_t^{1/2}, \tP_i \tP_i^\top \rangle}.
\end{align}
Substitute \Cref{eq:pf-key-6} and \Cref{eq:pf-key-7} into \Cref{eq:pf-key-5} and apply \Cref{lm:inequ-2}, we can get
\begin{align}\label{eq:pf-key-8}
    \frac{1}{\eta}\Tr [\A_t^{1/2} -(\A_t^{-1/2} + \eta \tI(x_i))^{-1}] &\geq \frac{\langle \A_t, \tD\rangle}{1 + \eta \langle \A_t^{1/2} ,\tD\rangle}+ \frac{\langle \B_t,\tP_i \tP_i^\top\rangle}{1+ \eta \langle\B_t^{1/2}, \tP_i \tP_i^\top \rangle}\nonumber\\
    & \geq \frac{\langle \A_t, \tD\rangle +\langle \B_t,\tP_i \tP_i^\top\rangle }{1 + \eta[\langle \A_t^{1/2} ,\tD\rangle +  \langle\B_t^{1/2}, \tP_i \tP_i^\top \rangle ]}.
\end{align}
Now by \Cref{lm:inequ-1} and \Cref{eq:pf-key-8}:
\begin{align}\label{eq:pf-key-9}
    &\max_{i\in[m]}\frac{1}{\eta}\Tr [\A_t^{1/2} -(\A_t^{-1/2} + \eta \tI(x_i))^{-1}] \geq \max_{i\in[m]} \frac{\langle \A_t, \tD\rangle +\langle \B_t,\tP_i \tP_i^\top\rangle }{1 + \eta[\langle \A_t^{1/2} ,\tD\rangle +  \langle\B_t^{1/2}, \tP_i \tP_i^\top \rangle ]} \nonumber\\
    \geq& \frac{\sum_{i \in[m]}z_{\diamond, i}  \langle \A_t,\tD\rangle +\sum_{i \in[m]}z_{\diamond, i}\langle \B_t, \tP_i \tP_i^\top\rangle }{\sum_{i \in[m]}z_{\diamond, i} + \eta[\sum_{i \in[m]}z_{\diamond, i} \langle \A_t^{1/2},\tD\rangle + \sum_{i \in[m]}z_{\diamond, i} \langle \B_t^{1/2}, \tP_i \tP_i^\top\rangle]}\nonumber\\
    =& \frac{\langle \A_t, b\tD\rangle + \langle \B_t, \b I- b\tD\rangle}{b + \eta [\langle \A_t^{1/2}, b\tD\rangle + \langle \B_t^{1/2}, \b I- b\tD\rangle]},
\end{align}
where the last equality follows by \Cref{eq:mtd-identity} and the fact that $\sum_{i\in[m]} z_{\diamond, i} = b$.
    
\textbf{step 3.} In this step, we will show that the numerator of  \Cref{eq:pf-key-9} is lower bounded by $1-\eta/2b$. First note that we have derived that $\B_t^{1/2} = \A_t^{1/2} - \eta  \A_t^{1/2} \b E  \A_t^{1/2}$ in \Cref{eq:pf-key-3}. Then
\begin{align}\label{eq:pf-key-10}
     \B_t &= (\A_t^{1/2} - \eta  \A_t^{1/2} \b E  \A_t^{1/2}) ^2\nonumber\\
     &= \A_t - \underbrace{(\eta \A_t \b E  \A_t^{1/2} + \eta \A_t^{1/2} \b E  \A_t - \eta^2 \A_t^{1/2} \b E   \A_t \b E  \A_t^{1/2})}_{\triangleq\b G} = \A_t - \b G.
\end{align}
Substitute this into the numerator of \eqref{eq:pf-key-9}, we have
\begin{align}\label{eq:pf-key-11}
    \langle \A_t, b\tD\rangle + \langle \B_t, \b I- b\tD\rangle &= \langle \A_t, b\tD\rangle + \langle\A_t - \b G, \b I -b\tD \rangle\nonumber\\
    &= \Tr(\A_t) - \langle \b G, \b I -b\tD\rangle\nonumber\\
    & = 1 - \langle \b G, \b I-b \tD\rangle,
\end{align}
where the last equality follows by $\Tr(\A_t) = 1$. Now we intend to find an upper bound for $\langle \b G, \b I - b\tD \rangle$. First note that since $\A_t^{1/2}\b E   \A_t \b E  \A_t^{1/2}\succeq \b 0 $, by the definition of $\b G$ in \Cref{eq:pf-key-10} we have
\begin{align}\label{eq:pf-key-12}
    \b G \preceq \eta \A_t \b E  \A_t^{1/2} + \eta \A_t^{1/2} \b E  \A_t.
\end{align}
Recall the definition of $\b E$ in \Cref{eq:pf-key-3}, we claim that $\b E \preceq \tD$. Indeed, since $(\bSigma_{\diamond})^{-1/2} \A_t^{1/2} (\bSigma_{\diamond})^{-1/2}$ is positive definite, we have 
\[
\D^{-1} + \eta (\bSigma_{\diamond})^{-1/2} \A_t^{1/2} (\bSigma_{\diamond})^{-1/2} \succeq \D^{-1},
\]
 Thus $\Big[ \D^{-1} + \eta (\bSigma_{\diamond})^{-1/2} \A_t^{1/2} (\bSigma_{\diamond})^{-1/2}\Big]^{-1} \preceq \D$ and therefore, 
\begin{align}\label{eq:pf-key-13}
    \b E\triangleq (\bSigma_{\diamond})^{-1/2}\Big[ \D^{-1} + \eta (\bSigma_{\diamond})^{-1/2} \A_t^{1/2} (\bSigma_{\diamond})^{-1/2}\Big]^{-1}  (\bSigma_{\diamond})^{-1/2} \preceq (\bSigma_{\diamond})^{-1/2}\D (\bSigma_{\diamond})^{-1/2}= \tD.
\end{align}
Now we have
\begin{align}\label{eq:pf-key-14}
    \langle \b G, \b I-b\tD\rangle &\stackrel{\Cref{eq:pf-key-12}}{\leq} \eta \langle \A_t \b E  \A_t^{1/2} +  \A_t^{1/2} \b E  \A_t , \b I-b\tD\rangle \nonumber\\
    &=\eta \langle \b E, \A_t^{1/2} (\b I-b\tD) \A_t\rangle + \eta \langle\b E, \A_t(\b I-b\tD) \A_t^{1/2} \rangle\nonumber\\
    &\stackrel{\Cref{eq:pf-key-13}}{\leq} \eta \langle \tD, \A_t^{1/2} (\b I-b\tD) \A_t + \A_t(\b I-b\tD) \A_t^{1/2}\rangle\nonumber\\
    & = 2 \eta \Tr(\A_t^{3/2} \tD) - 2\eta b\Tr(\A_t^{1/2} \tD \A_t \tD) \triangleq h(\tD),
\end{align}
where we define function $h:\mathbb{S}_{+}^{\dtilde}\rightarrow \mathbb{R}$. By \Cref{eq:mtd-identity}, $b \tD \preceq \b I$ and thus the domain of function $h$ is $\mathrm{dom}h = \{ \tD \in \mathbb{S}_{+}^{\dtilde} : \tD \preceq \frac{1}{b }\b I\}$.

We intend to find an upper bound for $h(\tD)$. First we prove that $h(\tD)$ is a concave function. We can verify its concavity by considering an arbitrary line, given by $\Z+t\V$, where $\Z, \V \in\mathbb{S}_{+}^{\td}$. Define $g(t): = h(\Z+t\V)$, where $t$ is restricted to the interval such that $\Z + t\V \in \mathrm{dom}h$. By convex analysis theory, it is sufficient to prove the concavity of function $g$. Note that
\begin{align}
    g(t) &= 2\eta \Tr[ \A_t^{3/2}(\Z+t\V)] - 2\eta b\Tr[\A_t^{1/2} (\Z +t\V) \A_t (\Z+t\V)]\nonumber\\
    &= -2\eta b t^2 \Tr(\A_t^{1/2} \V \A_t \V) + 2\eta t\Tr( \A_t^{3/2} \V)  \nonumber\\
    & -2\eta b t\Tr(\A_t^{1/2}\V \A_t \Z+\A_t^{1/2}\Z\A_t \V) + 2\eta \Tr(\Z \A_t^{3/2}) -2\eta b\Tr(\A_t^{1/2} \Z \A_t \Z).
\end{align}
Thus  $g^{\prime \prime}(t) = -4\eta b\Tr(\A_t^{1/2} \V \A_t \V) $ and $g^{\prime \prime}(t) \leq 0$ because $\A_t^{1/2} \V \A_t \V \succeq \b 0$. Therefore $g(t)$ is concave and so is $h(\tD)$. Now consider the gradient of  $h(\tD)$:
\begin{align}\label{eq:pf-key-15}
    \nabla h(\tD) = 2 \eta \A_t^{3/2} - 4\eta b\A_t^{1/2} \tD \A_t.
\end{align}
Let $\nabla h(\tD)=0$, we can get $\tD = \frac{1}{2b}\b I \in \text{dom}h$. Thus 
\begin{align}\label{eq:pf-key-16}
    \sup_{\tD \in \text{dom}h} h(\tD) = h\Big(\frac{1}{2b}\b I\Big)=\frac{\eta}{b}\Tr(\A_t^{3/2})- \frac{\eta}{2b}\Tr(\A_t^{3/2}) = \frac{\eta}{2b}\Tr(\A_t^{3/2}) \leq \frac{\eta}{2b},
\end{align}
where the last inequality follows by the fact that all eigenvalues of $\A_t$ lie in $[0,1]$ and $\Tr(\A_t) =1$.

Combining \Cref{eq:pf-key-11} , \Cref{eq:pf-key-14} and \Cref{eq:pf-key-16}, we can conclude that
\begin{align}\label{eq:pf-key-17}
    \langle \A_t, b\tD\rangle + \langle \B_t, \b I- b\tD\rangle \geq 1-\frac{\eta}{2b}.
\end{align}

\textbf{step 4.} Now we derive an upper bound for the denominator of the right hand side of \Cref{eq:pf-key-9}. By \Cref{eq:pf-key-3}, we have
\begin{align}\label{eq:pf-key-18}
      \langle \A_t^{1/2},b\tD\rangle + \langle \B_t^{1/2}, \b I -b\tD\rangle  &=  \langle \A_t^{1/2},b\tD\rangle + \langle \A_t^{1/2} - \eta\A_t^{1/2}\b E\A_t^{1/2}, \b I-b\tD\rangle\nonumber\\
      &=\Tr(\A_t^{1/2}) - \eta \langle\A_t^{1/2}\b E\A_t^{1/2}, \b I- b\tD \rangle\nonumber\\
      &\stackrel{(a)}{\leq}\Tr(\A_t^{1/2})  \stackrel{(b)}{\leq }\sqrt{\dtilde},
\end{align}
where (a) follows by the fact that both $\A_t^{1/2}\b E\A_t^{1/2}$ and $\b I- b\tD$ are positive semidefinite, (b) follows by the following property:
\begin{align}
    \Tr(\A_t^{1/2}) = \sum_{i\in[\td]} \lambda_i(\A_t^{1/2}) \leq \sqrt{\td}\sqrt{\sum_{i\in[\td]} \lambda_i^2(\A_t^{1/2})}  = \sqrt{\td}\sqrt{\sum_{i\in[\td]} \lambda_i(\A_t)}  =\sqrt{\dtilde}.
\end{align}
where $\lambda_i(\A_t)$ is the $i$-th eigenvalue of $\A_t$, the inequality follows by the Cauchy-Schwarz inequality, the last equality follows by $\Tr(\A_t)=1$.
    
\textbf{step 5.} Now substitute \Cref{eq:pf-key-17} and \Cref{eq:pf-key-18} into \Cref{eq:pf-key-9}, we have
    \begin{align}
        \max_{i\in[m]}\frac{1}{\eta}\Tr [\A_t^{1/2} -(\A_t^{-1/2} + \eta \tI(x_i))^{-1}] \geq \frac{\langle \A_t, b\tD\rangle + \langle \B_t, \b I- b\tD\rangle}{b + \eta [\langle \A_t^{1/2}, b\tD\rangle + \langle \B_t^{1/2}, \b I- b\tD\rangle]}\geq \frac{1-\frac{\eta}{2b}}{b +\eta\sqrt{\dtilde}} .
    \end{align}
    
\end{proof}

\subsection{Proof of Theorem~\ref{thm:mtd-selection}}
\begin{proof} \label{appendix:mtd-performance}

    Let $b = 32\dtilde/\epsilon^2 + 16\sqrt{\dtilde}/\epsilon^2$, $\eta = 8 \sqrt{\dtilde}/\epsilon$, by \Cref{prop:mtd-selection}, we have
    \begin{align}\label{eq:pf-thm-selection-1}
        &\sum_{t=1}^b \Tr[\A_t^{1/2} - (\A_t^{-1/2} + \eta \F_t)^{-1}]\nonumber\\
        \geq& \sum_{t=1}^b\frac{1-\frac{\eta}{2b}}{b +\eta\sqrt{\dtilde}} = \frac{b - \frac{\eta}{2}}{b + \eta\sqrt{\dtilde}} \geq \frac{32\dtilde/\epsilon^2 + 16\sqrt{\dtilde}/\epsilon^2 - 4\sqrt{\dtilde}/\epsilon}{32\dtilde/\epsilon^2 + 16\sqrt{\dtilde}/\epsilon^2  + 8 \dtilde/\epsilon}\nonumber\\
        \geq &\frac{32\dtilde/\epsilon^2 + 16\sqrt{\dtilde}/\epsilon^2+ 8 \dtilde/\epsilon - ( 8 \dtilde/\epsilon+4 \sqrt{\dtilde}/\epsilon )}{32\dtilde/\epsilon^2 + 16\sqrt{\dtilde}/\epsilon^2+ 8 \dtilde/\epsilon} = 1 - \frac{8 \dtilde/\epsilon+4 \sqrt{\dtilde}/\epsilon}{\frac{4}{\epsilon} (8 \dtilde/\epsilon+4 \sqrt{\dtilde}/\epsilon) + 8\sqrt{\dtilde}/\epsilon}\nonumber\\
        \geq & 1 - \frac{\epsilon}{4}.
    \end{align}
Substitute \Cref{eq:pf-thm-selection-1}  into \Cref{eq:mtd-FTRL-bound} in \Cref{prop:mtd-FTRL}, we have
\begin{align}
    \lambda_{\min} (\sum_{t=1}^b \F_t) 
      &\geq -\frac{2\sqrt{ \dtilde}}{\eta} +  \frac{1}{\eta} \sum_{t=1}^b \Tr[\A_t^{1/2} - (\A_t^{-1/2}+ \eta \F_t)^{-1}]\nonumber\\
      &\geq -\frac{2\sqrt{\dtilde}}{8 \sqrt{\dtilde}/\epsilon} + 1 - \frac{\epsilon}{4}= 1 - \frac{\epsilon}{2} \geq\frac{1}{1+\epsilon}.
\end{align}

By  Proposition~\ref{prop:mtd-min-eigen-f-relation}, we can get
\begin{align}
    f\Big(\sum_{t=1}^b \F_t\Big) \leq (1+\epsilon) f^*.
\end{align}
\end{proof}

\subsection{Proof of Theorem~\ref{thm:mtd-sub-performance}} \label{appendix:mtd-sub-performance}

In this section, we intend to prove \Cref{thm:mtd-sub-performance}. Our main approach is combining \Cref{thm:sub-thm} and \Cref{thm:mtd-selection}. In order to account for the effect of using ERM $\theta_0$ as surrogate for $\theta_*$, we first define optimal sampling over $\theta_*$ (\Cref{def:opt-hindsight}) and optimal sampling over $\theta_0$ (\Cref{def:opt-ERM}). \Cref{cor:mtd-selection} is a direct result from \Cref{prop:mtd-selection}. At the end of this section, we give the proof for \Cref{thm:mtd-sub-performance}.

\begin{definition}\label{def:opt-hindsight}[optimal sampling in hindsight] Suppose we know $\theta_*$,  we select points $X_*$ defined by
    \begin{align}
        X_* \in \argmin_{\substack{X\subset U \\ |X|=b}}\big \langle \hq(\theta_*)^{-1}, \hp(\theta_*) \big\rangle, \quad \mathrm{ where }\quad
        q(x) \triangleq \frac{1}{n_0+b} \sum_{x^\prime \in  X_{0}\cup X} \delta( x^\prime - x).
    \end{align}
Denote the empirical distribution on points $X_0\cup X_*$ by $q_*(x)$.
\end{definition}

\begin{definition}\label{def:opt-ERM}[optimal sampling over ERM] The optimal sampling over ERM $\theta_0$ is defined by
    \begin{align}
        \widehat{X}_* \in \argmin_{\substack{X\subset U \\ |X|=b}} \big \langle \hq(\theta_{0})^{-1}, \hp(\theta_{0}) \big\rangle,\, 
     \quad \mathrm{ where }\quad
        q(x) \triangleq \frac{1}{n_0+b} \sum_{x^\prime \in  X_{0}\cup X} \delta( x^\prime - x).
    \end{align}
Denote the empirical distribution on points $X_0\cup \widehat{X}_*$ by $\widehat{q}_*(x)$.
\end{definition}

\begin{corollary}\label{cor:mtd-selection}
Given $\epsilon\in(0,1) $, consider $\eta = 8\sqrt{\dtilde}/\epsilon$, $b\geq 32 \dtilde/\epsilon^2 + 16 \sqrt{\dtilde}/\epsilon^2$ in Algorithm~\ref{algo:round}. Then we have
\begin{align}\label{eq:pf-cor-near-optimal}
    \big \langle \big(\hq(\theta_{0})\big)^{-1}, \hp(\theta_{0}) \big \rangle\leq (1+\epsilon)  \big \langle \big(\b H_{\widehat{q}_*} (\theta_{0})\big)^{-1}, \hp(\theta_{0}) \big \rangle.
\end{align}
\end{corollary}
\begin{proof}
Let $X$ be the set of  points selected by Algorithm~\ref{algo:round}, by \Cref{eq:mtd-Ix} we have:
    \begin{align}
      \hq(\theta_{0})  = \frac{1}{n}\sum_{x\in  X} \b H(x), \quad
     \b H_{\widehat{q}_*} (\theta_0) = \frac{1}{n}\sum_{x\in \widehat{X}_*} \b H(x) ,\nonumber
    \end{align}
where $n = n_0 + b$, and thus
\begin{align}\label{eq:pf-cor-1}
    \big \langle \big(  \hq(\theta_{0})\big)^{-1}, \hp(\theta_{0})\big \rangle = n f\Big(\sum_{x \in  X} \b H(x)\Big).
\end{align}
By Definition~\ref{def:opt-ERM}, we know that $\widehat{X}_*$ is the optimal solution to optimization problem Eq.~\eqref{eq:mtd-obj-2}. Since $f_*$ is the optimal value of the objective function in \eqref{eq:mtd-obj-2}, we have
\begin{align}\label{eq:pf-cor-2}
    \big \langle \big(\b H_{\widehat{q}_*}(\theta_0)\big)^{-1}, \hp(\theta_{0}) \big \rangle = n \Big \langle \big(\sum_{x\in \widehat{X}_*} \b H (x) \big)^{-1} , \hp(\theta_0) \Big\rangle = n f_*.
\end{align}
By Theorem~\ref{thm:mtd-selection}, we have $f\big(\sum_{x \in X} \b H(x)\big)\leq (1+\epsilon) f_*$. Combining this with Eqs.~\eqref{eq:pf-cor-1} and \eqref{eq:pf-cor-2}, we can obtain Eq.~\eqref{eq:pf-cor-near-optimal}.
\end{proof}

\begin{proof}[proof of Theorem~\ref{thm:mtd-sub-performance}]

By Eq.~\eqref{eq:sub-thm-risk} we have 
\begin{align}\label{eq:pf-sub-1}
     \E[L_p(\theta_{0})] - L_p(\theta_*) \lesssim \frac{e^{\alpha_1} - \alpha_1 -1}{ \alpha_1^2} \cdot \frac{\big\langle (\hq(\theta_*))^{-1}, \hp(\theta_*) \big\rangle}{n_0 + b},
\end{align}
where
\begin{align}
    \alpha_1 =C_3 \sqrt{\sigma_1\rho} \sqrt{\big(\dtilde + \sqrt{\td}\log(e/\delta)\big)/(n_0+b)}  ,
\end{align}
where $\sigma_1 = \lambda_{\max}(\hq^{-1} \hp)$ . From the step 2 of the proof of Theorem~\ref{thm:sub-thm}, we have with probability at least $1-\delta$,
\begin{align}\label{eq:sub-algo-1}
    \frac{1}{\sqrt{2}} \hq(\theta_*) \preceq \hq(\theta_{r-1})\preceq  \sqrt{2}  \hq(\theta_*).
\end{align}
Combining results from step 6 in the proof of Theorem~\ref{thm:sub-thm} with Eq.~\eqref{eq:taylor-Hessian} in Proposition~\ref{prop:taylor-concordance}, we can obtain that with probability at least $1-\delta$, 
        \begin{align}\label{eq:sub-algo-2}
             e^{-\alpha_{0}} \hp(\theta_*) \preceq \hp(\theta_{0}) \preceq e^{\alpha_{0}} \hp(\theta_*),
         \end{align}
where 
\begin{align}
    \alpha_0 =C_3^\prime \sqrt{\sigma_0\rho} \sqrt{\big(\dtilde + \sqrt{\td}\log(e/\delta)\big)/n_0}  ,
\end{align}
where $\sigma_0 = \lambda_{\max}(\b H_{q_0}^{-1} \hp)$ , $q_0(x)$ is the empirical distribution over the inital labeled points, i.e. $q_0(x) \triangleq \sum_{x^\prime \in X_0} \delta(x - x^\prime)$.

Therefor we have
\begin{align}\label{eq:sub-algo-3}
        \Big \langle  \big(\hq (\theta_*)\big)^{-1}, \hq(\theta_*)    \Big \rangle &\stackrel{(a)}{\leq} \sqrt{2} e^{\alpha_{0}}    \Big  \langle  \big(\hq (\theta_{0})\big)^{-1}, \hp(\theta_{0})    \Big \rangle\nonumber\\
    &\stackrel{(b)}{\leq} \sqrt{2} e^{\alpha_{0}}(1+\epsilon)     \Big  \langle \big(\b H_{\widehat{q}_*}(\theta_{0})\big)^{-1}, \hp(\theta_{0})    \Big \rangle\nonumber\\
    &\stackrel{(c)}{\leq} \sqrt{2} e^{\alpha_{0}}(1+\epsilon)     \Big  \langle  \big(\b H_{q_*}(\theta_{0})\big)^{-1}, \hp(\theta_{0})    \Big \rangle\nonumber\\
    &\stackrel{(d)}{\leq} 2 e^{2\alpha_{0}}(1+\epsilon)     \Big \langle  \big(\b H_{q_*}(\theta_*)\big)^{-1}, \hp(\theta_*)    \Big \rangle \nonumber\\
   & = 2 e^{2\alpha_{0}}(1+\epsilon) OPT,
\end{align}
where (a) and (d) follow by Eqs.~\eqref{eq:sub-algo-1} and \eqref{eq:sub-algo-2}, (b) follows by Corollary~\ref{cor:mtd-selection}, (c) follows by the fact that $\widehat{q}_*$ is the optimal sampling distribution to minimize $\langle (\hq(\theta_{0}))^{-1}, \hp(\theta_{0})\rangle$ (see the definition of optimal sampling over ERM  in Definition~\ref{def:opt-ERM}).

By \Cref{eq:sub-algo-3,eq:pf-sub-1}, we can obtain Eq.~\eqref{eq:sub-algo-risk}.
    
\end{proof}

\section{Additional experimental details}\label{appendix:experiments}

\subsection{Synthetic experiments}\label{appendix:synthetic}

\begin{figure}[t]
\centering
  \footnotesize
\begin{tikzpicture}
\node[inner sep=0pt] (a1) at (0,0) {\includegraphics[width=3.2cm]{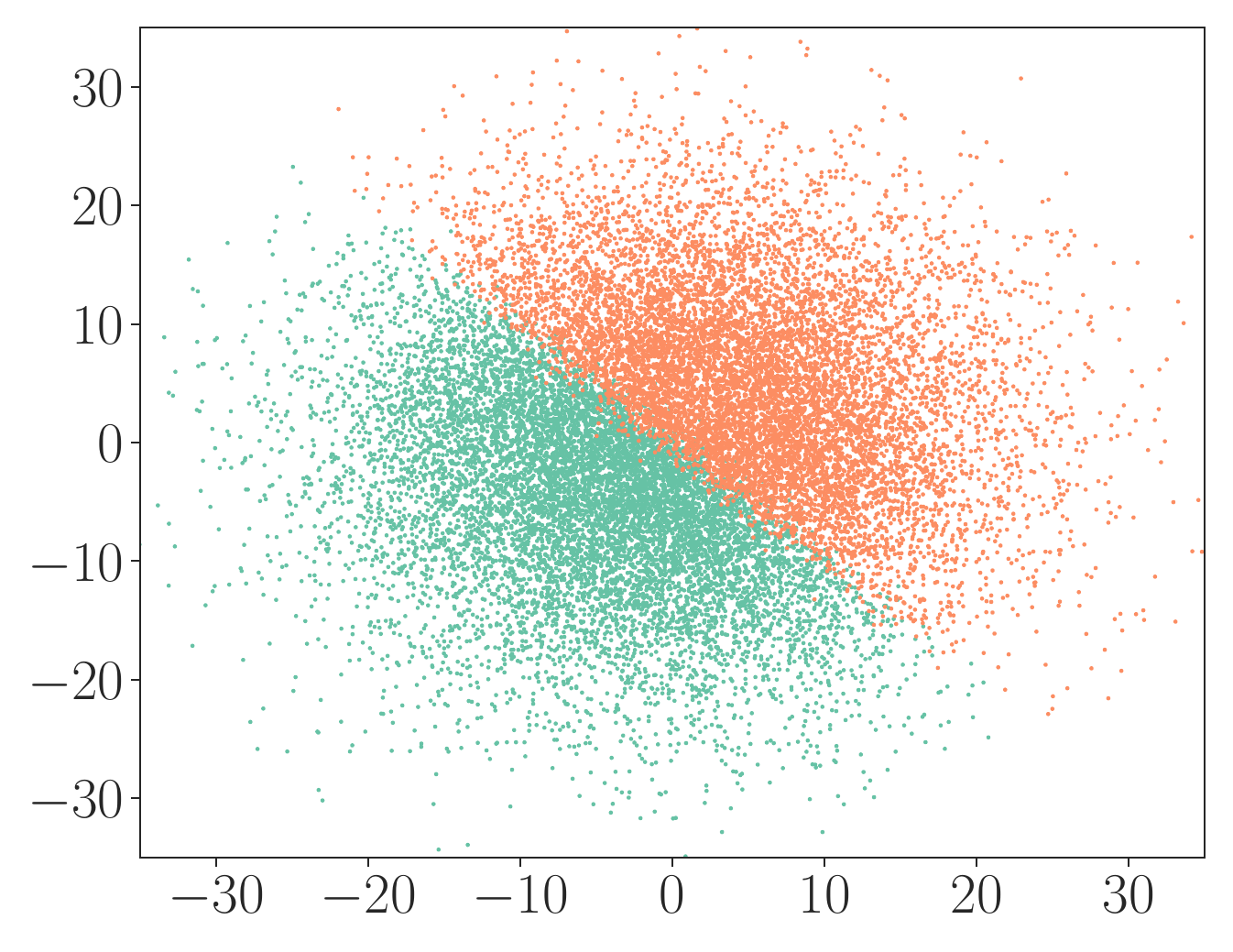}};
\node[inner sep=0pt] (a2) at (3.4,0) {\includegraphics[width=3.2cm]{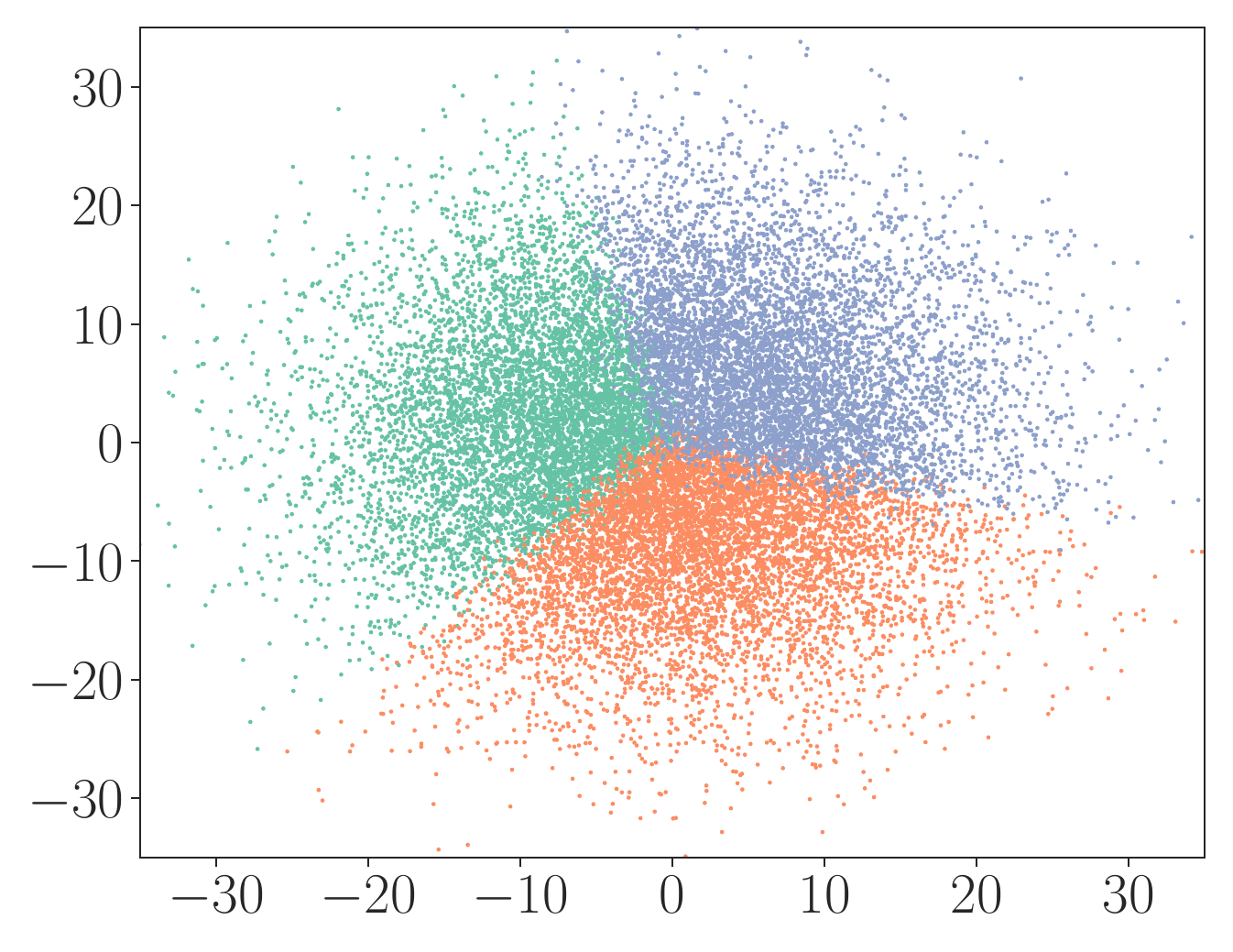}};
\node[inner sep=0pt] (a3) at (6.8,0) {\includegraphics[width=3.2cm]{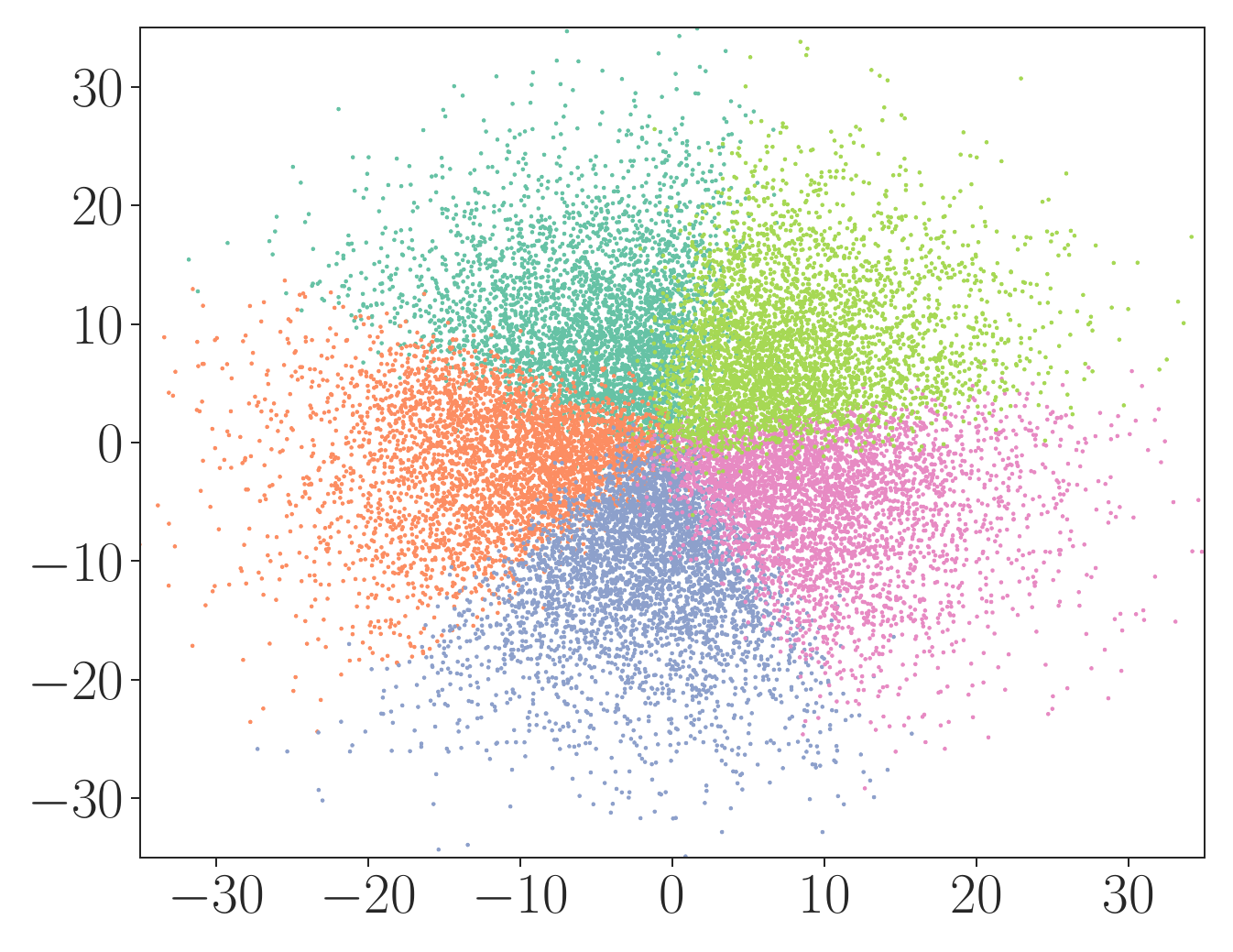}};
\node[inner sep=0pt] (a4) at (10.2,0) {\includegraphics[width=3.2cm]{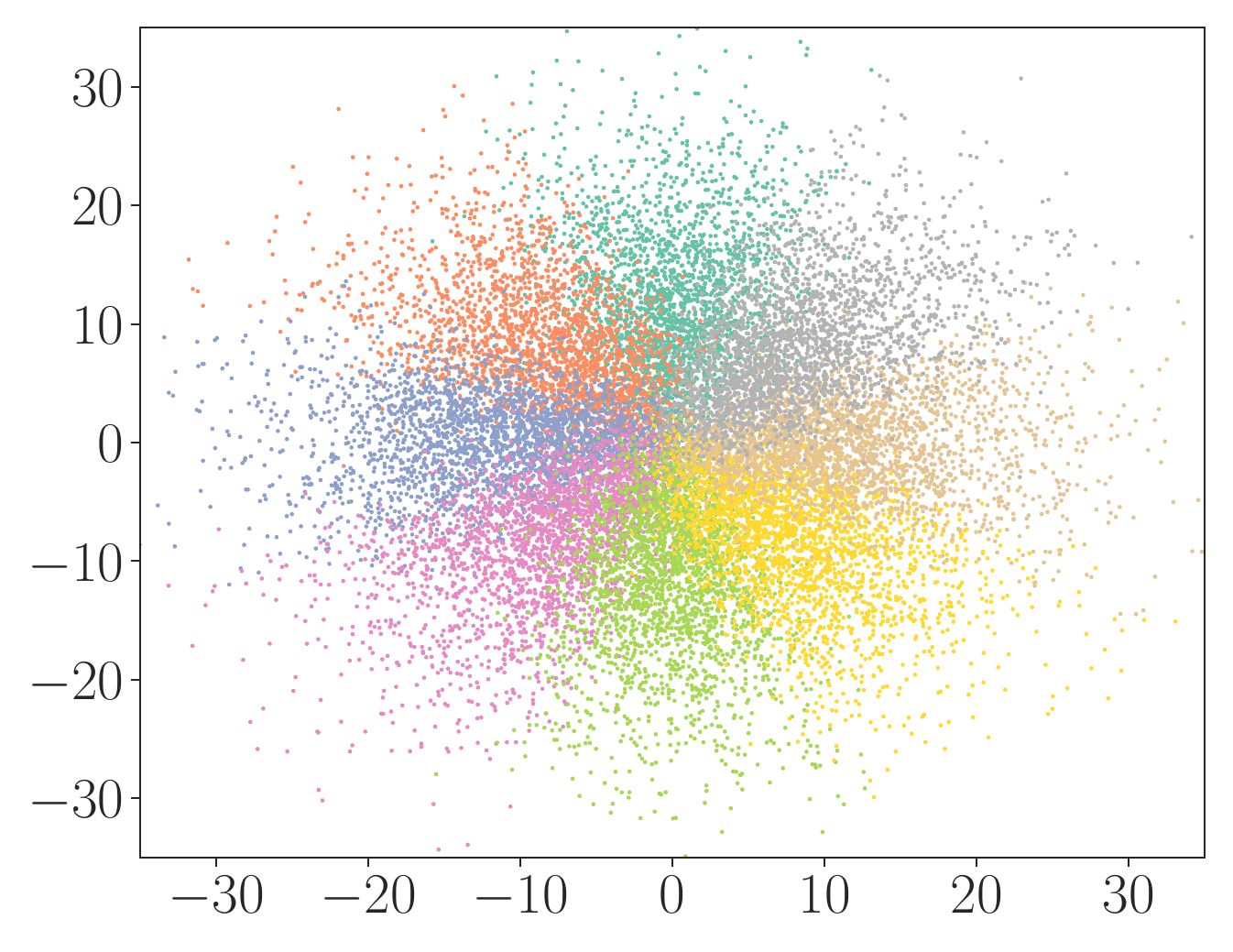}};
\end{tikzpicture}
    \caption{Plots of first two coordinates of points draw from the joint distribution $pi_p(x,y)$. }
    \label{fig:2d-plot}
\end{figure}

We use numerical tests on synthetic datasets to demonstrate the two excess risk bounds derived in \Cref{thm:sub-thm-detail} (detailed version of \Cref{thm:sub-thm}): \Cref{eq:sub-thm-erm-detail} and \Cref{eq:sub-thm-risk-detail}.

\paragraph{Gaussian Setup.} For a given dimension $d$, we choose $p(x) \sim \mathcal{N}(\b 0, \V_p)$, where $\V_p = 100\b I_d $. For a given class number $c$,  we define $\theta_*\in \mathbb{R}^{(c-1) \times d}$ such that points generated by $p(x)$ are almost equally distributed across the $c$ classes. Besides, we normalize the row of $\theta_*$, i.e. $\| \theta_{*,i}\|_2 = 1$. In \Cref{fig:2d-plot}, we plot the first two coordinates of the points draw from the joint distribution $pi_p(x,y)$, where each point is colored by its class id. 

We use Monte Carlo method to approximate the risk of $p(x)$ at a given parameter $\theta$, i.e. $L_p(\theta) = \E_{(x,y)\sim \pi_p(x,y)}[\ell_{(x,y)}(\theta)]$. In specific, we draw $N=50,000$ i.i.d. points $\{ x_i\}_{i \in[N]}$ from $p(x)$, for each $x_i$, we draw $M=100$ i.i.d. labels $\{y_{ij}\}_{j\in[M]}$ from $p(y|x_i, \theta_*)$, then we can estimate the risk by
\begin{align}
    L_p(\theta) &\triangleq \E_{(x,y) \sim \pi_p(x,y)} [\ell_{(x,y)}(\theta)] = \E_{x\sim p(x)} \E_{y \sim p(y|x, \theta_*}[\ell_{x,y}(\theta)]\nonumber\\
    & \approx \frac{1}{N}\frac{1}{M}\sum_{i\in[N]}\sum_{j\in[M]} \ell_{(x_i, y_{ij})}(\theta).
\end{align}

\begin{figure}[!t]
\centering
  \footnotesize
\begin{tikzpicture}
\node[inner sep=0pt] (a1) at (0,0) {\includegraphics[width=4.6cm]{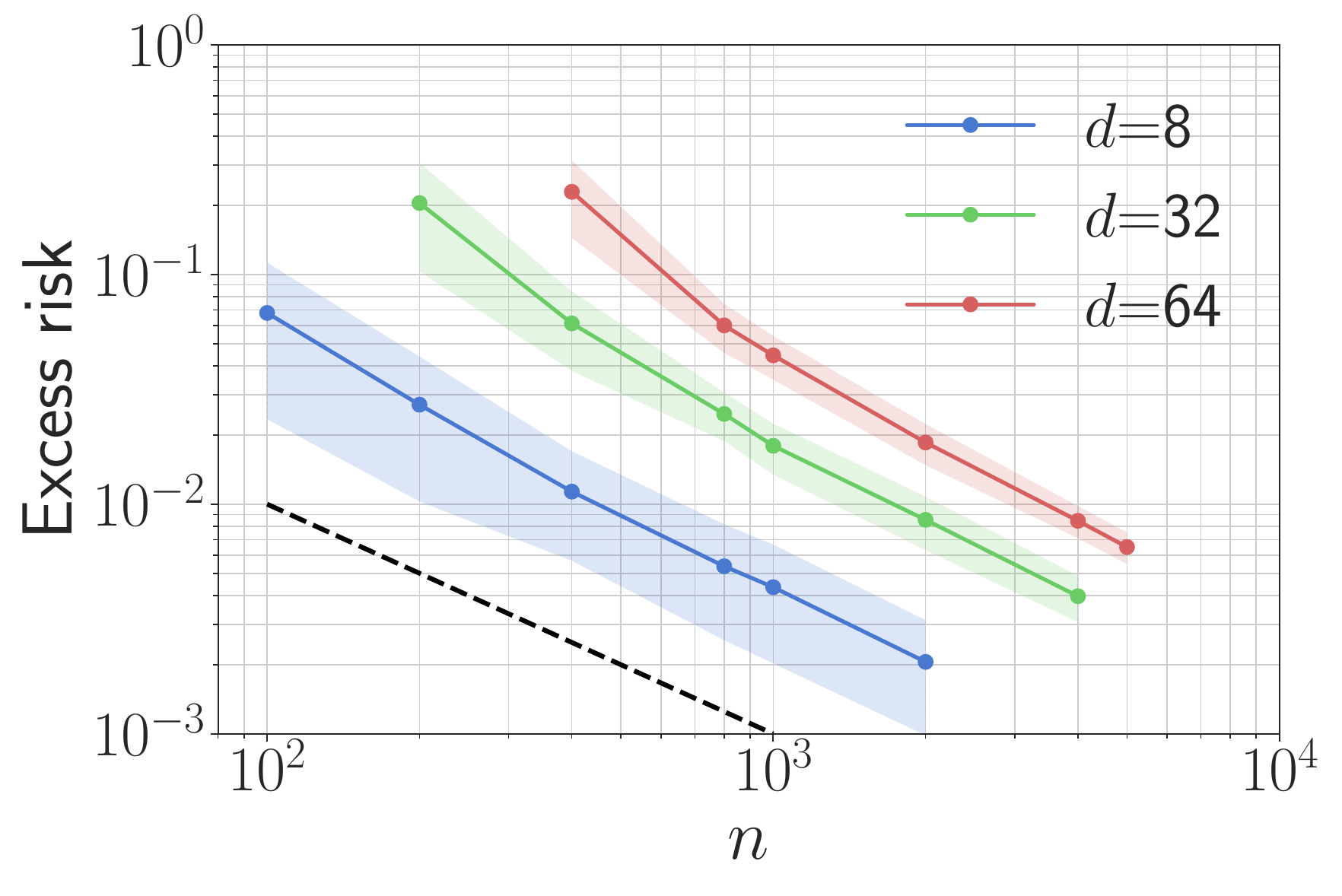}};
\node[inner sep=0pt] (a2) at (4.6,0) {\includegraphics[width=4.6cm]{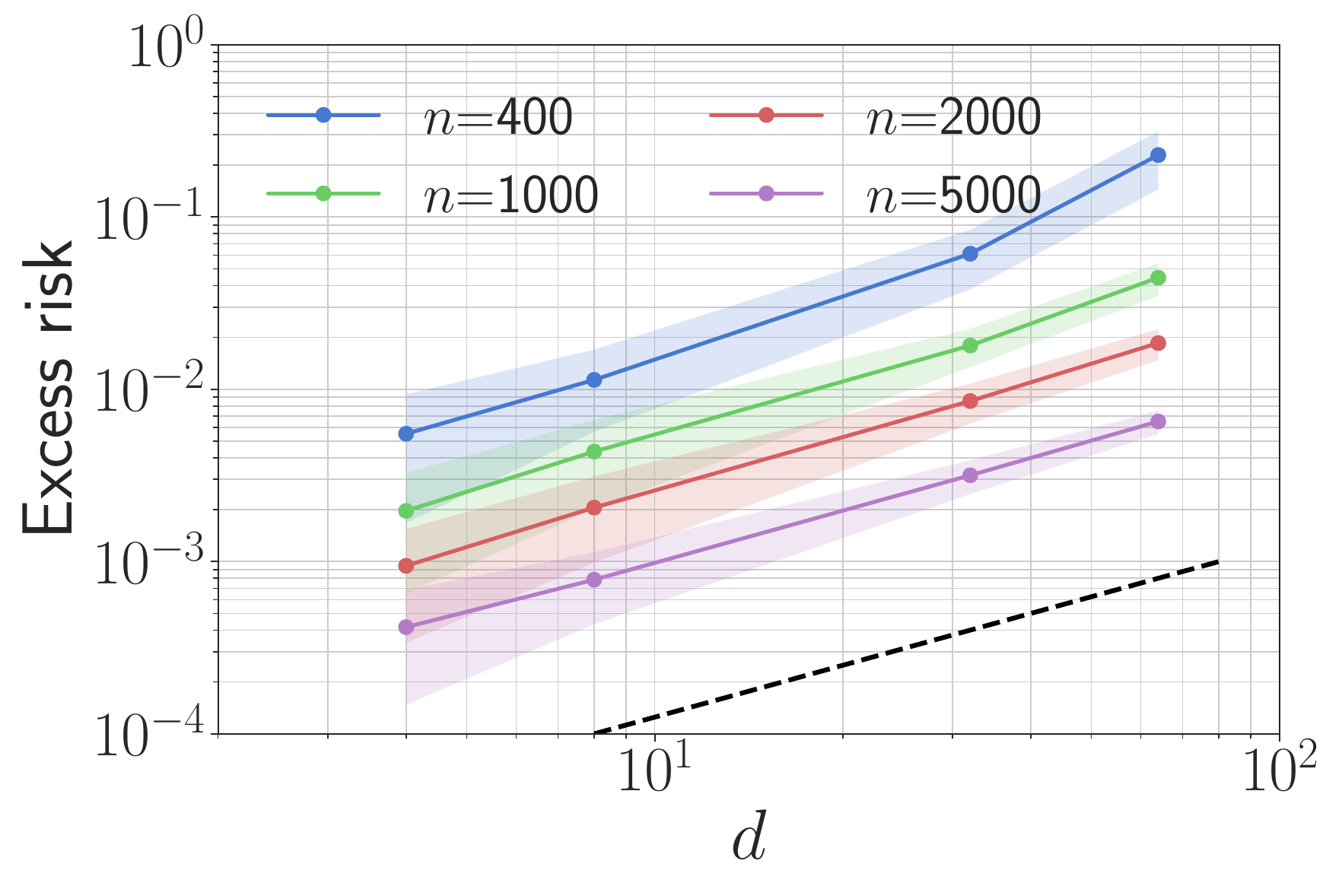}};
\node[inner sep=0pt] (a3) at (9.2,0) {\includegraphics[width=4.6cm]{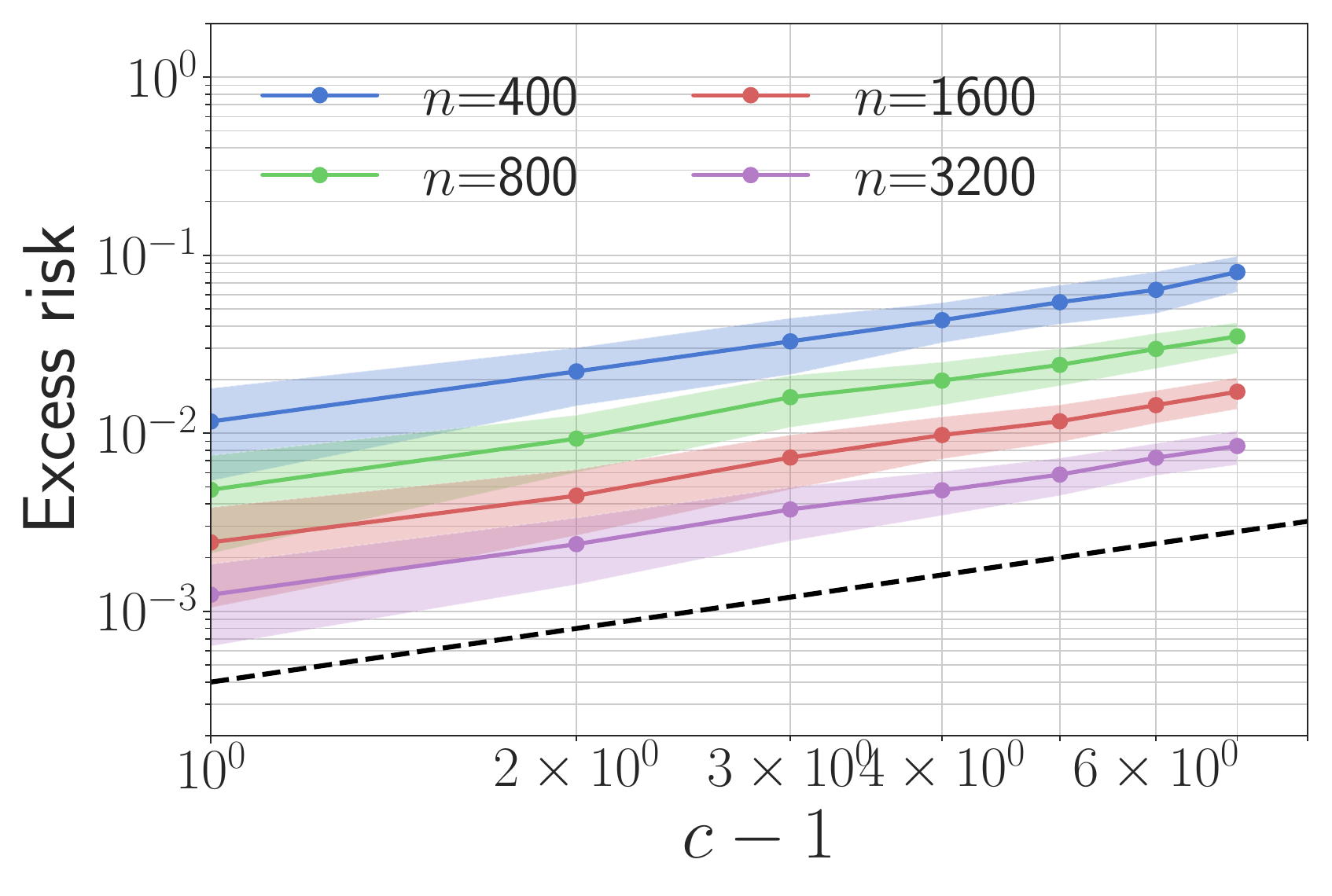}};
\end{tikzpicture}
    \caption{Excess risk of  $q(x)$ as a function of $n$, $d$ and $c-1$. The dashed black line in the left plot indicates inversely linear relation. The dashed black lines in the center and right plots indicate linear relations.}
    \label{fig:gaussian-erm}
\end{figure}

\paragraph{Demonstration of excess risk bound for $q(x)$ (\Cref{eq:sub-thm-erm-detail}).} We use $q(x)\sim \mathcal{N}(\b 0, 100 \b I_d)$ to demonstrate \Cref{eq:sub-thm-erm-detail}. Let $\{(x_i,y_i)\}_{i\in[n]}$ be samples i.i.d draw from $\pi_q(x,y)$. Denote the ERM estimate as  $\theta_n$ defined by \Cref{eq:ERM-q}. In \Cref{fig:gaussian-erm}, we plot the excess risk with respect to $q(x)$ (i.e. $L_q(\theta_n) - L_q(\theta_*)$) against $n$, $d$ and $c-1$. From theses plots, we can observe that the excess risk almost linearly depends on $\frac{1}{n}$, $d$ and $c-1$ respectively.
This observation is consistent to our upper bound derived in \Cref{eq:sub-thm-erm-detail}.

\begin{figure}[!t]
\centering
  \footnotesize
\begin{tikzpicture}
\node[inner sep=0pt] (a1) at (0,-1) {\includegraphics[width=4.6cm]{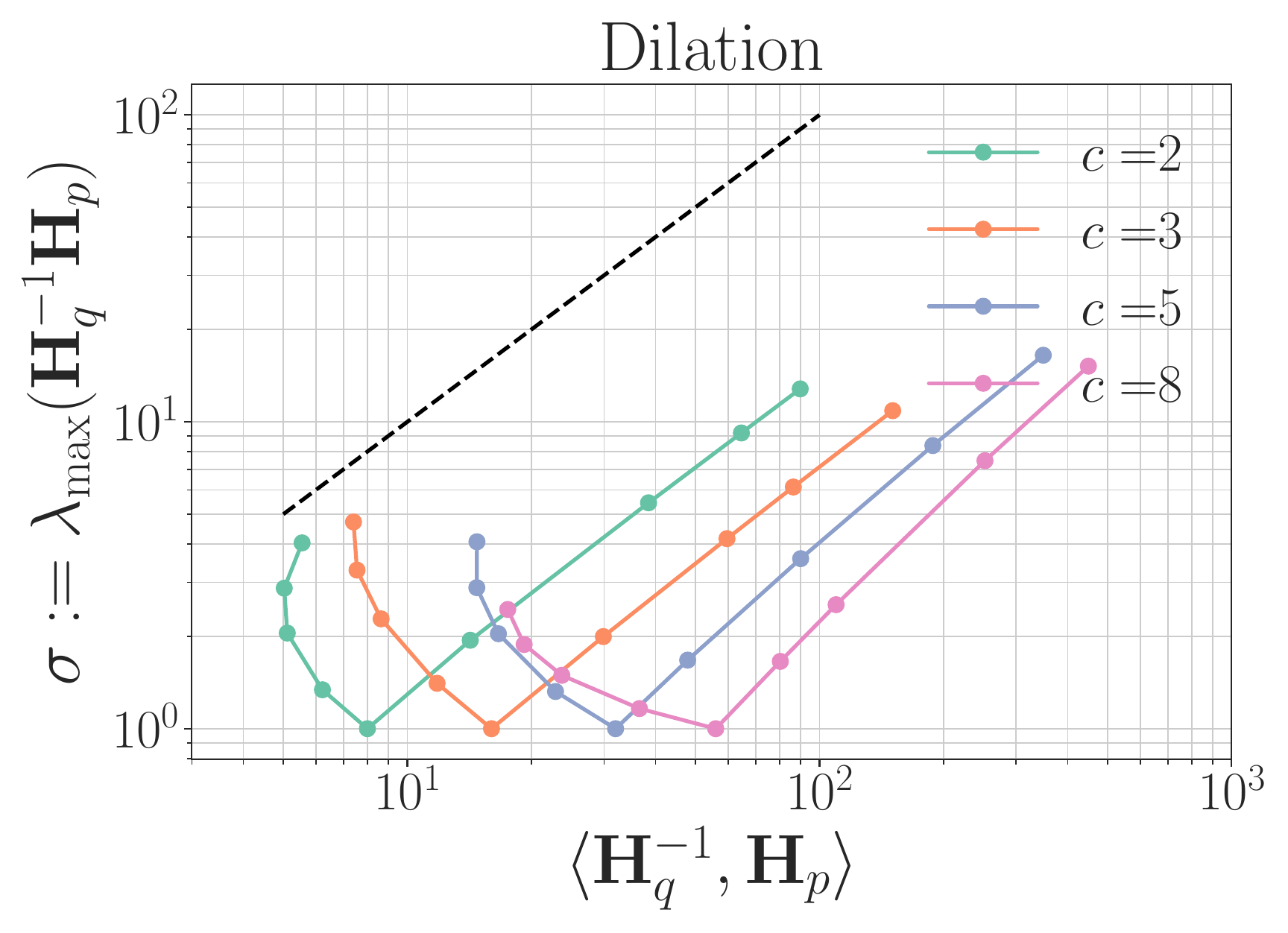}};
\node[inner sep=0pt] (a2) at (7,-1) {\includegraphics[width=4.6cm]{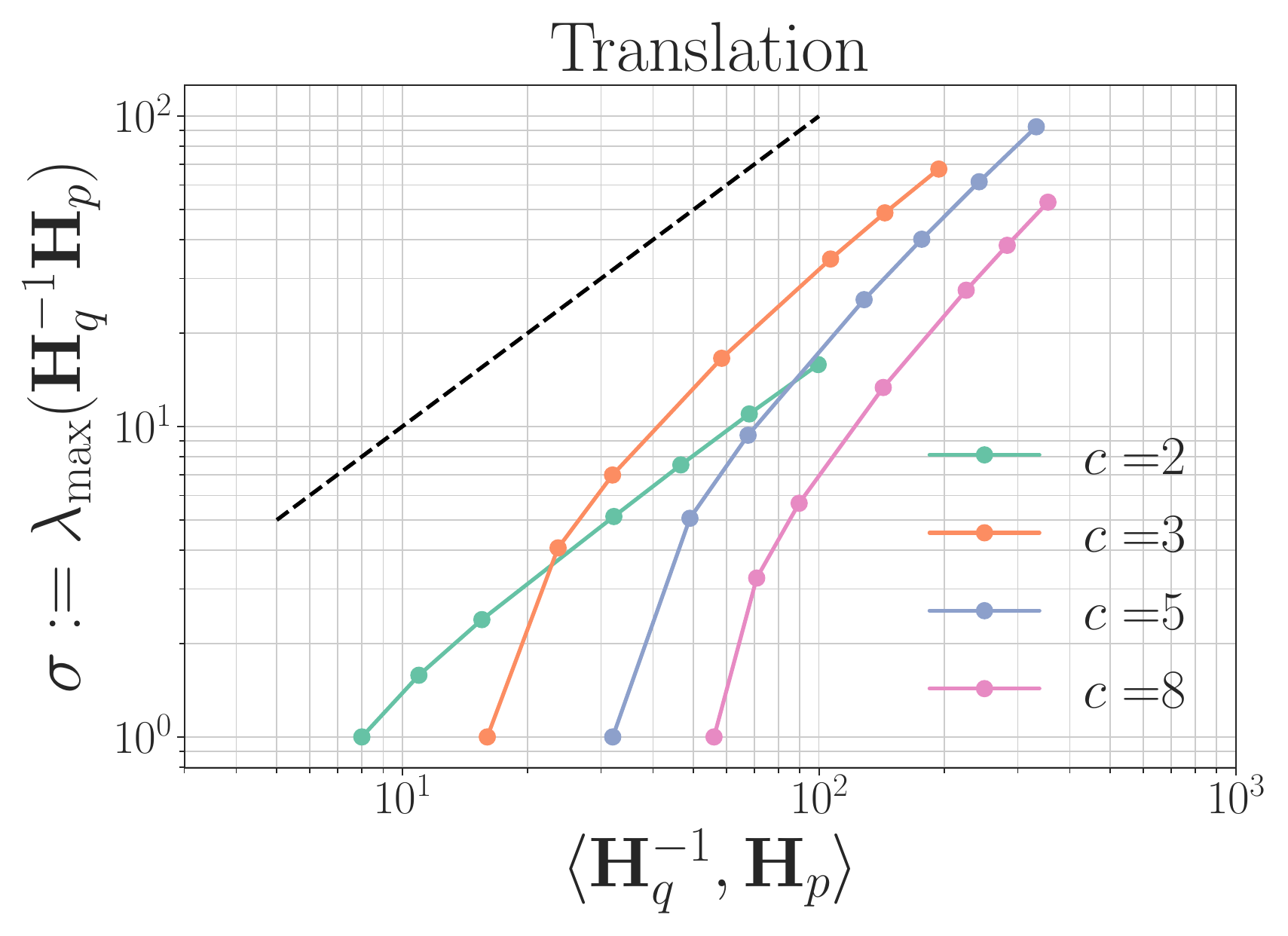}};
\end{tikzpicture}
    \caption{$\lambda_{\max}(\hq^{-1}\hp)$ vs $\langle \hq^{-1}, \hp \rangle$ in dilation tests (left plot) and translation tests (right plot). }
    \label{fig:gaussian-sigma-trace}
\end{figure}

\begin{figure}[!t]
\centering
  \footnotesize
\begin{tikzpicture}
\node[inner sep=0pt] (a1) at (0,0) {\includegraphics[width=4.5cm]{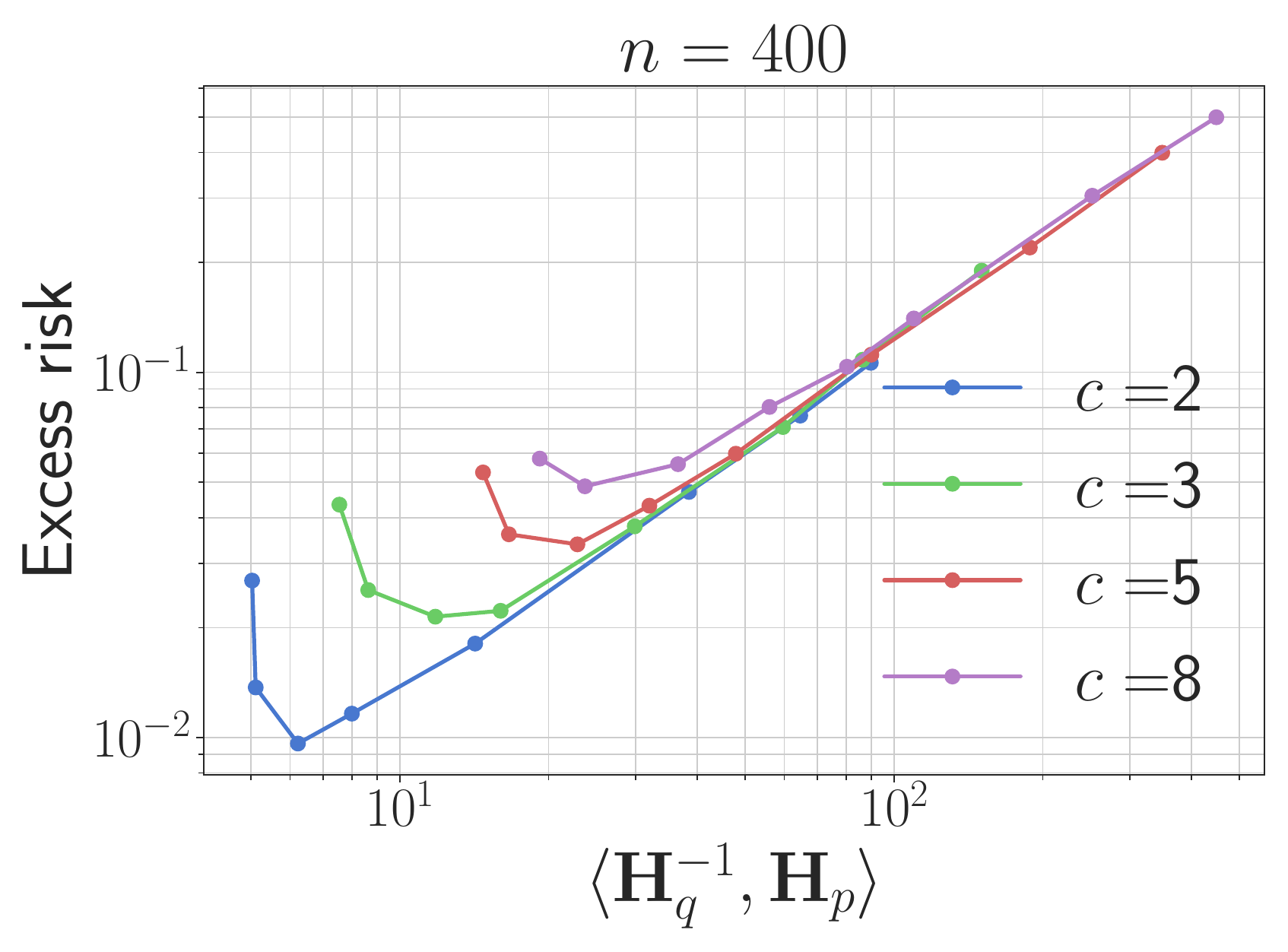}};
\node[inner sep=0pt] (a2) at (4.6,0) {\includegraphics[width=4.5cm]{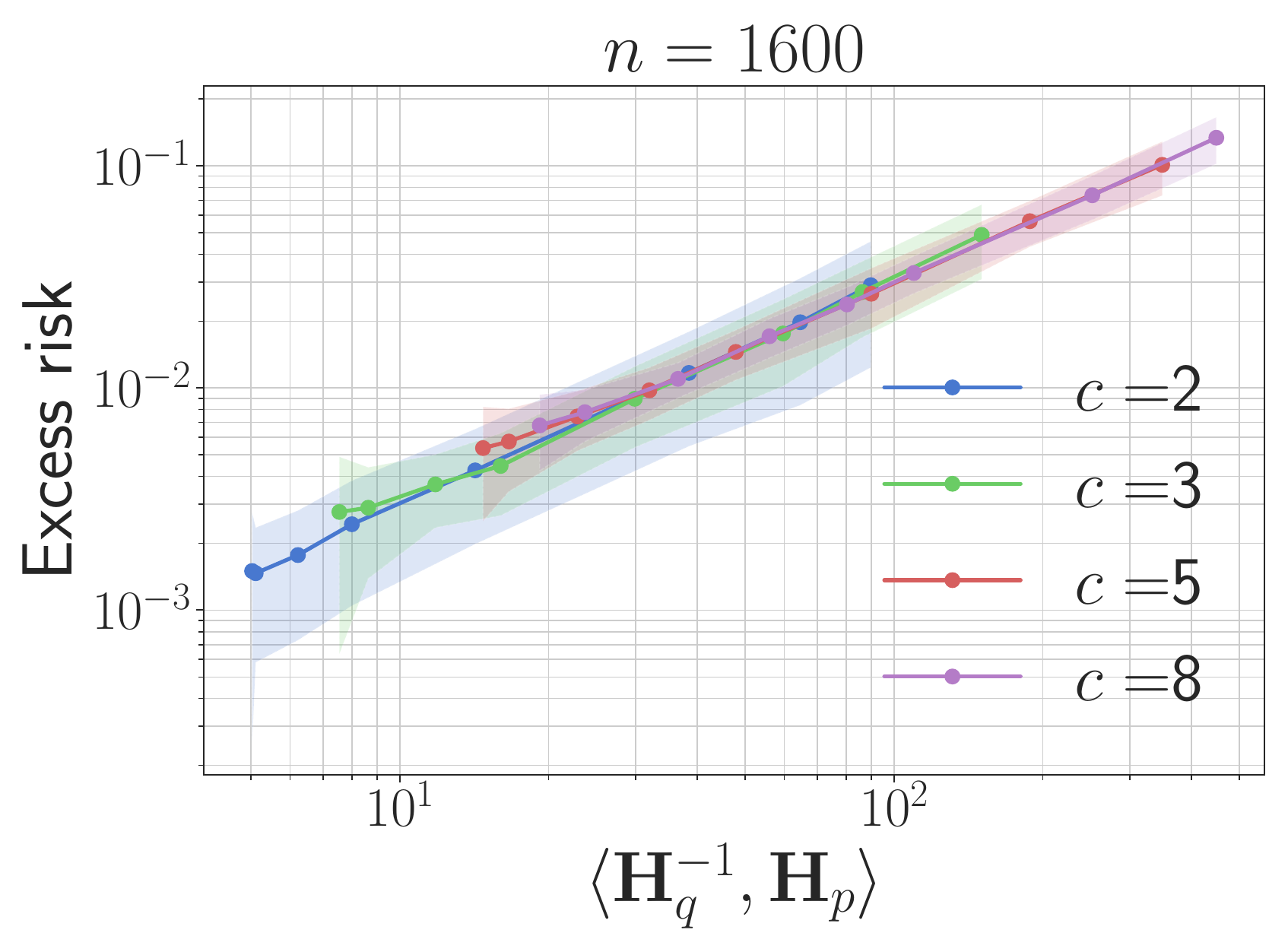}};
\node[inner sep=0pt] (a3) at (9.2,0) {\includegraphics[width=4.5cm]{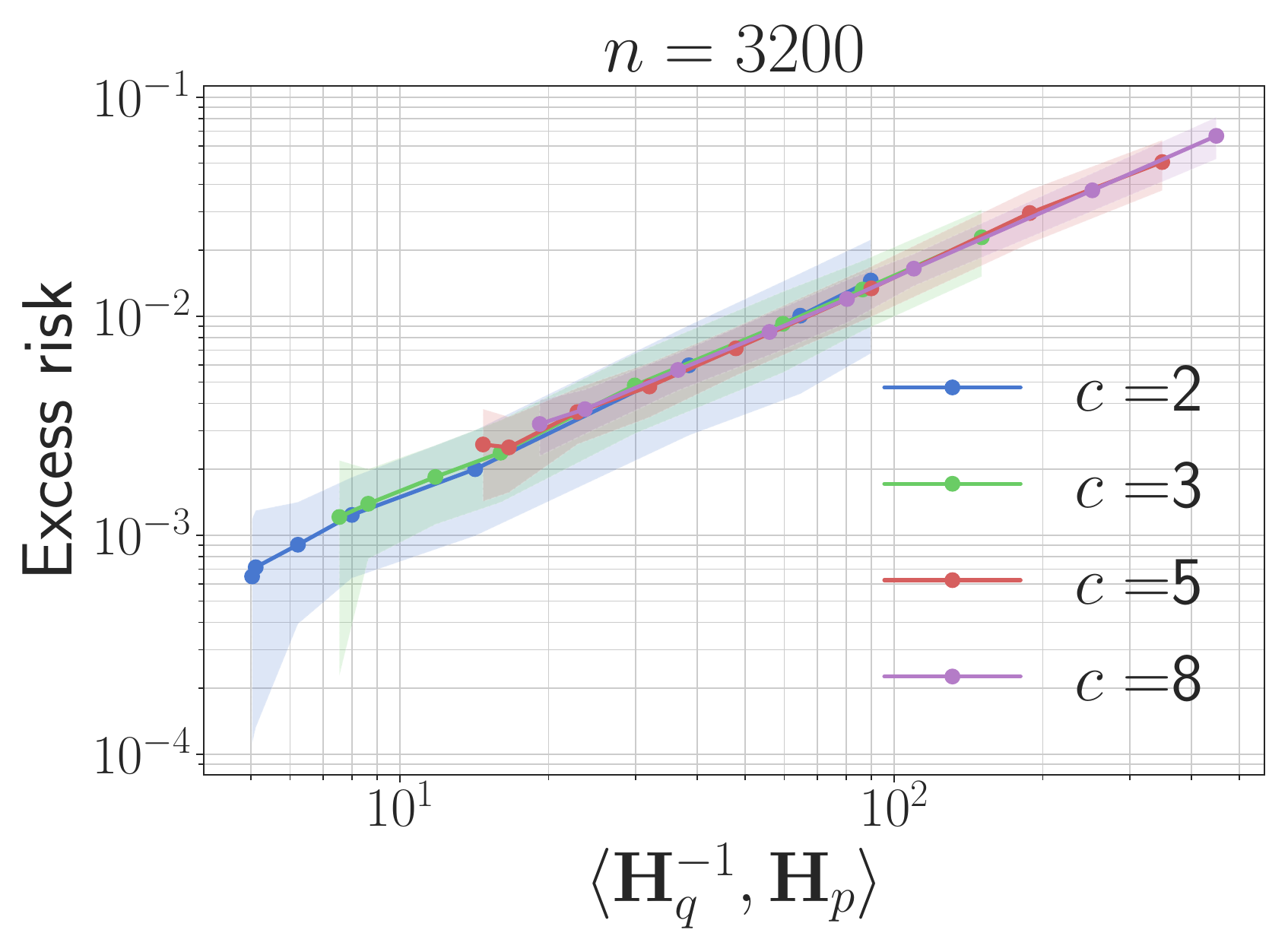}};
\node[inner sep=0pt] (b1) at (0,-3.2) {\includegraphics[width=4.5cm]{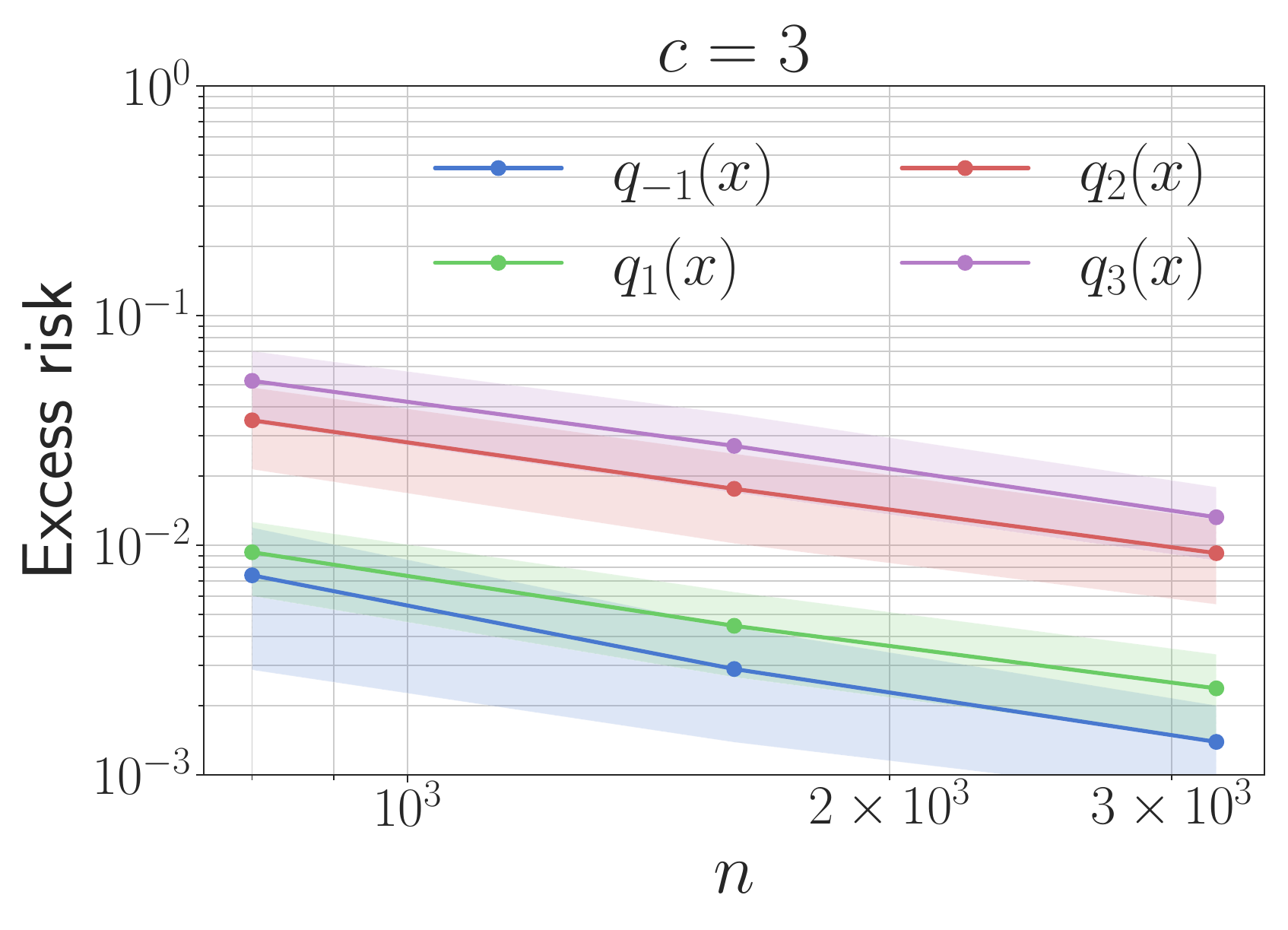}};
\node[inner sep=0pt] (b2) at (4.6,-3.2) {\includegraphics[width=4.5cm]{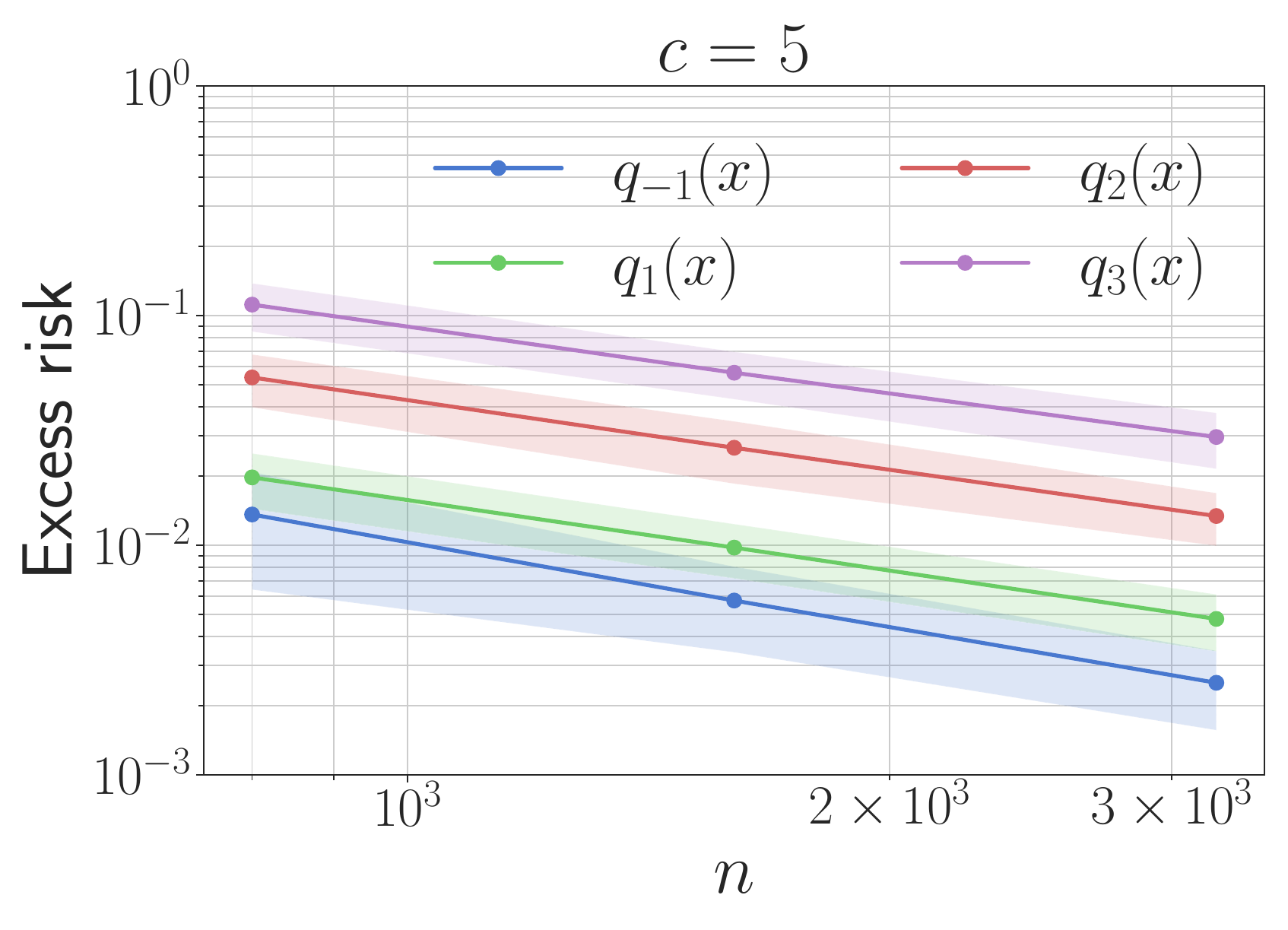}};
\node[inner sep=0pt] (b3) at (9.2,-3.2) {\includegraphics[width=4.5cm]{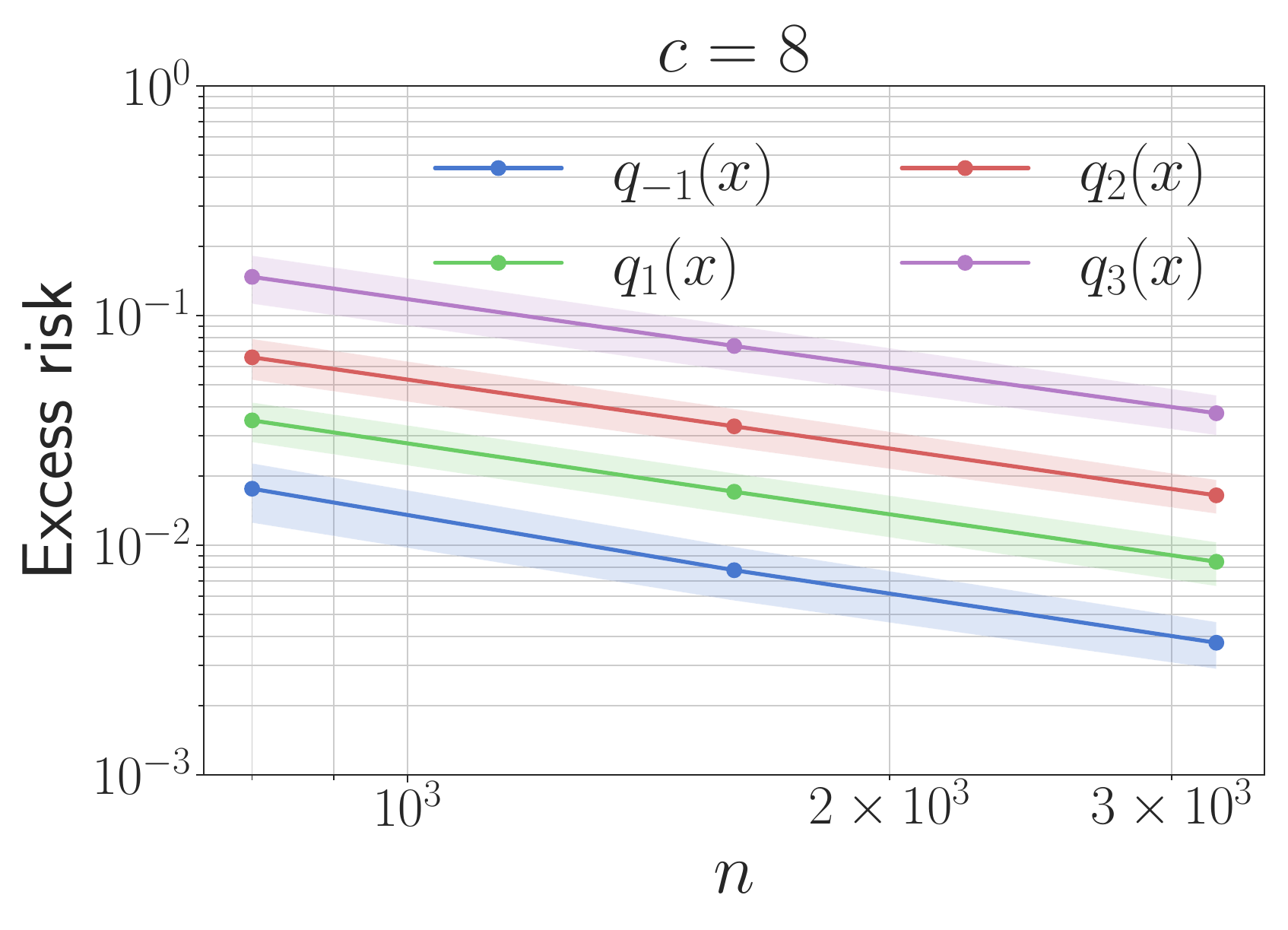}};
\node[inner sep=0pt] (c1) at (0,-6.5) {\includegraphics[width=4.5cm]{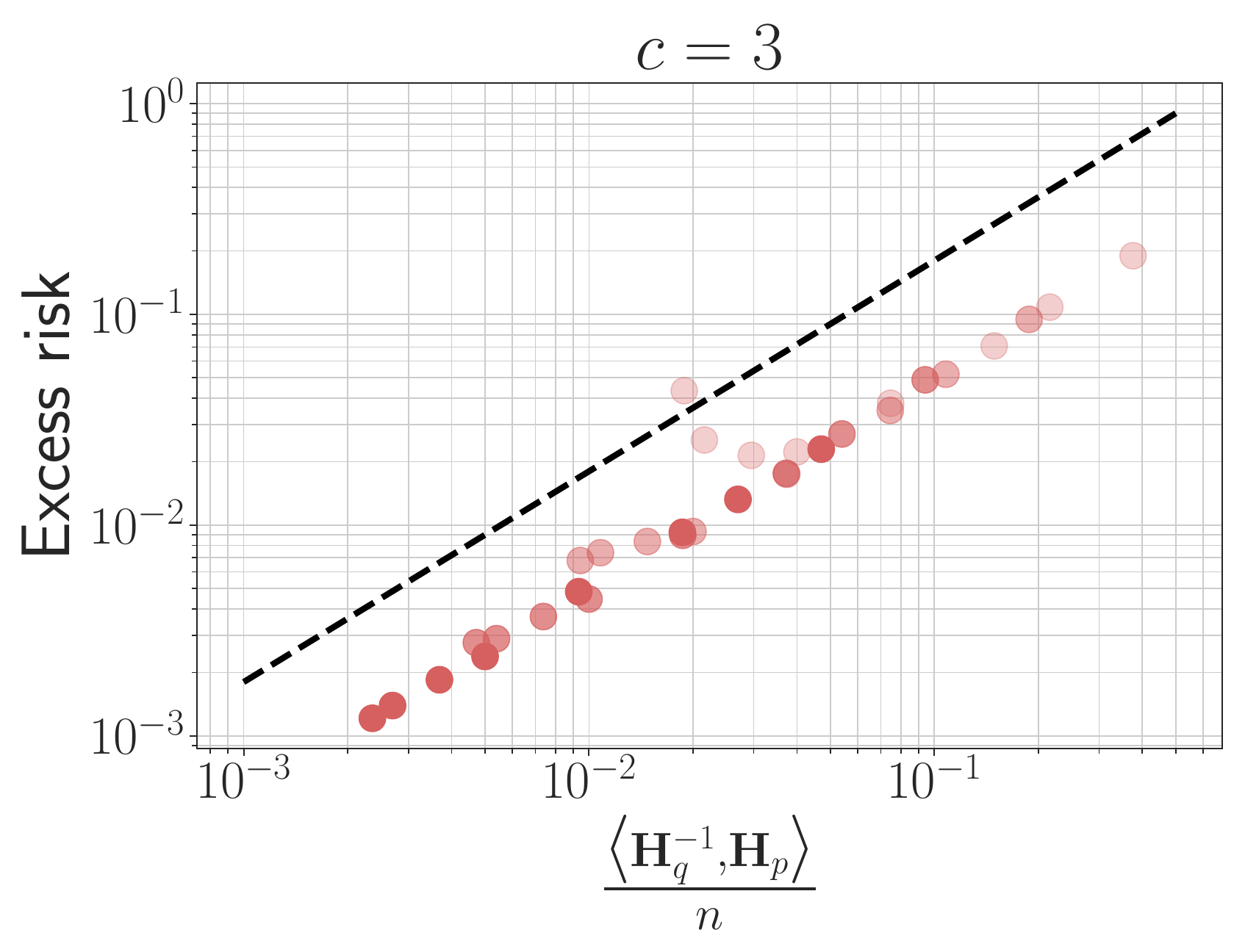}};
\node[inner sep=0pt] (c2) at (4.6,-6.5) {\includegraphics[width=4.5cm]{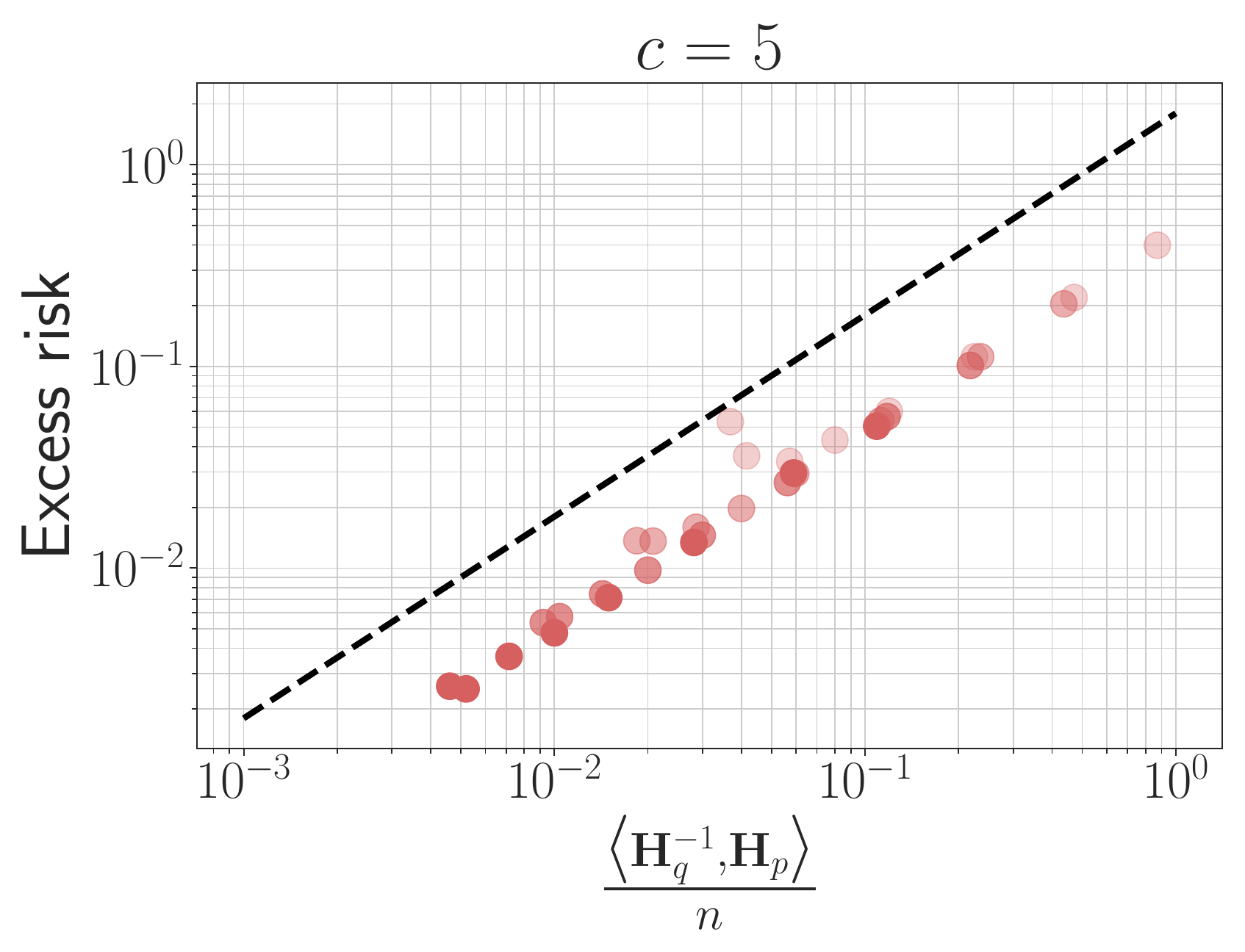}};
\node[inner sep=0pt] (c3) at (9.2,-6.5) {\includegraphics[width=4.5cm]{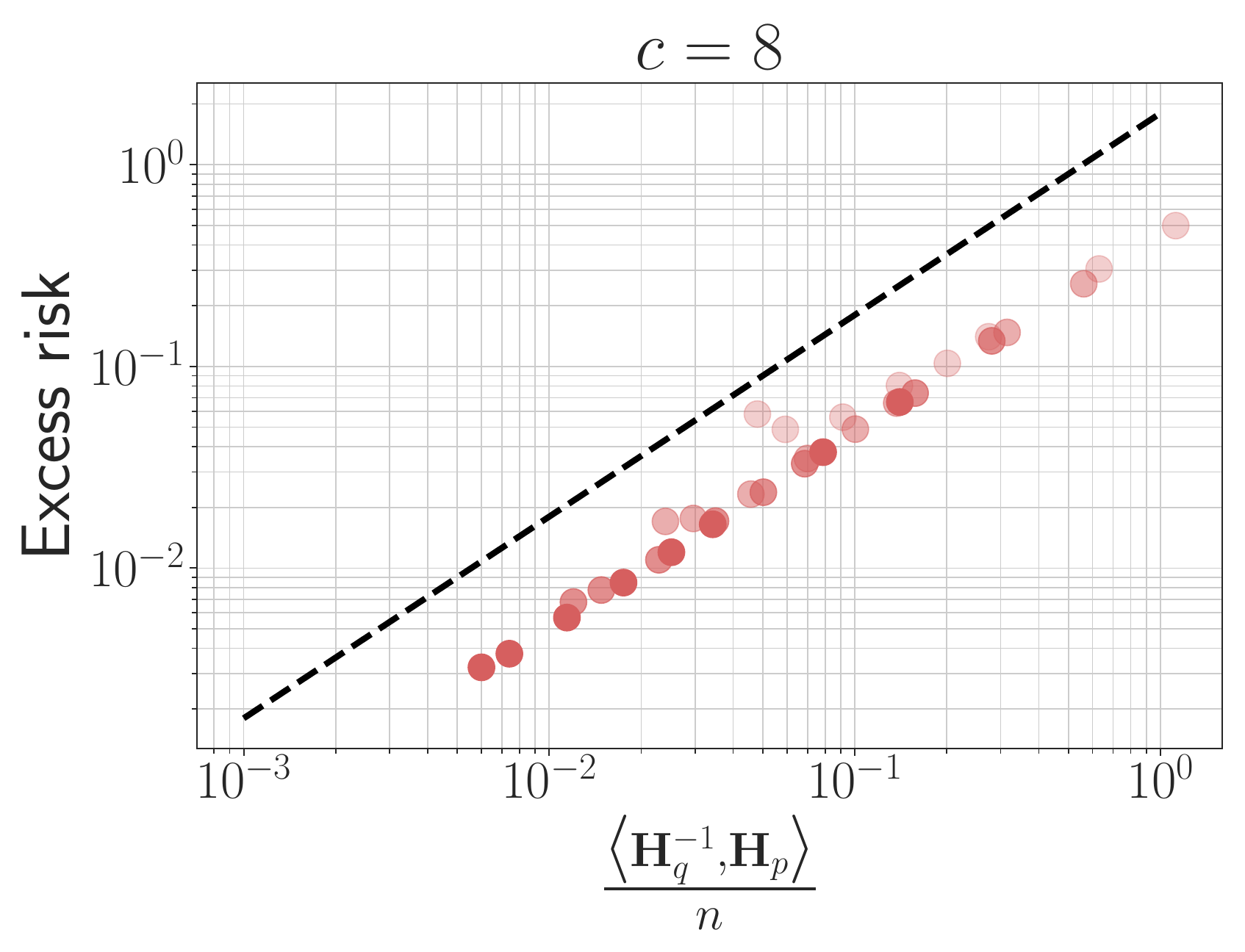}};
\end{tikzpicture}
\caption{Gaussian \textit{dilation} tests: excess risk of $p(x)$ vs FIR (upper row), $n$ (middle row) and $\mathrm{FIR}/n$ (lower row). For all plots in the lower row, the less transparent dots represent the larger sample size $n$, the black dashed lines represent linear relation $y=\frac{9}{5} x$. }
\label{fig:gaussian-dilation}
\end{figure}

\begin{figure}[!t]
\centering
  \footnotesize
\begin{tikzpicture}
\node[inner sep=0pt] (a1) at (0,0) {\includegraphics[width=4.6cm,height = 3.3cm]{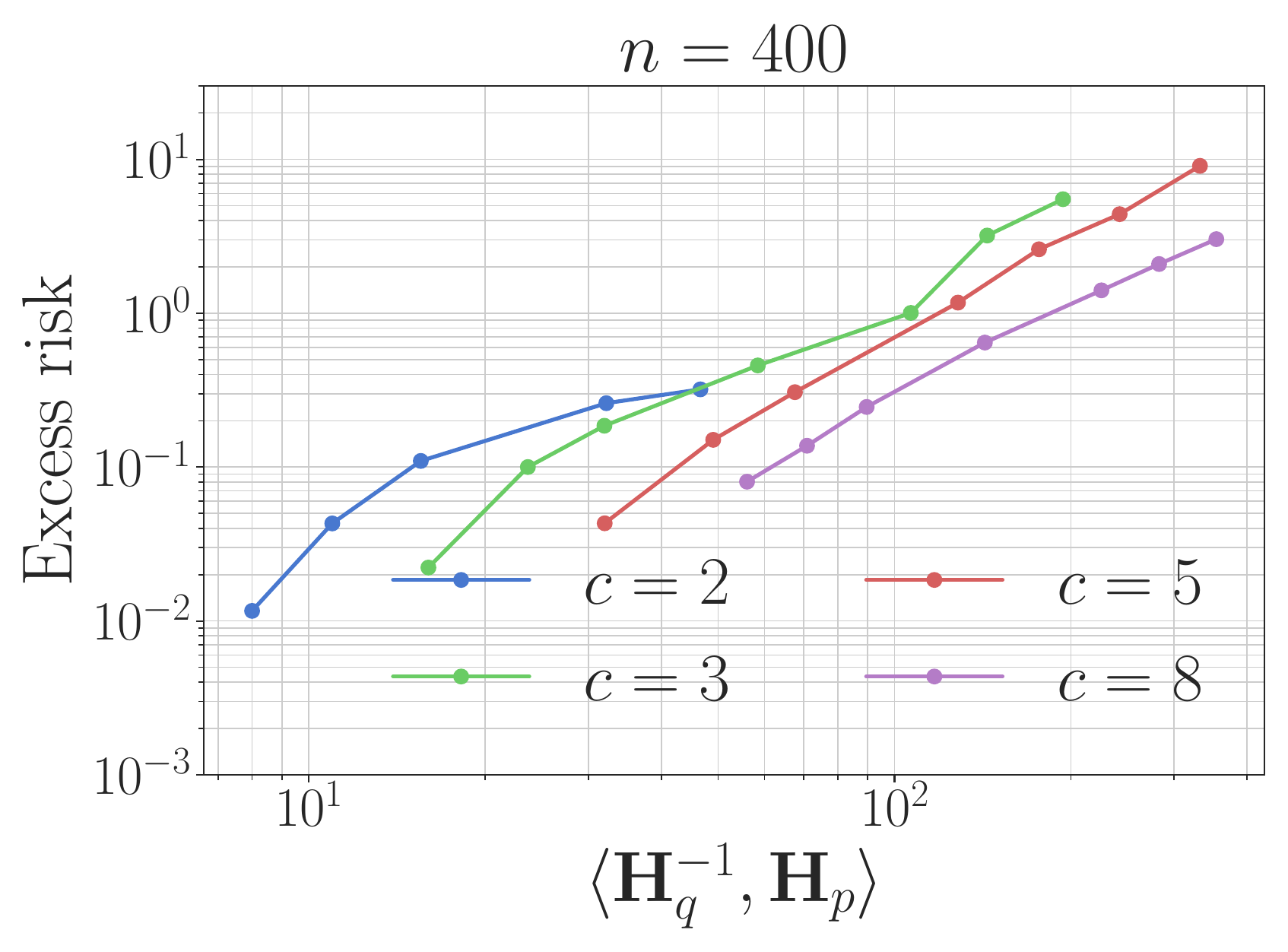}};
\node[inner sep=0pt] (a2) at (4.6,0) {\includegraphics[width=4.6cm,height = 3.3cm]{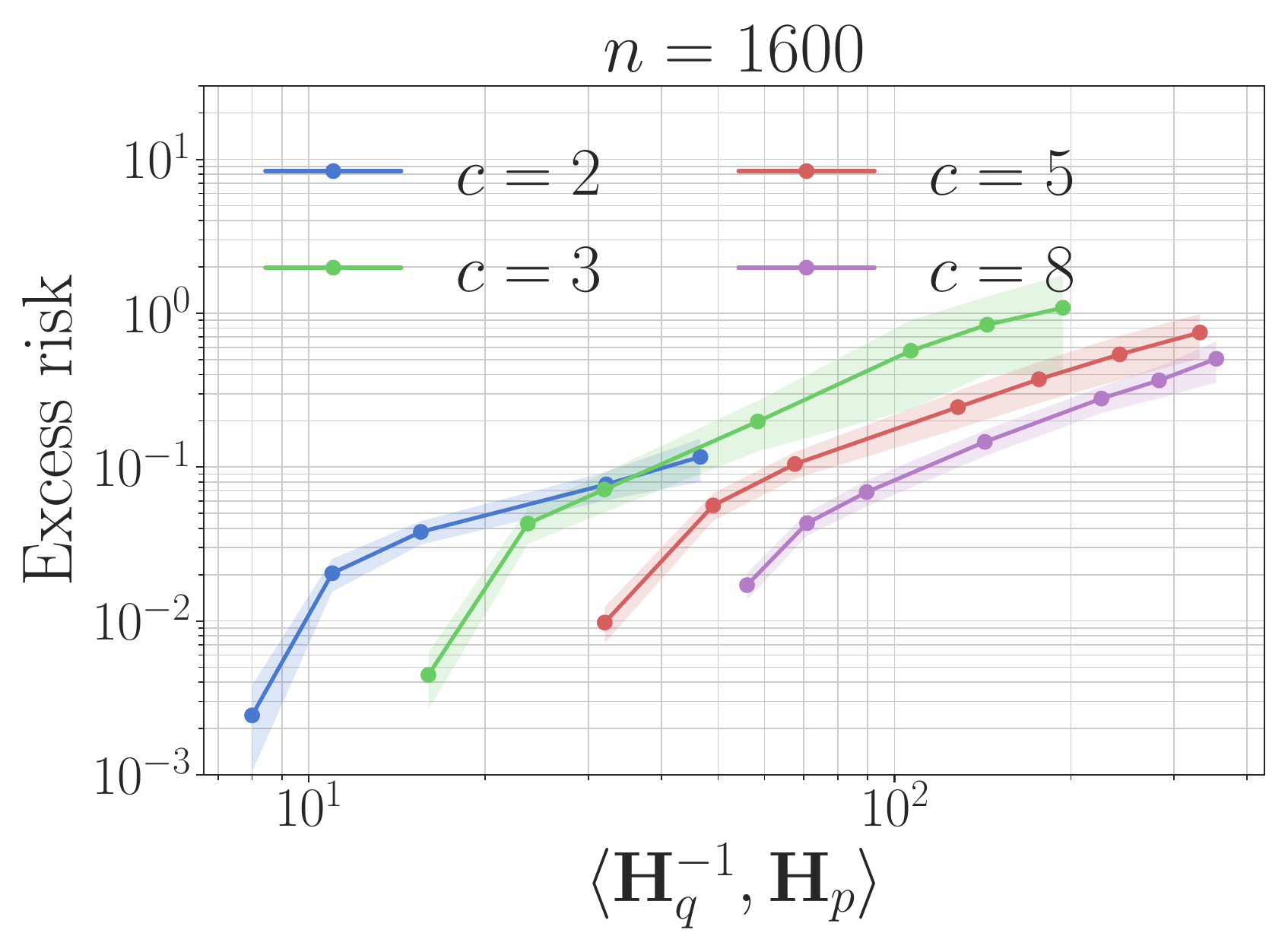}};
\node[inner sep=0pt] (a3) at (9.2,0) {\includegraphics[width=4.6cm,height = 3.3cm]{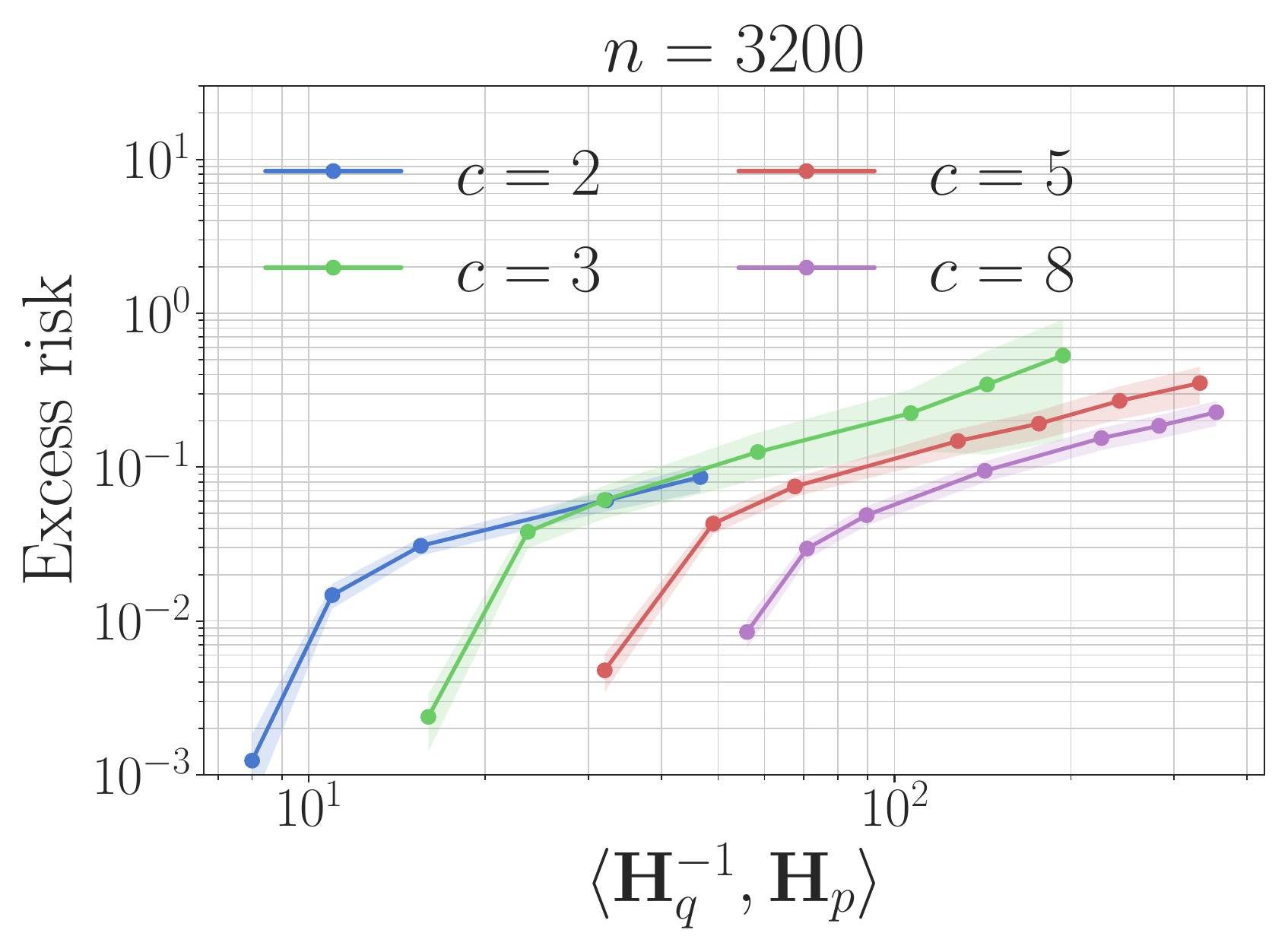}};
\node[inner sep=0pt] (b1) at (0,-3.2) {\includegraphics[width=4.6cm,height = 3.3cm]{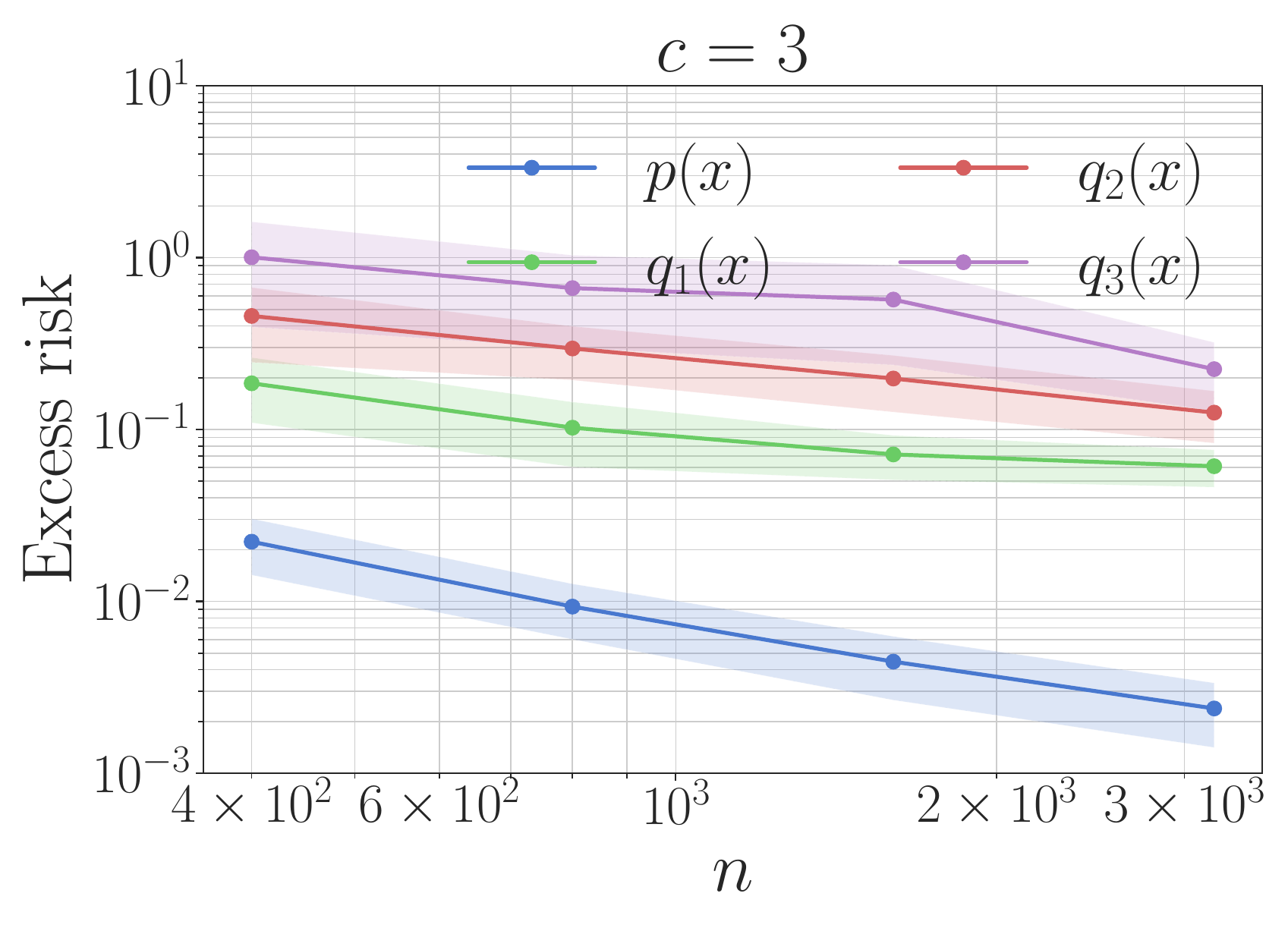}};
\node[inner sep=0pt] (b2) at (4.6,-3.2) {\includegraphics[width=4.6cm,height = 3.3cm]{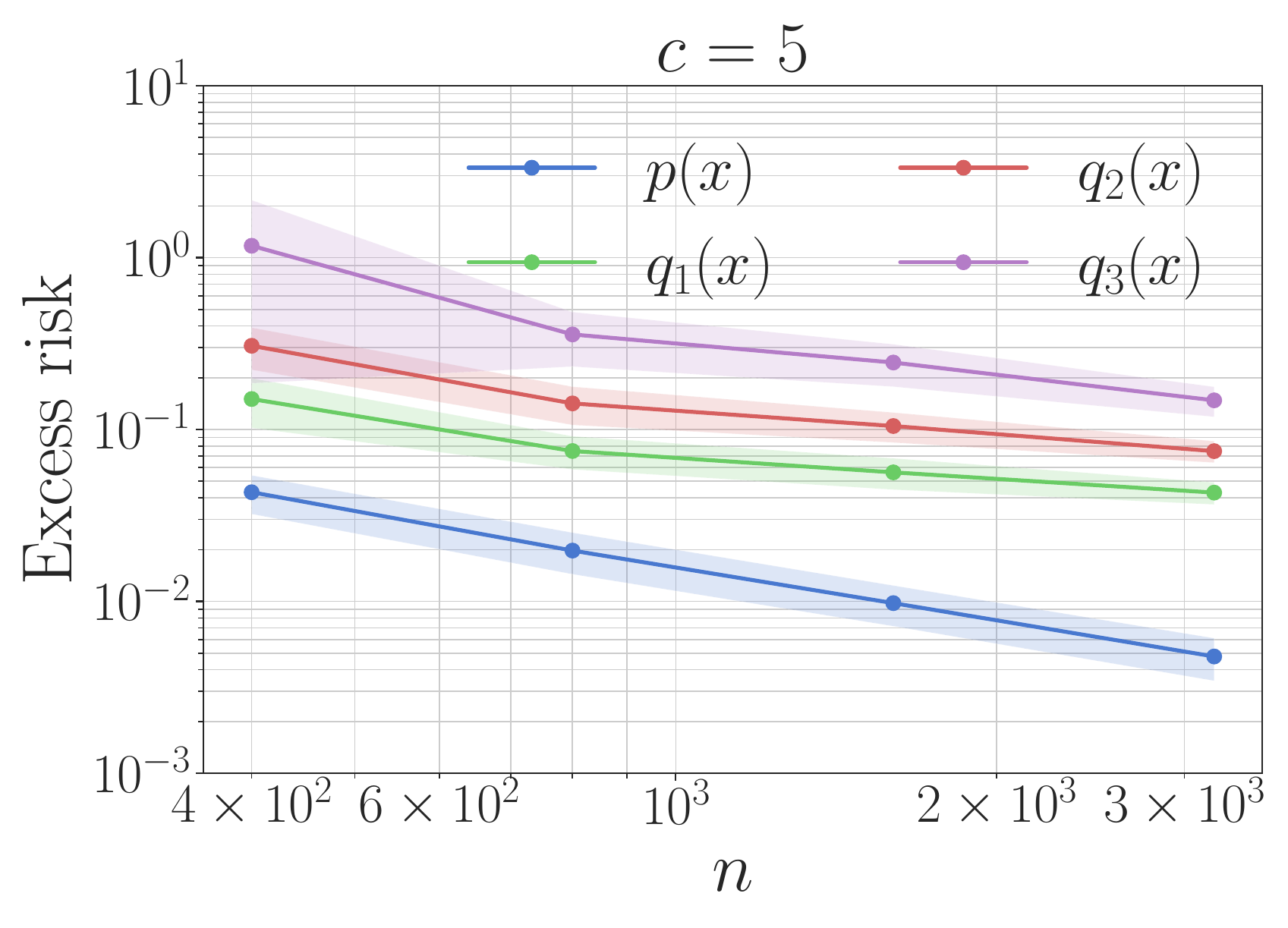}};
\node[inner sep=0pt] (b3) at (9.2,-3.2) {\includegraphics[width=4.6cm,height = 3.3cm]{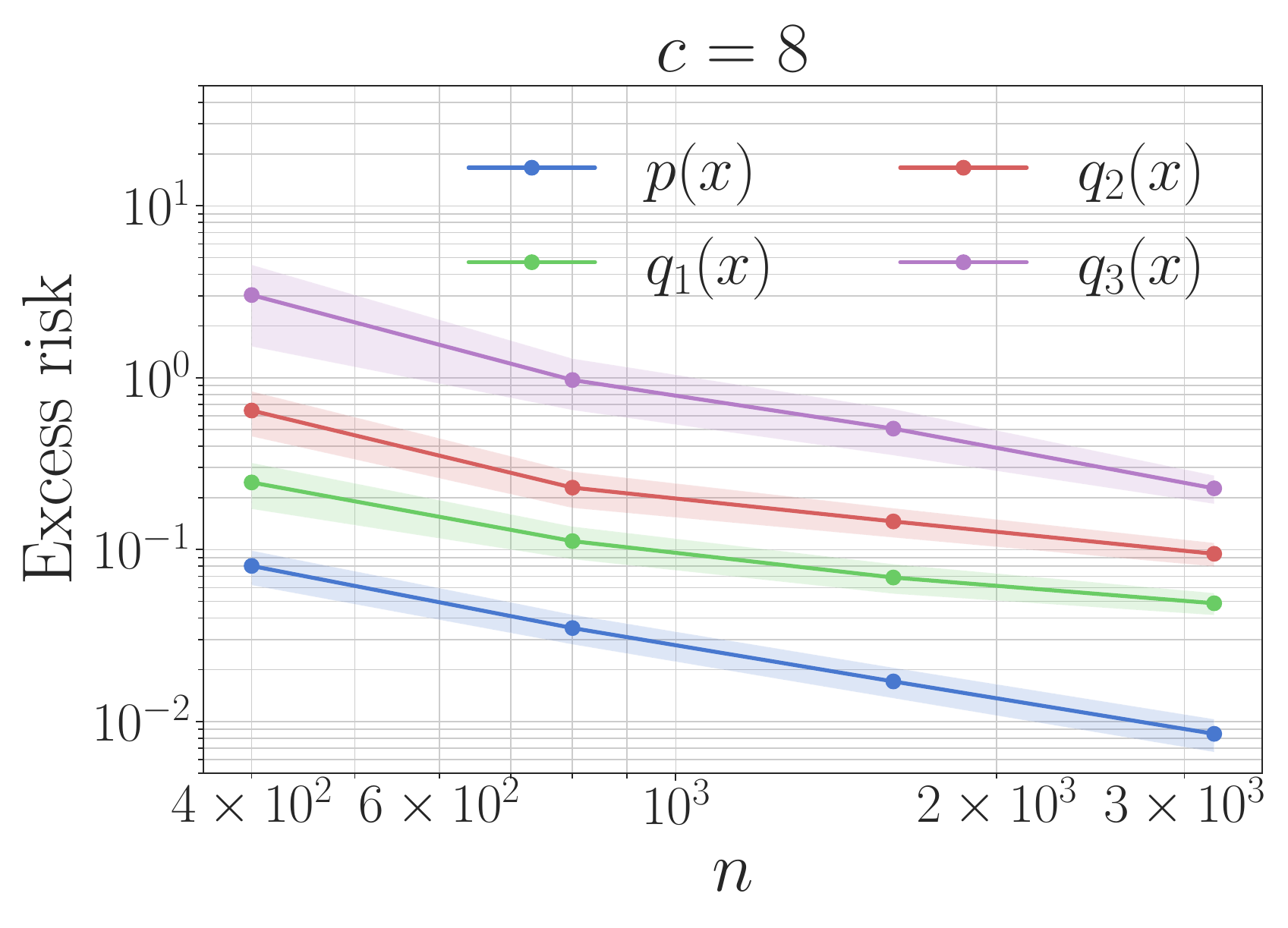}};
\node[inner sep=0pt] (c1) at (0,-6.5) {\includegraphics[width=4.6cm,height = 3.3cm]{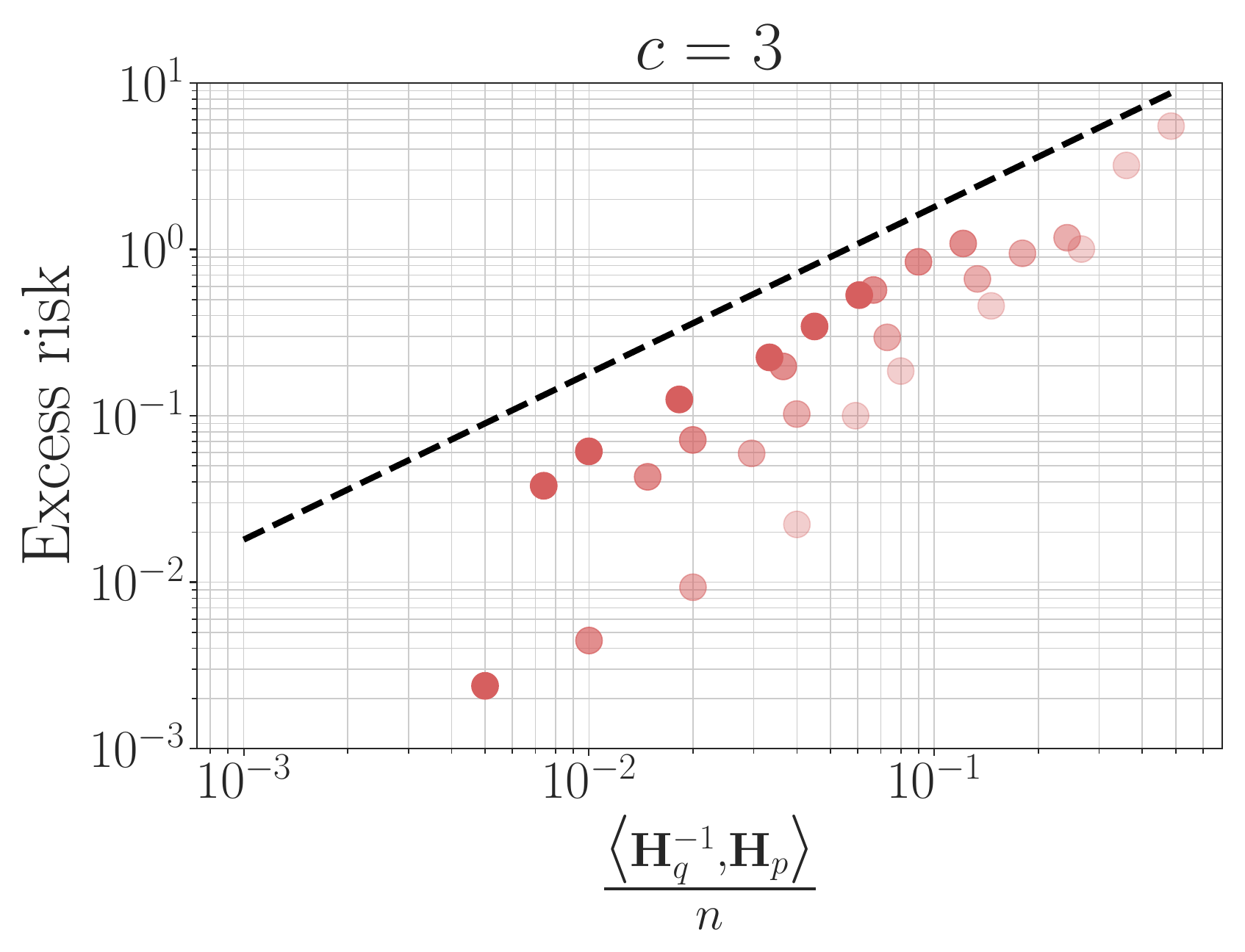}};
\node[inner sep=0pt] (c2) at (4.6,-6.5) {\includegraphics[width=4.6cm,height = 3.3cm]{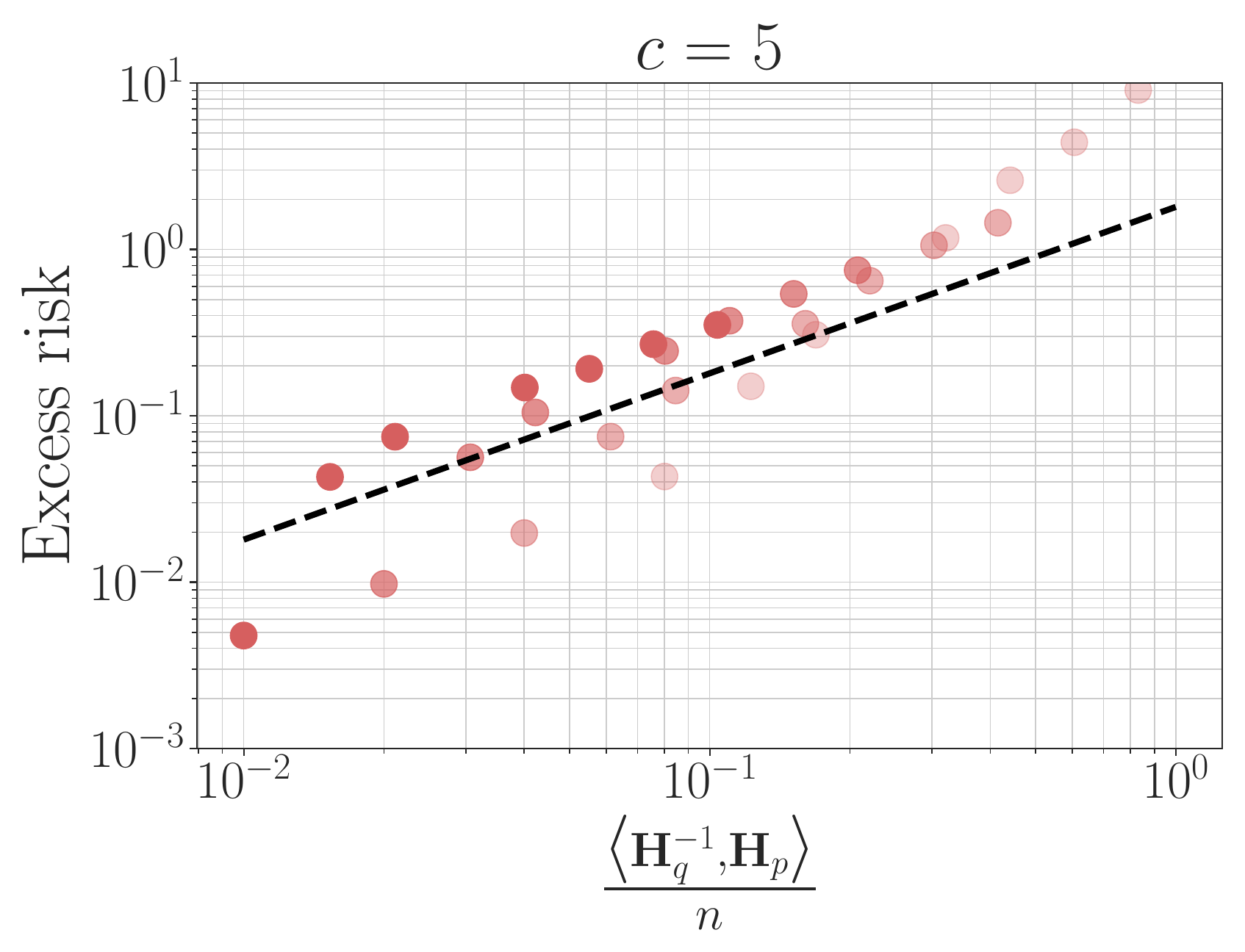}};
\node[inner sep=0pt] (c3) at (9.2,-6.5) {\includegraphics[width=4.6cm,height = 3.3cm]{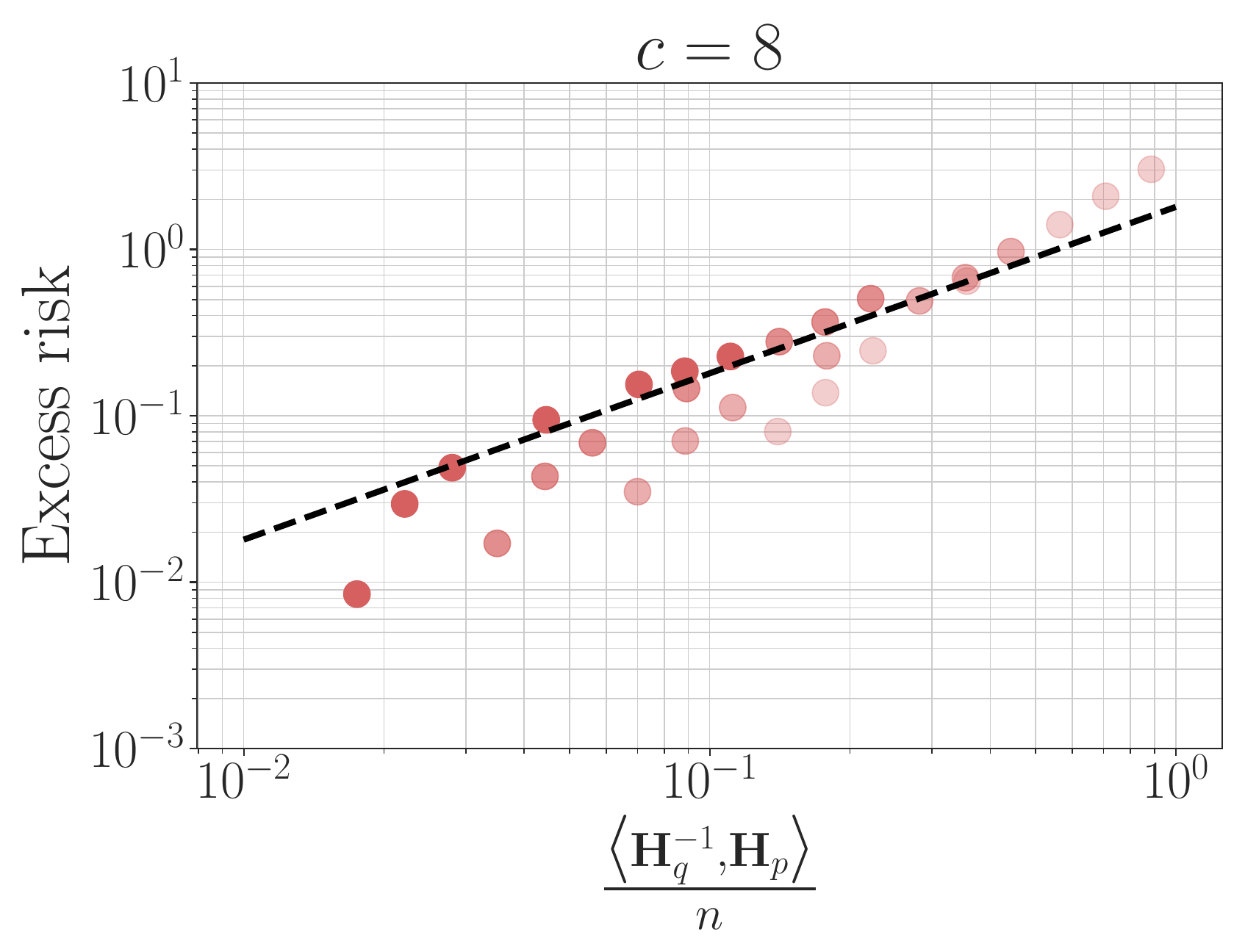}};
\end{tikzpicture}
\caption{Gaussian \textit{translation} tests: excess risk of $p(x)$ vs FIR (upper row), $n$ (middle row) and $\mathrm{FIR}/n$ (lower row). For all plots in the lower row, the less transparent dots represent the larger sample size $n$, the black dashed lines represent linear relation $y=\frac{9}{5} x$. }
\label{fig:gaussian-translation}
\end{figure}

\paragraph{Demonstration of excess risk bounds for $p(x)$ (\Cref{eq:sub-thm-risk-detail}).} In \Cref{sec:experiments}, we have introduced the different types of $q(x)$ used in dilation and translation tests. In \Cref{fig:gaussian-sigma-trace}, we plot the relations of $\lambda_{\max}(\hq^{-1} \hp)$ (which is $\sigma$ in \Cref{thm:sub-thm-detail}) and FIR ($\langle  \hq^{-1}, \hp\rangle$. For the dilation tests, we present the plots of  excess risk of $p(x)$ vs FIR, $n$, and $\mathrm{FIR}/n$ respectively in \Cref{fig:gaussian-dilation}.  We plot the results for translation tests in \Cref{fig:gaussian-translation}. 
As mentioned in Section \ref{sec:experiments}, these results are consistent to the bounds we derived in \Cref{eq:sub-thm-risk-detail}. One interesting finding is that from the lower rows of \Cref{fig:gaussian-dilation,fig:gaussian-translation}, the excess risk is upper bounded by $\frac{9}{5}\frac{\mathrm{FIR}}{n}$ when $n$ is large. This observation is consistent with the upper bound we derived in the bounded domain case (\Cref{eq:bound-domain-9-5bound} in \Cref{appendix:bounded-domain}).

\begin{figure}[!t]
    \centering
\begin{tikzpicture}
    \node[inner sep=0pt] (a1) at (0,0) {\includegraphics[width=4.5cm]{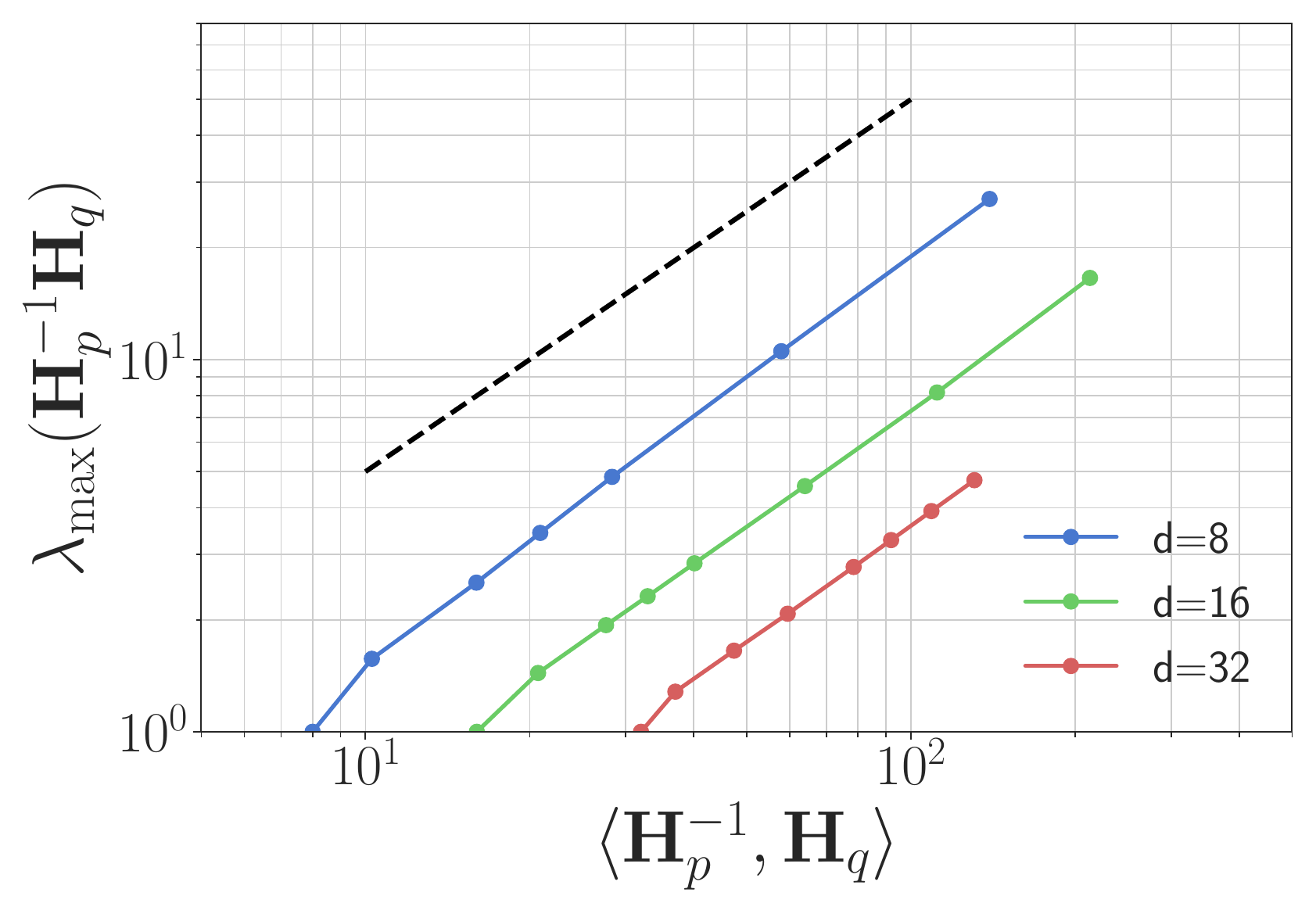}};
   \node[inner sep=0pt] (a2) at (6,0)  {\includegraphics[width=4.5cm]{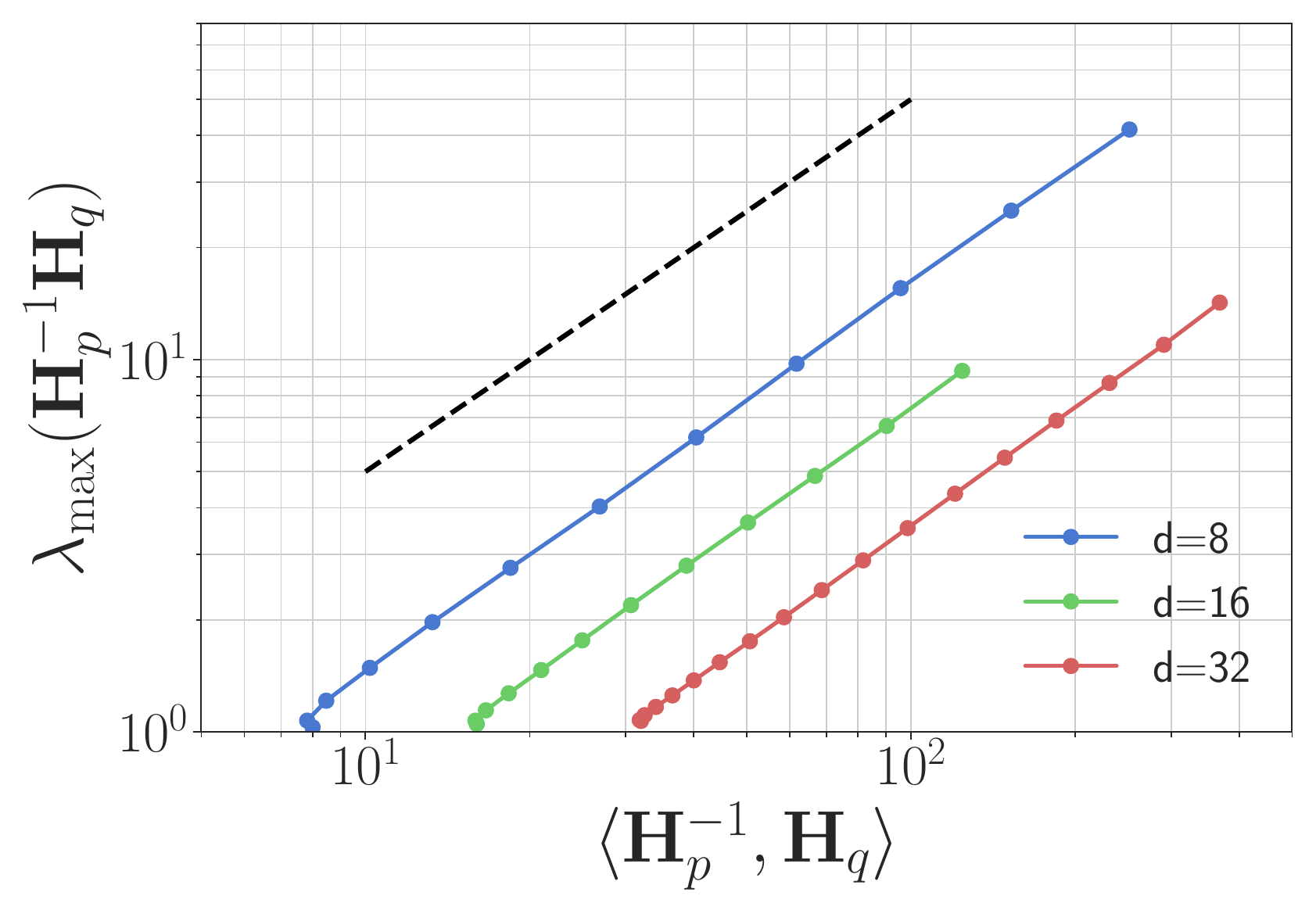}};
\end{tikzpicture}
    \caption{ $\lambda_{\max} (\hq^{-1} \hp)$ vs  $\Tr(\hq^{-1} \hp)$ in Multivariate Laplace tests and t-distribution tests.}
    \label{fig:trace_sigma_laplace_t}
\end{figure}

\begin{figure}[!t]
\centering
  \footnotesize
\begin{tikzpicture}
\node[inner sep=0pt] (a1) at (0,0) {\includegraphics[width=4.6cm]{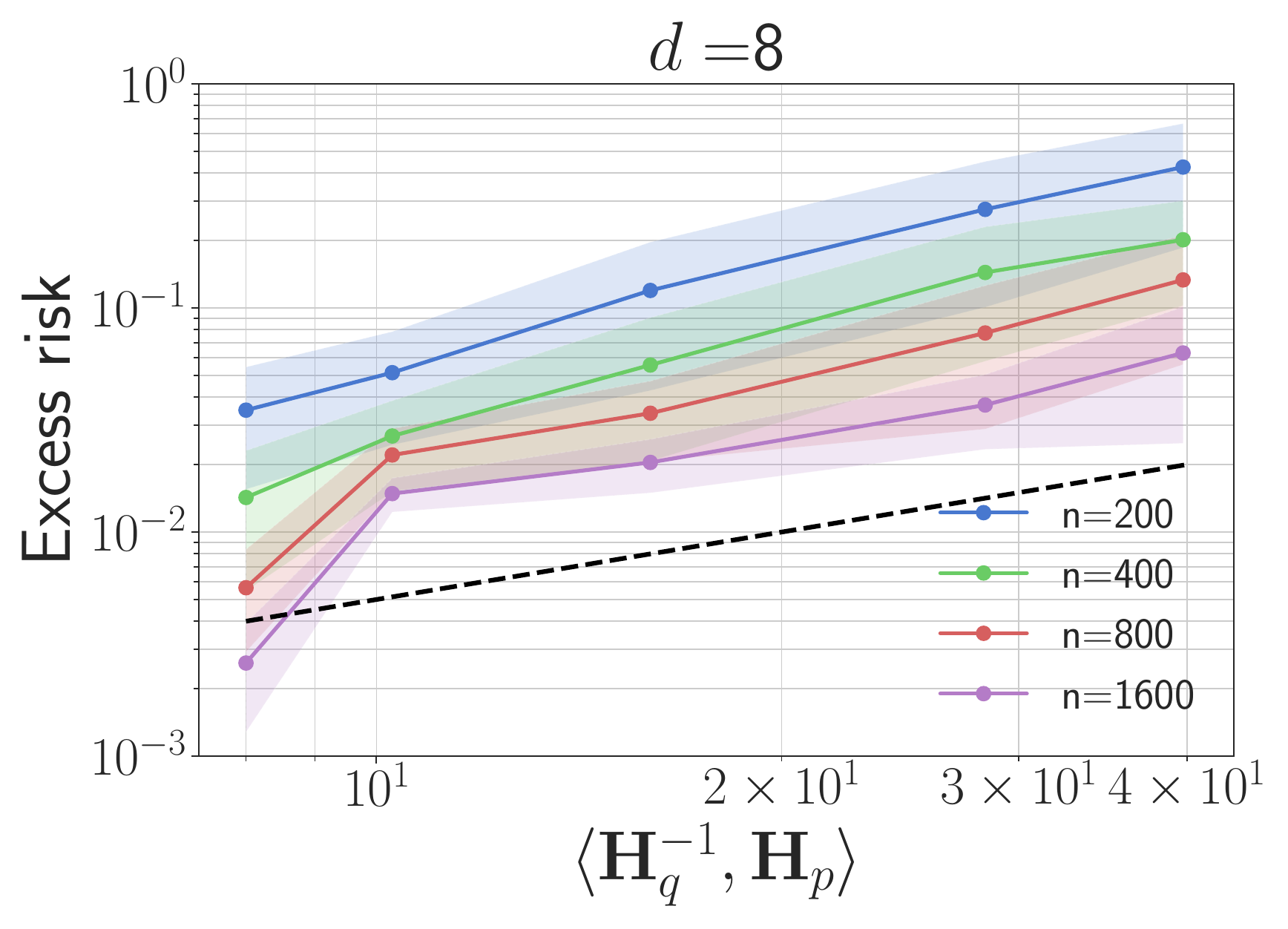}};
\node[inner sep=0pt] (a2) at (4.6,0) {\includegraphics[width=4.6cm]{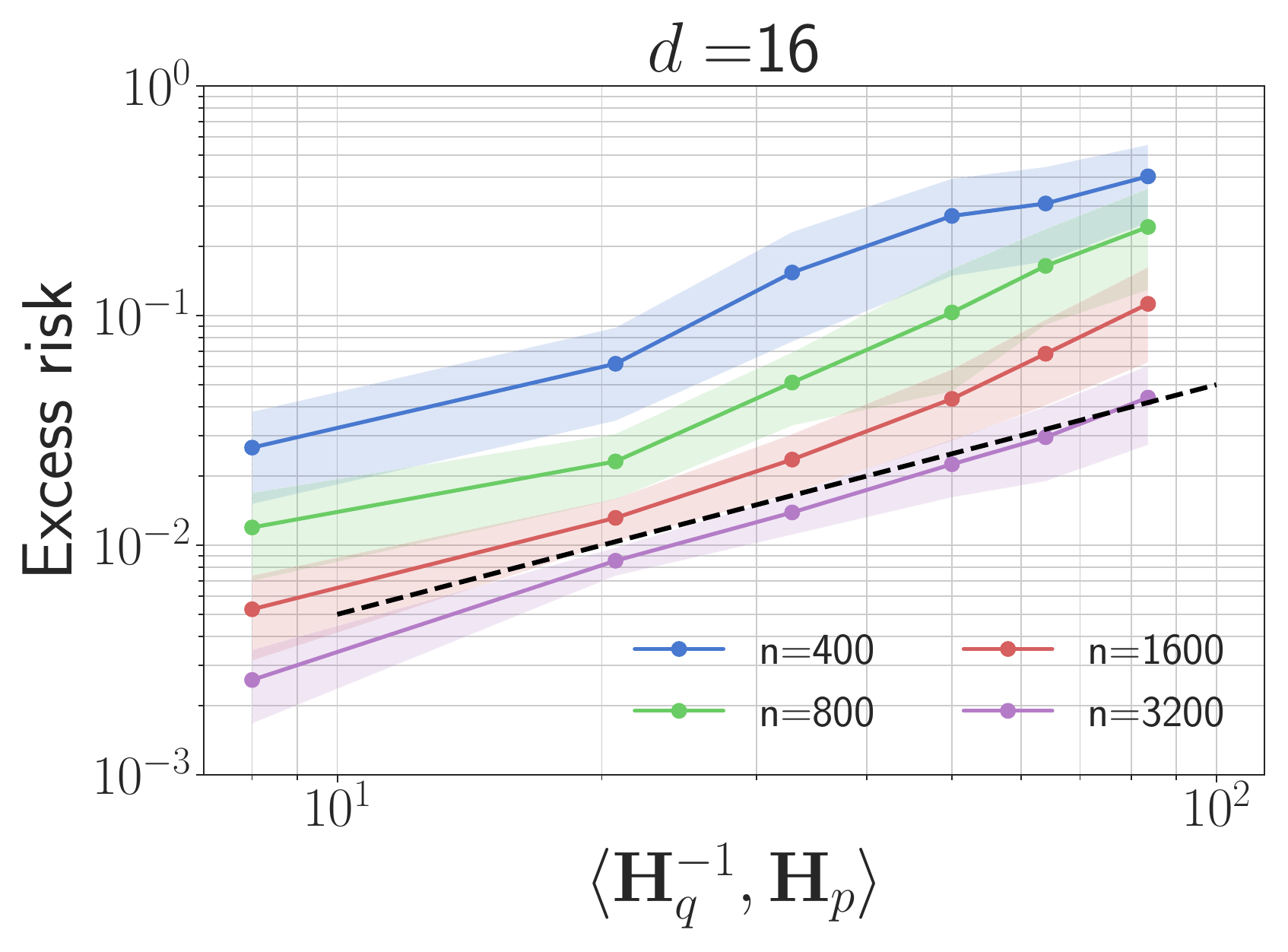}};
\node[inner sep=0pt] (a3) at (9.2,0) {\includegraphics[width=4.6cm]{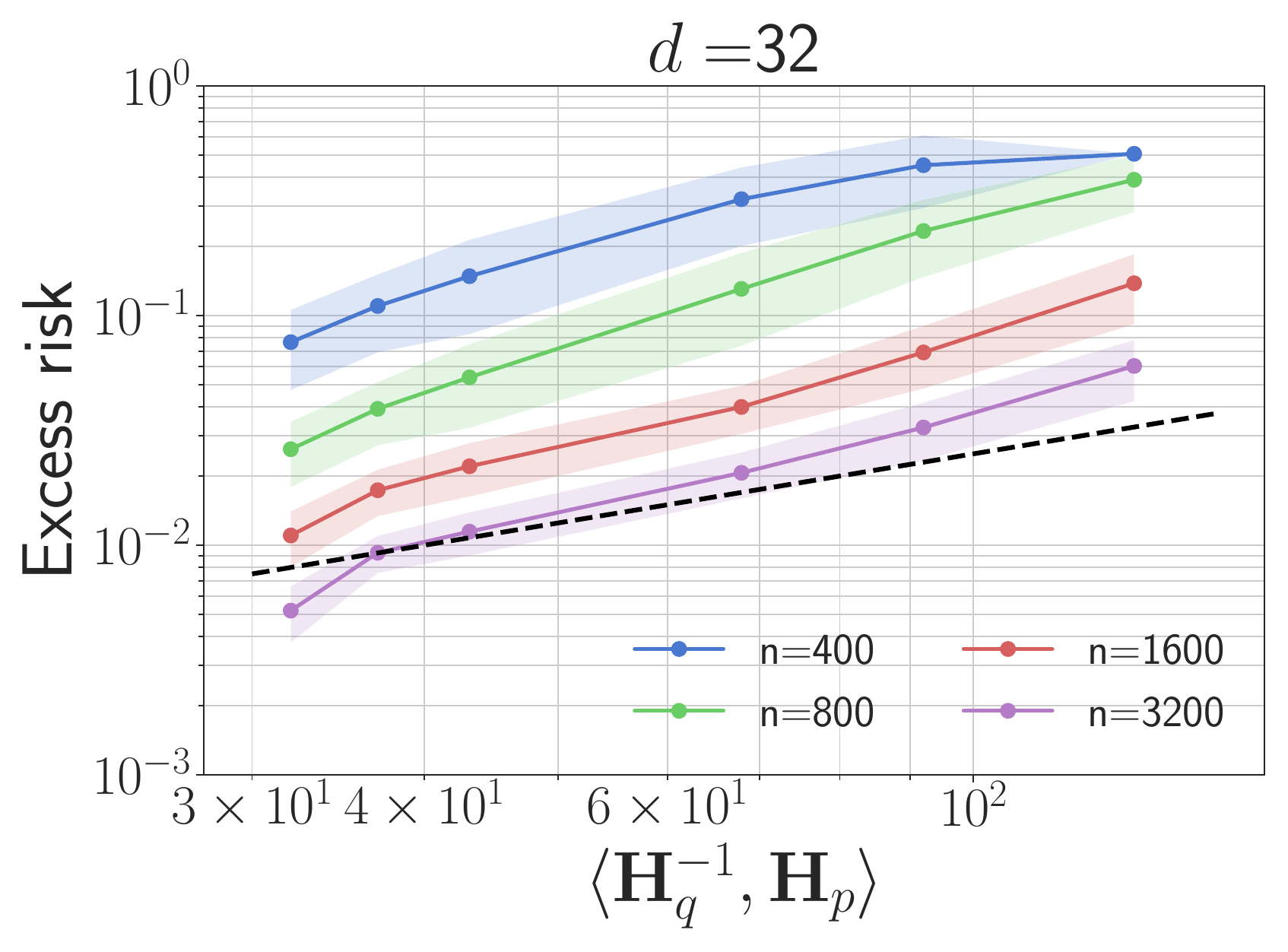}};
\node[inner sep=0pt] (b1) at (0,-3.2){\includegraphics[width=4.6cm]{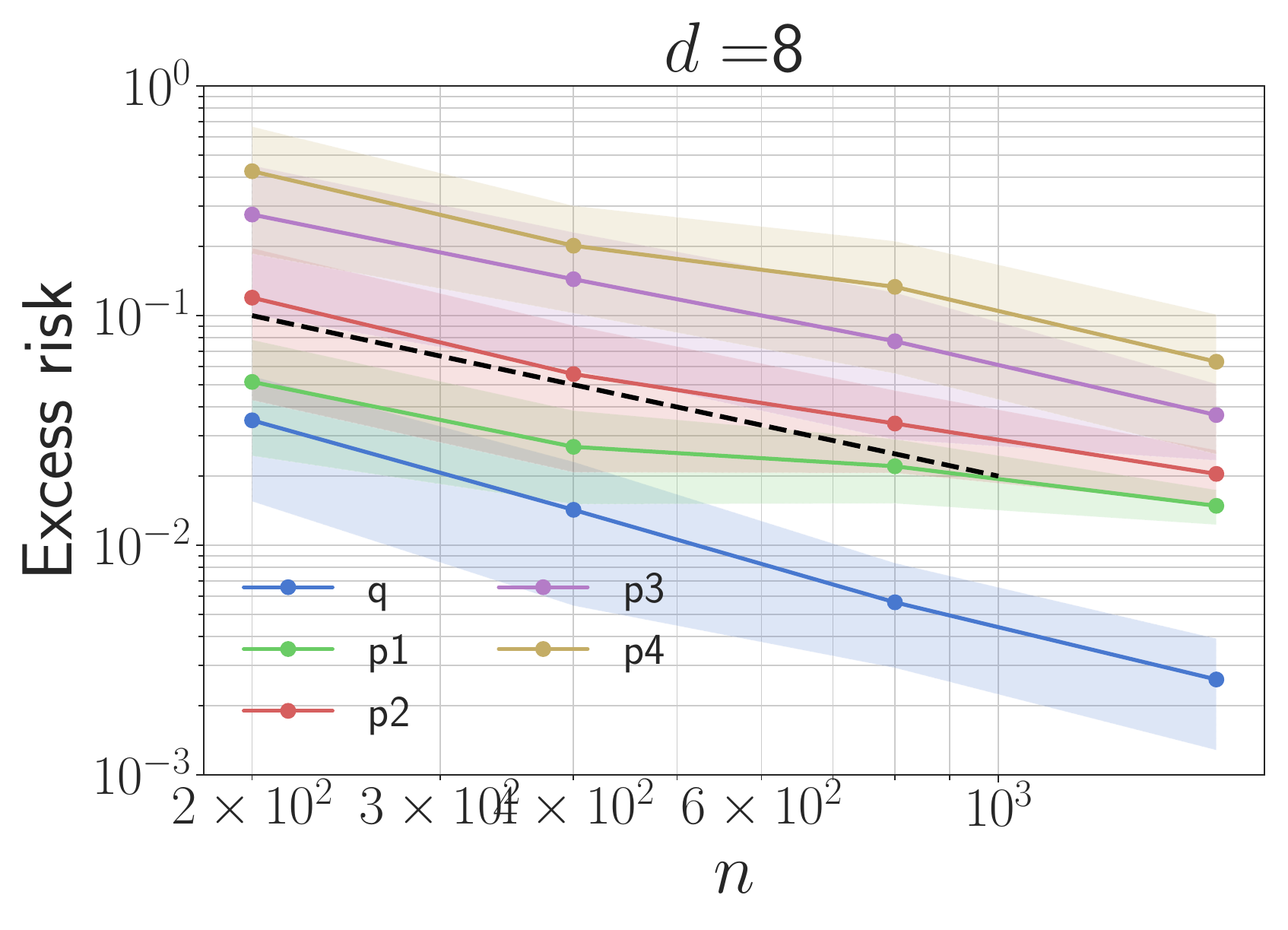}};
\node[inner sep=0pt] (b2) at (4.6,-3.2) 
{\includegraphics[width=4.6cm, ]{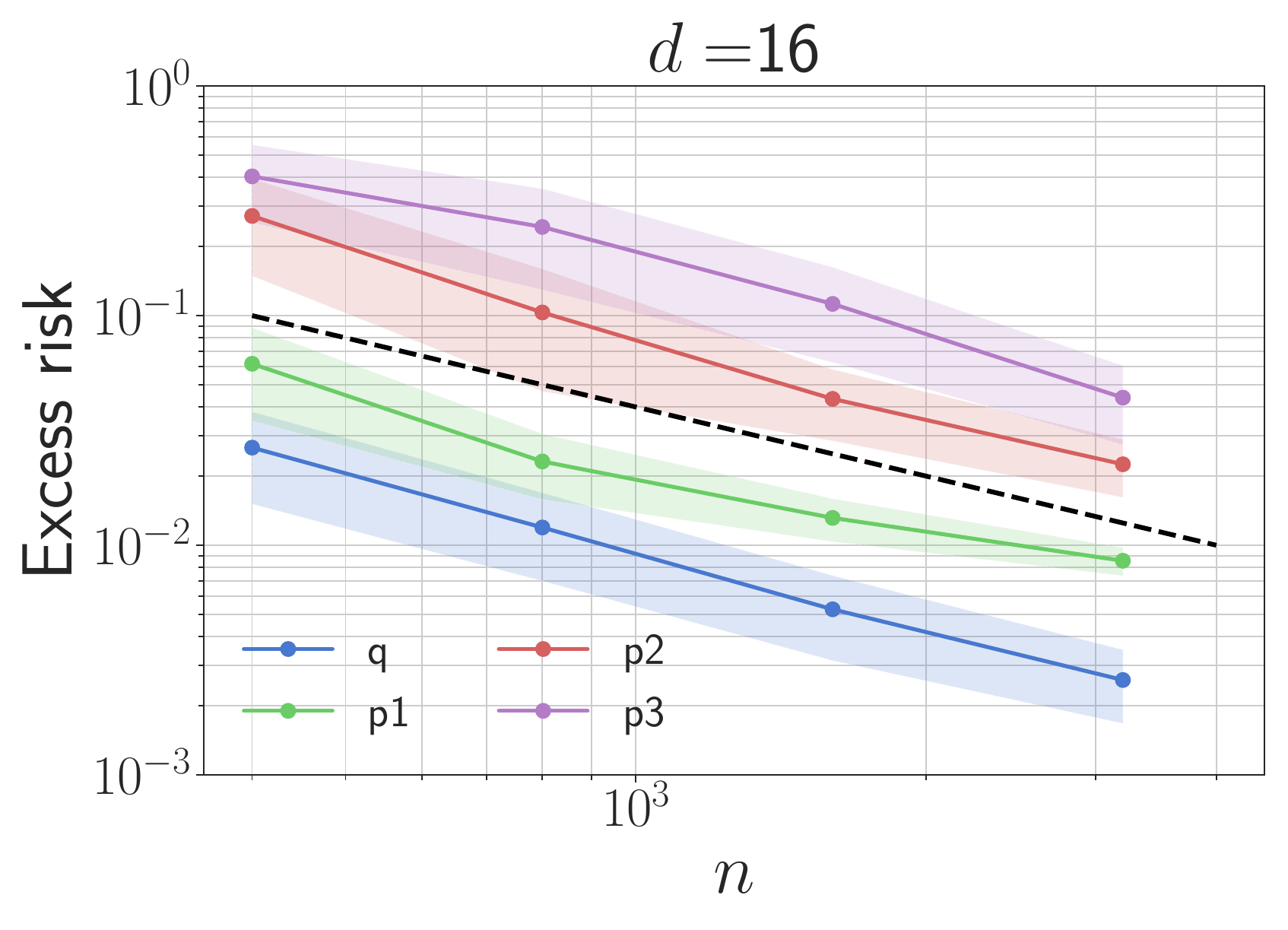}};
\node[inner sep=0pt] (b3) at (9.2,-3.2) {\includegraphics[width=4.6cm]{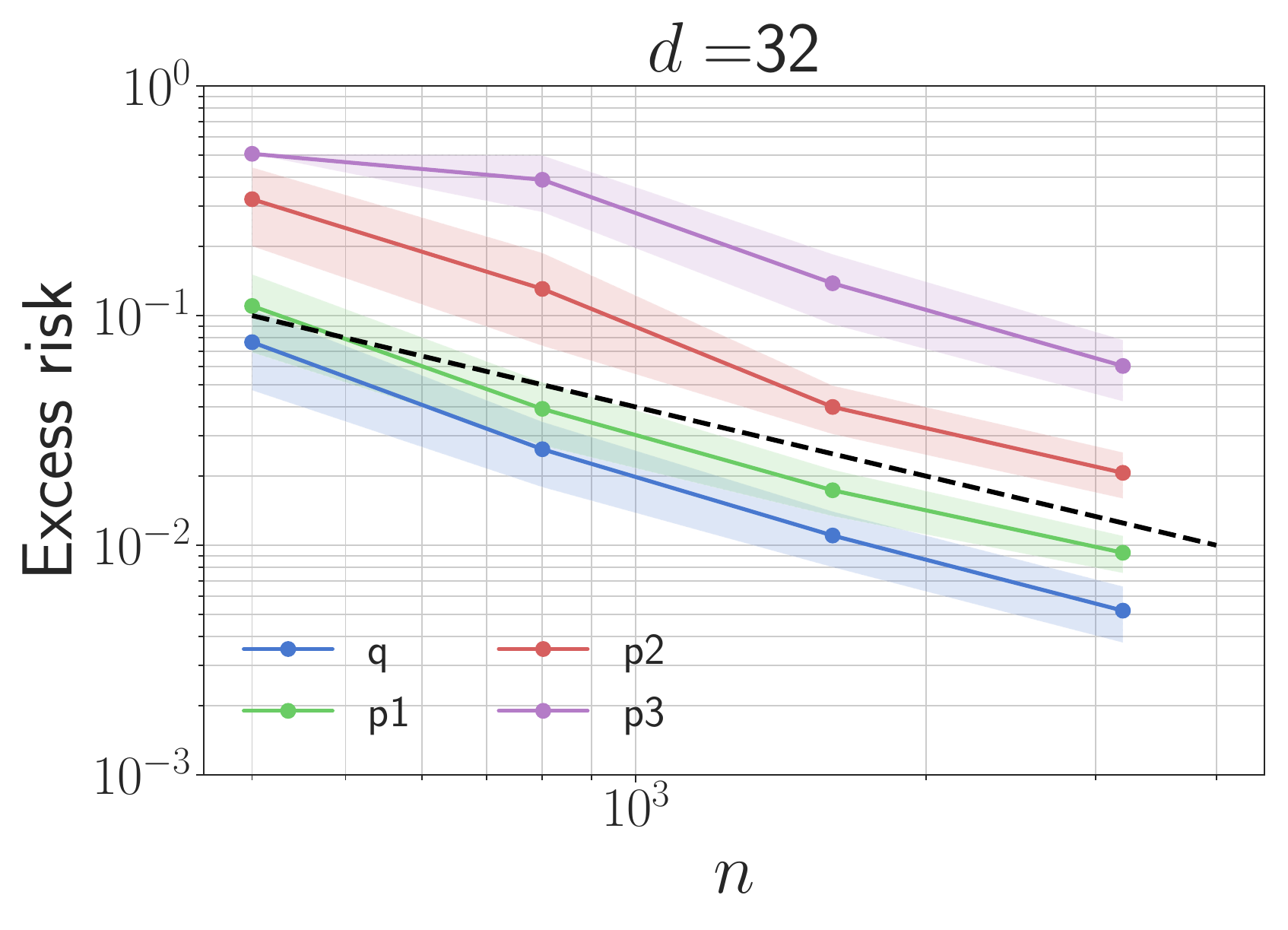}};
\node[inner sep=0pt] (C1) at (0,-6.4) {\includegraphics[width=4.6cm]{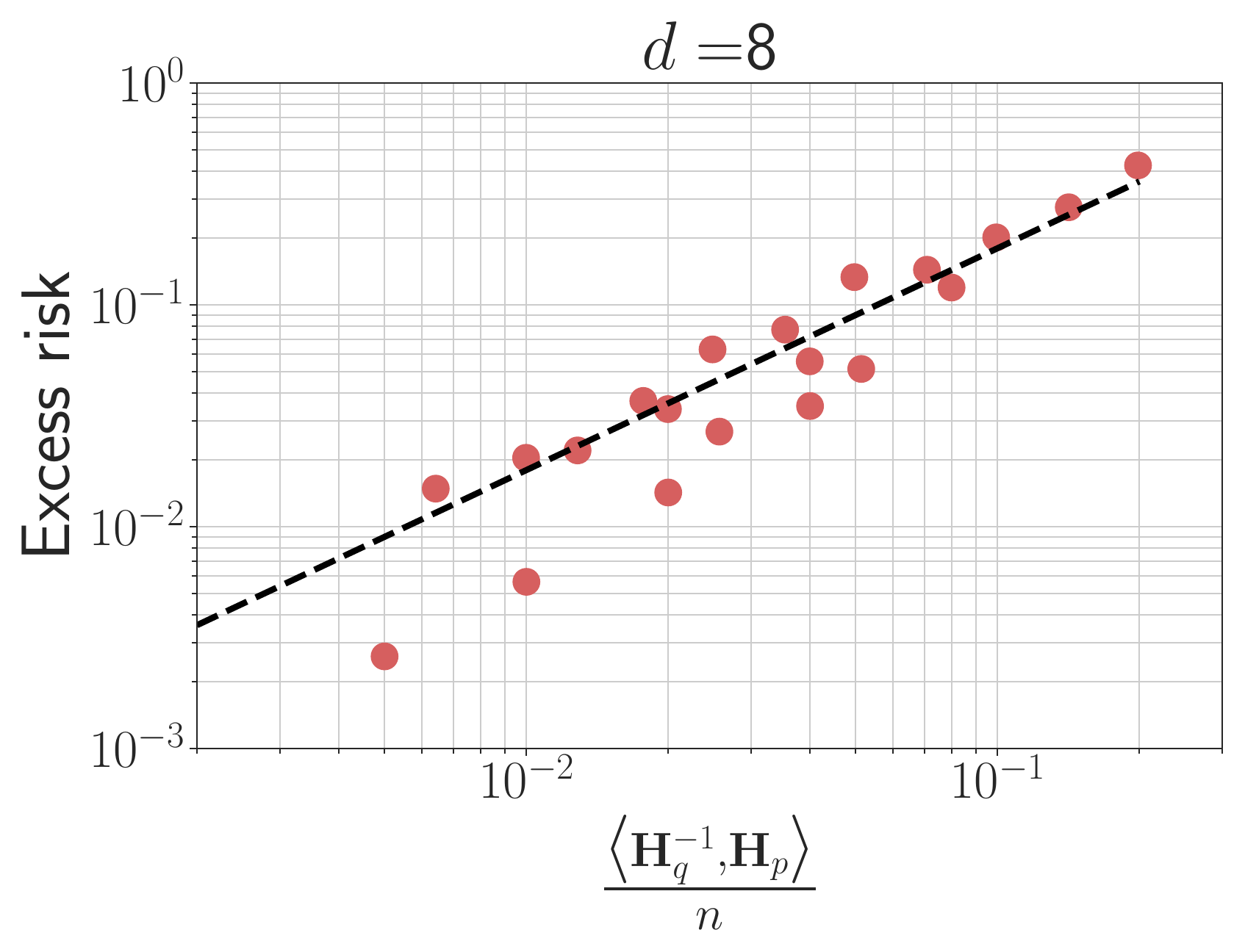}};
\node[inner sep=0pt] (C2) at (4.6,-6.4) {\includegraphics[width=4.6cm]{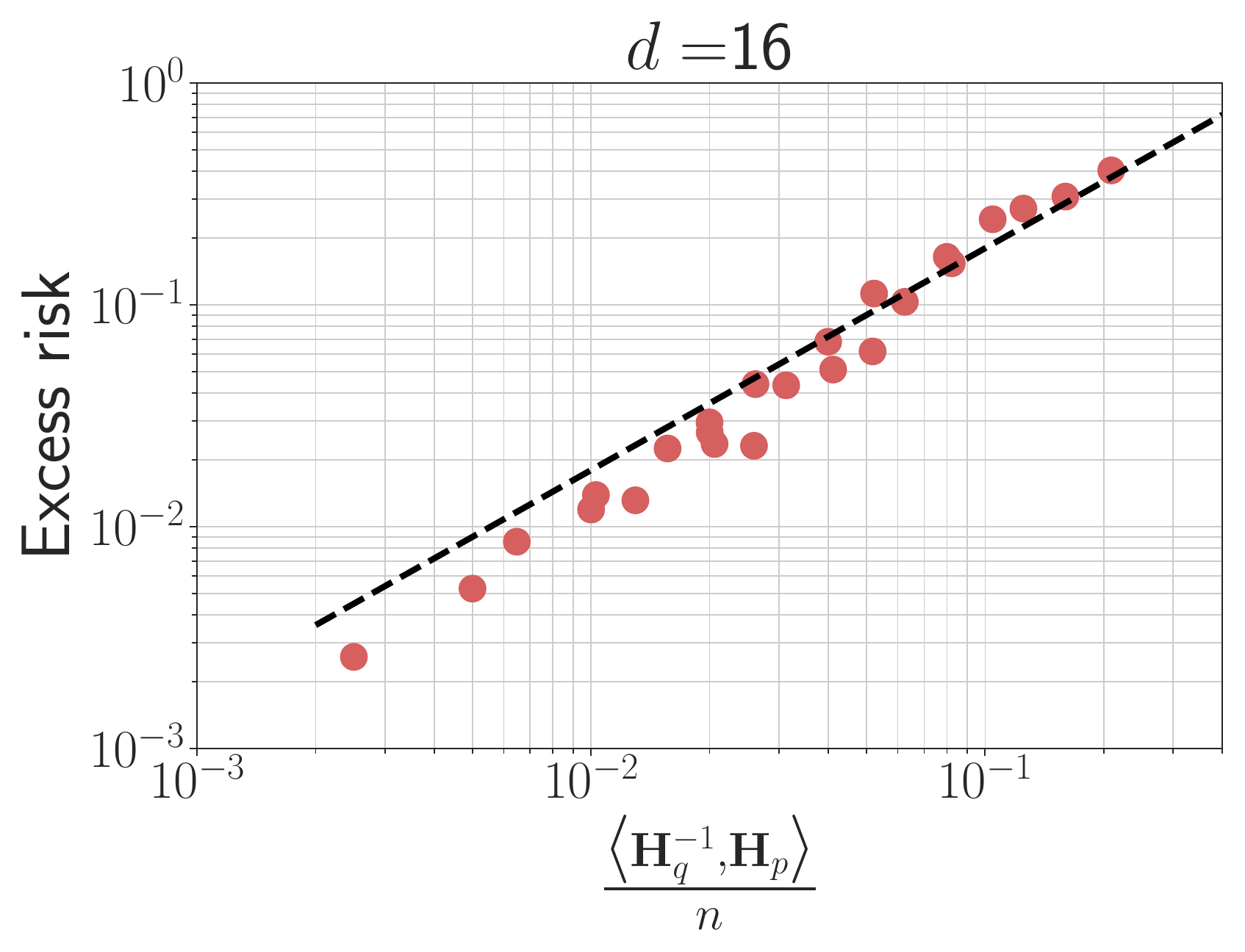}};
\node[inner sep=0pt] (C3) at (9.2,-6.4) {\includegraphics[width=4.6cm]{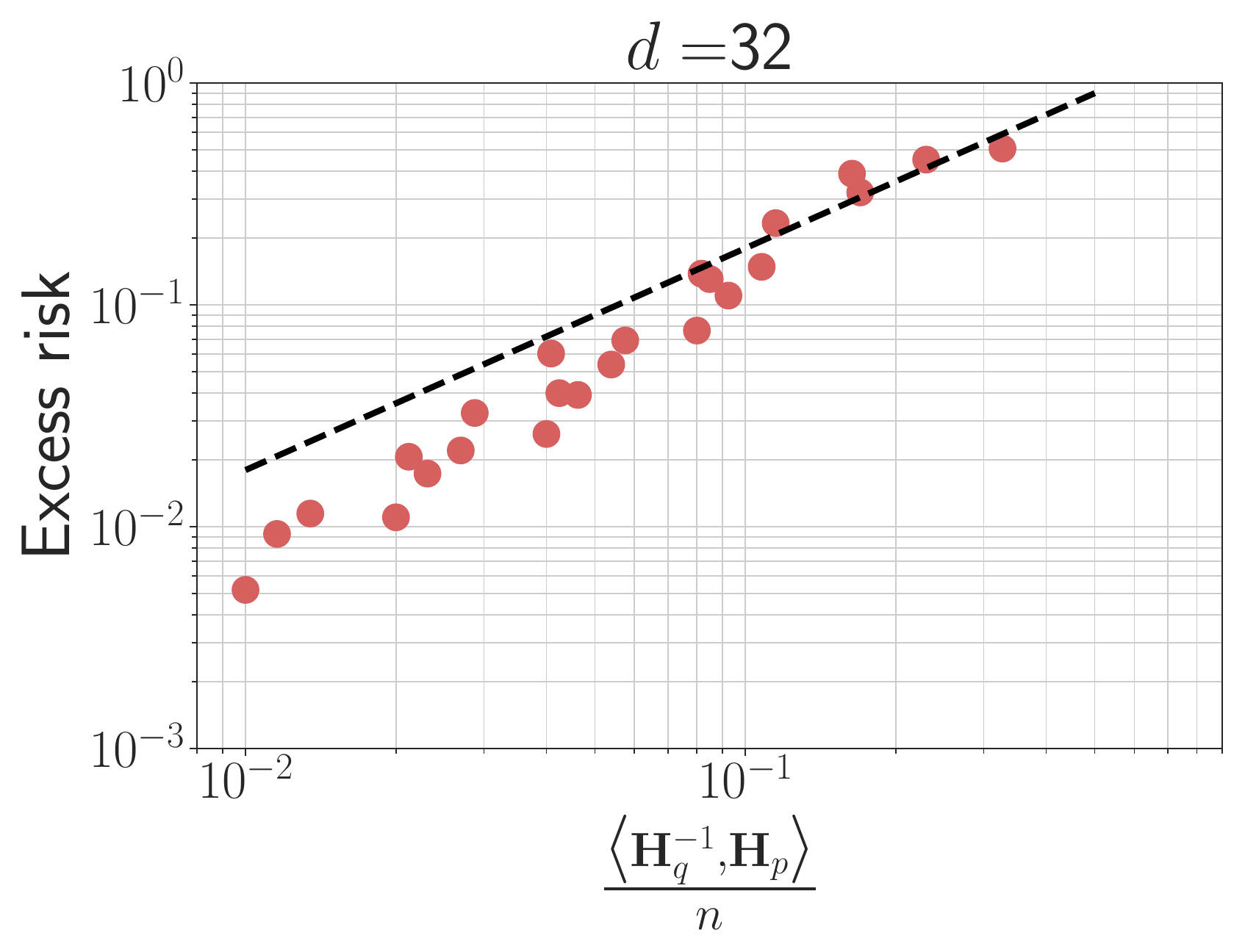}};
\end{tikzpicture}
     \caption{Multivariate Laplace distribution test: excess risk of $p(x)$  vs  FIR (upper), $n$ (middle),  and $\frac{\mathrm{FIR}}{n}$ (lower), the black dashed lines have slope 1 in upper and lower rows , and slope -1 in the middle row. }
          \label{fig:syn-risk-laplace}
\end{figure}

\begin{figure}[!t]
\centering
  \footnotesize
\begin{tikzpicture}
\node[inner sep=0pt] (a1) at (0,0) {\includegraphics[width=4.6cm]{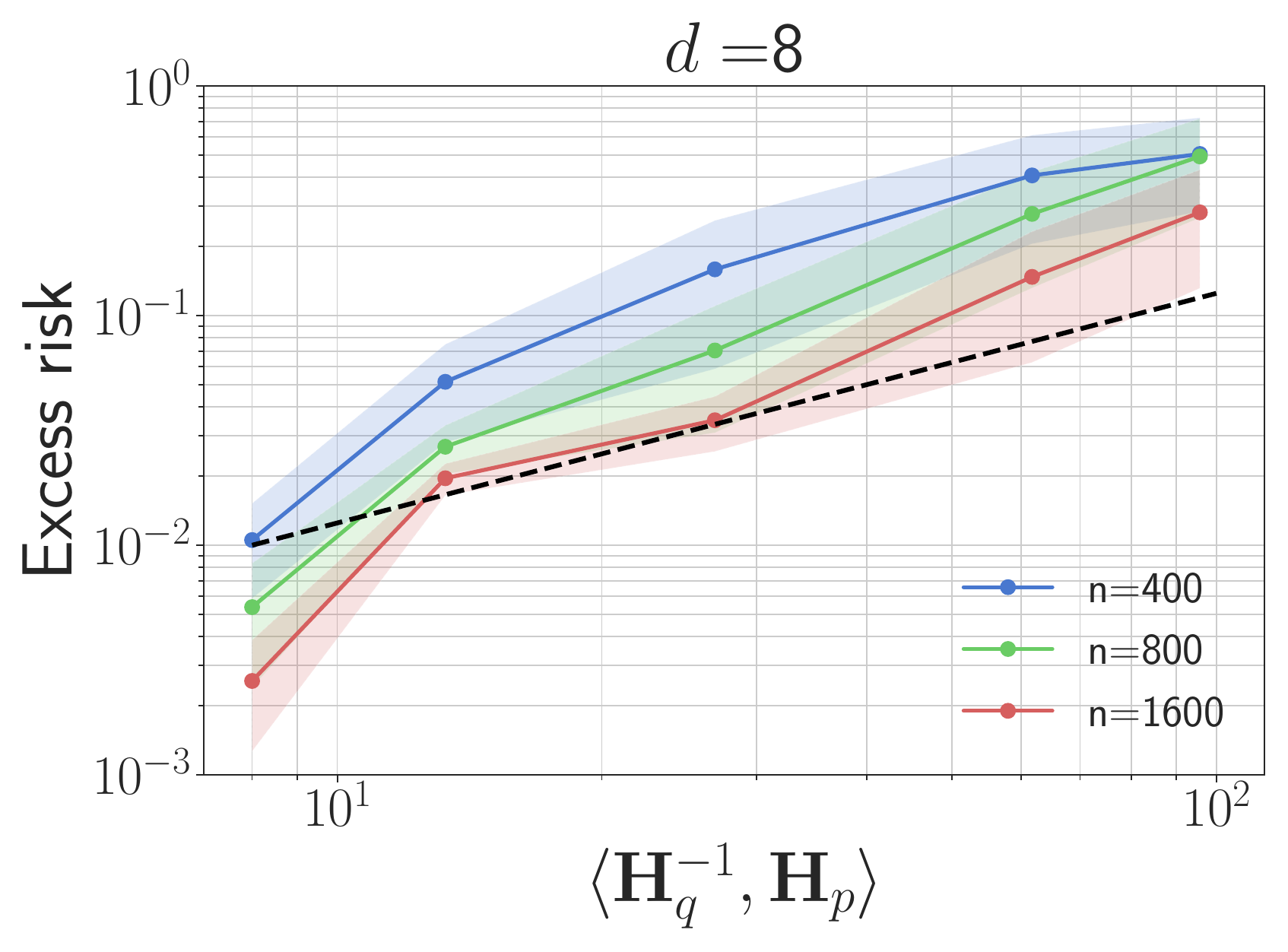}};
\node[inner sep=0pt] (a2) at (4.6,0) {\includegraphics[width=4.6cm]{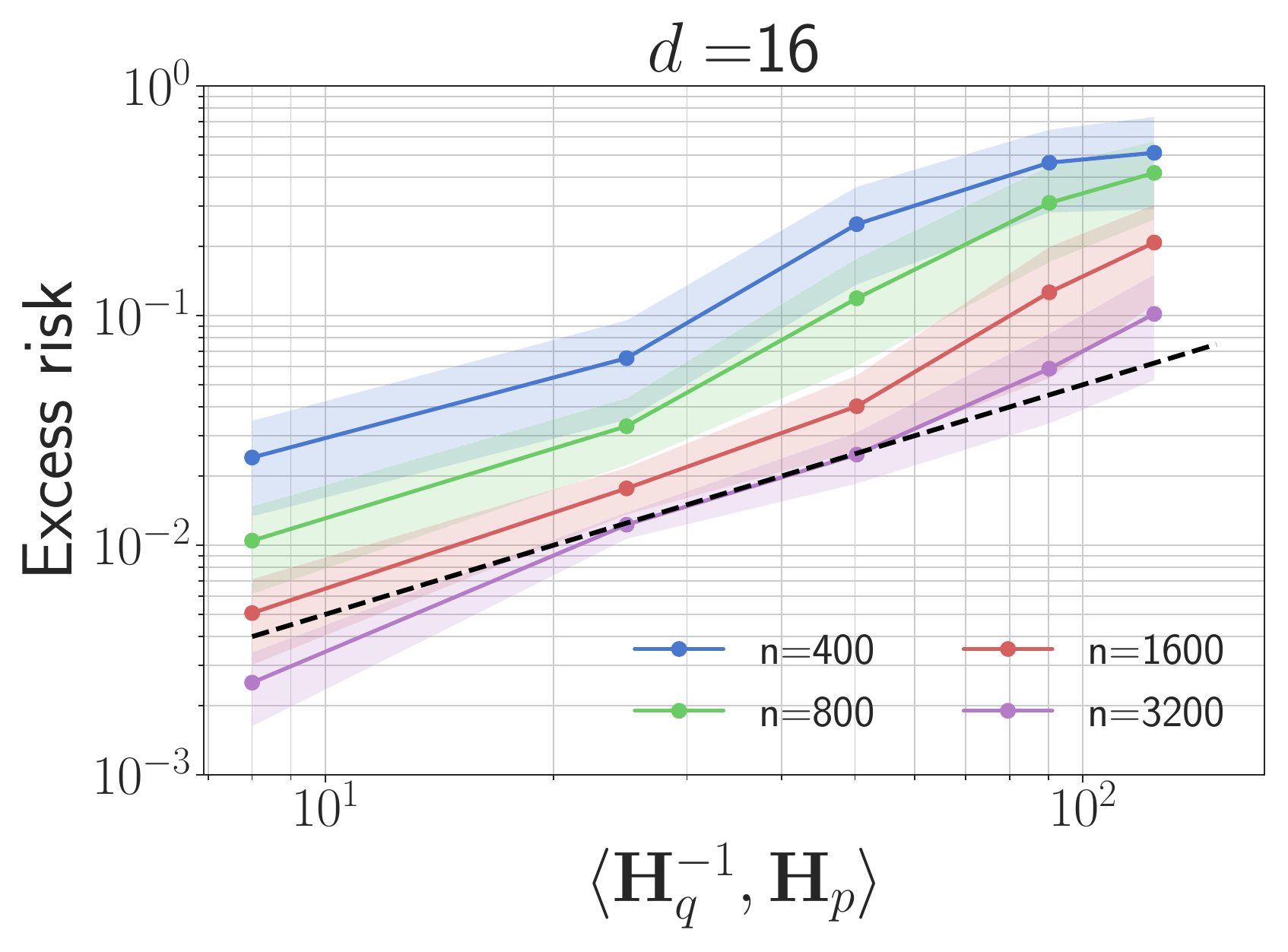}};
\node[inner sep=0pt] (a3) at (9.2,0) {\includegraphics[width=4.6cm]{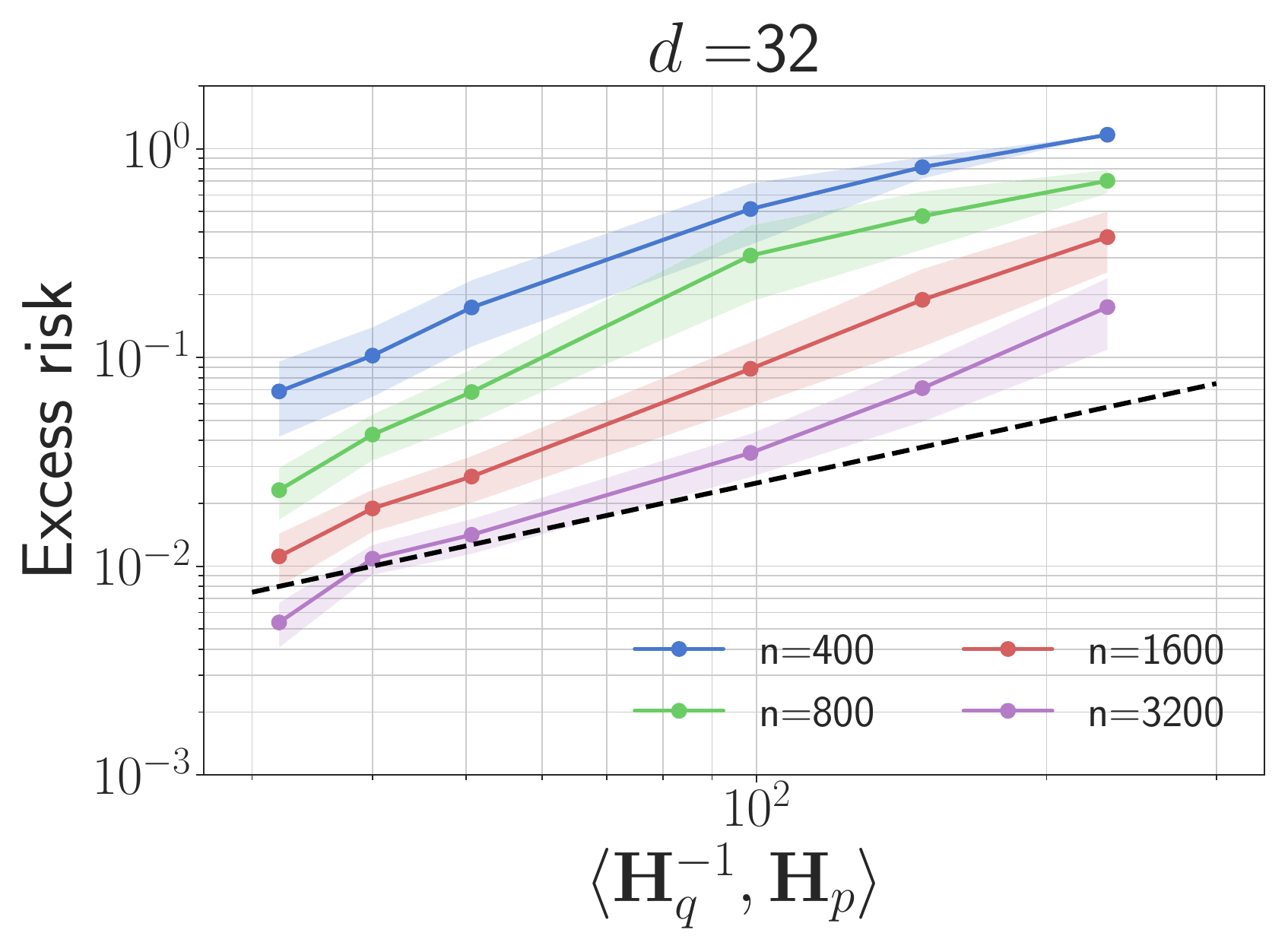}};
\node[inner sep=0pt] (b1) at (0,-3.2){\includegraphics[width=4.6cm]{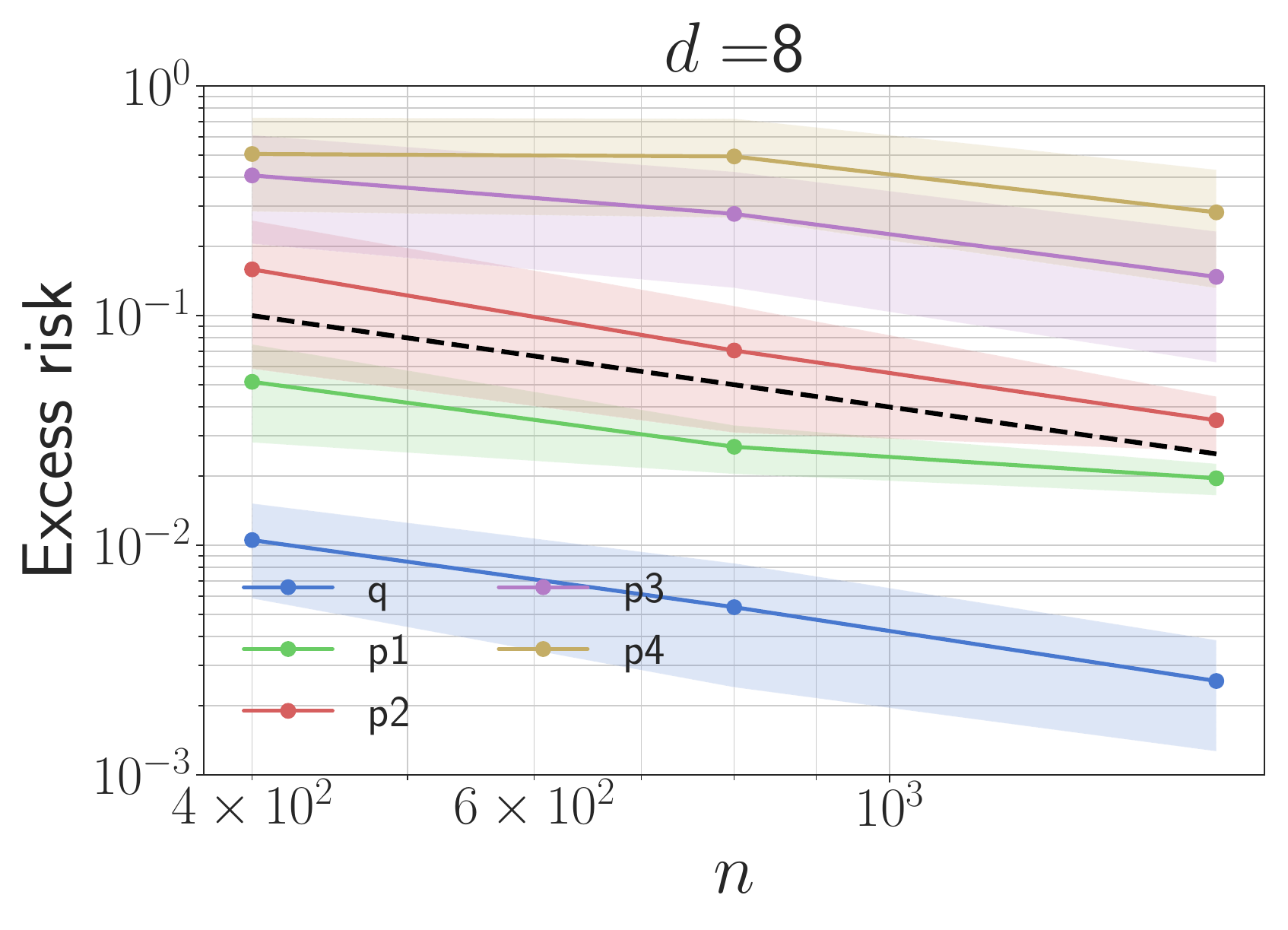}};
\node[inner sep=0pt] (b2) at (4.6,-3.2) 
{\includegraphics[width=4.6cm, ]{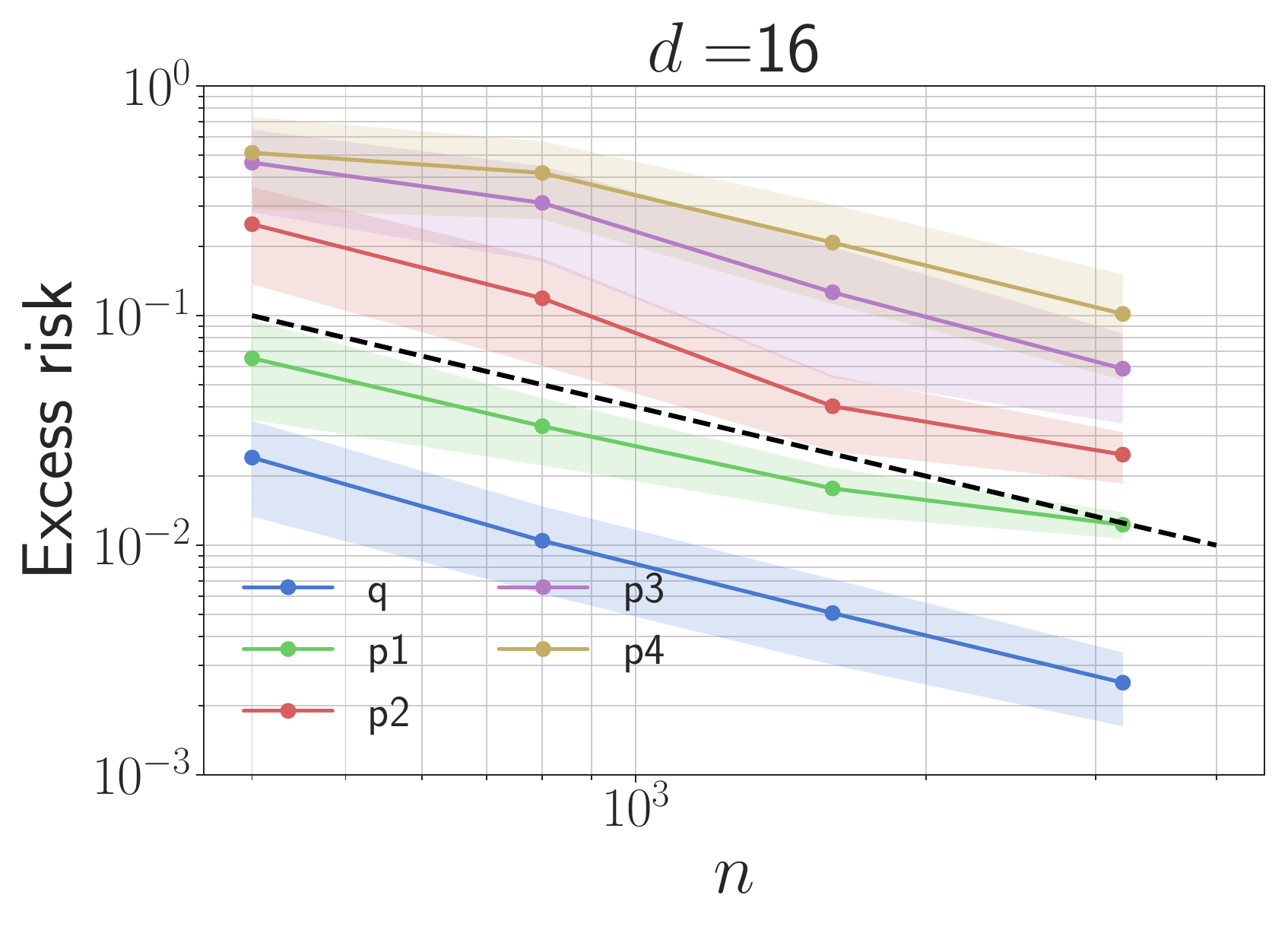}};
\node[inner sep=0pt] (b3) at (9.2,-3.2) {\includegraphics[width=4.6cm]{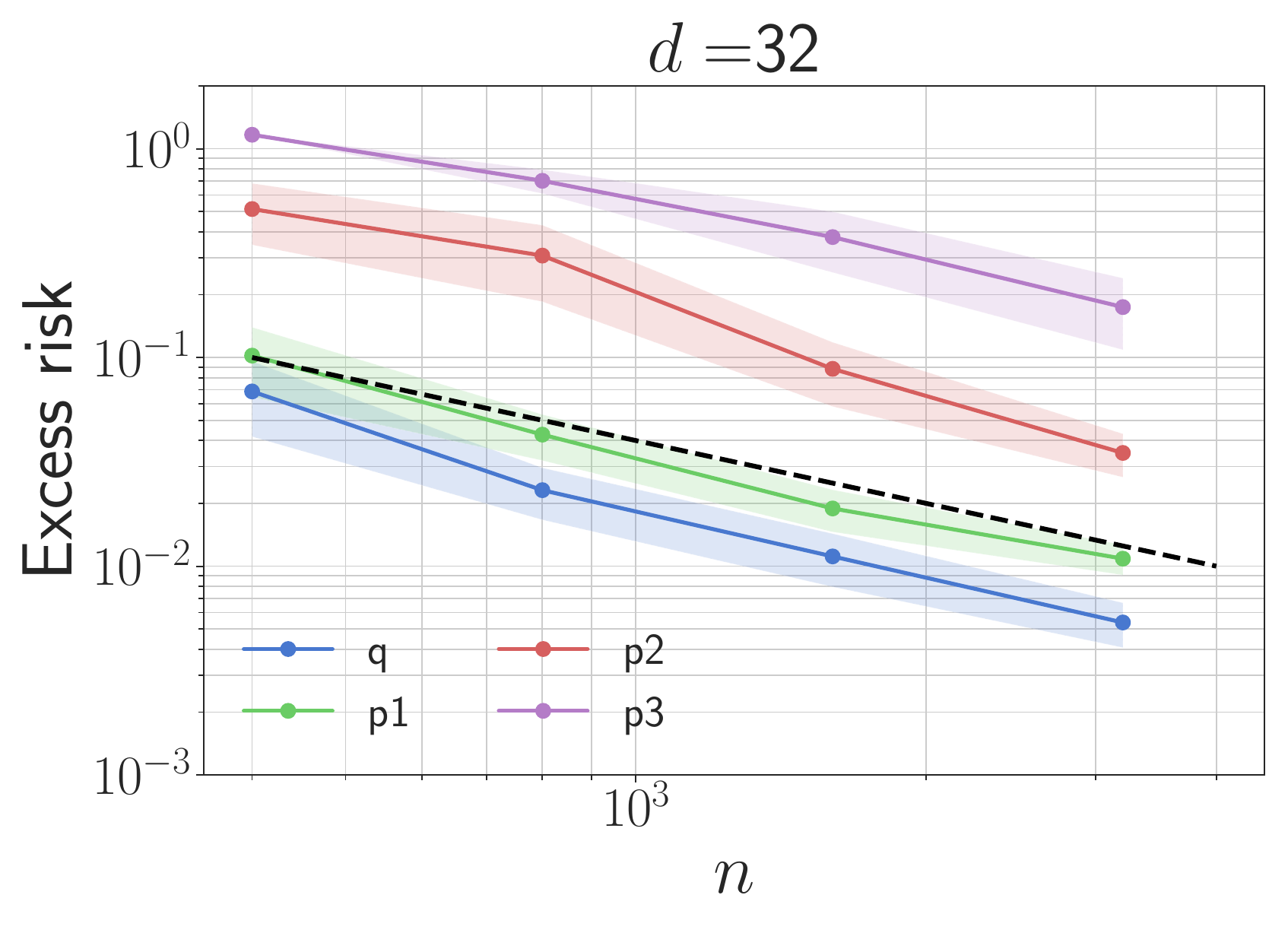}};
\node[inner sep=0pt] (C1) at (0,-6.4) {\includegraphics[width=4.6cm]{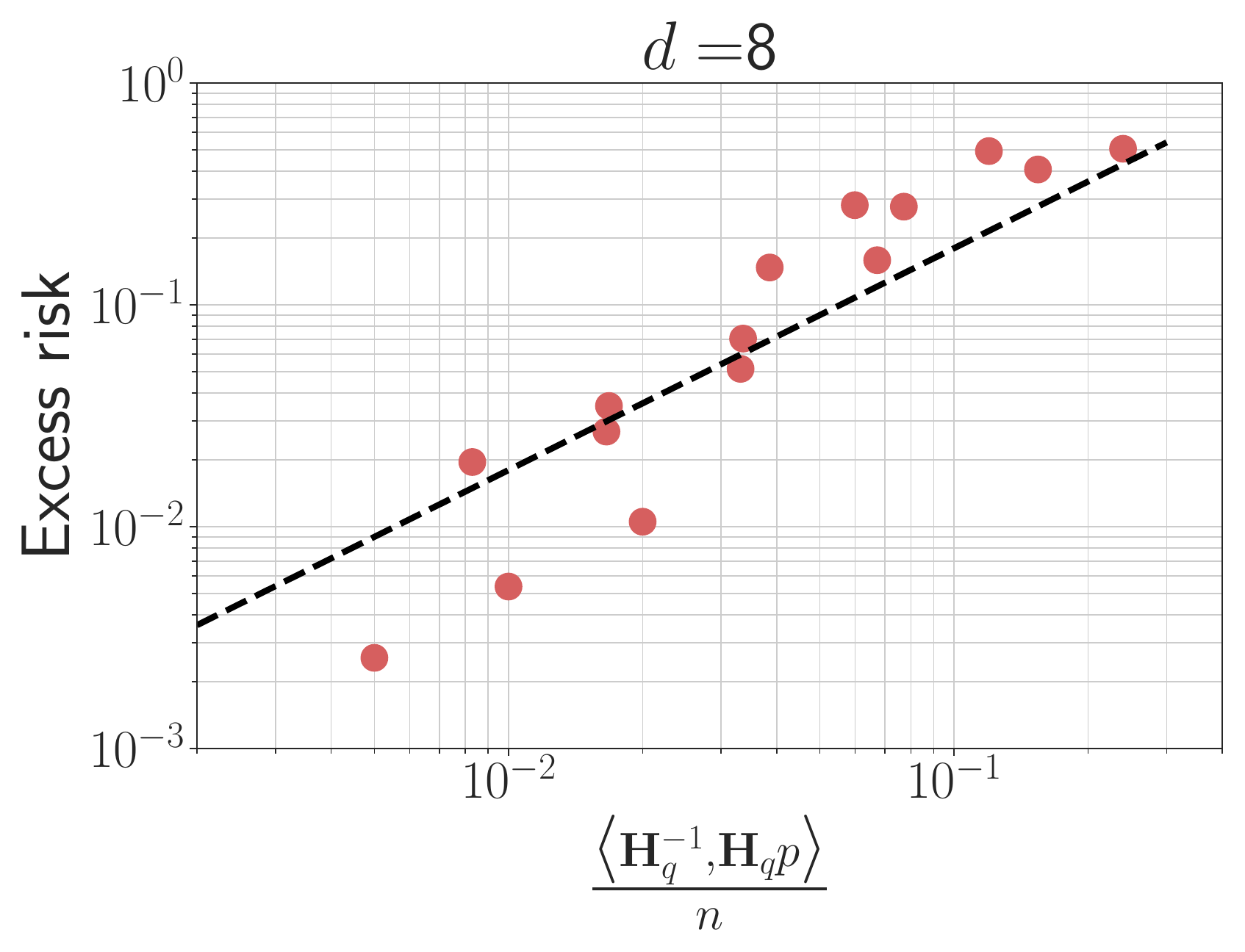}};
\node[inner sep=0pt] (C2) at (4.6,-6.4) {\includegraphics[width=4.6cm]{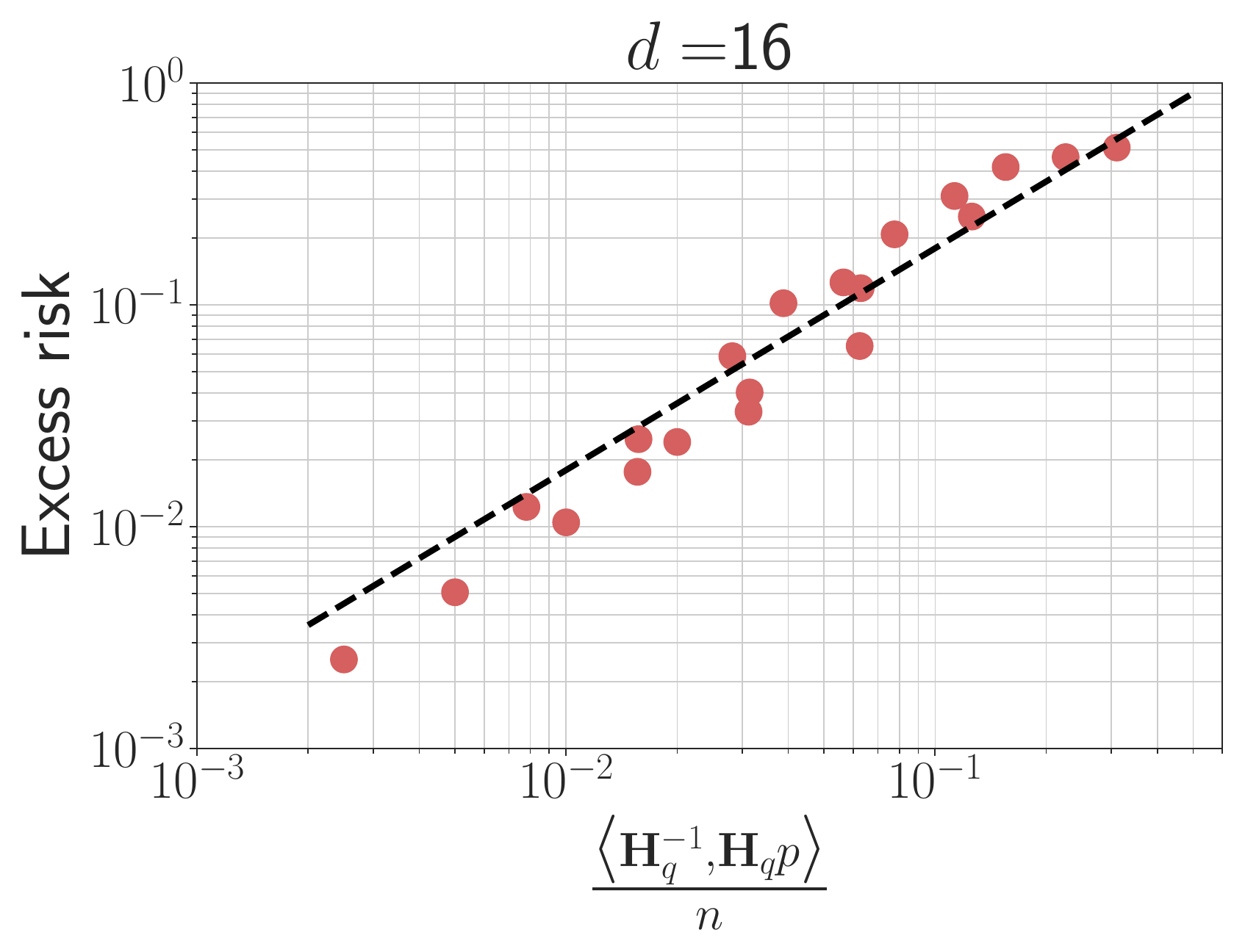}};
\node[inner sep=0pt] (C3) at (9.2,-6.4) {\includegraphics[width=4.6cm]{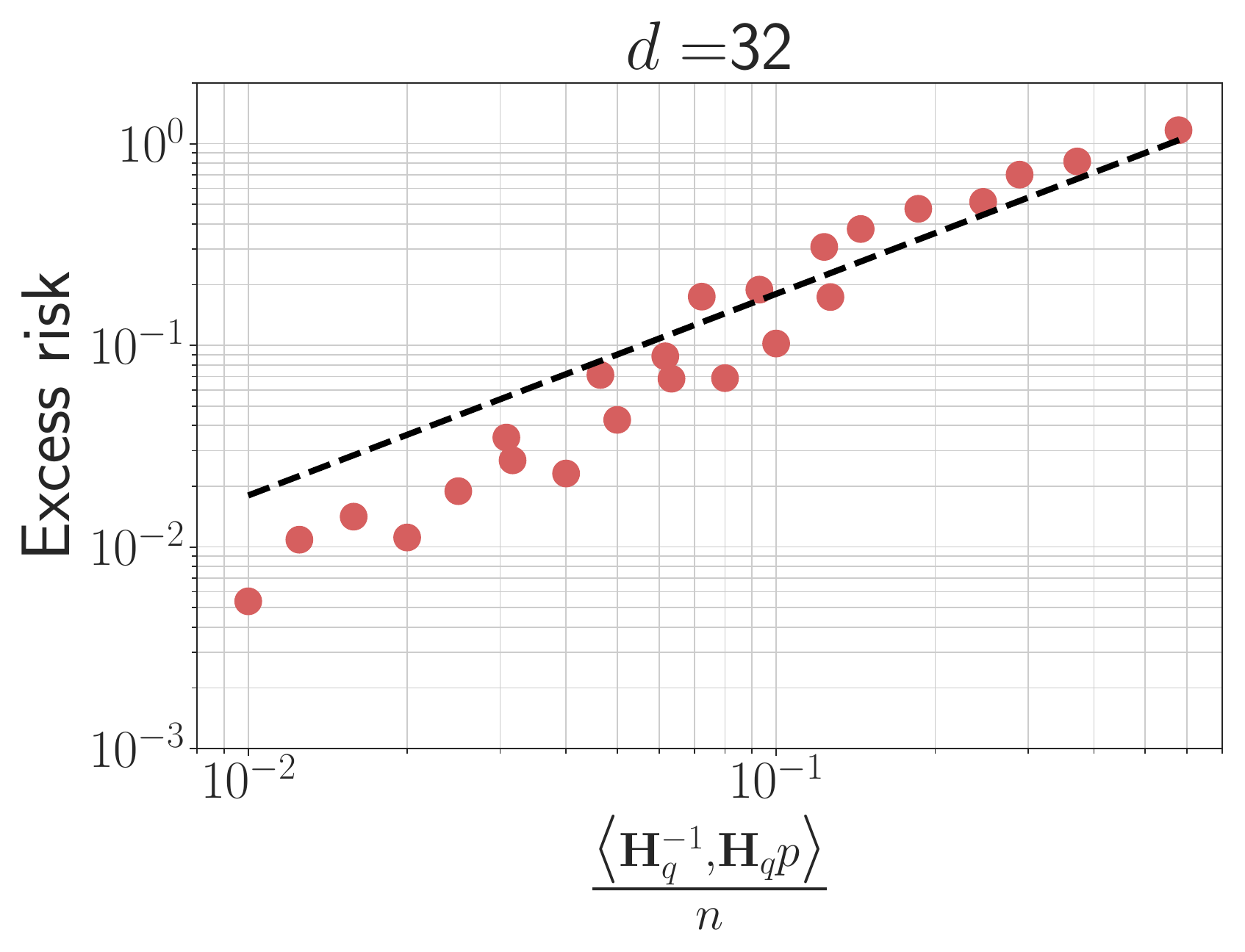}};
\end{tikzpicture}
     \caption{Multivariate t-distribution test: excess risk of $p(x)$  vs  FIR (upper), $n$ (middle),  and $\frac{\mathrm{FIR}}{n}$ (lower), the black dashed lines have slope 1 in upper and lower rows , and slope -1 in the middle row.}
          \label{fig:syn-risk-t}
\end{figure}

\paragraph{Non-sub-Gaussian distributions.} We consider two non-sub-Gaussian distributions: multivariate Laplace distribution and t-distribution. For $q(x)$, we only consider the translation case. We fix $c=2$ and vary $d$, $n$ and $q(x)$. In \Cref{fig:trace_sigma_laplace_t}, we plot $\lambda_{\max}(\hq^{-1} \hp)$ vs FIR in different distributions. For multivariate Laplace distribution tests, we plot excess risk of $p(x)$ vs FIR, $n$ and $\mathrm{FIR}/n$ respectively in \Cref{fig:syn-risk-laplace}. We plot results for the multivariate t-distribution in \Cref{fig:syn-risk-t}. We can observe that the results are consistent to the excess risk bound derived in \Cref{eq:sub-thm-risk}, even though we have sub-Gaussian distribution assumption in \Cref{thm:sub-thm}.

\clearpage
\subsection{Real-world Datasets}\label{appendix:real-world}

\begin{algorithm}[!t]
\caption{Spectral embedding via normalized graph Laplacian}
\label{algo:laplacian}
\hspace*{\algorithmicindent} \textbf{Input:} \nolinebreak
 data points $\X \in \mathbb{R}^{N \times D}$, nearest neighbor number $k$, target out put dimension $d$   \\
 \hspace*{\algorithmicindent} \textbf{Output:} \nolinebreak $\widehat{\X} \in\mathbb{R}^{N\times d}$
\begin{algorithmic}[1]
    \State Obtain $k$-nearest neighbor graph $\mathcal{G}$ on $\X$.
    \State Obtain adjacency matrix $\A$ and its degree matrix $\b D$ from $\mathcal{G}$ (using ones as weights).
    \State Calculate normalized Laplacian $\b L\gets \b I - \D^{-1/2} \A \D^{-1/2}$.
    \State Calculate the first $d$ eigenvectors of $\b L$ (corresponding to the $d$ smallest eigenvalues of $\b L$): $\{ v_i \}_{i\in[d]}$.
    \State Form matrix $\widehat{\X}$ by stacking  $\{v_i\}_{i\in[d]}$ column-wise.
\end{algorithmic}

\end{algorithm}

\paragraph{Data pre-processing.} 
We use unsupervised learning to find an appropriate feature space that we can then use for multi-class logistic regression.  {SimCLR~\cite{simclr}} is a framework for contrastive learning of visual representations. It learns representations by maximizing agreement between differently augmented views of the same data example via a contrastive loss in the latent space. We also employ a spectral embedding using  the normalized nearest-neighbor  graph Laplacian to extract features. We present the algorithm in \Cref{algo:laplacian}, where we use $k=256$ as the  number of nearest neighbor for all three datasets. Below, we provide a more detailed description of the preprocessing steps performed for each dataset.

\begin{itemize}[leftmargin=*]
    \item MNIST. We use the normalized Laplacian to reduce the dimension of the input data to dimension of 20. In \Cref{algo:laplacian}, $N=60,000$, $D=  784$, and $d=20$. For the active learning runs, we randomly select $m =3,000$ points (with 300 points in each class id) to form the unlabeled data set $U$.
    \item CIFAR-10. First, we use \href{https://github.com/google-research/simclr}{pre-trained SimCLR model } on the whole training data and extract the feature maps from the last layer (with dimension 512). Second, we use the normalized Laplacian to reduce the dimension of the training data to dimension of 20. In \Cref{algo:laplacian}, $N=50,000$, $D=  512$, and $d=20$. For the active learning tests, we randomly select $m =3,000$ points (with 300 points in each class id) to form the unlabeled data set $U$.
    \item ImageNet-50.  We first randomly select 50 classes from the training set of ImageNet. We use pre-trained SimCLR model and extract the features with dimension 2048.  Then we use the normalized Laplacian to reduce the dimension of the training data to dimension of 40.  In \Cref{algo:laplacian},  $D=  2048$, and $d=40$. or the active learning tests, we randomly select $m =5,000$ points (with 100 points in each class id) to form the unlabeled data set $U$.
\end{itemize}

\paragraph{Tuning hyperparameter $\eta$.} In  \Cref{algo:round}, we have to set the learning rate $\eta$. We try different $\eta$ and select the one that maximizes $\lambda_{\min}(\sum_{t=1}^b \tI(x_{i_t}))$ since this is our goal of the sparsification step (lines 3-11 in \Cref{algo:round}). Note that for each round of active learning, we only need to solve the relaxed problem~\Cref{eq:mtd-obj-relax} once. Furthermore,  tuning $\eta$ does not require labeling information.

\begin{figure}[!b]
\centering
  \footnotesize
\begin{tikzpicture}
\node[inner sep=0pt] (a) at (0,0) {\includegraphics[width=4.5cm]{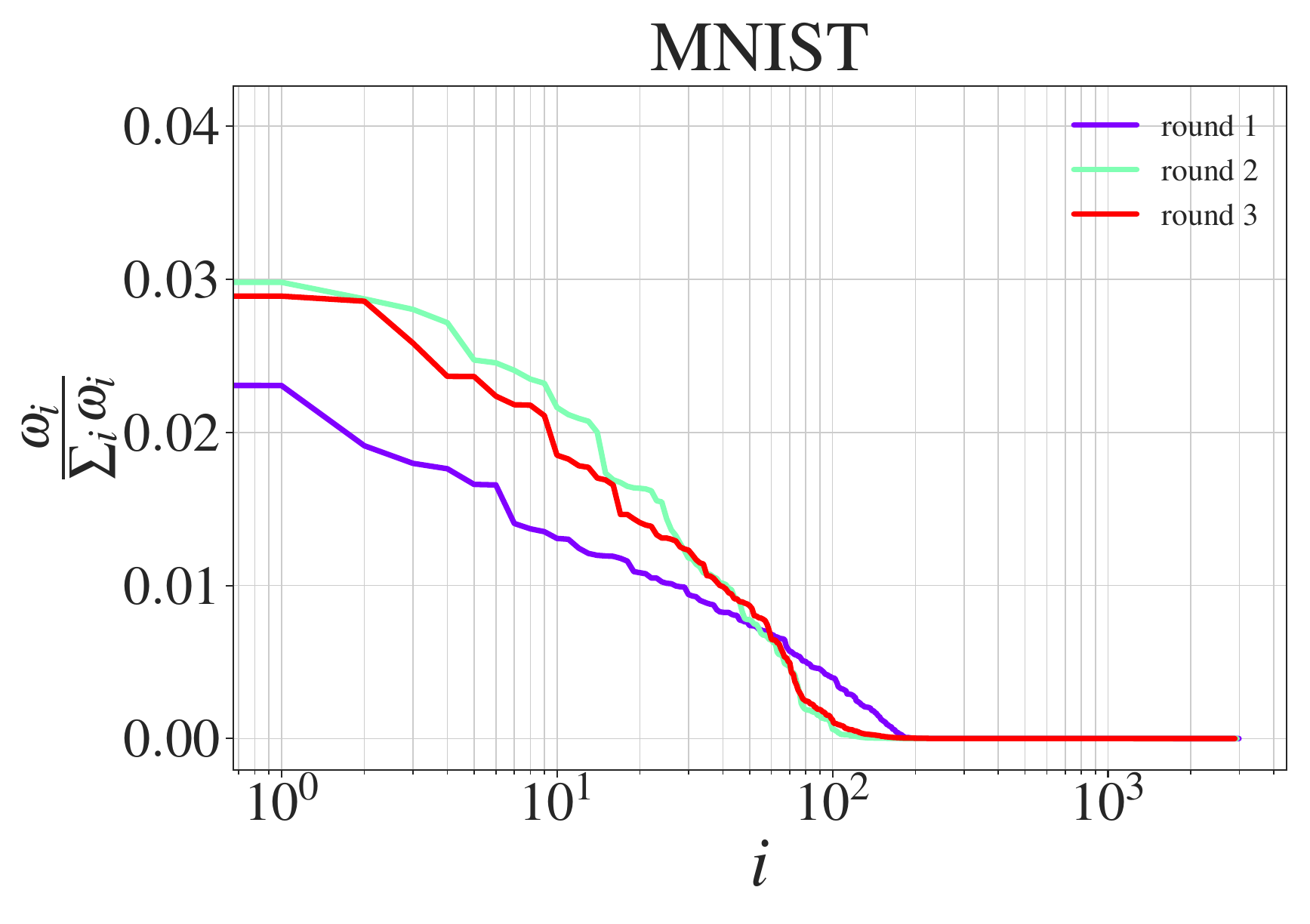}};
\node[inner sep=0pt] (b) at (4.7,0) {\includegraphics[width=4.5cm]{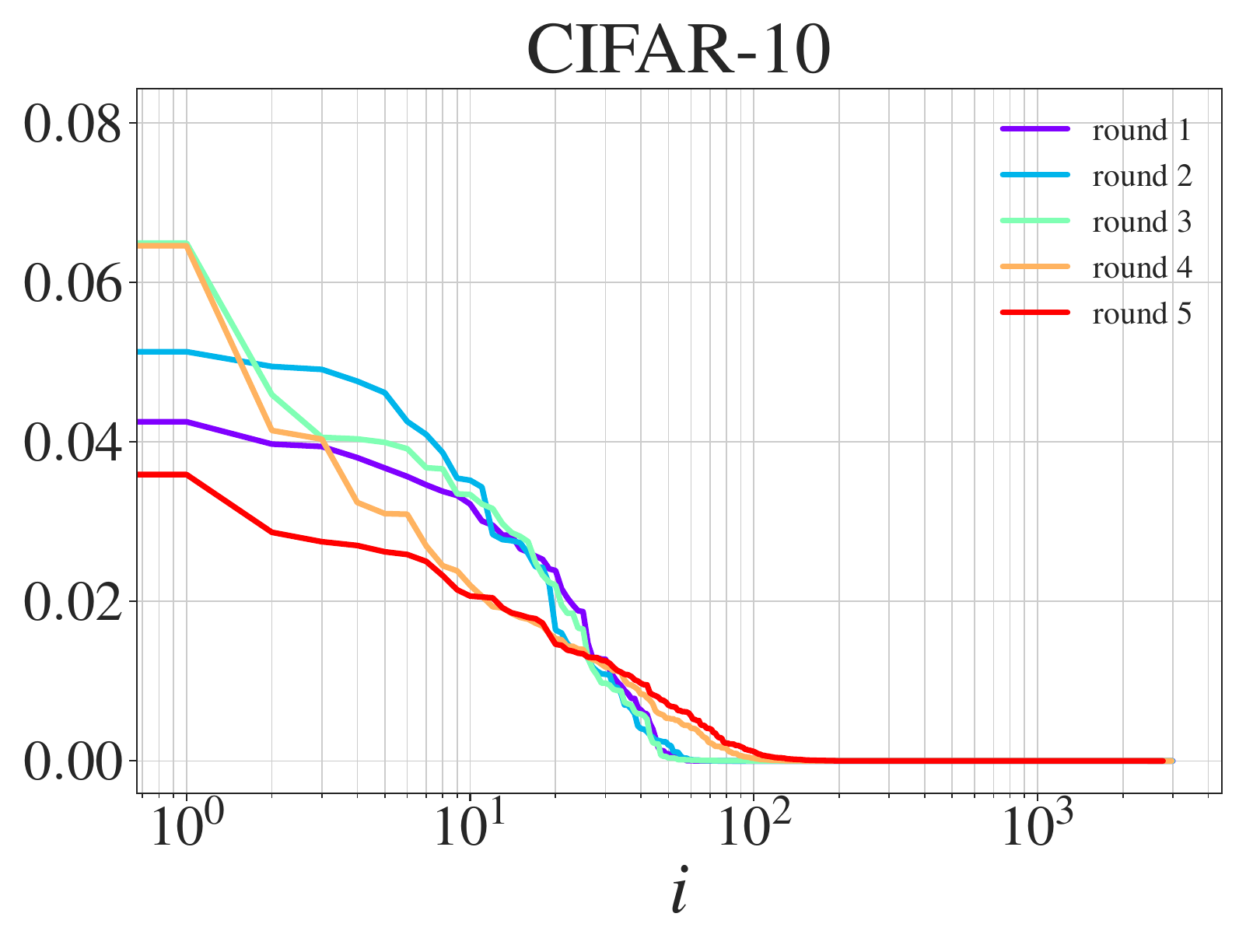}};
\node[inner sep=0pt] (c) at (9.2,0) {\includegraphics[width=4.5cm]{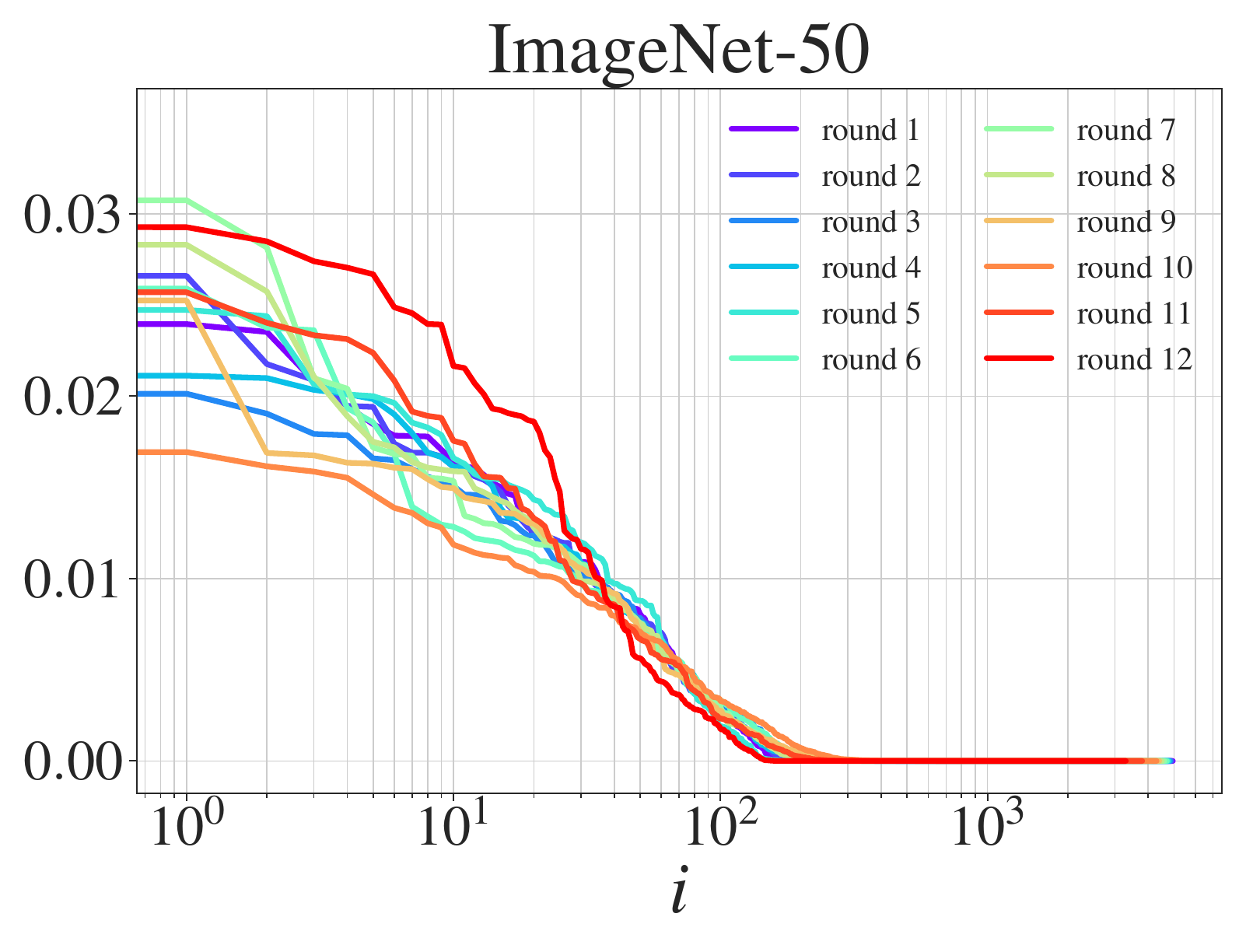}};
\end{tikzpicture}
\caption{Normalized weights $z_{\diamond}$ (solution of \Cref{eq:mtd-obj-relax}) at each round of active learning tests. }
\label{fig:real-weights}
\end{figure}

\begin{figure}[!b]
\centering
  \footnotesize
\begin{tikzpicture}
    \node[]  at (0, 2) {\textbf{Random}};
    \node[]  at (7, 2) {\textbf{K-means}};
    \node[]  at (0, -2) {\textbf{Entropy}};
    \node[]  at (7, -2) {\textbf{Var Ratios}};
    \node[]  at (0, -6) {\textbf{BAIT}};
    \node[]  at (7, -6) {\textbf{FIRAL}};
\node[inner sep=0pt] (a) at (0,0) {\includegraphics[width=6.8cm,height = 3.5cm]{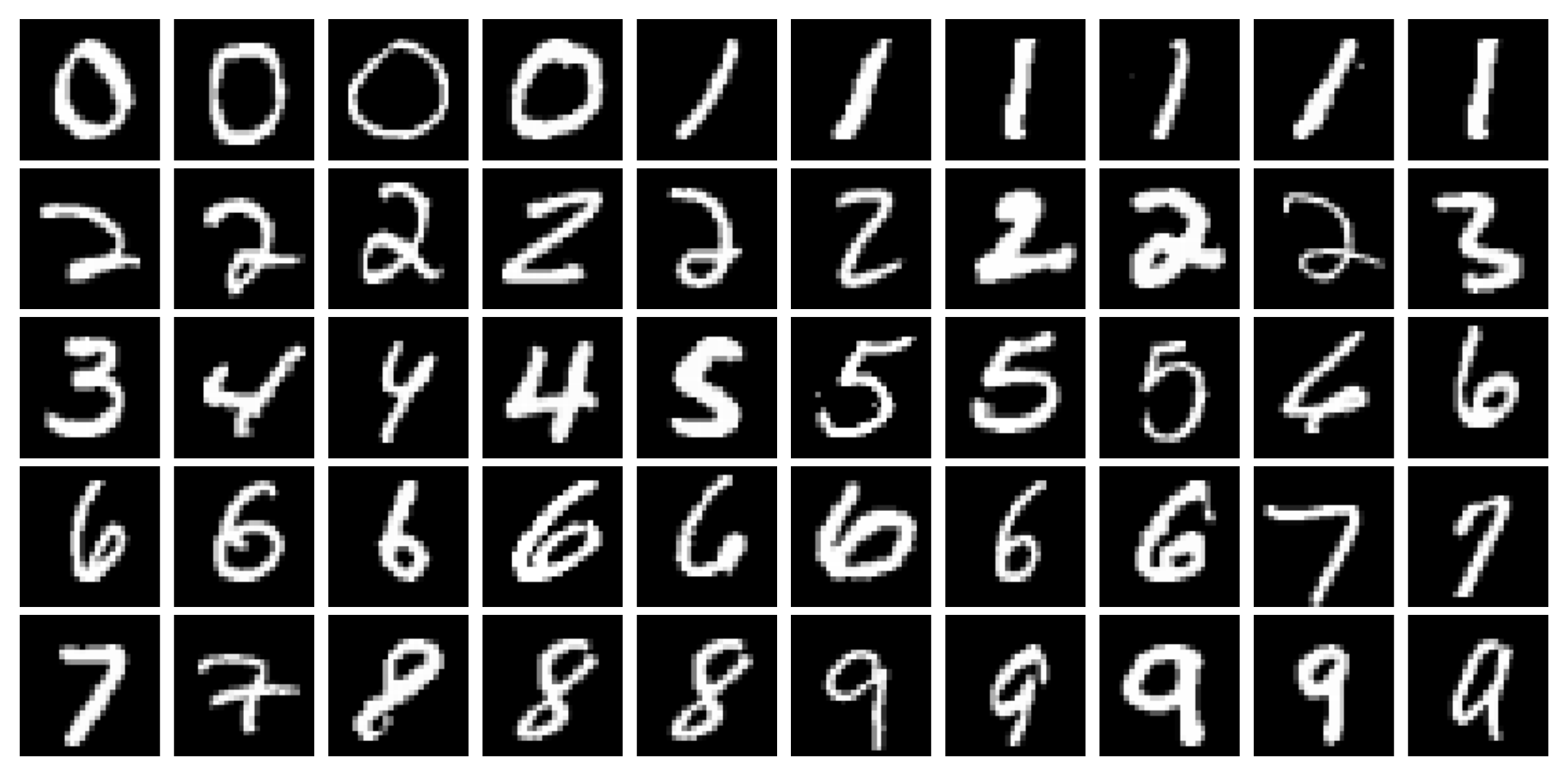}};
\node[inner sep=0pt] (b) at (7.1,0) {\includegraphics[width=6.8cm,height = 3.5cm]{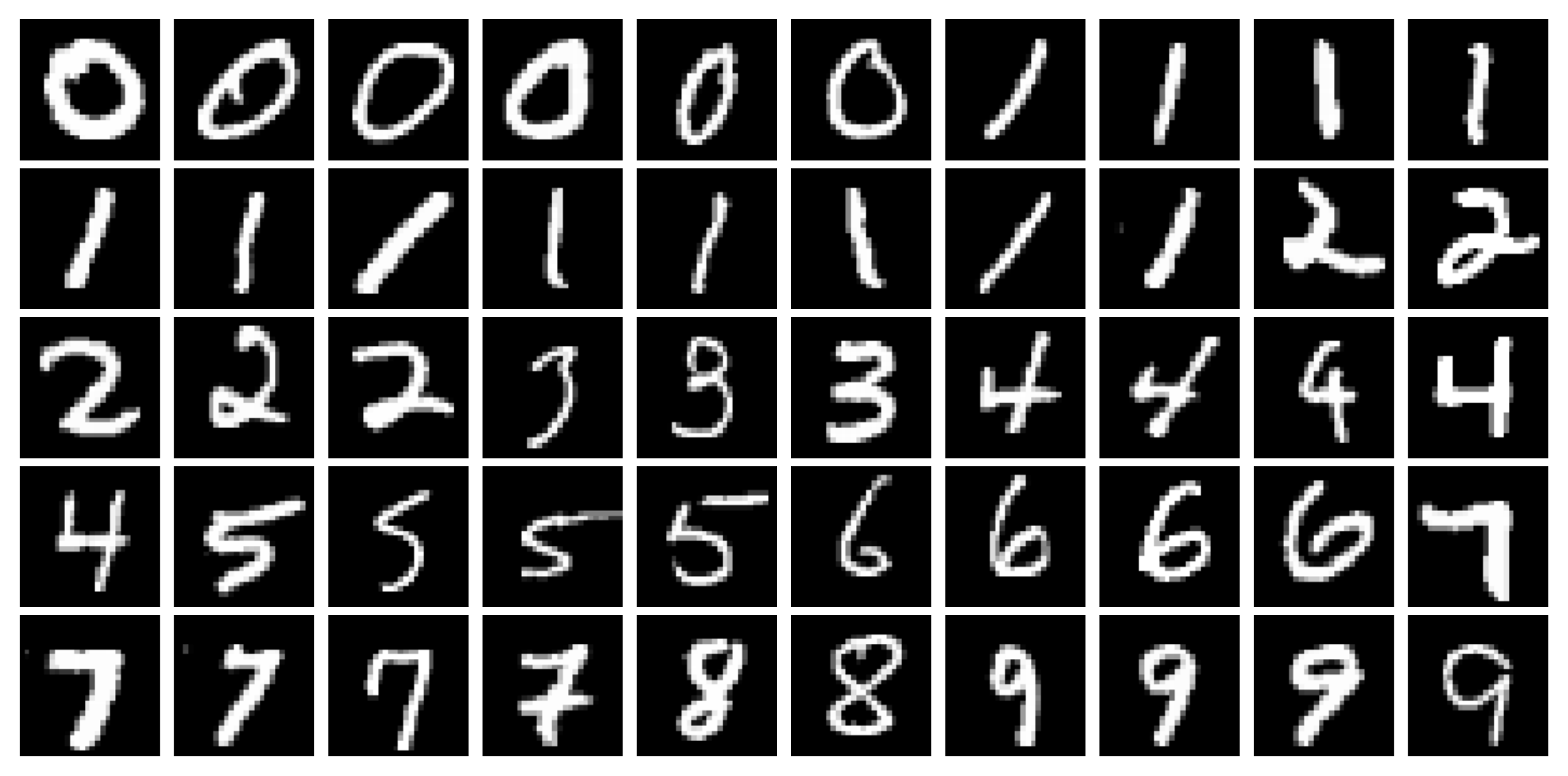}};
\node[inner sep=0pt] (c) at (0,-4) {\includegraphics[width=6.8cm,height = 3.5cm]{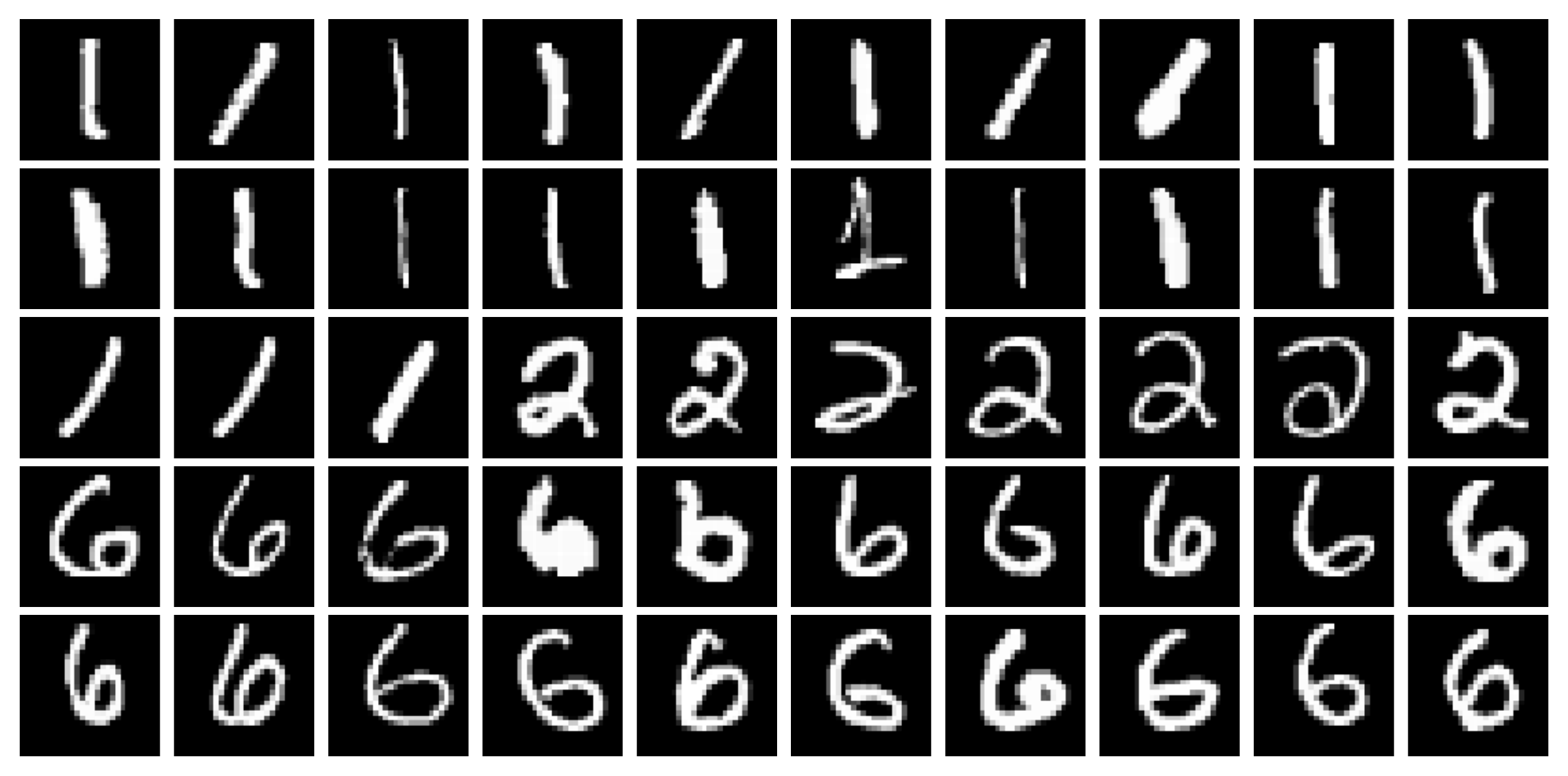}};
\node[inner sep=0pt] (d) at (7.1,-4) {\includegraphics[width=6.8cm,height = 3.5cm]{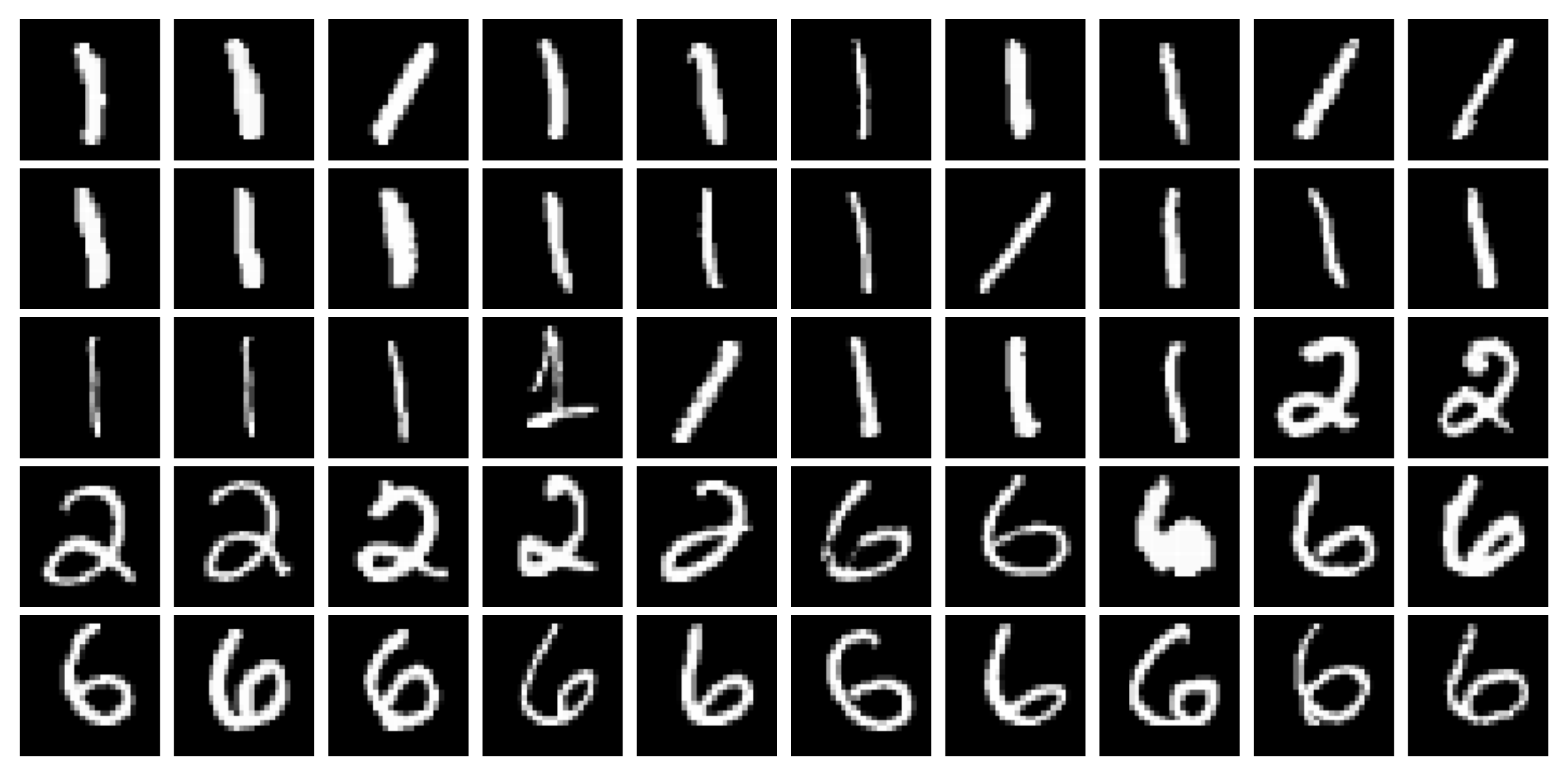}};
\node[inner sep=0pt] (e) at (0,-8) {\includegraphics[width=6.8cm,height = 3.5cm]{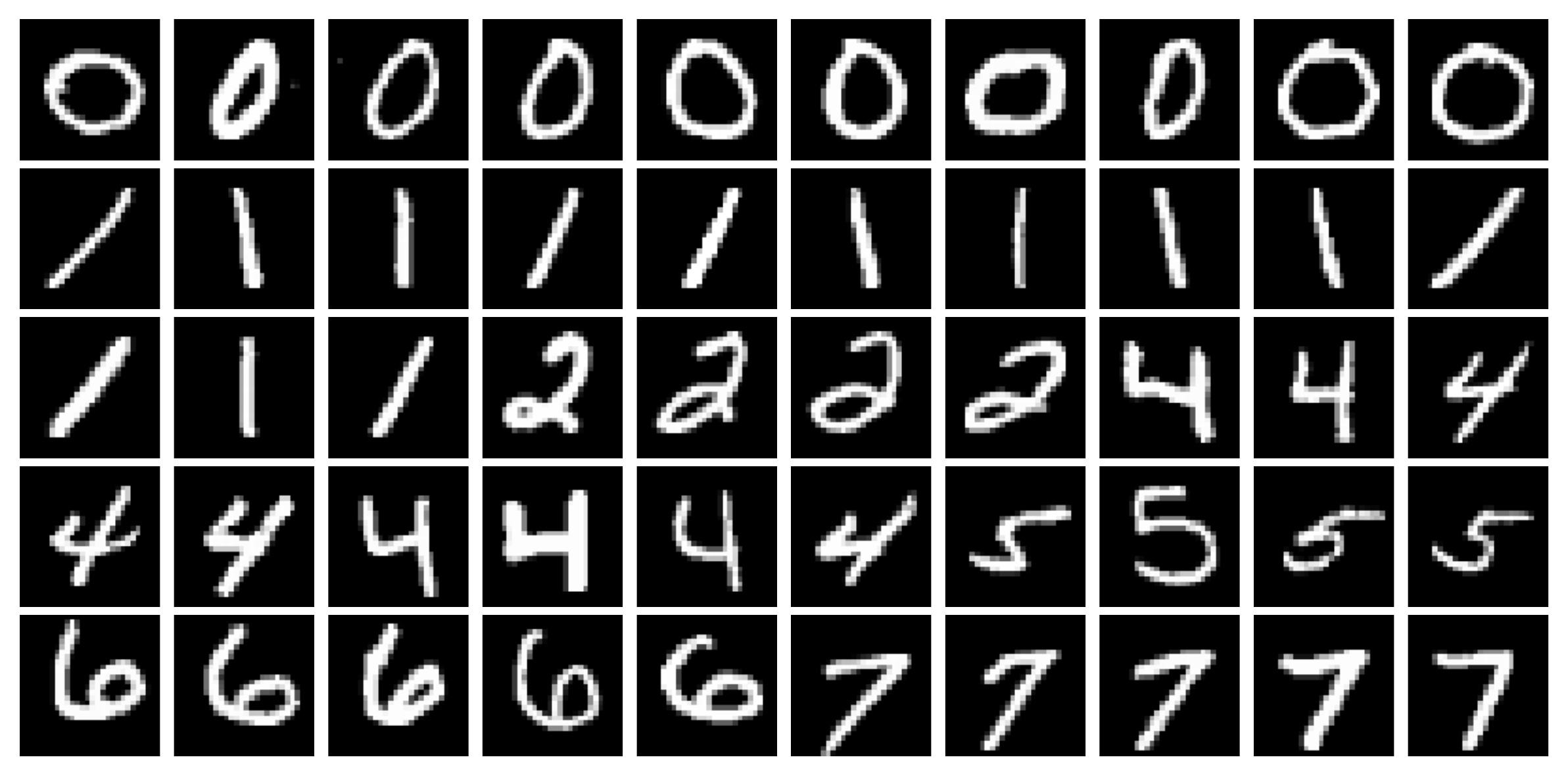}};
\node[inner sep=0pt] (f) at (7.1,-8) {\includegraphics[width=6.8cm,height = 3.5cm]{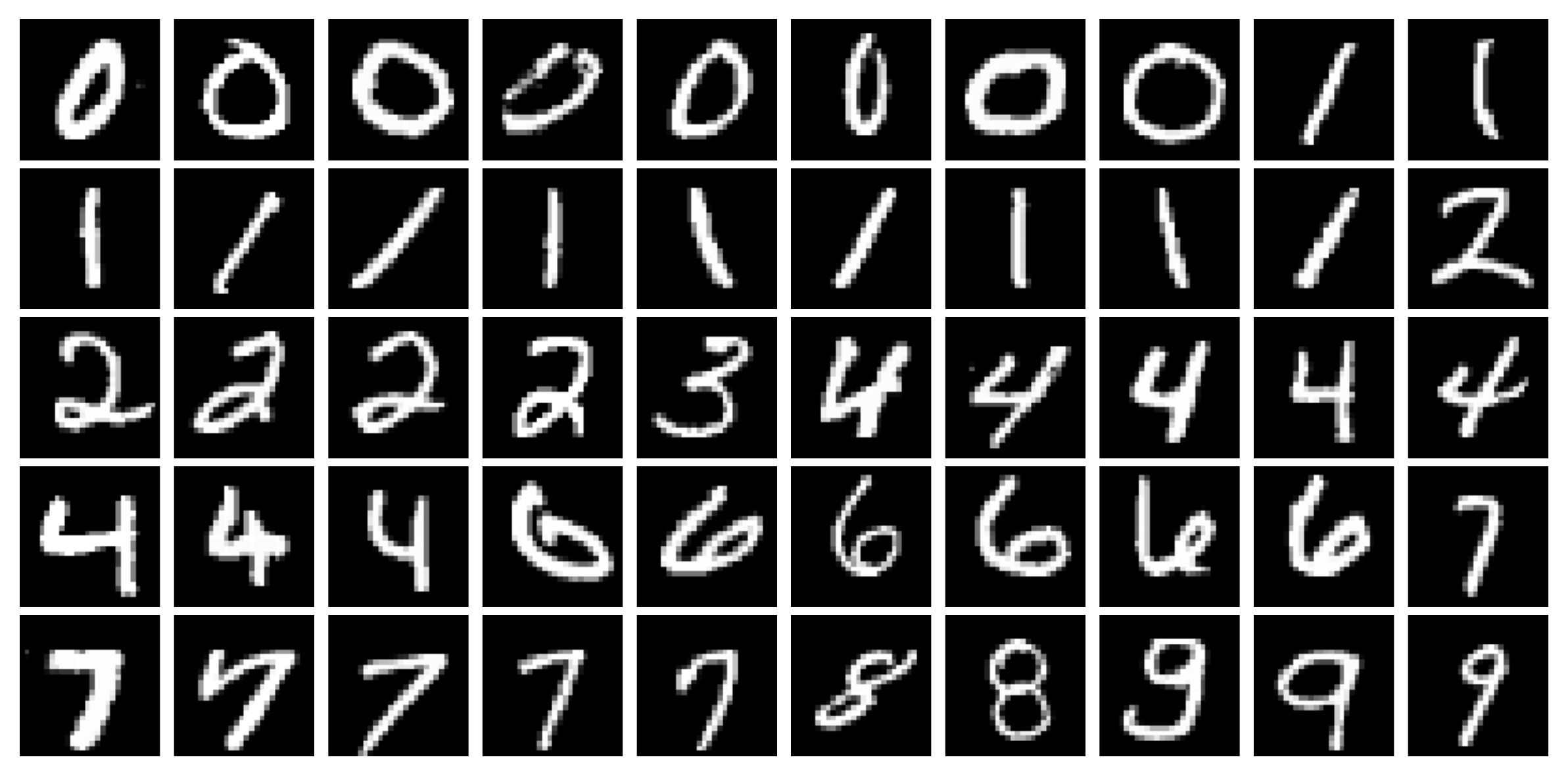}};
\end{tikzpicture}
\caption{Selected samples for MNIST at the first round of active learning test. }
\label{fig:real-mnist}
\end{figure}
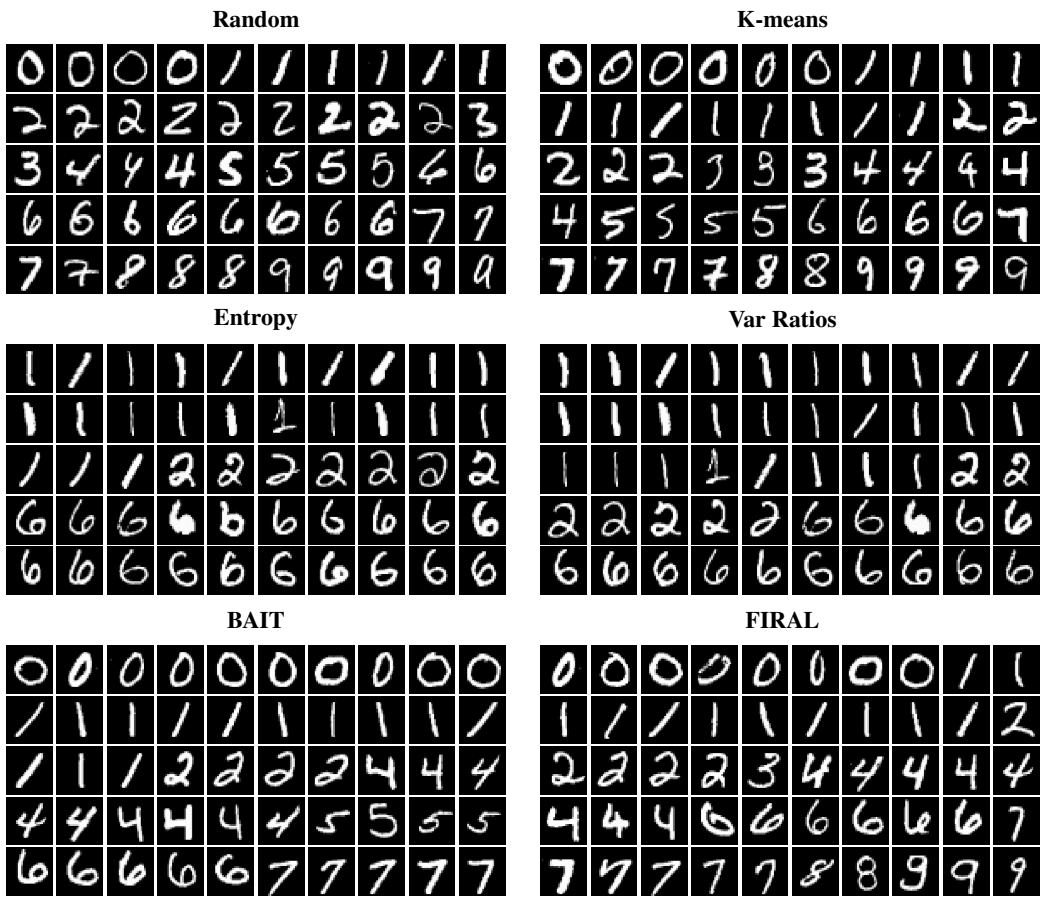

\begin{figure}[!b]
\centering
  \footnotesize
\begin{tikzpicture}
\node[inner sep=0pt] (a) at (0,0) {\includegraphics[width=6.8cm,height = 5.5cm]{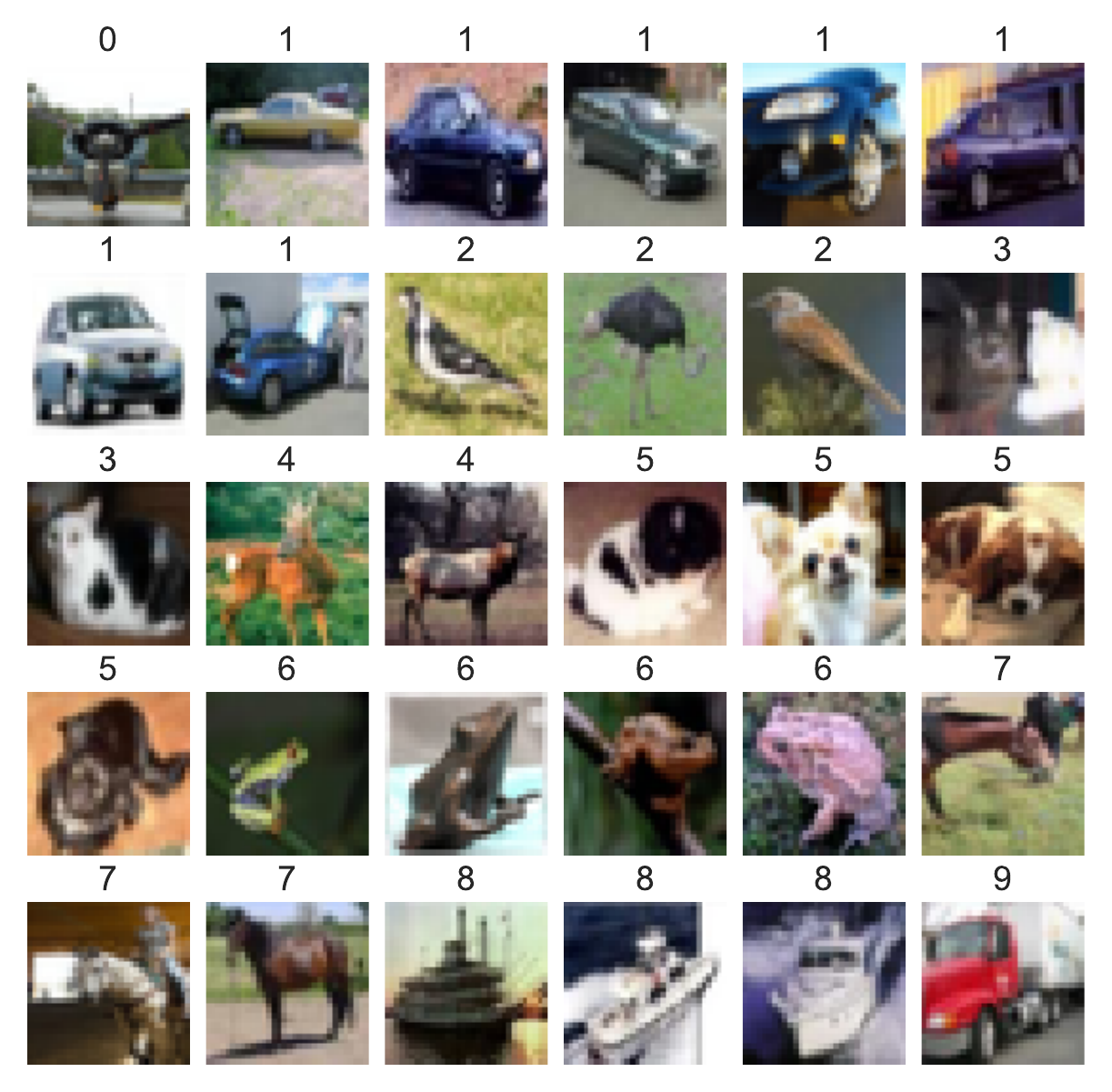}};
\node[inner sep=0pt] (b) at (7,0) {\includegraphics[width=6.8cm,height = 5.5cm]{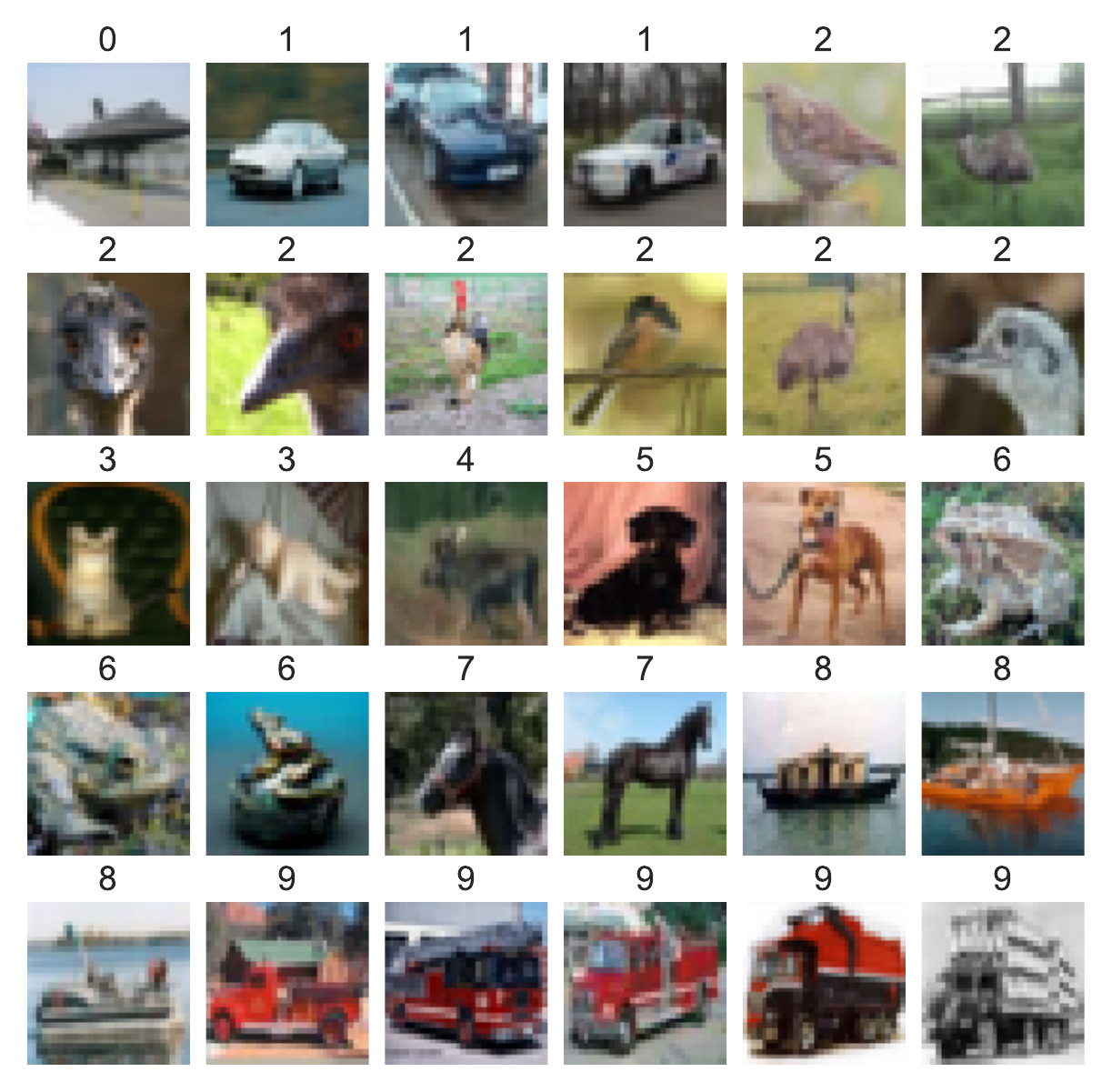}};
\node[inner sep=0pt] (c) at (0,-6) {\includegraphics[width=6.8cm,height = 5.5cm]{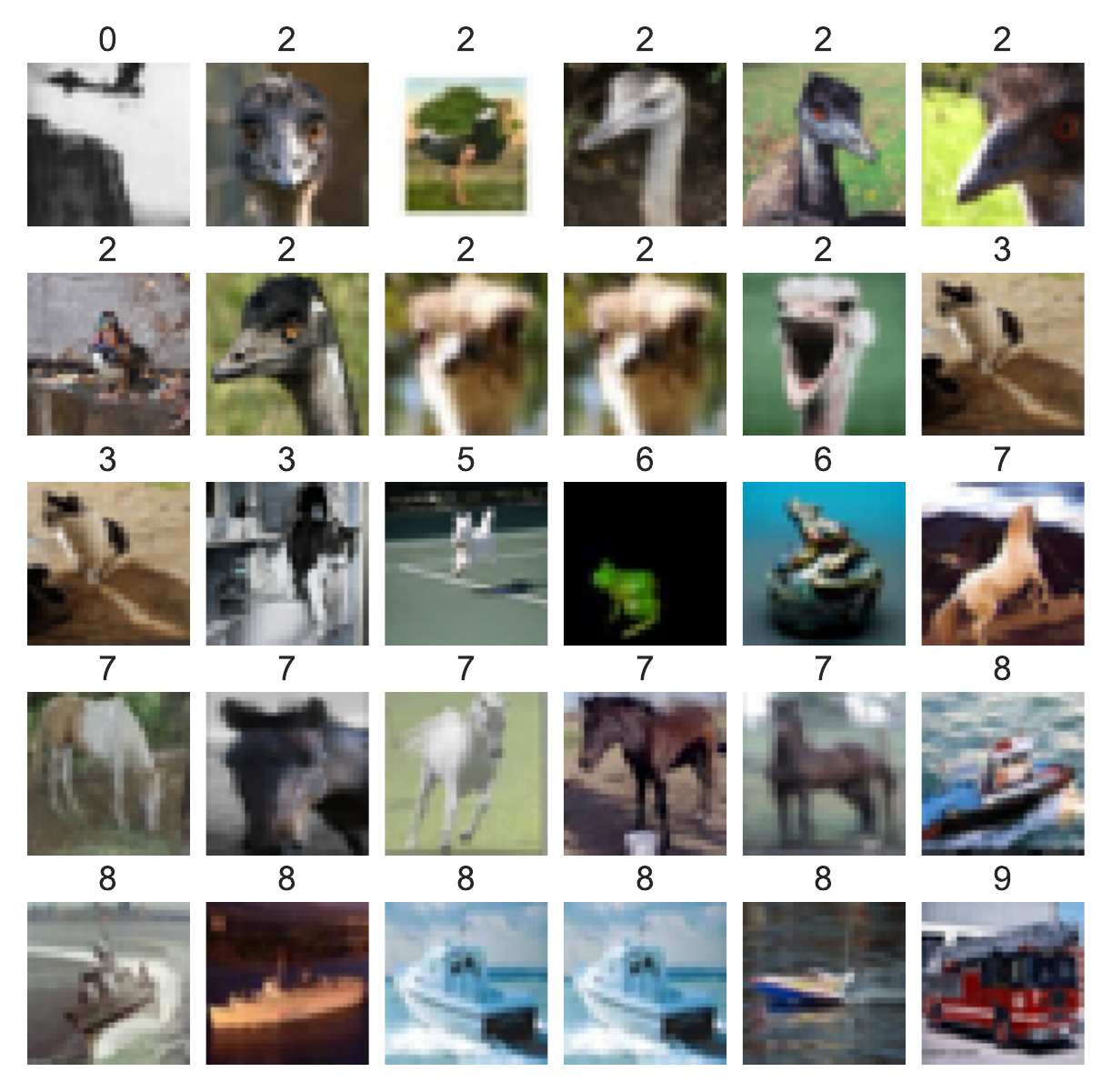}};
\node[inner sep=0pt] (d) at (7,-6) {\includegraphics[width=6.8cm,height = 5.5cm]{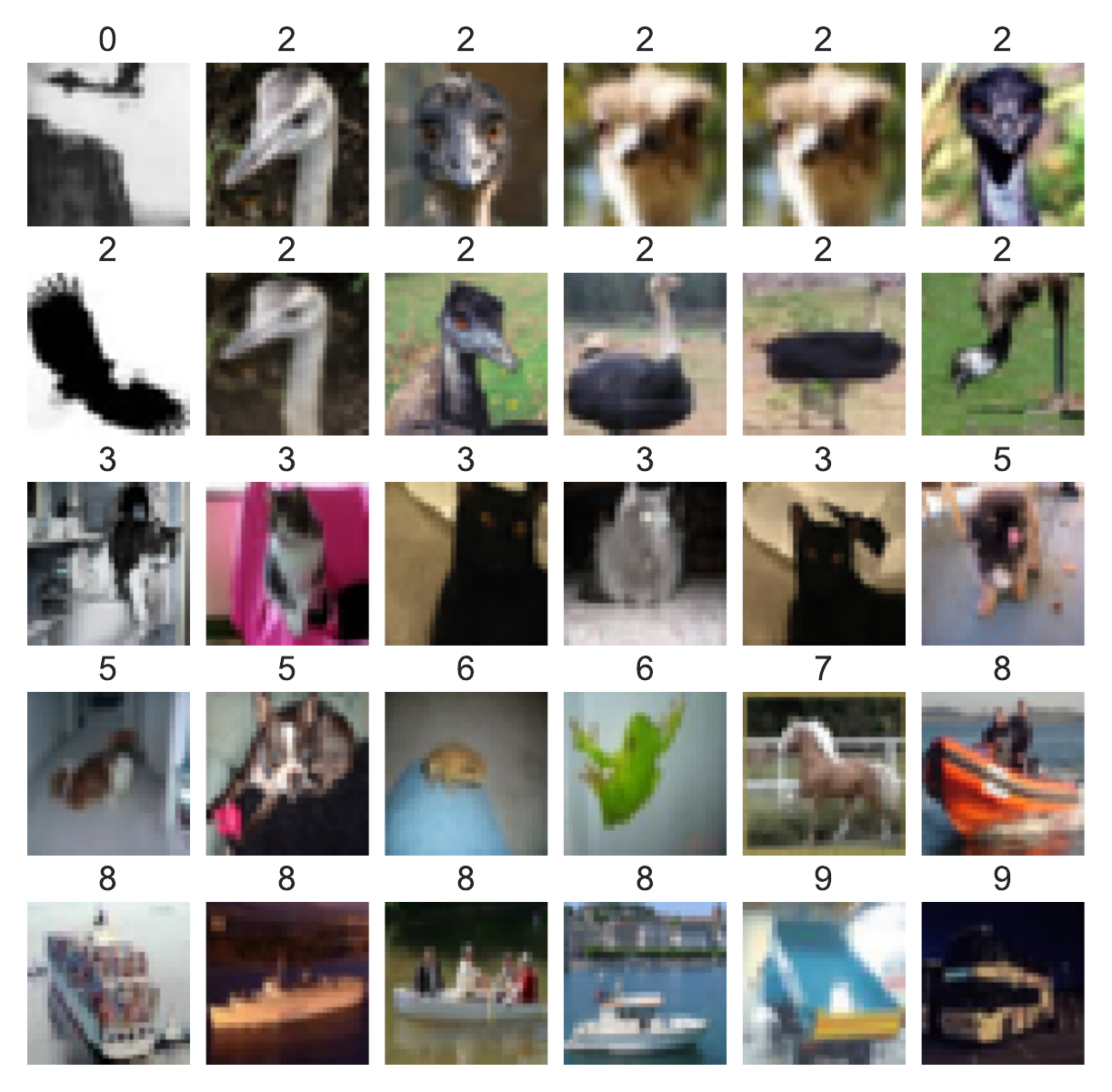}};
\node[inner sep=0pt] (e) at (0,-12) {\includegraphics[width=6.8cm,height = 5.5cm]{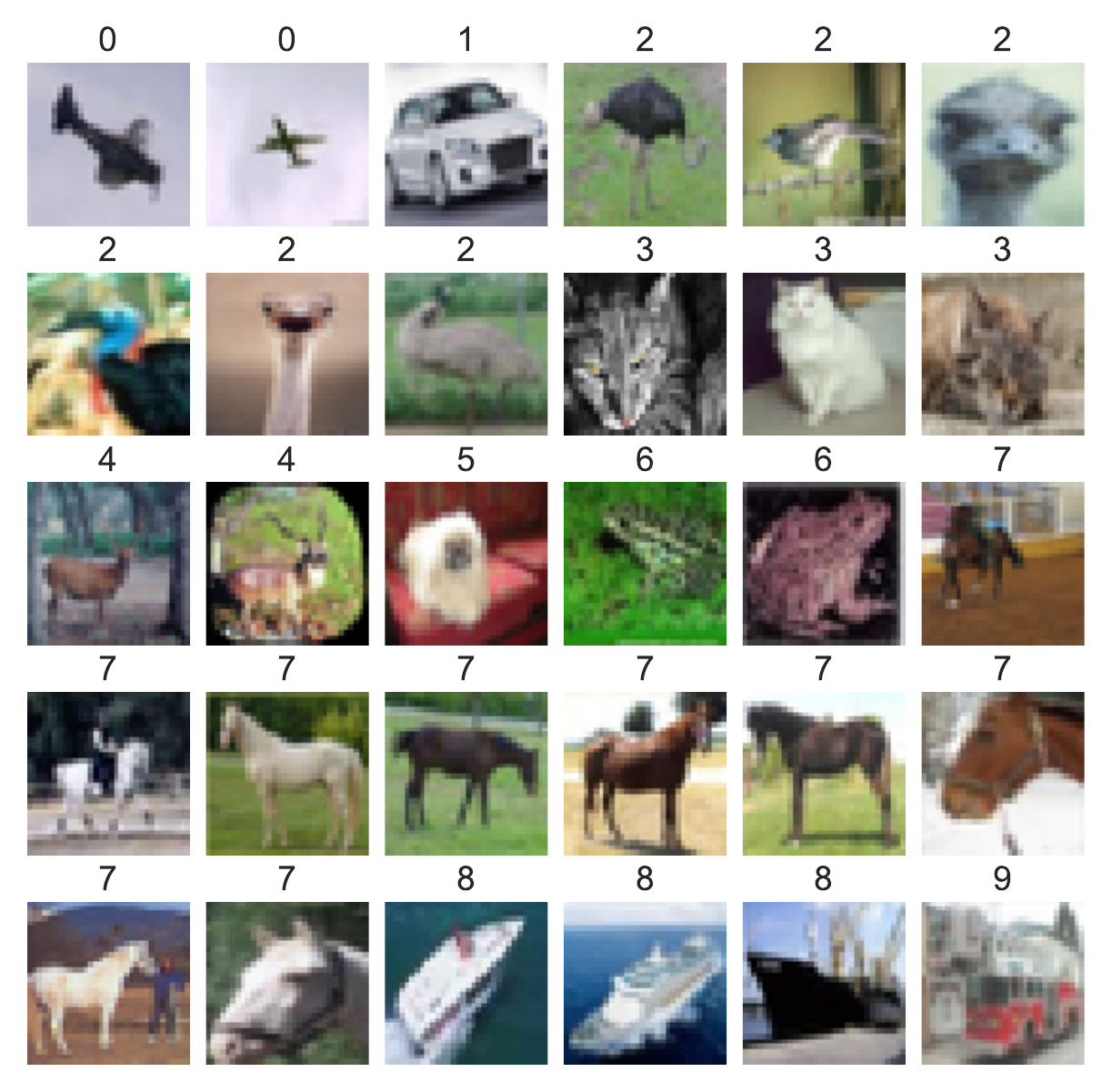}};
\node[inner sep=0pt] (f) at (7,-12) {\includegraphics[width=6.8cm,height = 5.5cm]{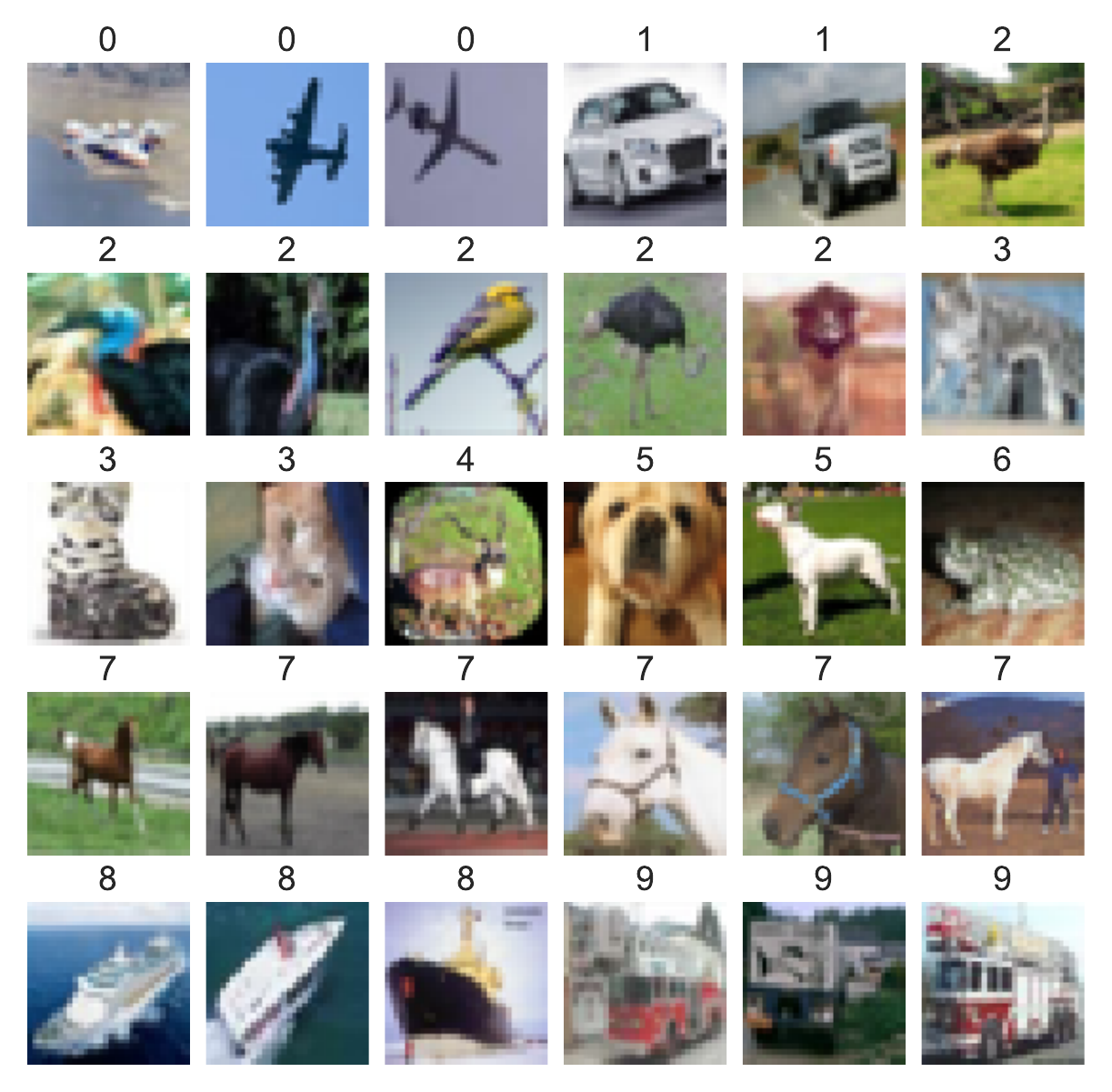}};
    \node[]  at (0, 2.9) {\textbf{Random}};
    \node[]  at (7, 2.9) {\textbf{K-means}};
    \node[]  at (0, -3.1) {\textbf{Entropy}};
    \node[]  at (7, -3.1) {\textbf{Var Ratios}};
    \node[]  at (0, -9.1) {\textbf{BAIT}};
    \node[]  at (7, -9.1) {\textbf{FIRAL}};
\end{tikzpicture}
\caption{Selected samples for CIFAR10 at the first 
 three rounds of active learning test. }
\label{fig:real-cifar10}
\end{figure}
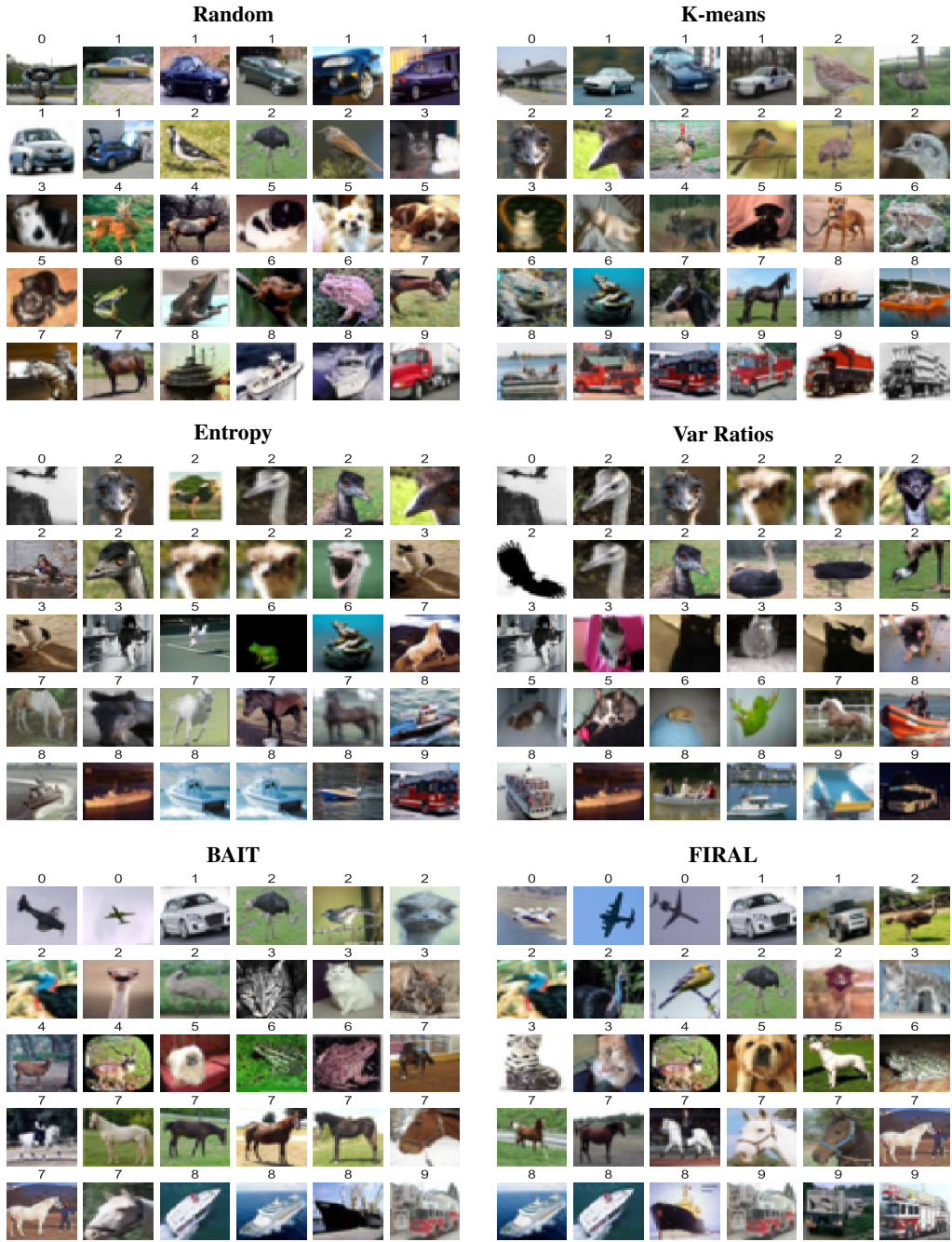

\begin{figure}[!b]
\centering
  \footnotesize
\begin{tikzpicture}
    \node[]  at (7, 3.5) {\textbf{Random}};
    \node[]  at (7, -3.5) {\textbf{K-means}};
    \node[]  at (7, -10.5) {\textbf{Entropy}};
\node[inner sep=0pt] (a) at (7,0) {\includegraphics[width=13cm,height = 6.7cm]{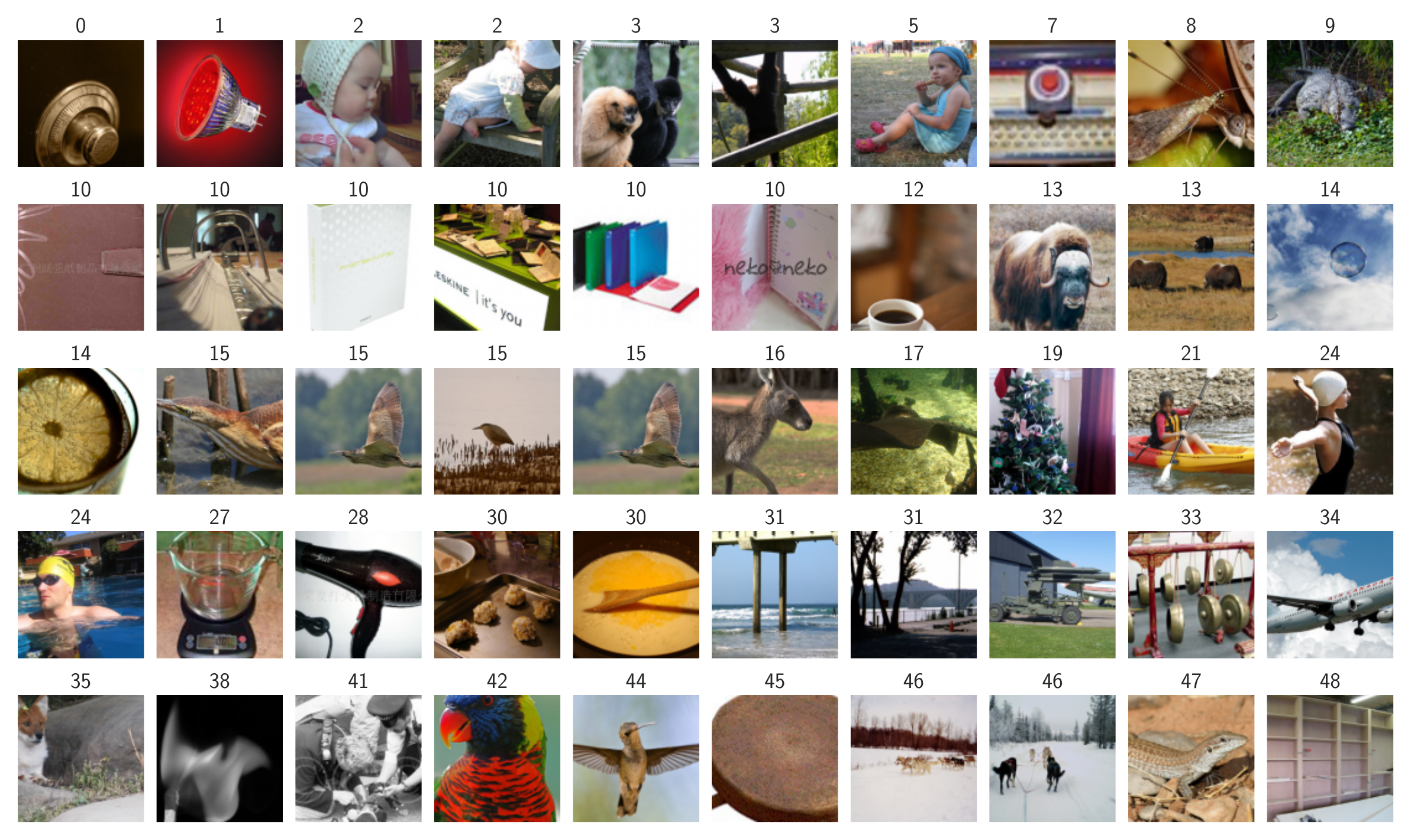}};
\node[inner sep=0pt] (b) at (7,-7) {\includegraphics[width=13cm,height = 6.7cm]{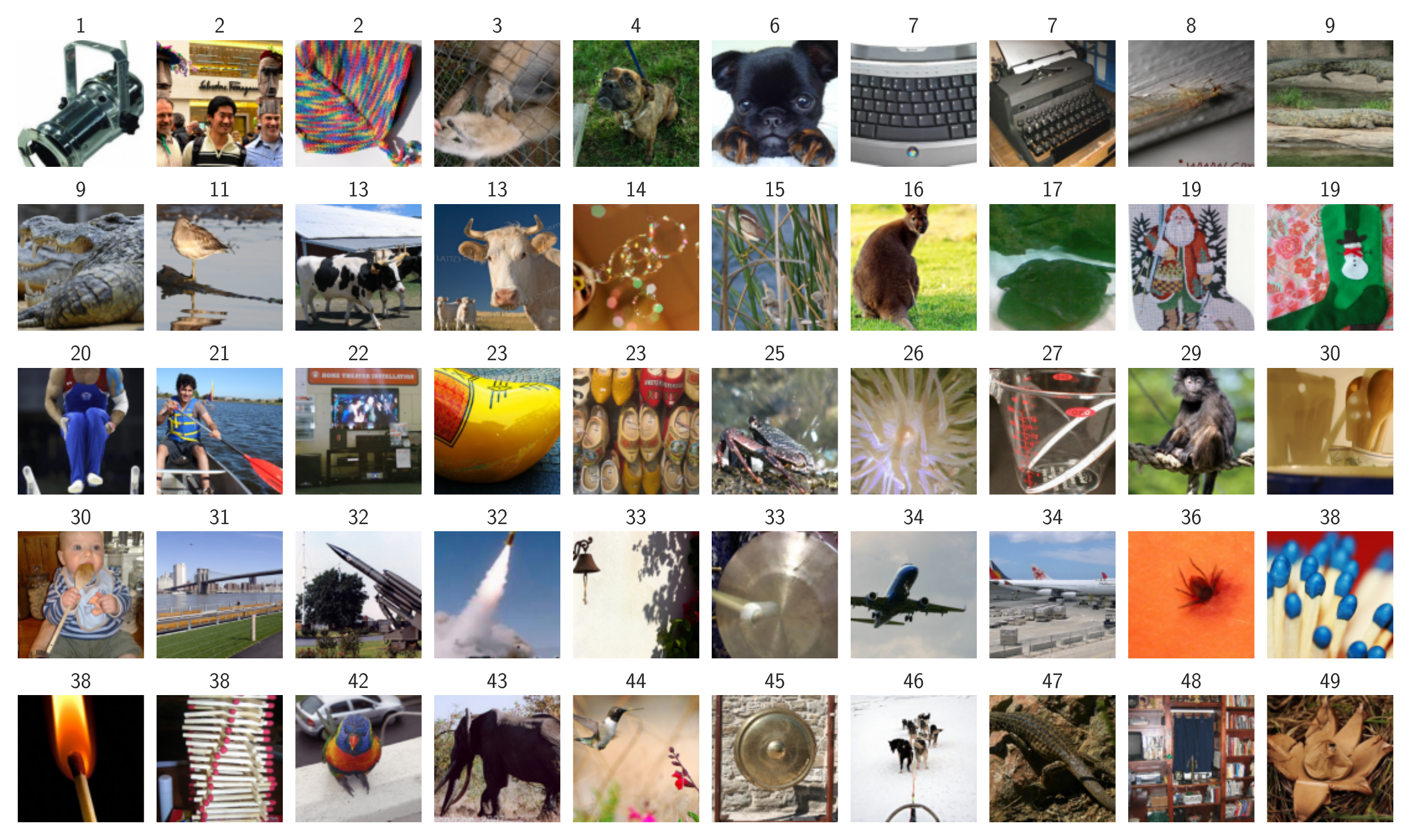}};
\node[inner sep=0pt] (c) at (7,-14) {\includegraphics[width=13cm,height = 6.7cm]{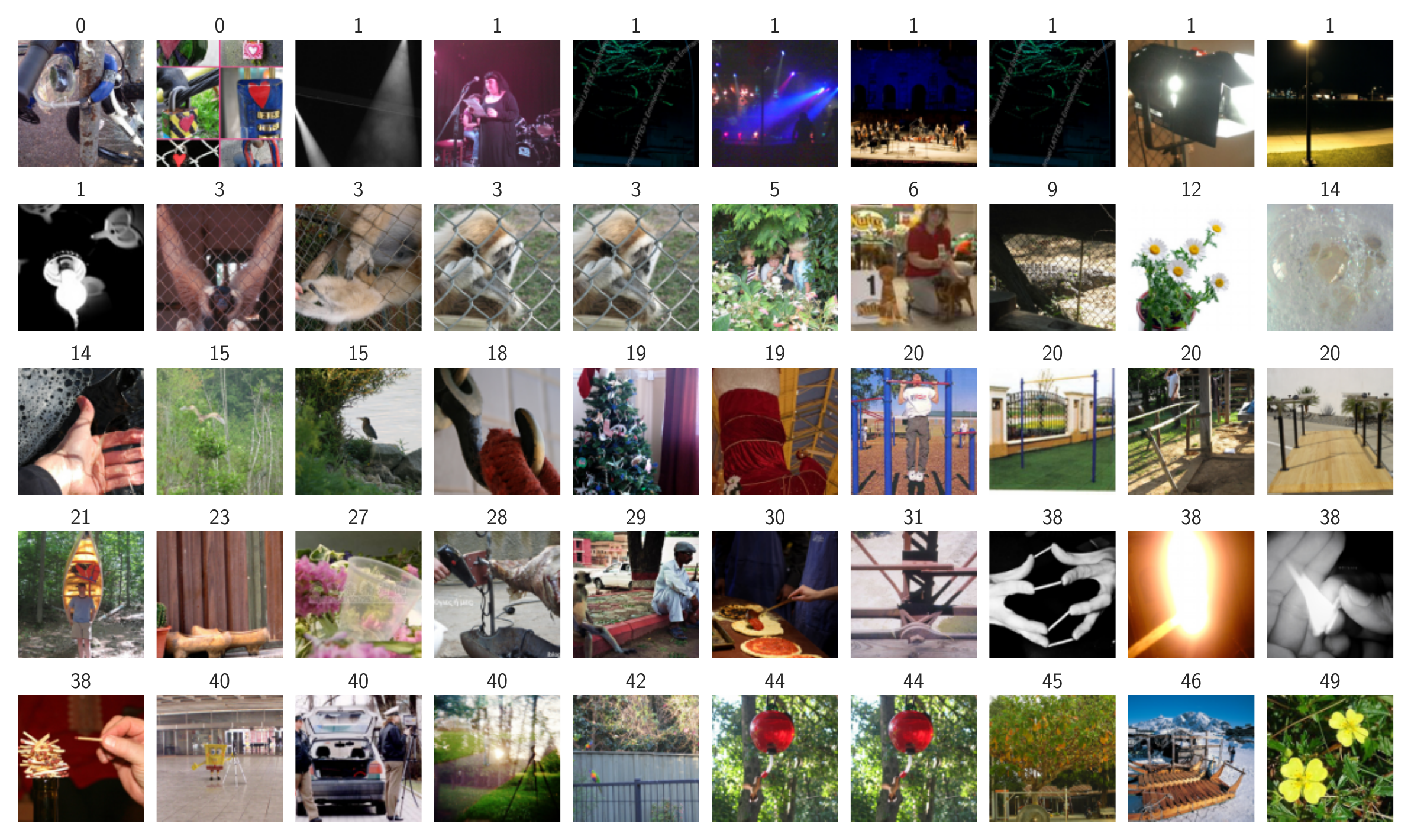}};
\end{tikzpicture}
\caption{Selected samples for ImageNet-50 at the first round of active learning test. }
\label{fig:real-imagenet-1}
\end{figure}
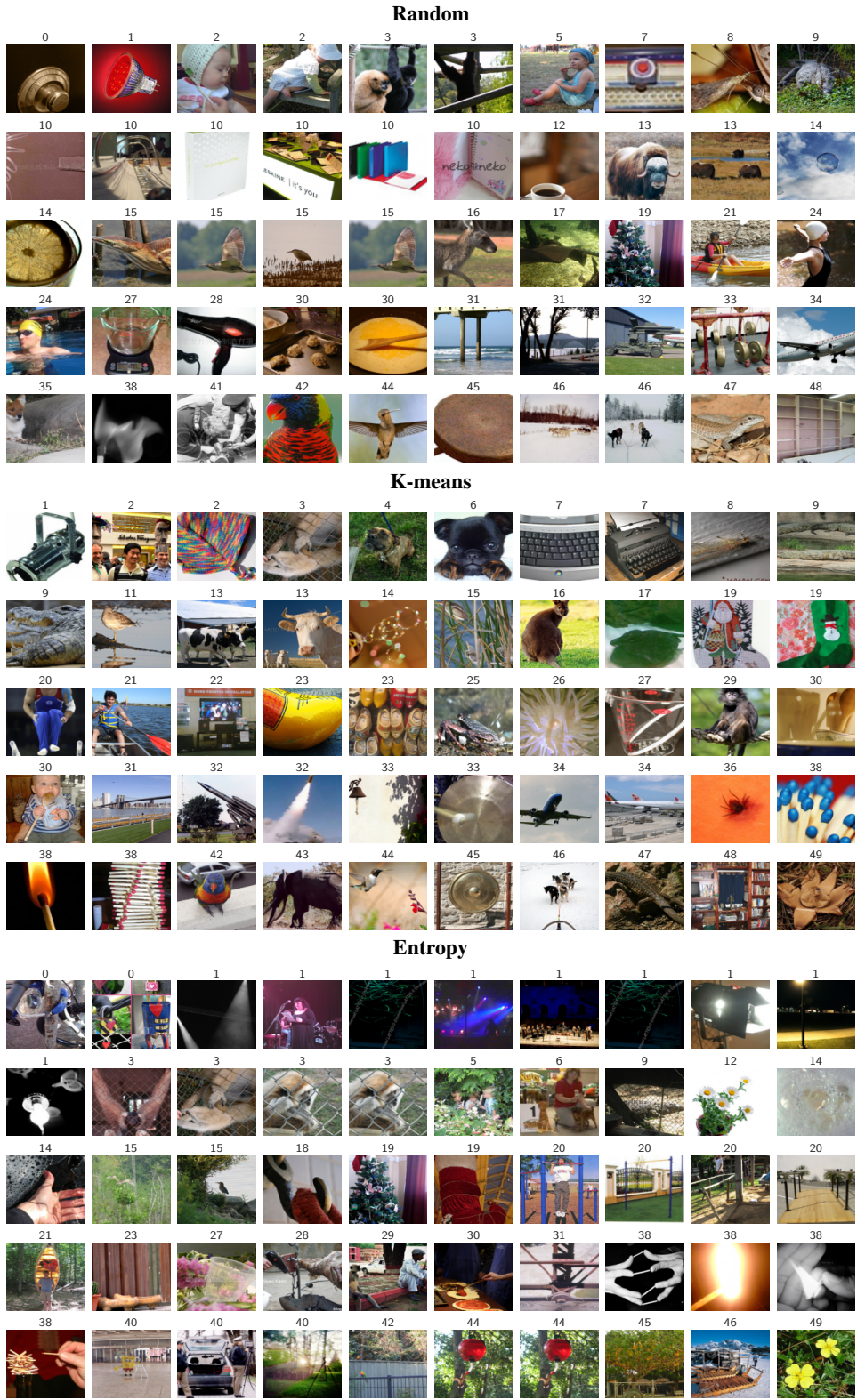

\begin{figure}[!b]
\centering
  \footnotesize
\begin{tikzpicture}
    \node[]  at (7, 3.5) {\textbf{Var Ratios}};
    \node[]  at (7, -3.5) {\textbf{BAIT}};
    \node[]  at (7, -10.5) {\textbf{FIRAL}};
\node[inner sep=0pt] (a) at (7,0) {\includegraphics[width=13cm,height = 6.7cm]{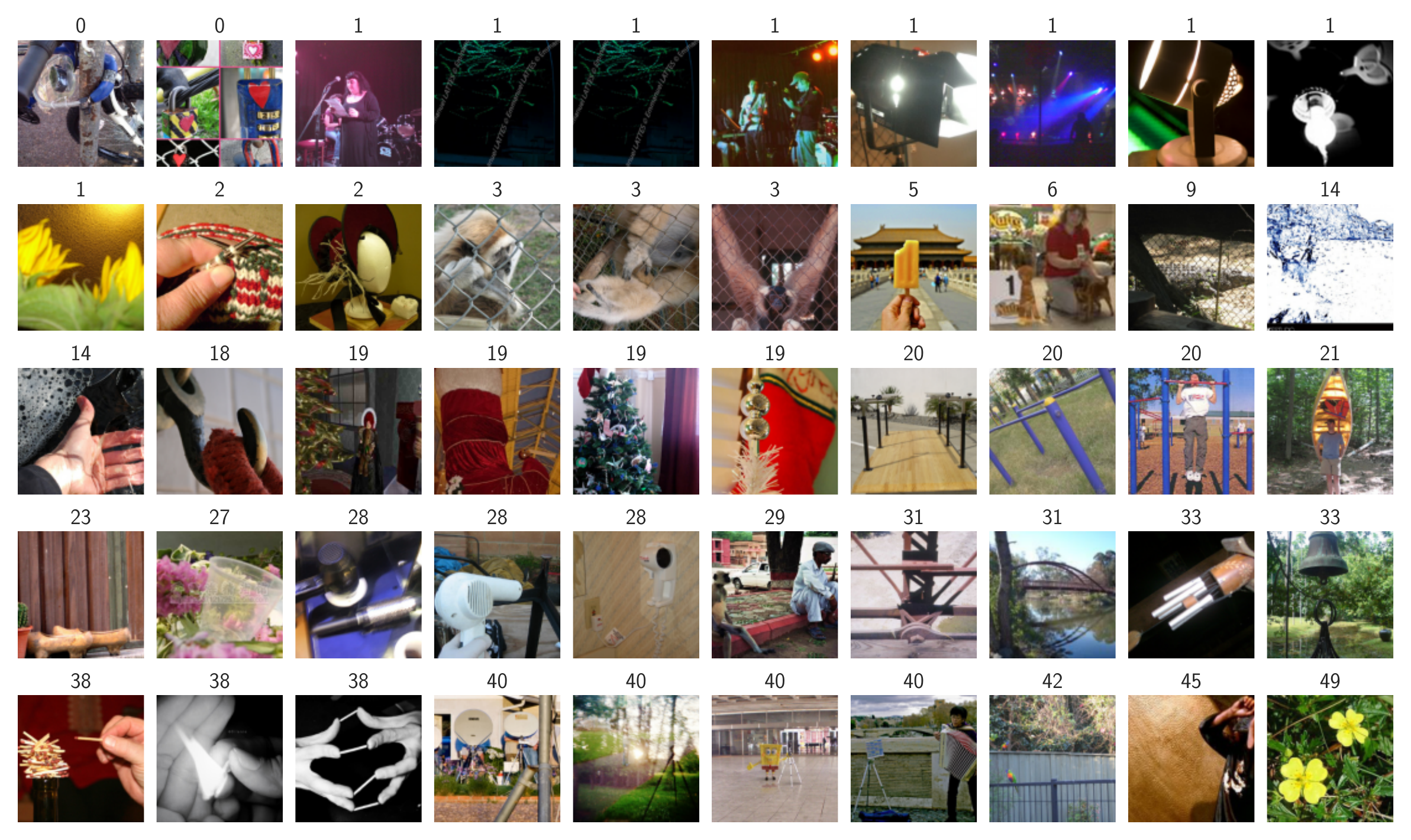}};
\node[inner sep=0pt] (b) at (7,-7) {\includegraphics[width=13cm,height = 6.7cm]{plots/imagenet/bait.pdf}};
\node[inner sep=0pt] (c) at (7,-14) {\includegraphics[width=13cm,height = 6.7cm]{plots/imagenet/fisher.pdf}};
\end{tikzpicture}
\caption{Selected samples for ImageNet-50 at the first round of active learning test. }
\label{fig:real-imagenet-2}
\end{figure}

\paragraph{Additional results.} We have presented the classification accuracy on unlabeled set in \Cref{fig:real-accuracy}. In \Cref{fig:real-weights}, we plot the normalized weights $z_{\diamond}$ (i.e. the solution of the relaxed problem \Cref{eq:mtd-obj-relax}) at each round of active learning tests.  We present the images selected by different active learning methods for MNIST (\Cref{fig:real-mnist}), CIFAR-10 (\Cref{fig:real-cifar10}), and ImageNet-50 (\Cref{fig:real-imagenet-1,fig:real-imagenet-2}).

\end{document}